\documentclass{article}
\usepackage{algpseudocode}
\usepackage{algorithm}
\usepackage{amsthm}
\usepackage{amstext}
\usepackage{amsmath}
\usepackage{amssymb}
\usepackage{array}
\usepackage{appendix}
\usepackage{booktabs}
\usepackage{caption}
\usepackage{courier}
\usepackage{color,soul}
\usepackage{color,caption}
\usepackage{color,multirow}
\usepackage{epsfig}
\usepackage{epstopdf}
\usepackage{geometry}
\usepackage{graphics}
\usepackage{graphicx}
\usepackage{helvet}
\usepackage{latexsym}
\usepackage{mathrsfs}
\usepackage{multirow}
\usepackage{setspace}
\usepackage{subfigure}
\usepackage[pagewise]{lineno}

\newcommand{\argmin}{\mathop{\mathrm{arg\,min}}}

\usepackage{url}
\usepackage{xcolor}
\begin{document}
\title{Tensor train rank minimization with nonlocal self-similarity for tensor completion}

\author{Meng Ding$^{a}$\footnote{E-mail: dingmeng56@163.com},\mbox{ } Ting-Zhu Huang$^{a}$\footnote{Corresponding author. E-mail: tingzhuhuang@126.com},\mbox{ } Xi-Le Zhao$^{a}$\footnote{Corresponding author. E-mail: xlzhao122003@163.com},\mbox{ } Michael K. Ng$^{b}$\footnote{E-mail: mng@maths.hku.hk},\mbox{ } Tian-Hui Ma$^{c}$\footnote{E-mail: nkmth0307@126.com}\\
{\small \it a. School of Mathematical Sciences,}\\
{\small \it University of Electronic Science and Technology of China, Chengdu, Sichuan, 611731, P.R. China}\\
{\small \it b. Department of Mathematics, The University of Hong Kong, Pokfulam, Hong Kong}\\
{\small \it c. School of Mathematics and Statistics,}\\
{\small \it Xi'an Jiaotong University, Xi'an, Shaanxi, 710049, P.R.China.}}
\date{}
\maketitle


\begin{abstract}
The tensor train (TT) rank has received increasing attention in tensor completion due to its ability to capture the global correlation of high-order tensors ($\textrm{order} >3$). For third order visual data, direct TT rank minimization has not exploited the potential of TT rank for high-order tensors. The TT rank minimization accompany with \emph{ket augmentation}, which transforms a lower-order tensor (e.g., visual data) into a higher-order tensor, suffers from serious block-artifacts. To tackle this issue, we suggest the TT rank minimization with nonlocal self-similarity for tensor completion by simultaneously exploring the spatial, temporal/spectral, and nonlocal redundancy in visual data. More precisely, the TT rank minimization is performed on a formed higher-order tensor called group by stacking similar cubes, which naturally and fully takes advantage of the ability of TT rank for high-order tensors. Moreover, the perturbation analysis for the TT low-rankness of each group is established. We develop the alternating direction method of multipliers tailored for the specific structure to solve the proposed model. Extensive experiments demonstrate that the proposed method is superior to several existing state-of-the-art methods in terms of both qualitative and quantitative measures.
\end{abstract}

Key words: low-rank tensor completion, tensor train rank, nonlocal self-similarity, alternating direction method of multipliers.
\section{Introduction}
\label{sec:Int}
Tensor completion aims at estimating missing entries or damaged parts in high-dimensional data and plays an important role in computer vision, e.g., color image inpainting \cite{Bertalmio2000Image-inpainting,Komodakis2006Image-Completion,Liu2019Image,Zhao2015Bayesian}, video inpainting \cite{Chan2011An,Chen2014STDC,Zhang2018Nonconvex}, hyperspectral images recovery \cite{Li2012Coupled,Xing2012Dictionary,Zhao2013Deblurring}, higher-order web link analysis \cite{Kolda2005Higher,Liu2015Trace}, and seismic data reconstruction \cite{Ely2015Seismic}. As a typical ill-posed inverse problem, stable tensor completion processes usually rely on prior knowledge of the underlying tensor. Recently, the low-rankness prior has demonstrated to be a powerful tool for tensor completion, namely low-rank tensor completion (LRTC). A common way to characterize the low-rankness of tensors is to decompose them into several lower-dimensional multilinear spaces. Representative works on tensor rank include CANDECOMP/PARAFAC (CP) rank, Tucker rank \cite{Kolda2009Tensor}, and tubal rank \cite{Kilmer2013Third-Order}; see Section \ref{section:Related works} for a brief review on these related works.

Recently, the tensor train (TT) rank has achieved great success in LRTC. Given a $j$-th order tensor $\mathcal{X}\in \mathbb{R}^{n_{1}\times \cdots \times n_{j}}$, TT decomposition \cite{Oseledets2011Tensor-Train-Decomposition} models each element of $\mathcal{X}$ by
\begin{equation}\label{TT decomposition}
x_{i_{1},\ldots,i_{j}} = \mathcal{G}_{1}(:,i_{1},:) \cdots \mathcal{G}_{j}(:,i_{j},:),
\end{equation}
where $\mathcal{G}_{k}\in \mathbb{R}^{r_{k-1} \times n_{k} \times r_{k}}$, $k=1, \cdots, j$ with $r_{0} = r_{j} =1$. The TT rank corresponding to \eqref{TT decomposition} is defined as $(r_{1},\ldots,r_{j-1})$. The TT rank has shown to be effective in tensor completion due to its ability of capturing the intrinsic structure within higher-order tensors. Some optimization methods has been proposed for TT rank minimization, such as alternating minimization \cite{Grasedyck2015TT,Wang2016TT}, simple low-rank tensor completion via tensor train (SiLRTC-TT), and tensor completion by parallel matrix factorization via tensor train (TMac-TT) \cite{Bengua2017Efficient}; see Section \ref{section:Related works} for more details.

\begin{figure}[!ht]
\scriptsize\setlength{\tabcolsep}{0.9pt}
\begin{center}
\begin{tabular}{c}
\includegraphics[width=0.9\textwidth]{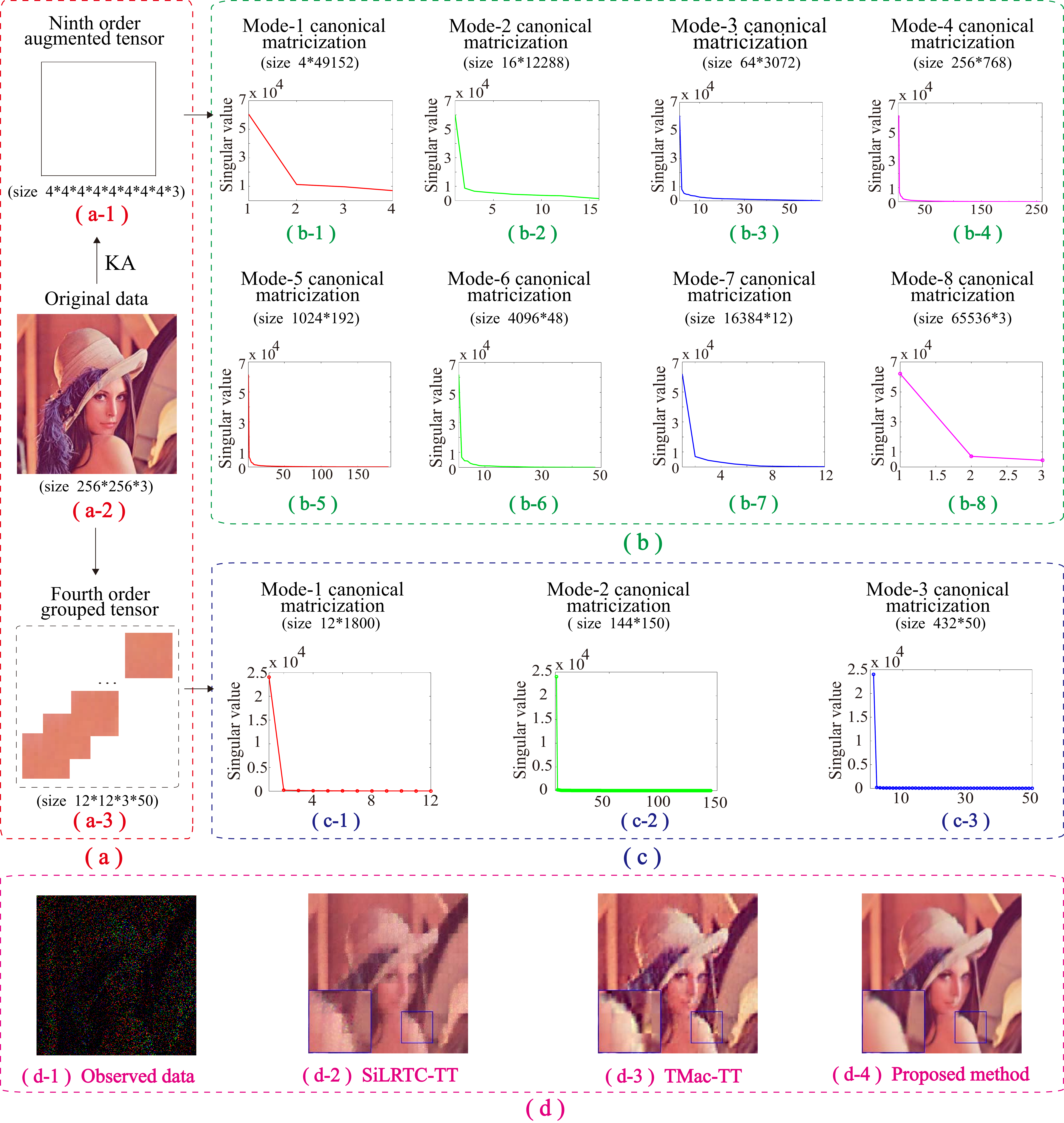}\vspace{0.1cm}\\
\end{tabular}
\caption{\small{Comparison of TT low-rankness of higher-order tensors generated by KA and NSS. (a-1, 2, 3) the augmented tensor, the original data, and an example of grouped tensors. (b-1) to (b-8) the distribution of singular values of the mode-1 to mode-8 canonical matricizations of the augmented tensor (a-1) and the average ratio of singular values larger than $1\%$ of the corresponding largest ones is $25.8\%$. (c-1, 2, 3) the distribution of singular values of the mode-1, mode-2, and mode-3 canonical matricizations of the grouped tensor (a-3) and the average ratio of singular values larger than $1\%$ of the corresponding largest ones is $1.5\%$. (d-1, 2, 3, 4) the observed data, the recovered results by SiLRTC-TT, TMac-TT, and the proposed method.}}
  \label{fig:motivation}
  \end{center}\vspace{-0.3cm}
\end{figure}


Nevertheless, existing TT-based LRTC methods still leave much room for further improvement. Since most multidimensional visual data are of third order in practical applications, direct TT-rank minimization has not yet excavated the estimation potential of TT rank minimization. When handling third order visual data, such as color images and multispectral images, most existing methods use $ket \ augmentation$ (KA) \cite{Latorre2005Image} as a tensor order increment preprocessing to transform a lower-order tensor into a higher-order tensor. However, KA uses a fixed rule to stack blocks extracted from the original data, without considering the inherent correlation between different blocks, which leads to serious block-artifacts on restored images \cite{Ding2019TTTV}; see Fig. \ref{fig:motivation} (d) for an example. To overcome this issue, our previous work \cite{Ding2019TTTV} used the total variation regularizer to depict the spatial local smoothness prior of the underlying data. Although alleviating the block-artifacts to some extent, this method is still a palliative one, due to the ignorance of the intrinsic structural redundancy of real-world data. These motivate us to find a more adaptive method to retain the strength of TT rank and alleviate block-artifacts.


In this paper, we propose a novel scheme to adaptively generate higher-order tensors with TT low-rankness, by exploring the nonlocal self-similarity (NSS) prior of tensor data. As a significant intrinsic prior of natural images, NSS depicts the redundancy of repeated similar structures across a natural image, which has been demonstrated to be powerful in various image processing applications \cite{Chang2017Hyper,Dabov2007Image,Gu2014Weighted,Huang2018rank,MA2016Group,Zhang2012Single}. Our main idea is to stack similar cubes into a higher-order tensor called a group. The motivation behind is that the similarity between cubes naturally implies the TT low-rankness of each group, which is more natural and effective than the fixed KA scheme. In fact, the perturbation analysis for the TT low-rankness of each group is established in Section \ref{section:Proposed Method}. To explain our motivation, in Figure \ref{fig:motivation}, we compare the TT low-rankness of two higher-order tensors generated by KA and NSS. We obtain two insights from Fig. \ref{fig:motivation}. First, the canonical matricizations of the grouped higher-order tensor obtained by NSS exhibits low-rank property more significantly than that obtained by KA, which can be visually verified by the singular value curves shown in Fig. \ref{fig:motivation} (c-1, 2, 3) and Fig. \ref{fig:motivation} (b-1) to (b-8). Second, the proposed method can effectively alleviate block-artifacts compared with SiLRTC-TT and TMac-TT.

\begin{figure}[!th]
\scriptsize\setlength{\tabcolsep}{0.9pt}
\begin{center}
\begin{tabular}{c}
\includegraphics[width=0.9\textwidth]{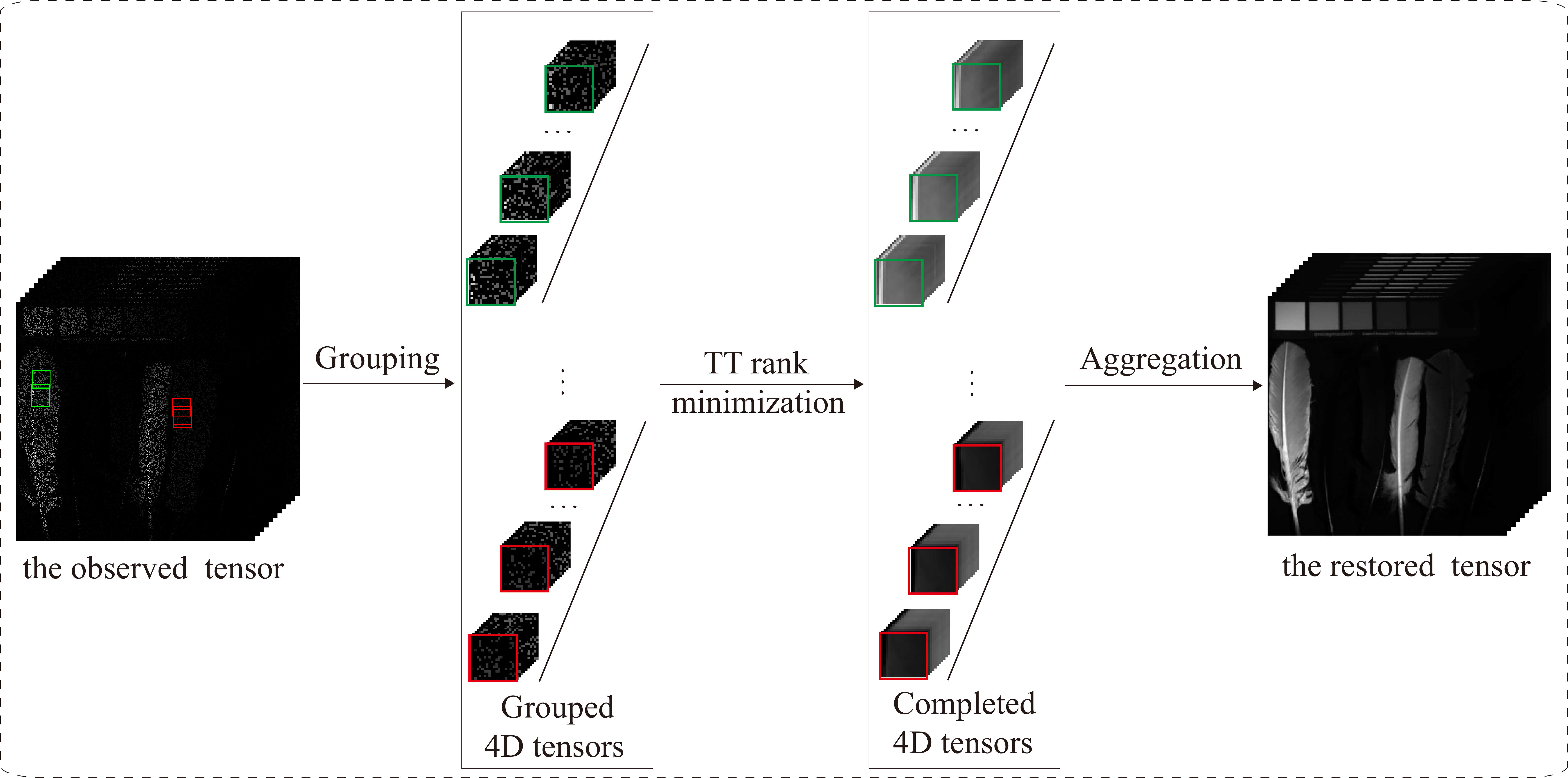} \vspace{0.1cm}\\
\end{tabular}
\caption{\small{Flowchart of the proposed completion framework.}}
  \label{fig:flowchart}
  \end{center}\vspace{-0.3cm}
\end{figure}

Once established, we develop a group-based TT rank minimization framework for tensor completion, which simultaneously exploits various data prior knowledge, such as spatial, temporal/spectral, and NSS redundancy. The main idea is to consider group as the basis unit of completion, and to impose the low-TT-rank constraint on each group to learn the correlations of all modes. Specifically, the proposed framework involves three steps. First, we stack $j$-th order similar cubes together into an $(j+1)$-th order tensor called a group and consider the group as the basis unit of completion. Second, we complete each group by solving a TT rank minimization model. Third, we calculate the final result by aggregating the final completed tensor by averaging all completed groups. The flowchart of the proposed method is illustrated in Fig. \ref{fig:flowchart}.


In summary, our contributions are mainly three folds: (1) we propose an adaptive strategy based on the NSS prior to fully exploit the potential of TT rank for high-order tensors, and establish a perturbation analysis for the TT low-rankness of groups consisting of similar cubes; (2) we propose a group-based TT rank minimization framework for tensor completion to simultaneously exploit the spatial, temporal/spectral, and NSS priors of tensor data; (3) experiments show that our method outperforms several state-of-the-art methods in handling multiple tensor completion problems.

The outline of this paper is as follows. Section \ref{section:Preliminary} states some basic knowledge about tensors. Section \ref{section:Related works} reviews some works about LRTC. Section \ref{section:Proposed Method} gives the details of the proposed method and establishes a perturbation analysis of the TT low-rankness of each group. Section \ref{section:Proposed Algorithm} develops an alternating direction method of multipliers (ADMM)-based solving algorithm. Section \ref{section:Experiments} presents extensive numerical experiments. Section \ref{section:Discussion} discusses parameters study and the numerical convergence of the proposed algorithm. Section \ref{section:Conclusion} summarizes this paper.
\section{Preliminary}
\label{section:Preliminary}
We denote scalars, vectors, matrices, and tensors as lowercase letters (e.g., $z$), boldface lowercase letters (e.g., \textbf{z}), boldface capital letters (e.g., $\textbf{Z}$), and calligraphic letters (e.g., $\mathcal{Z}$), respectively. A tensor is a high-dimensional array and its order (or mode) is the number of its dimensions. Given a $j$-th order tensor $\mathcal{Z}\in \mathbb{R}^{n_{1}\times \cdots \times n_{j}}$, its $(i_{1},\ldots,i_{j})$-th component is denoted as $z_{i_{1},\ldots,i_{j}}$. A mode-$k$ fiber of $\mathcal{Z}$ is a vector $\textbf{z}_{i_{1},\ldots,i_{k-1},:,i_{k+1},\ldots,i_{j}}$ obtained by varying index $i_{k}$ in while keeping the others fixed. The Frobenius norm of $\mathcal{Z}$ is $\|\mathcal{Z}\|_{F} = \sqrt{\langle\mathcal{Z}, \mathcal{Z}\rangle}$, where $\langle\mathcal{X}, \mathcal{Y}\rangle = \sum_{i_{1},\ldots,i_{j}} x_{i_{1},\ldots,i_{j}}\cdot y_{i_{1},\ldots,i_{j}}$ is the inner product of two tensors $\mathcal{X}$, $\mathcal{Y}$.

The mode-$k$ canonical matricization of $\mathcal{Z}$ is defined as $\textbf{Z}_{[k]}\in \mathbb{R}^{(\prod_{d=1}^{k}n_{d}) \times (\prod_{d=k+1}^{j}n_{d})}$, where the element $(i_{1},\ldots,i_{j})$ of $\mathcal{Z}$ maps to the element $(a, b)$ of $\textbf{Z}_{[k]}$ satisfying
\begin{equation}\label{mode-$k$ canonical matricization}
\begin{split}
& a = 1+\sum_{d=1}^{k}\big((i_{d}-1)\prod_{t=1}^{d-1}n_{t}\big) \\
\textrm{and} \quad &b = 1+\sum_{d=k+1}^{j}\big((i_{d}-1)\prod_{t=k+1}^{d-1}n_{t}\big).
\end{split}
\end{equation}
In MATLAB, it can be implemented by the reshape function
\[
\textbf{Z}_{[k]}=\text{reshape}(\mathcal{Z}, \Pi_{d=1}^{k}n_{d}, \Pi_{d=k+1}^{j}n_{d}).
\]
We denote the inverse operator as $\text{fold}_{[k]}(\textbf{Z}_{[k]})=\mathcal{Z}$. The TT nuclear norm of $\mathcal{Z}$ is defined as $\| \mathcal{Z}\|_{\ast} = \sum_{k=1}^{j-1}\alpha_{k}\|\textbf{Z}_{[k]}\|_{\ast}$, where $\{\alpha_{k}\}_{k=1}^{j-1}$ are positive constants satisfying {\small$\sum_{k=1}^{j-1}\alpha_{k}=1$.}

\section{Related works}
\label{section:Related works}
In this section, we briefly introduce some related works on the definition of tensor rank, including CP rank, Tucker rank \cite{Kolda2009Tensor}, tubal rank \cite{Kilmer2013Third-Order}, and TT rank \cite{Oseledets2011Tensor-Train-Decomposition}.

The CP rank is defined as the smallest number of rank-one tensors that generate the target tensor. There exist some heuristic CP-based LRTC methods \cite{Liu2015Trace,Liu2019Robust,Papalexakis2016TDM,Sidiropoulos2017Tensor,Yokota2016Smooth,Zhao2015Bayesian}, with promising performance. Nevertheless, computing CP rank is generally NP-hard \cite{Hillar2013Most}, which limits its application. The Tucker rank is defined as a vector composed of ranks of unfolding matrices of the target tensor \cite{Kolda2009Tensor,Wang2018Hyperspectral}. Existing Tucker rank minimization methods include convex relaxation methods  \cite{Du2019Exploiting,Gandy2011low-n-rank,Liu2013tensor,Wang2018Sparse} and low-rank matrix factorization methods \cite{Ji2016Tensor,Xu2017Parallel,Zheng2019smooth}. The limitation of Tucker rank is that it only captures the correlation between one mode and all the rest modes of the tensor, due to the unbalanced matricization scheme in the unfolding operator \cite{Bengua2017Efficient}. The tensor tubal rank is defined as the number of nonzero singular tubes under the tensor singular value decomposition (tSVD) of the target tensor (please see \cite{Kilmer2013Third-Order,Lu2016TRPCA} for more details). Zhang et al. \cite{Zhang2017tSVD} proposed tensor nuclear norm (TNN) as a convex surrogate of tubal rank for LRTC. Moreover, Lu et al. \cite{lu2018exact} established the theoretical guarantee of TNN minimization for low tubal rank tensor recovery from Gaussian measurements.

For TT rank, according to \cite{Oseledets2011Tensor-Train-Decomposition}, there exists decomposition that makes $r_{k}$ equal to the rank of the canonical matrix $X_{[k]}\in \mathbb{R}^{(\prod_{d=1}^{k}n_{d}) \times (\prod_{d=k+1}^{j}n_{d})}$, a well-balanced matricization characterizing the correlation between the first $k$ and the rest $j-k$ dimensions of $\mathcal{X}$ \cite{Bengua2017Efficient}. Compared with existing works, the TT decomposition and TT rank have two significant advantages. First, the TT decomposition is free from the ``curse of dimensionality" \cite{Oseledets2011Tensor-Train-Decomposition}, which enables its applications to large-scale problems. Second, the TT rank, approximately calculated by the rank of canonical matrices, can capture the global correlation of higher-order tensors due to its well-balanced matricization scheme, i.e., matricizing the tensor along permutations of modes.

Researchers have developed a series of methods for TT rank minimization. Grasedyck et al. \cite{Grasedyck2015TT} and Wang et al. \cite{Wang2016TT} proposed an iterative algorithm by alternatively updating each core tensor. However, \cite{Grasedyck2015TT} fixed the TT rank as $r_{1}=\cdots=r_{j-1}$, and \cite{Wang2016TT} assumed that the TT rank is given. Recently, Bengua et al. \cite{Bengua2017Efficient} proposed two TT minimization models for LRTC. The first one, simple low-rank tensor completion via tensor train (SiLRTC-TT), minimizes the sum of nuclear norm of canonical matrices, i.e.,

{\small
\begin{equation}\label{TT nuclear norm}
\begin{split}
\min_{\mathcal{X}} \quad & \sum_{k=1}^{j-1}\alpha_{k}\|\textbf{X}_{[k]}\|_{\ast}\\
s.t.     \quad & \mathcal{P}_{\Omega}(\mathcal{X})=\mathcal{P}_{\Omega}(\mathcal{T}),
\end{split}
\end{equation}}
where $\textbf{X}_{[k]}$ is the mode-$k$ canonical matricization of the tensor $\mathcal{X}$ and $\{\alpha_{k}\}_{k=1}^{j-1}$ are positive constants satisfying $\sum_{k=1}^{j-1}\alpha_{k}=1$, $\mathcal{T}\in \mathbb{R}^{n_{1}\times \ldots \times n_{j}}$ is the observed tensor, $\Omega$ is the index of observed entries, and $\mathcal{P}_{\Omega}(\cdot)$ is the projection operator that keeps entries in $\Omega$ and zeros out others. Another one, tensor completion by parallel matrix factorization via tensor train (TMac-TT), uses matrix factorization to approximate the TT rank, i.e.,
{\small\begin{equation}\label{TT matrix factorization}
\begin{split}
\min_{\{\textbf{W}_{k}\}_{k=1}^{j-1}, \{\textbf{Z}_{k}\}_{k=1}^{j-1}, \mathcal{X}} \quad & \sum_{k=1}^{j-1}\frac{\alpha_{k}}{2}\|\textbf{W}_{k}\textbf{Z}_{k}-\textbf{X}_{[k]}\|_{F}^{2}\\
s.t.     \quad & \mathcal{P}_{\Omega}(\mathcal{X})=\mathcal{P}_{\Omega}(\mathcal{T}),
\end{split}
\end{equation}}where $\textbf{W}_{k}\in \mathbb{R}^{(\prod_{d=1}^{k}n_{d})\times r_{k}}$, $\textbf{Z}_{k}\in \mathbb{R}^{r_{k}\times (\prod_{d=k+1}^{j}n_{d})}$, and $r_{k}$ is the rank of the matrix $\textbf{X}_{[k]}$.

\section{Tensor completion via nonlocal TT rank minimization}
\label{section:Proposed Method}
The proposed method, called tensor completion via \textbf{n}on\textbf{l}ocal \textbf{TT} rank minimization (NL-TT), involves three main steps: grouping, completion, and aggregation. Below we detail each step.

\textbf{Grouping.} We use a third order tensor $\mathcal{T}\in \mathbb{R}^{n_{1}\times n_{2} \times n_{3}}$ as an example to show how to construct groups, which can be easily extended to higher-order tensors. We extract reference cubes with size $s \times s \times n_{3}$ with overlapped size $o$, denoted as $\{\mathcal{\hat{T}}_{p}\}_{p=1}^{t}$, where the total number of reference cubes is $t=((n_{1}-s)/(s-o)+1)\times ((n_{2}-s)/(s-o)+1)$. We use block-matching \cite{Dabov2007Image} to find the locations of similar cubes and adopt the Euclidean distance to measure the similarity between two cubes. A smaller distance indicates a higher similarity. For each reference cube $\mathcal{\hat{T}}_{p}$, we assume that $h$ cubes $\{\mathcal{\hat{T}}_{p}^{a}\}$ ($a =1,2,\ldots,h$) similar to $\mathcal{\hat{T}}_{p}$ are found in spatial domain. These cubes are chosen to overlap to avoid possible block effects. Then we stack the similar cubes to form a group $\mathcal{T}_{p} \in \mathbb{R}^{s\times s \times n_{3}\times h} $ satisfying $\mathcal{T}_{p}(:,:,:,a)=\mathcal{\hat{T}}_{p}^{a}$.

\textbf{Completion}. Let $\mathcal{T}_{p}$ be a $j$-th order grouped tensor, and $\Omega_{p}$ be the indicating known pixels of $\mathcal{T}_{p}$. To complete $\mathcal{T}_{p}$, we consider the following TT nuclear norm minimization model:
\begin{equation}\label{proposed model}
\begin{split}
\min_{\mathcal{X}_{p}} \quad & \|\mathcal{X}_{p}\|_{*}:=\sum_{k=1}^{j-1}\alpha_{k}\|\textbf{X}_{p,[k]}\|_{\ast}\\
s.t.     \quad & \mathcal{P}_{\Omega_{p}}(\mathcal{X}_{p})=\mathcal{P}_{\Omega_{p}}(\mathcal{T}_{p}),
\end{split}
\end{equation}
where $\textbf{X}_{p,[k]}$ is the mode-$k$ canonical matricization of the tensor $\mathcal{X}_{p}$ and $\{\alpha_{k}\}_{k=1}^{j-1}$ are positive constants satisfying $\sum_{k=1}^{j-1}\alpha_{k}=1$. The following proposition shows the existence of the solution of the proposed model \eqref{proposed model}.

\newtheorem{proposition}{\bf Proposition}
\begin{proposition} \label{Proposition}
The model \eqref{proposed model} has at least one minimizer.
\end{proposition}

The proof is provided in Appendix \ref{Appendix A}.

\textbf{Aggregation.} After completing each group, the obtained estimates actually form an over-complete representation of the final completion result. Since the cubes are overlapped and one cube can appear in more than one group, each pixel may be covered by several completed groups. The final completion result is calculated by first returning completed groups to their original positions and then averaging all covered cubes pixel-by-pixel.

\subsection{Perturbation analysis}
We establish a perturbation analysis for the TT low-rankness of the group consisting of similar cubes. Assume that $\mathcal{X}$ is a group with size $s\times s\times n_{3}\times h$. Next, we first show that $\mathcal{X}$ can be approximated by a series of TT low-rank tensors. Then, we present a perturbation analysis of the TT nuclear norm of $\mathcal{X}$ and its TT low-rank approximations.

Now we can establish an upper bound between $\mathcal{X}$ and a TT rank-$(1,1,1)$ tensor.

\newtheorem{theorem} {\bf Theorem}
\begin{theorem} \label{theorem1}
Let $\mathbf{x}$ be the average column fiber of $\mathcal{X}$, i.e., $\mathbf{x} = (\sum_{i_{2}=1}^{s}\sum_{i_{3}=1}^{n_{3}}\sum_{i_{4}=1}^{h}\mathcal{X}(:,i_{2},i_{3},i_{4}))/(sn_{3}h)$, and $\mathcal{Y}(:,i_{2},i_{3},i_{4})=\mathbf{x}, i_{2}=1,\ldots,s,i_{3}=1,\ldots,n_{3}$, and $i_{4}=1,\ldots,h$. Then the TT rank of $\mathcal{Y}$ is $(1,1,1)$. Suppose that
\[
\max_{i_{2},i_{3},i_{4}}\{\|\mathcal{X}(:,i_{2},i_{3},i_{4})-\mathbf{x}\|_{2}\}\leq \varepsilon.
\]
Let $\mathcal{X}=\mathcal{Y}+\mathcal{E}$. Then $\|\mathcal{E}\|_{F}\leq\sqrt{sn_{3}h}\varepsilon$.
\end{theorem}

We consider the TT rank-$(r_{1},r_{2},1)$ approximation of $\mathcal{X}$.

\begin{theorem} \label{theorem2}
Let $\mathcal{\hat{X}}$ be the average row-cube of $\mathcal{X}$, i.e., $\mathcal{\hat{X}} = (\sum_{i_{4}=1}^{h}\mathcal{X}(:,:,:,i_{4}))/h$, and $\mathcal{Y}(:,:,:,i_{4})=\mathcal{\hat{X}}, i_{4}=1,\ldots,h$. Then, $\mathcal{Y}$ is a TT rank-$(r_{1},r_{2},1)$ tensor, $r_{1}\leq \min\{s,sn_{3}h\}$, and $r_{2}\leq \min\{s^{2},n_{3}h\}$. Suppose that
\[
\max_{i_{4}}\{\|\mathcal{X}(:,:,:,i_{4})-\mathcal{\hat{X}}\|_{F}\}\leq \hat{\varepsilon}.
\]
Let $\mathcal{X}=\hat{\mathcal{Y}}+\hat{\mathcal{E}}$. Then $\|\hat{\mathcal{E}}\|_{F}\leq\sqrt{h}\hat{\varepsilon}$.
\end{theorem}


Next, we consider the TT rank-$(\tilde{r}_{1},\tilde{r}_{2},r)$ approximation of $\mathcal{X}$. Let $\mathcal{R}:=\{1,2,\ldots,r\}$.

\begin{theorem} \label{theorem3}
Let $\mathcal{X}=\mathcal{\tilde{Y}}+\mathcal{\tilde{E}}$, where $\mathcal{\tilde{Y}}(:,:,:,a)=\mathcal{X}(:,:,:,a), a=1,\ldots,r$, $\mathcal{\tilde{Y}}(:,:,:,a)=\mathcal{X}(:,:,:,1), a=r+1,\ldots,h$, i.e., $\mathcal{\tilde{Y}}=[\mathcal{X}(:,:,:,1),\mathcal{X}(:,:,:,2),\ldots,\mathcal{X}(:,:,:,r),\mathcal{X}(:,:,:,1),\ldots,\mathcal{X}(:,:,:,1)]\in \mathbb{R}^{s\times s \times n_{3} \times h}$. The TT rank of $\mathcal{Y}$ is $(\tilde{r}_{1},\tilde{r}_{2},r)$. Suppose that $\{\|\mathcal{X}(:,:,:,b)-\mathcal{X}(:,:,:,i_{4})\|_{F}\}\leq \tilde{\varepsilon}$ for $b\in \mathcal{R}$ and $i_{4}=1,\ldots,h$. Then
\[
\|\mathcal{\tilde{E}}\|_{F}\leq\sqrt{h-r}\tilde{\varepsilon}.
\]
\end{theorem}


Last, we display a perturbation analysis of the TT nuclear norm of $\mathcal{X}$ using the following lemma.

\newtheorem{lemma}{Lemma}
\begin{lemma} \label{lemma}
(Corollary 4.31, \cite{Stewart2001Matrix}) Let $\textbf{X},\textbf{E}\in \mathbb{C}^{m\times n} (m\geq n)$. Then
\[
|\sigma_{i}(\textbf{X}+\textbf{E})-\sigma_{i}(\textbf{X})|\leq \|\textbf{E}\|_{F},
\]
where $\sigma_{i}(\textbf{X})$ is the $i$-th largest singular value, $i=1,\ldots,n$.
\end{lemma}

\begin{theorem} \label{theorem4}
Let $\mathcal{X}=\mathcal{Y}+\mathcal{E}$, $s(\mathcal{X})=\sum_{k=1}^{3}\alpha_{k}\|\textbf{X}_{[k]}\|_{\ast}$, where $\|\textbf{X}_{[k]}\|_{\ast}=\sum_{j_{k}}\sigma_{j_{k}}(\textbf{X}_{[k]})$ and $\{\alpha_{k}\}_{k=1}^{3}$ are positive constants satisfying $\sum_{k=1}^{3}\alpha_{k}=1$. Then
\[
|s(\mathcal{X})-s(\mathcal{Y})|\leq c\|\mathcal{E}\|_{F},
\]
where $c$ is a positive constant.
\end{theorem}

The proofs are provided in Appendix \ref{Appendix B}.

\newtheorem{rem}{\bf Remark}
\begin{rem} \label{remark}
We can regard the stacked higher-order tensor as a TT low-rank tensor plus a perturbation noisy tensor. In this case, Theorems \ref{theorem1}, \ref{theorem2}, and \ref{theorem3} show that $\mathcal{X}$ can be approximated by the TT rank-$(1,1,1)$, rank-$(r_{1},r_{2},1)$, and rank-$(\tilde{r}_{1},\tilde{r}_{2},r)$ tensors, respectively. Theorem \ref{theorem4} illustrates that the singular values are insensitive to the perturbation in the tensor. When $\mathcal{Y}$ is a TT low-rank tensor and $\mathcal{E}$ is small, the stacked higher-order tensor approximates a TT low-rank tensor.
\end{rem}

The above results can be easily extended to higher-order tensors.

\section{The ADMM solver}
\label{section:Proposed Algorithm}
We develop ADMM \cite{Chan2013Constrained,Tom2014Fast-Alternating-Direction,Mei2017Cauchy} to solve the convex optimization problem \eqref{proposed model}. By introducing auxiliary variables $\{\mathcal{M}_{k}\}_{k=1}^{j-1}$, we obtain the equivalent constrained problem
\begin{equation}\label{constrained model}
\begin{split}
\argmin_{\mathcal{X},\mathcal{M}_{k}} & \ \sum_{k=1}^{j-1} \alpha_{k}\|\mathcal{M}_{k[k]}\|_{*}\\
s.t.    & \ \mathcal{P}_{\Omega}(\mathcal{X})=\mathcal{P}_{\Omega}(\mathcal{T}),\\
        & \ \mathcal{X}=\mathcal{M}_{k}, k=1,\ldots,j-1.
\end{split}
\end{equation}

The augmented Lagrangian function of \eqref{constrained model} is defined as
\begin{equation}\label{Lagrangian function}
\begin{split}
\mathcal{L}_{\beta}(\mathcal{X},\mathcal{M}_{k},\mathcal{Y}_{k})=\sum_{k=1}^{j-1} \alpha_{k} & \Big(\|\mathcal{M}_{k[k]}\|_{*}+\langle\mathcal{X}-\mathcal{M}_{k},\mathcal{Y}_{k}\rangle\\
&+\frac{\beta}{2}\|\mathcal{X}-\mathcal{M}_{k}\|_{F}^{2}\Big),
\end{split}
\end{equation}

\noindent where {\small$\{\mathcal{Y}_{k}\}_{k=1}^{j-1}$} are Lagrangian multipliers of the linear constraint and $\beta$ is the penalty parameter. We use the following iterative scheme to solve \eqref{Lagrangian function}:

\begin{equation}\label{iterative scheme}
\left\{
\begin{array}{l}
\begin{split}
\vspace{0.1cm}
&\mathcal{M}_{k}^{l+1} =\argmin_{\mathcal{M}_{k}}\mathcal{L}(\mathcal{X}^{l}, \mathcal{M}_{k}, \mathcal{Y}_{k}^{l}),\\
\vspace{0.2cm}
&\mathcal{X}^{l+1}=\argmin_{\mathcal{P}_{\Omega}(\mathcal{X}) = \mathcal{P}_{\Omega}(\mathcal{T}) }\mathcal{L}(\mathcal{X}, \mathcal{M}_{k}^{l+1}, \mathcal{Y}_{k}^{l}),\\
\vspace{0.1cm}
&\mathcal{Y}_{k}^{l+1}=\mathcal{Y}_{k}^{l}+\beta(\mathcal{X}^{l+1}-\mathcal{M}_{k}^{l+1}).\\
\end{split}
\end{array}
\right.
\end{equation}

We give the details for solving the first two subproblems in \eqref{iterative scheme}.

  (1)\ \textbf{$\{\mathcal{M}_{k}\}$-subproblem.} \ The optimal $\mathcal{M}_{k}$ is given by
  \begin{equation}\label{M subproblem}
  \mathcal{M}_{k}^{l+1} = \argmin_{\mathcal{M}_{k}} \|\mathcal{M}_{k[k]}\|_{*}+\frac{\beta}{2}\|\mathcal{X}^{l+1}-\mathcal{M}_{k}+\frac{\mathcal{Y}_{k}^{l}}{\beta}\|_{F}^{2}.
  \end{equation}
  Using the equation $\|\textbf{X}_{[k]}\|_{F} = \|\mathcal{X}\|_{F}$, we rewrite the \eqref{M subproblem} as the following problem:
  {\small
  \begin{equation}
  \mathcal{M}_{k}^{l+1}=\argmin_{\mathcal{M}_{k}} \|\mathcal{M}_{k[k]}\|_{*}+\frac{\beta}{2}\|\mathcal{X}^{l+1}_{[k]}-\mathcal{M}_{k[k]}+\frac{\mathcal{Y}_{k[k]}^{l}}{\beta}\|_{F}^{2}.
  \end{equation}}

  Using the singular value thresholding operator \cite{Cai2008SVD}, $\mathcal{M}_{k}$ has the closed-form solution
  \begin{equation}\label{M subproblem solver}
  \mathcal{M}_{k}^{l+1} = \textrm{fold}_{[k]}\big[ \textbf{U}_{k}S_{\frac{1}{\beta}}(\Sigma_{k}) \textbf{V}_{k}^{T}\big],
  \end{equation}
  where {\small $\mathcal{X}^{l+1}_{[k]}+\frac{\mathcal{Y}_{k[k]}^{l}}{\beta}=\textbf{U}_{k}\Sigma_{k} \textbf{V}_{k}^{T}$} and {\small $S_{\frac{1}{\beta}}(\Sigma_{k})= \textrm{diag}(\textrm{max}(\sigma_{k}-\frac{1}{\beta}))$}. The $\mathcal{M}_{k}$-subproblem involves the SVD of the matrix $\mathcal{M}_{k[k]}$ with size {\small $p_{k}\times q_{k}$ \big($p_{k}=\prod_{d=1}^{k}n_{d}, \ q_{k}=\prod_{d=k+1}^{j}n_{d}, \ k=1,\ldots,j-1$\big)}, whose complexity is $O\big(\textrm{min}\big(p_{k}^{2} q_{k},\ p_{k} q_{k}^{2}\big)\big)$.

  (2)\ \textbf{$\mathcal{X}$-subproblem.} \ The optimal $\mathcal{X}$ is the solution of the following quadratic problem:
  \begin{equation}\label{X subproblem}
  \mathcal{X}^{l+1} = \argmin_{ \mathcal{P}_{\Omega}(\mathcal{X}) = \mathcal{P}_{\Omega}(\mathcal{T}) } \sum_{k=1}^{j-1}\frac{\alpha_{k}\beta}{2}\|\mathcal{X}-\mathcal{M}_{k}^{l}+\frac{\mathcal{Y}_{k}^{l}}{\beta}\|_{F}^{2}.
  \end{equation}

  Then $\mathcal{X}$ can be calculated by
  \begin{equation}\label{X subproblem solver}
  \mathcal{X}^{l+1} = \mathcal{P}_{\Omega^{c}} \Bigg( \sum_{k=1}^{j-1}\alpha_{k}\Big(\mathcal{M}_{k}-\frac{1}{\beta}\mathcal{Y}_{k}\Big)\Bigg)+\mathcal{P}_{\Omega}(\mathcal{T}).
  \end{equation}

The cost of computing $\mathcal{X}$ is $O(\prod_{k=1}^{j}n_{k})$.

The proposed ADMM-based algorithm is summarized in Algorithm \ref{algorithm 2}. The minimization problem \eqref{constrained model} fits the framework of ADMM, and the proposed model is convex, thus the proposed algorithm is theoretically convergent \cite{Eckstein1992Douglas,He2012convergence}. At each iteration, the total cost of computing all variables is

\[
O\bigg(t\sum_{k=1}^{j-1}\textrm{min}\big(p_{k}^{2} q_{k},\ p_{k} q_{k}^{2}\big)\bigg),
\]
where $t$ is the number of reference cubes, $p_{k}=\prod_{d=1}^{k}n_{d}$, and $q_{k}=\prod_{d=k+1}^{j}n_{d}, \ k=1,\ldots,j-1.$

\renewcommand{\algorithmicrequire}{\textbf{Input:}} 
\renewcommand{\algorithmicensure}{\textbf{Output:}} 
\begin{algorithm}
\caption{ADMM-based algorithm for solving \eqref{proposed model}.}\label{algorithm 2}
\begin{algorithmic}[1]
      \Require The observed tensor $\mathcal{T}$, index set $\Omega$.
      \State \textbf{Grouping:} Perform block-matching to get $\{\mathcal{T}_{p}\}_{p=1}^{t}$.
	  \State \textbf{Out loop: For} $p=1, \cdots, t$, \textbf{do}
      \State  \quad \textbf{Parameters:} $\{\alpha_{k}\}_{k=1}^{j-1}$, $\beta$, and  inner iteration $l_{max}$.
      \State  \quad \textbf{Initialize:} $\mathcal{X}^{0}_{p}=\mathcal{T}_{p}$ and $\mathcal{Y}^{0}_{p,k}=0$.
      \State  \quad \textbf{Inner loop: While} {\small $l\leq l_{max}$} or {\footnotesize $\frac{\|\mathcal{X}^{l+1}_{p}-\mathcal{X}^{l}_{p}\|_{F}}{\|\mathcal{X}^{l}_{p}\|_{F}}\geq 10^{-4}$}, \textbf{do}
	  \State  \quad \qquad  \textbf{for} $k=1$ to $j-1$ \textbf{do};
	  \State  \quad \qquad \quad \quad update $\mathcal{M}_{k}$ via \eqref{M subproblem solver};
      \State  \quad \qquad  \textbf{end for};
	  \State  \quad \qquad  update $\mathcal{X}_{p}$ via \eqref{X subproblem solver};
      \State  \quad \textbf{end while}, and output $\mathcal{X}_{p}^{l+1}$.
	  \State \textbf{end for}, and output completed $\{\mathcal{X}_{p}\}_{p=1}^{t}$.
	  \Ensure Recovered data $\mathcal{X}$ via completed groups $\{\mathcal{X}_{p}\}_{p=1}^{t}$.
\end{algorithmic}
\end{algorithm}

\begin{figure}[!t]
\scriptsize\setlength{\tabcolsep}{0.9pt}
\begin{center}
\begin{tabular}{cccc}
\includegraphics[width=0.24\textwidth]{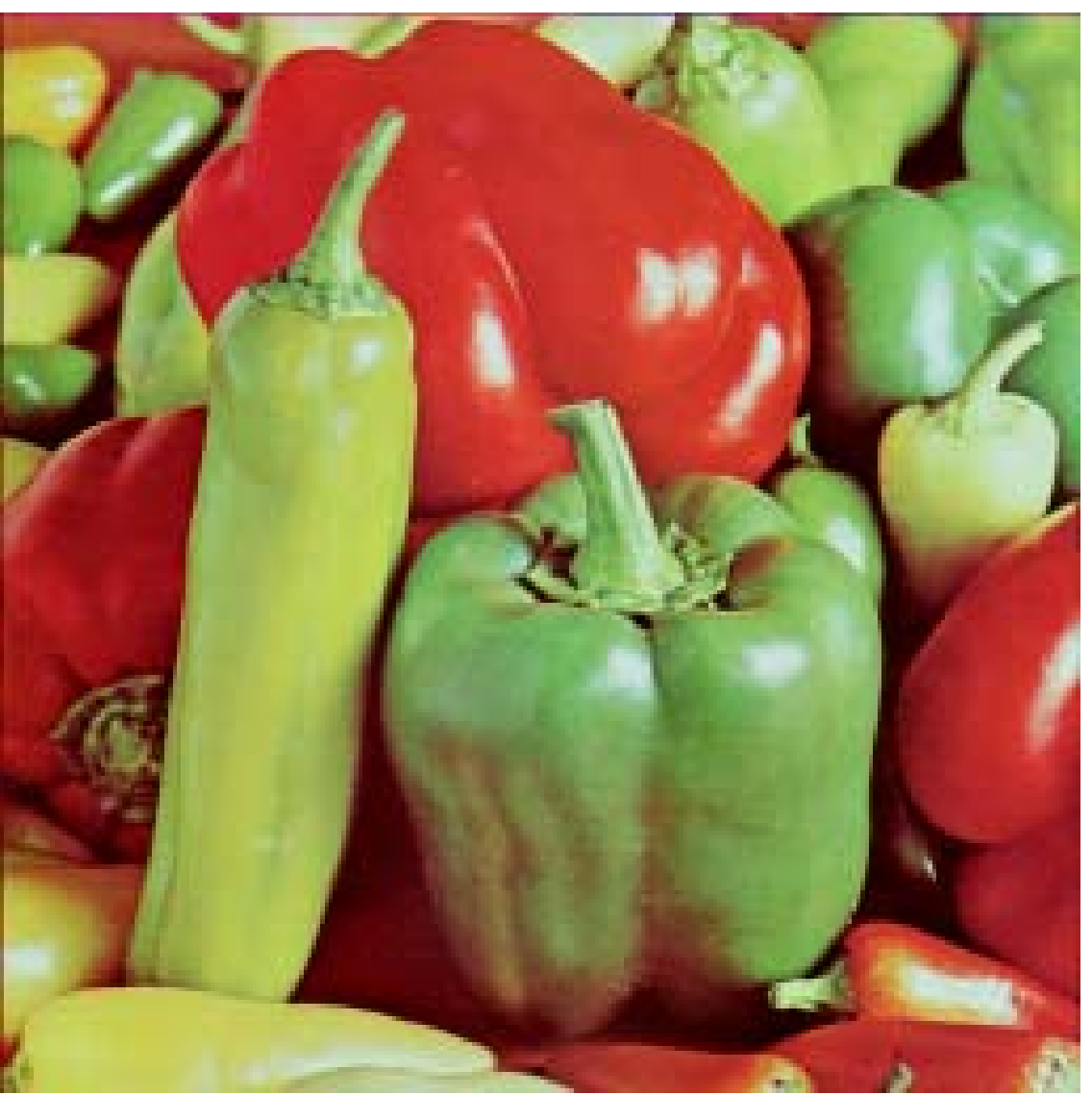}&
\includegraphics[width=0.24\textwidth]{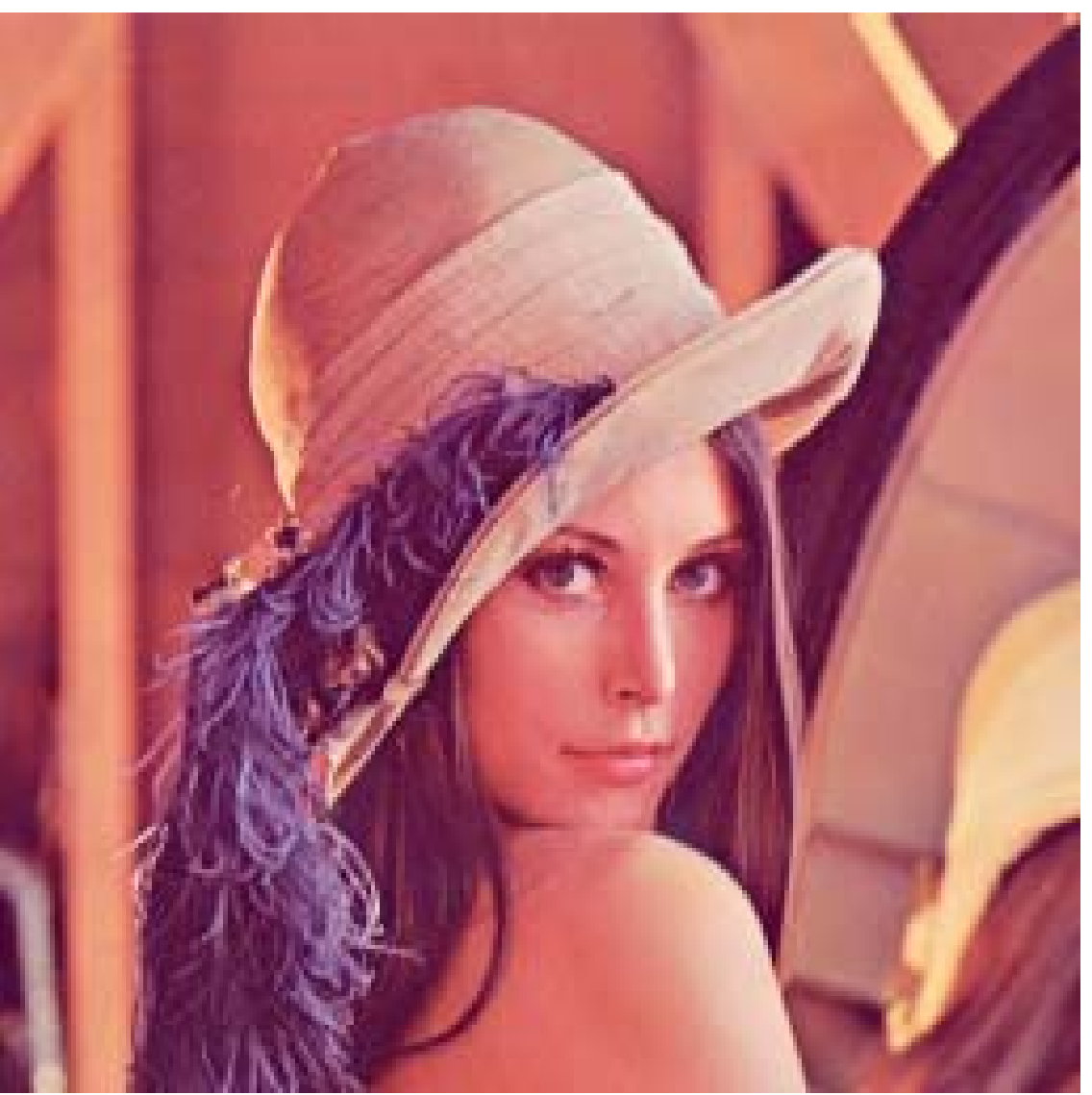}&
\includegraphics[width=0.24\textwidth]{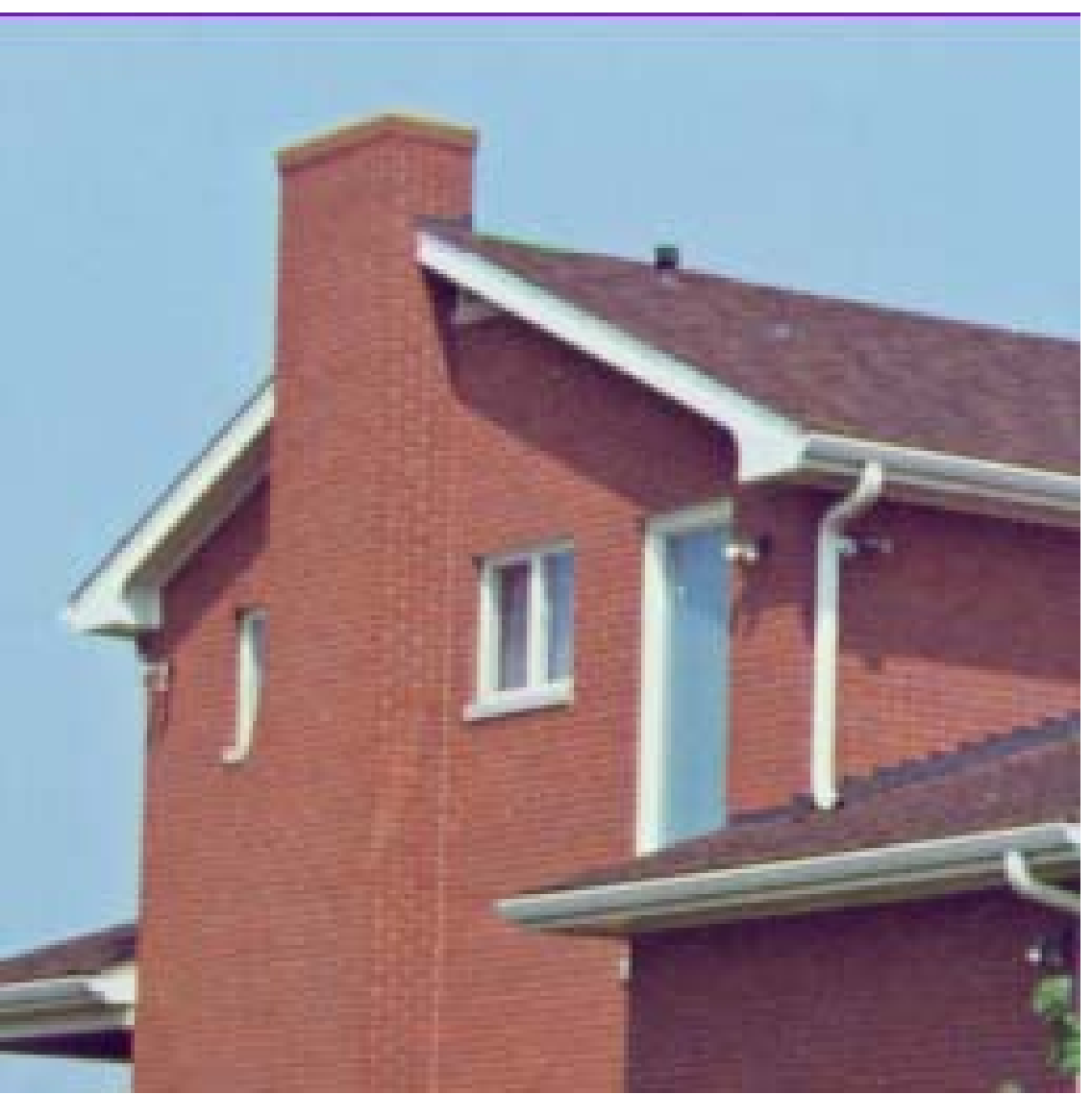}&
\includegraphics[width=0.24\textwidth]{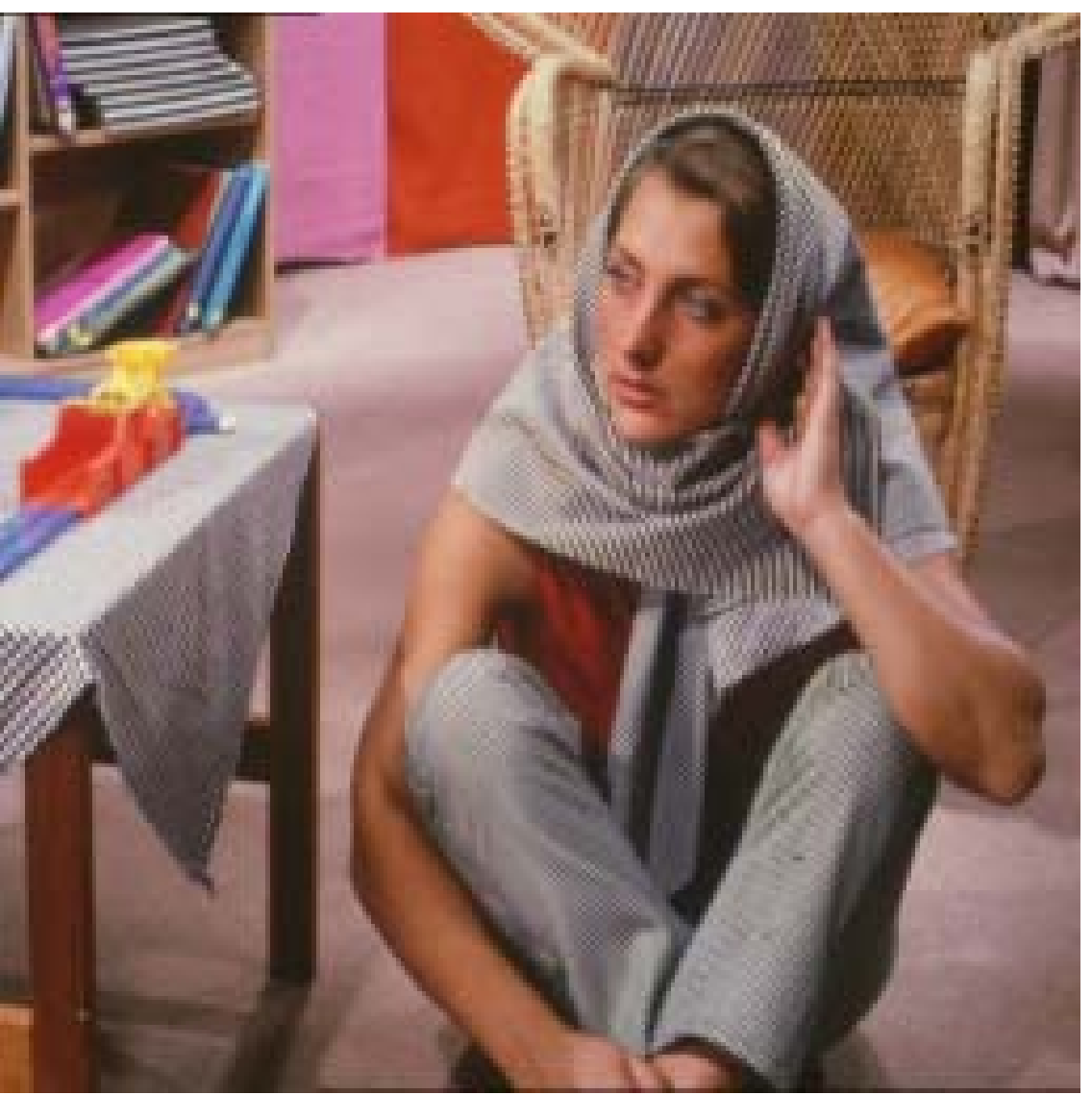}\\
{\footnotesize(a)} {\small \emph{peppers}} &  {\footnotesize(b)} {\small \emph{lena}} & {\footnotesize(c)} {\small \emph{house}} &  {\footnotesize(d)} {\small \emph{barbara}} \\
\includegraphics[width=0.24\textwidth]{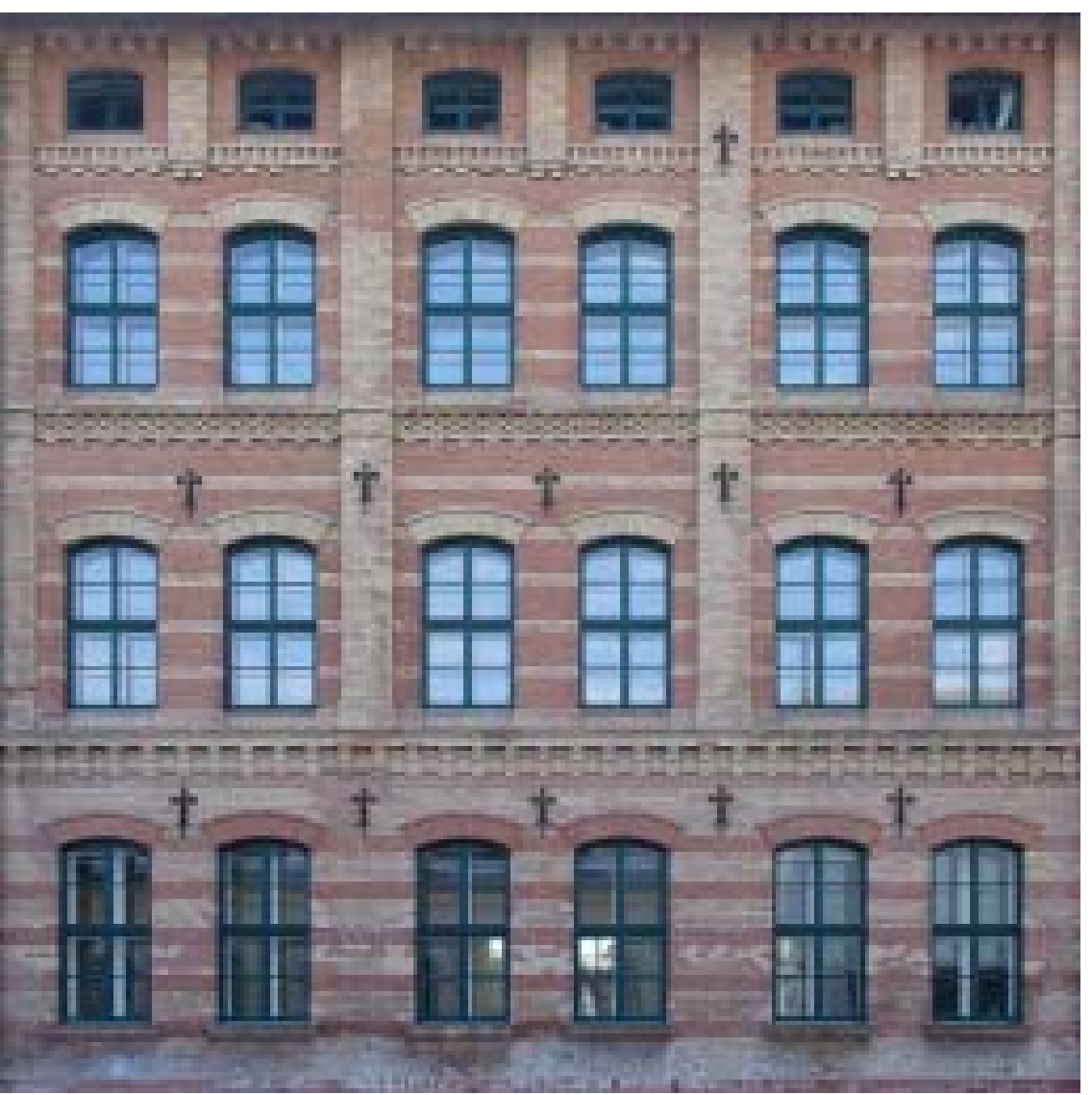}&
\includegraphics[width=0.24\textwidth]{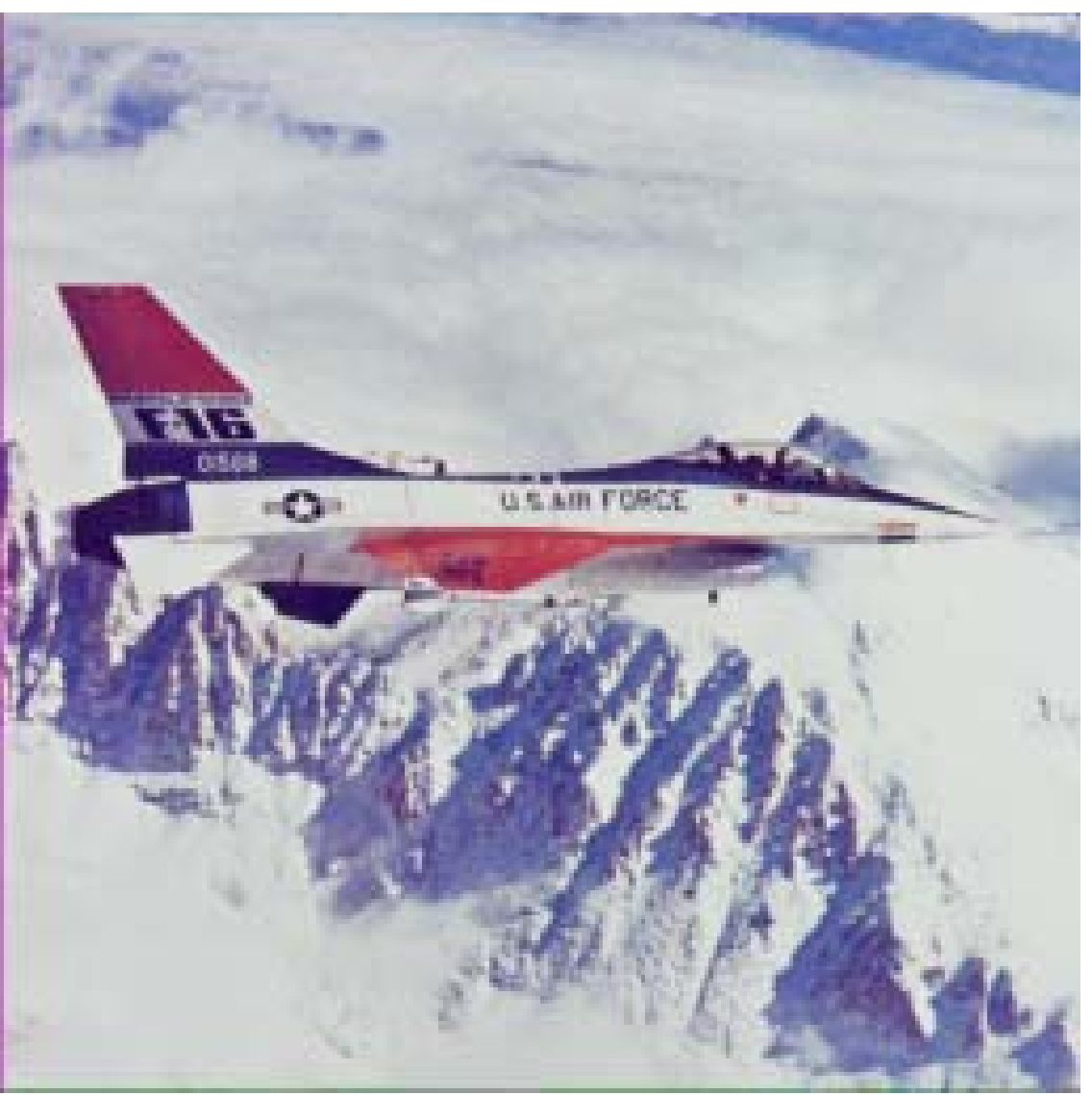}&
\includegraphics[width=0.24\textwidth]{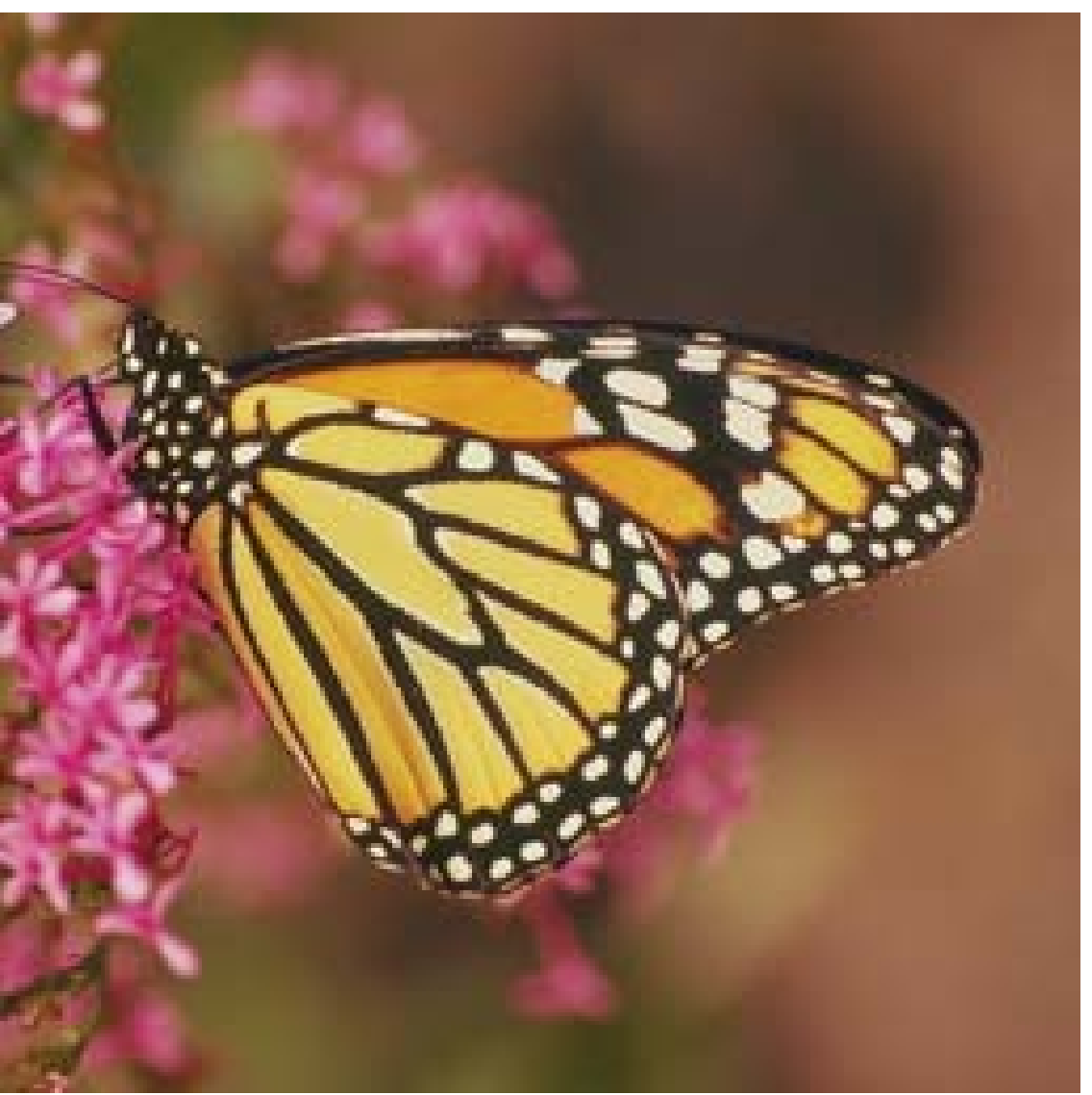}&
\includegraphics[width=0.24\textwidth]{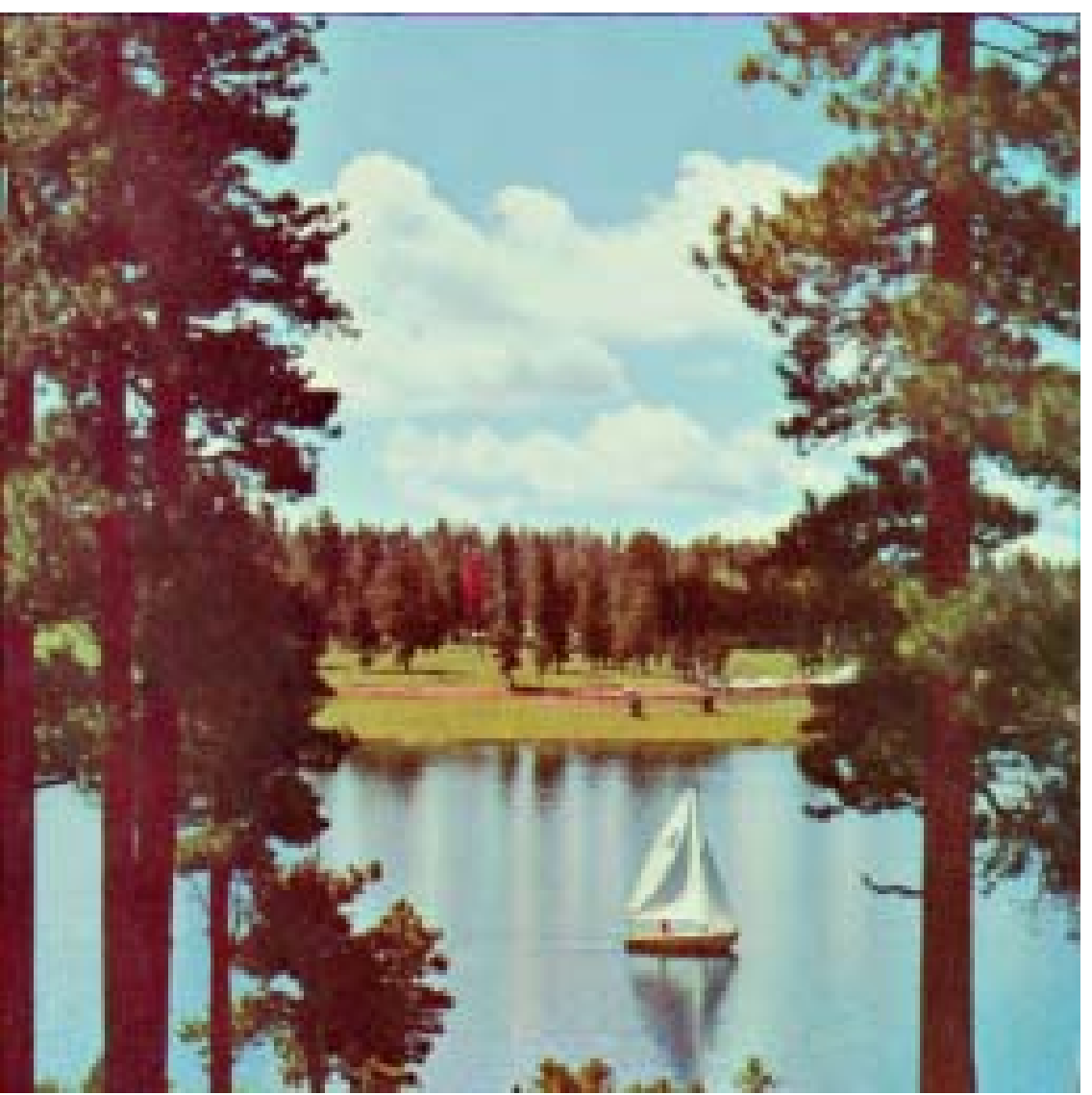}\\
{\footnotesize(e)} {\small \emph{facade}} &  {\footnotesize(f)} {\small \emph{airplane}} &  {\footnotesize(g)} {\small \emph{ monarch}} & {\footnotesize(h)} {\small \emph{sailboat}}\\
\end{tabular}
\caption{\small{Original images.}}
  \label{fig:ori_image}
  \end{center}\vspace{-0.3cm}
\end{figure}

\section{Experiments}
\label{section:Experiments}
In this section, we evaluate the performance of the proposed method by extensive experiments on color images, multispectral images (MSIs), and color videos. We compare our method (NL-TT) with four well-known methods: HaLRTC \cite{Liu2013tensor}, tSVD \cite{Zhang2017tSVD}, SiLRTC-TT, and TMac-TT \cite{Bengua2017Efficient}. The range of entry values for all test tensors are scaled into the interval $[0,255]$. In color videos tests, as tSVD is only applicable to third order tensors, we perform it on each frame separately. All numerical experiments are performed on Windows 10 64-bit and MATLAB R2012a running on a desktop equipped with an Intel(R) Core(TM) i7-6700M CPU with 3.40 GHz and 8 GB of RAM.

We use the peak signal-to-noise ratio (PSNR) and the structural similarity index (SSIM) to evaluate the quality of restored results. PSNR (dB)  and SSIM measure the similarity between the original tensor and the recovered tensor based on the distance and structural consistency, respectively. By calculating average PSNR and SSIM values for all bands, we obtain the PSNR and SSIM values of a higher-order tensor. Higher PSNR and SSIM values indicate better image quality.

\begin{figure*}[!ht]
\scriptsize\setlength{\tabcolsep}{0.9pt}
\begin{center}
\begin{tabular}{cccccccc}
\includegraphics[width=0.14\textwidth]{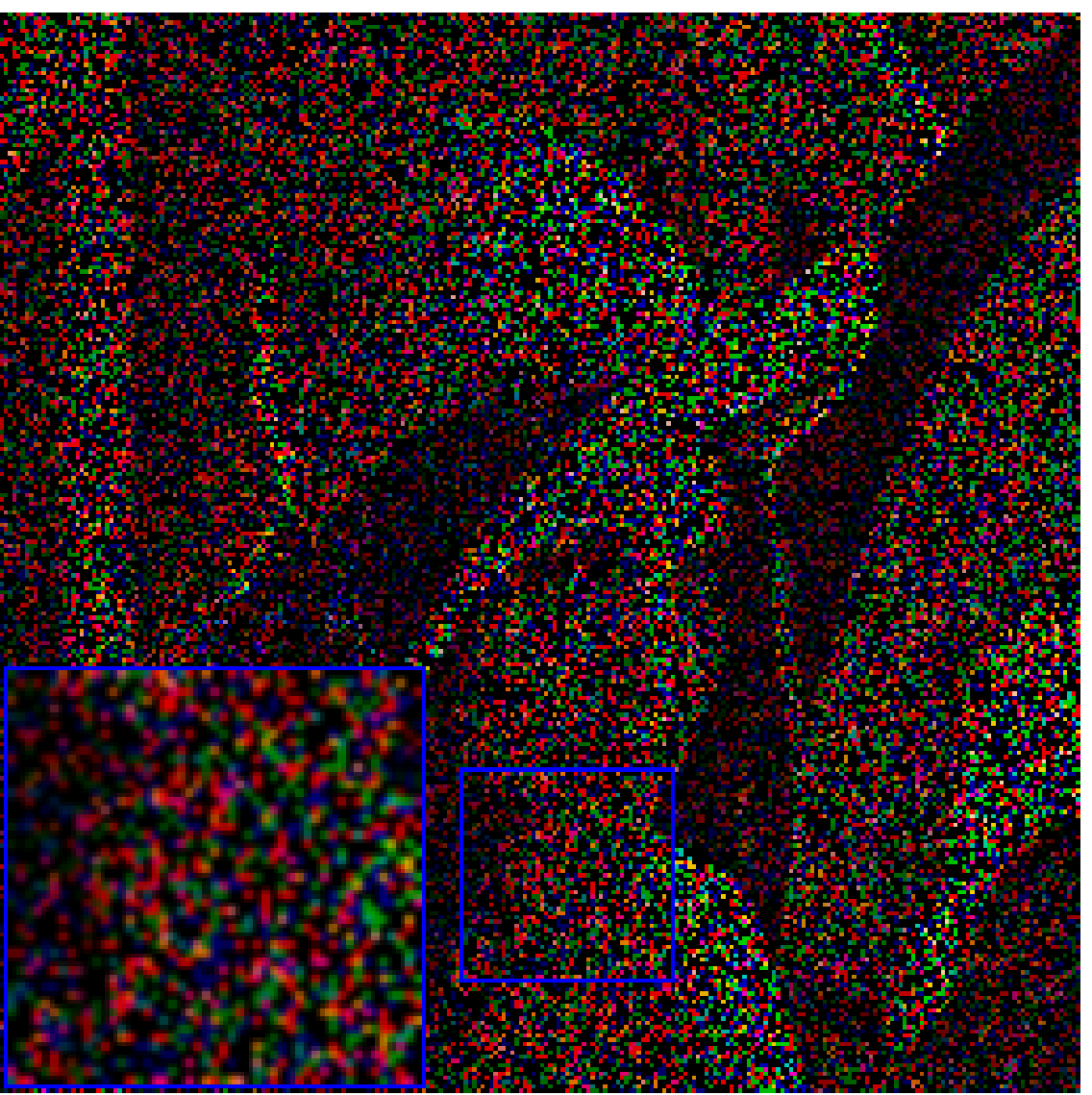}&
\includegraphics[width=0.14\textwidth]{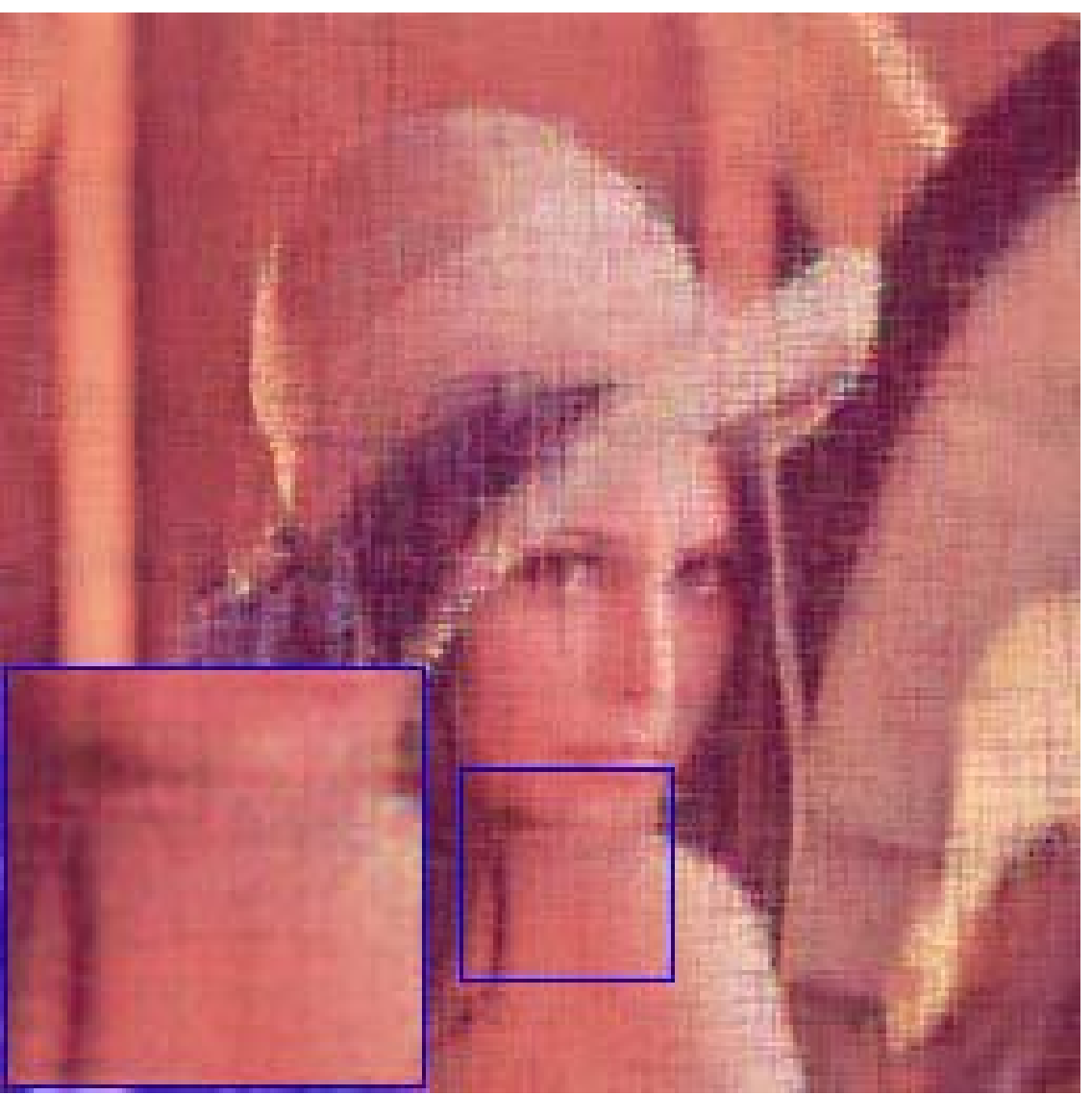}&
\includegraphics[width=0.14\textwidth]{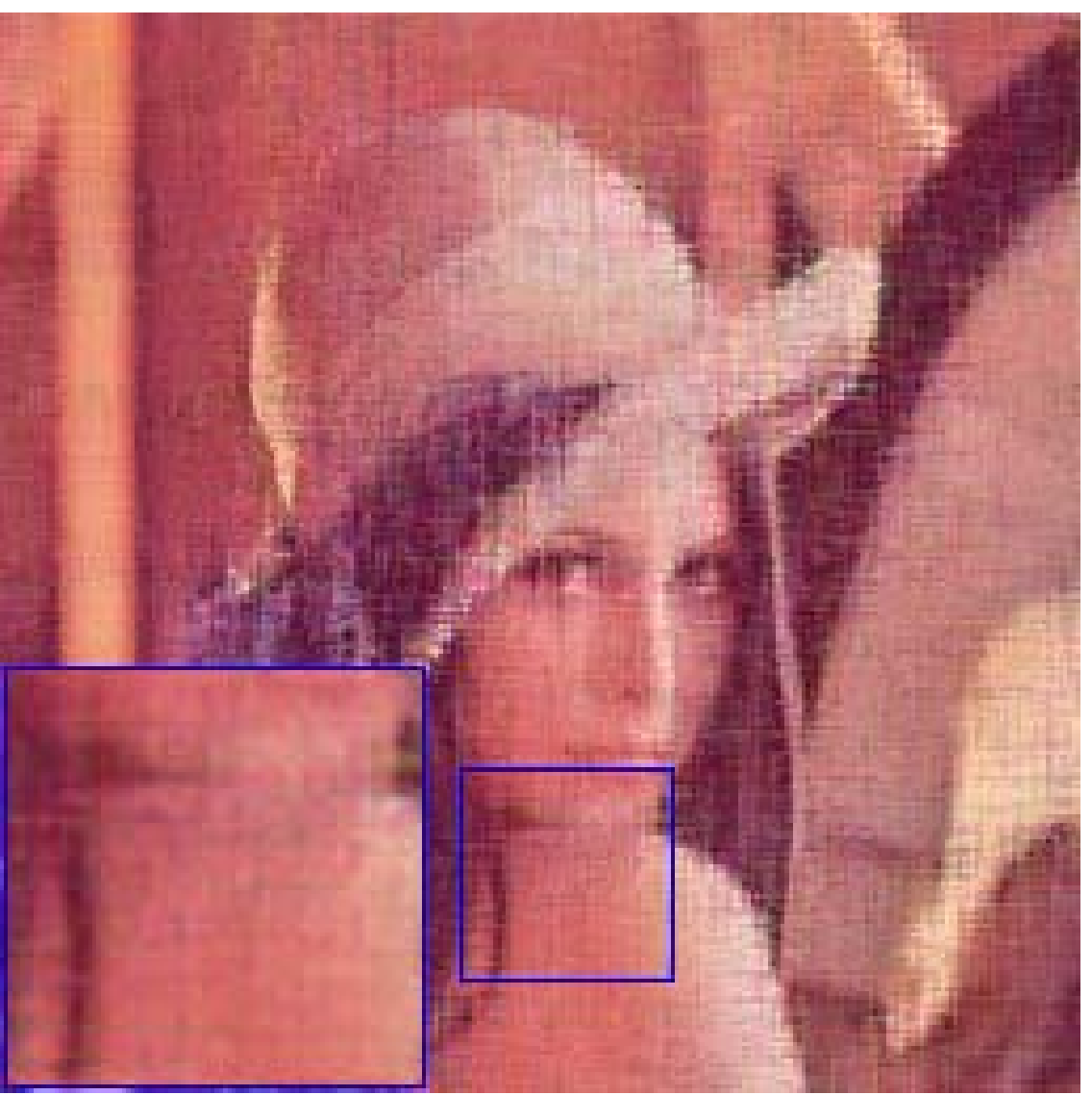}&
\includegraphics[width=0.14\textwidth]{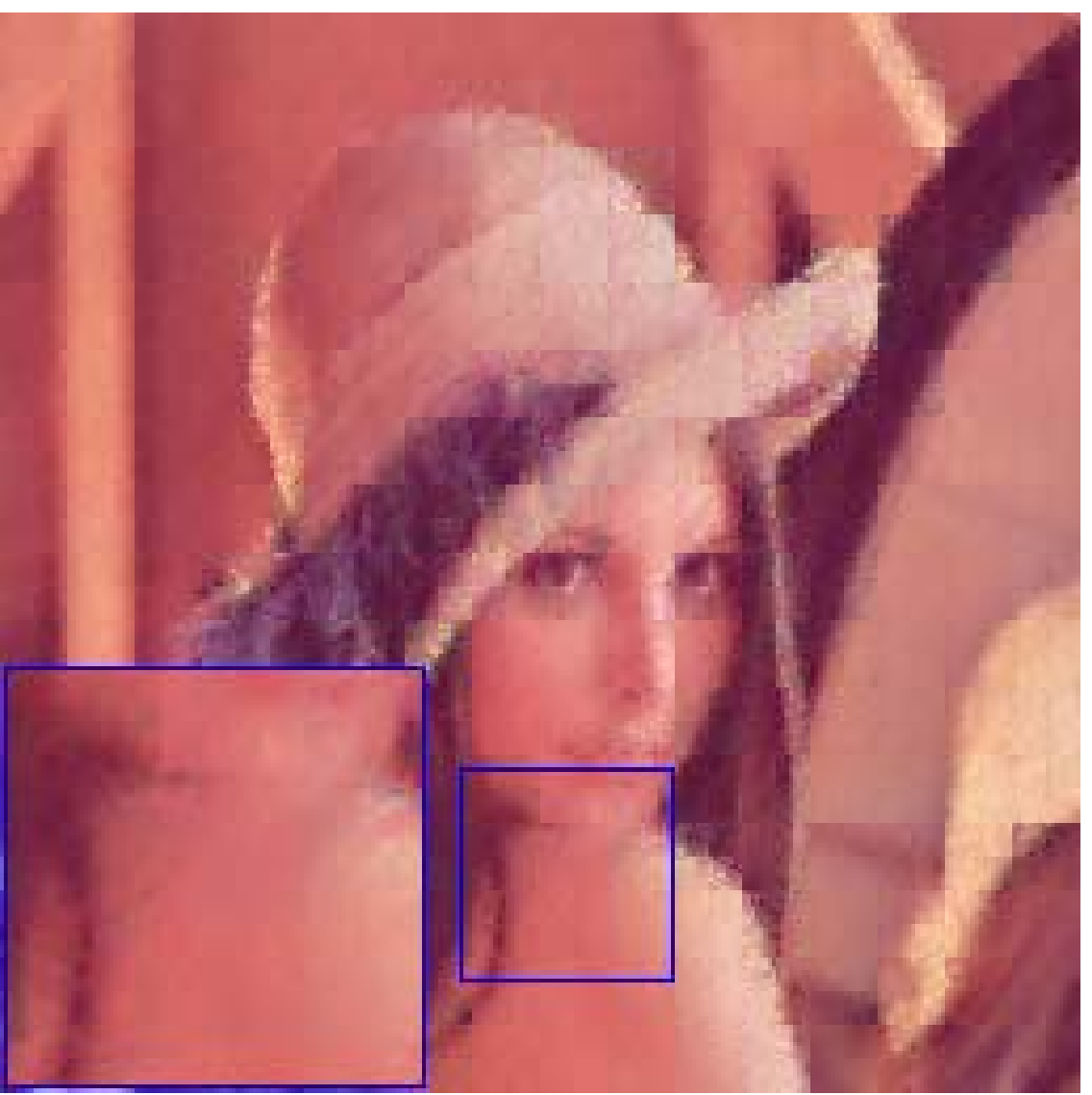}&
\includegraphics[width=0.14\textwidth]{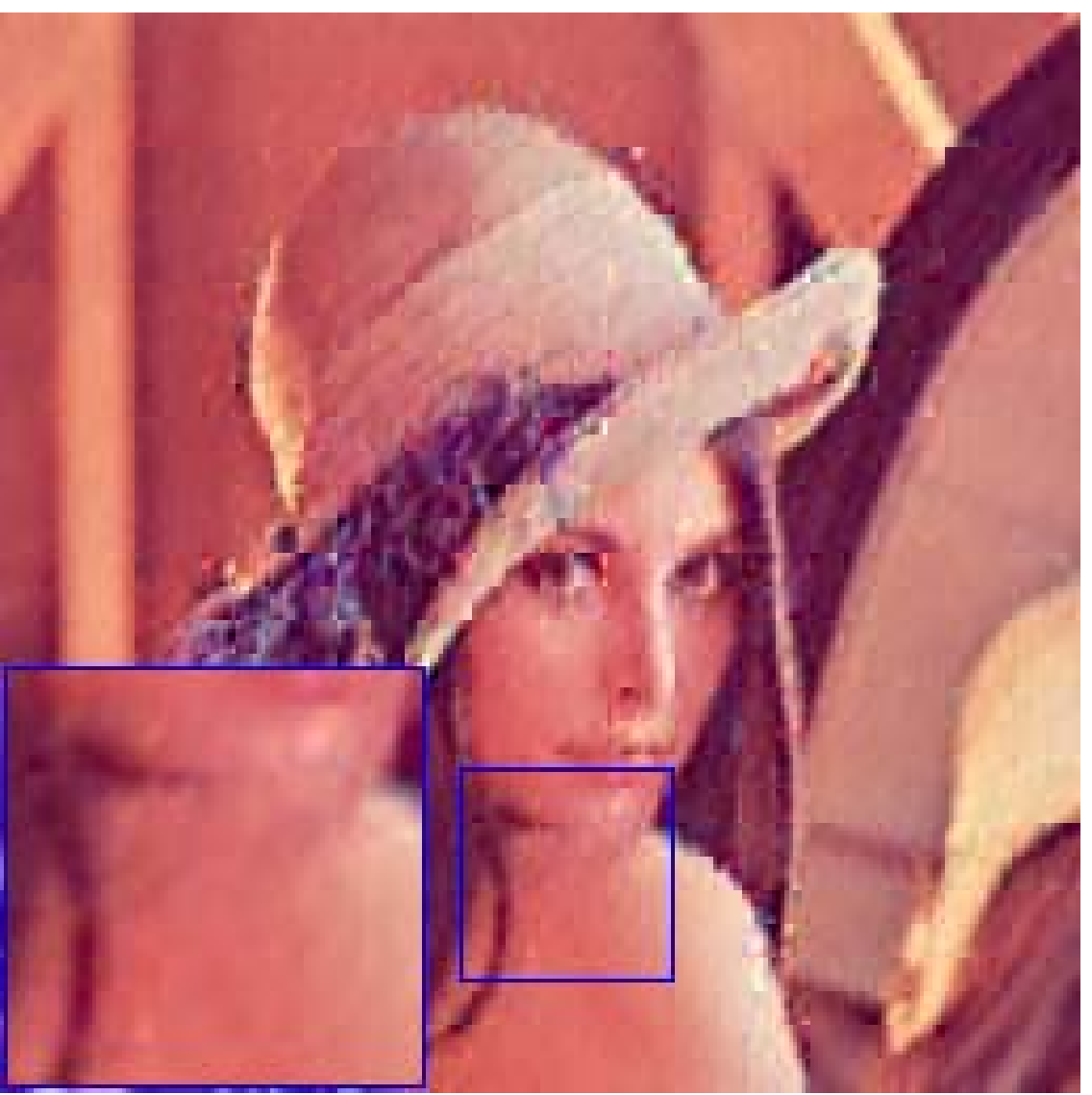}&
\includegraphics[width=0.14\textwidth]{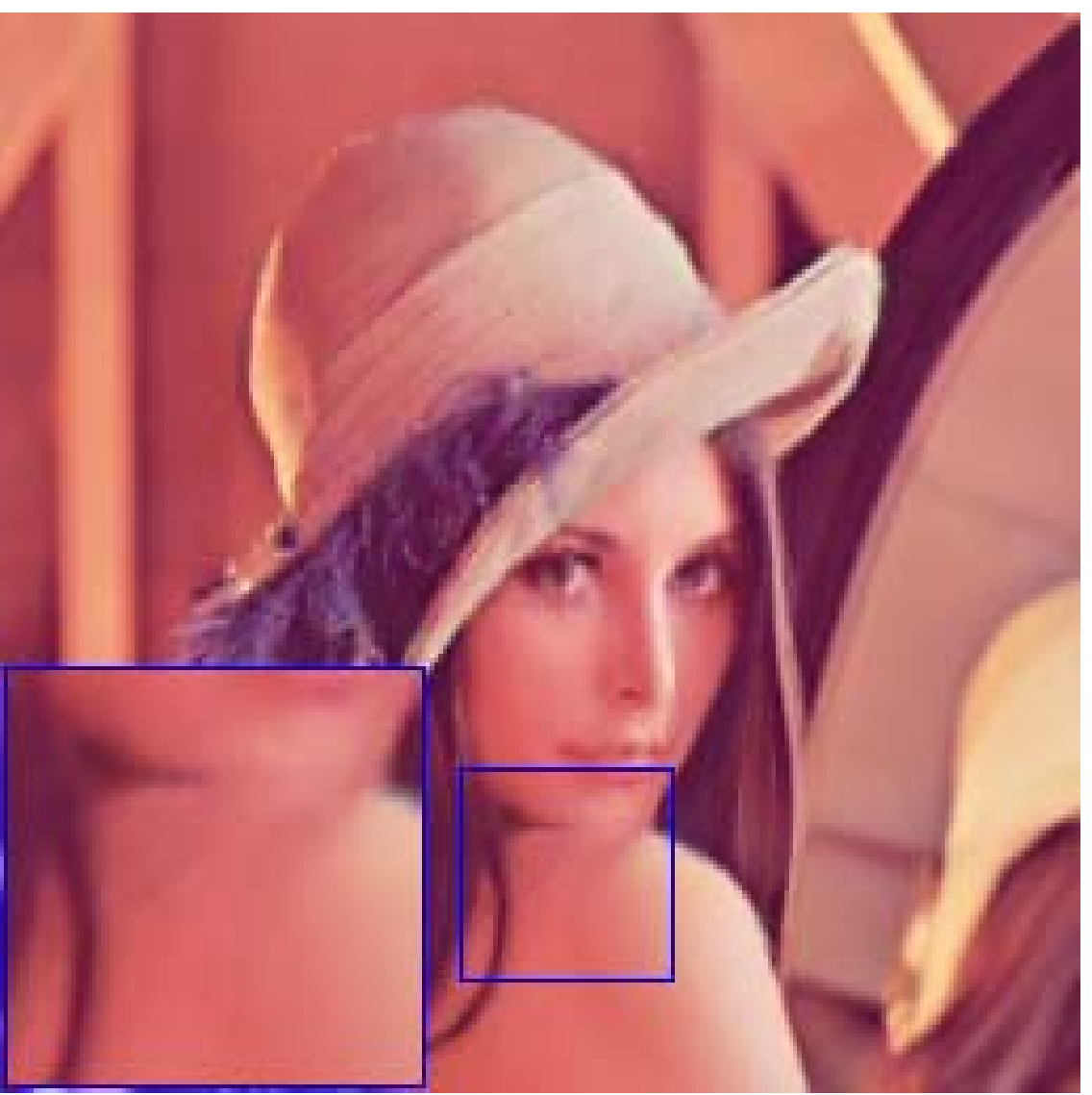}&
\includegraphics[width=0.14\textwidth]{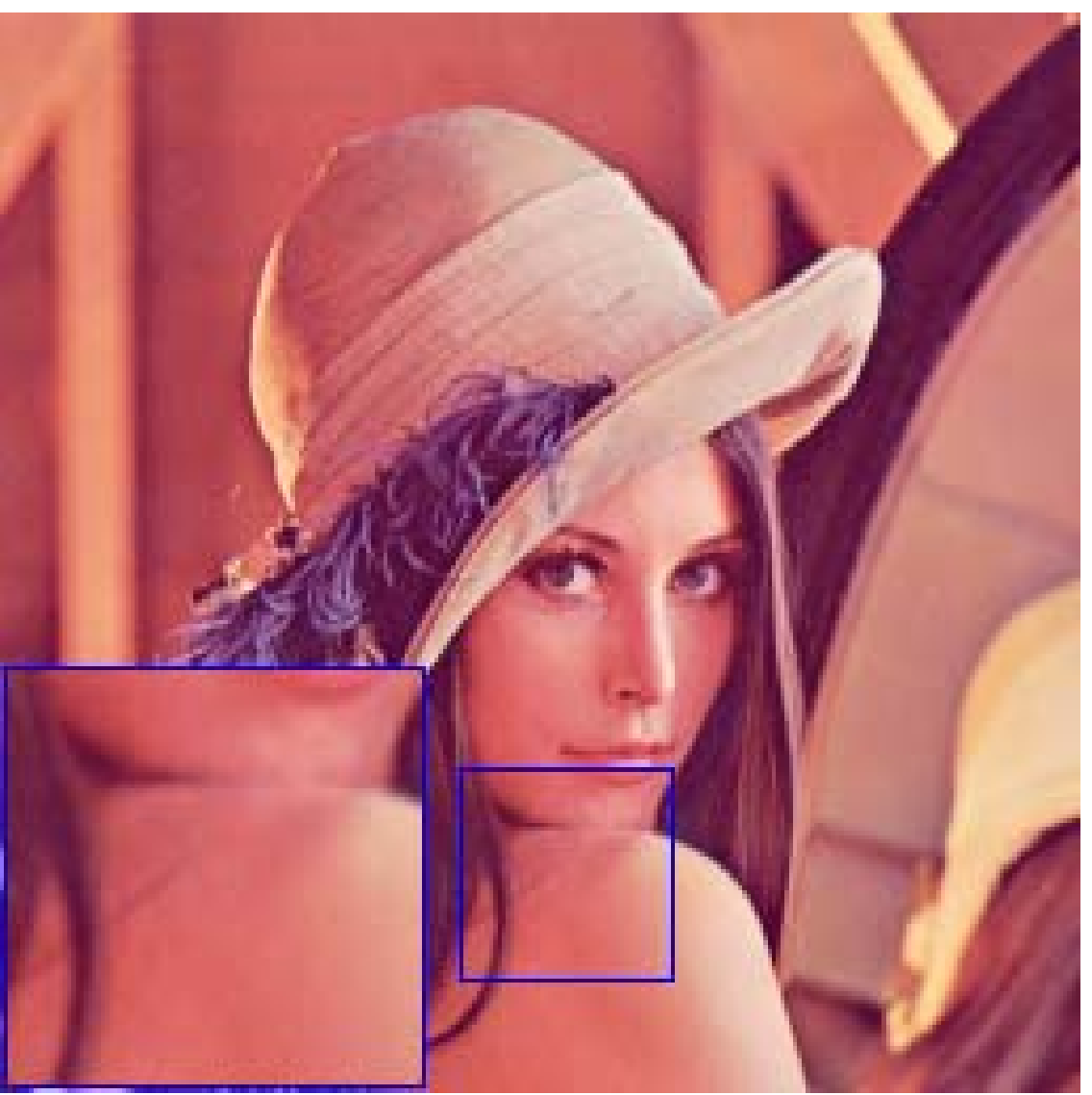}\\
\includegraphics[width=0.14\textwidth]{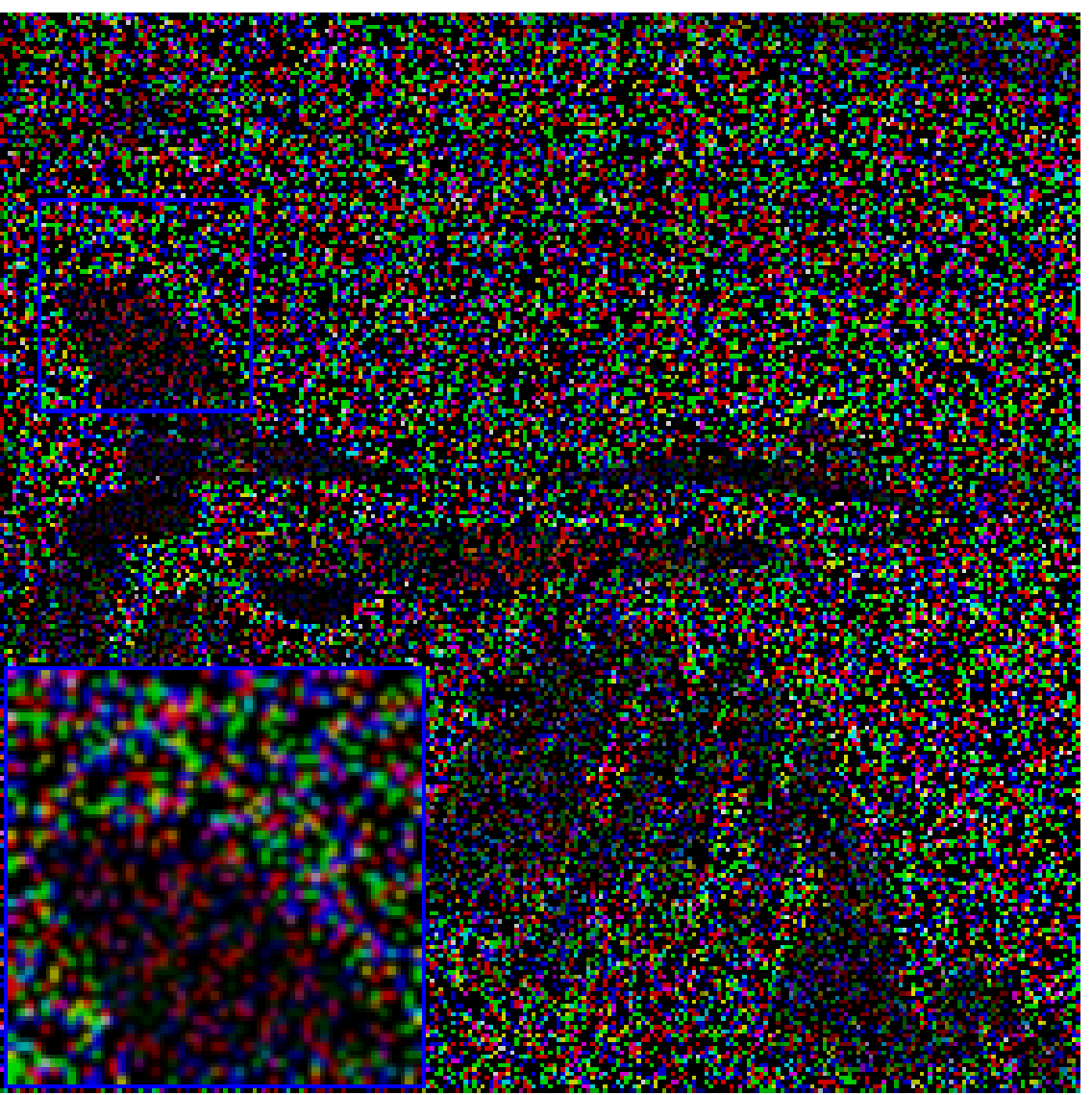}&
\includegraphics[width=0.14\textwidth]{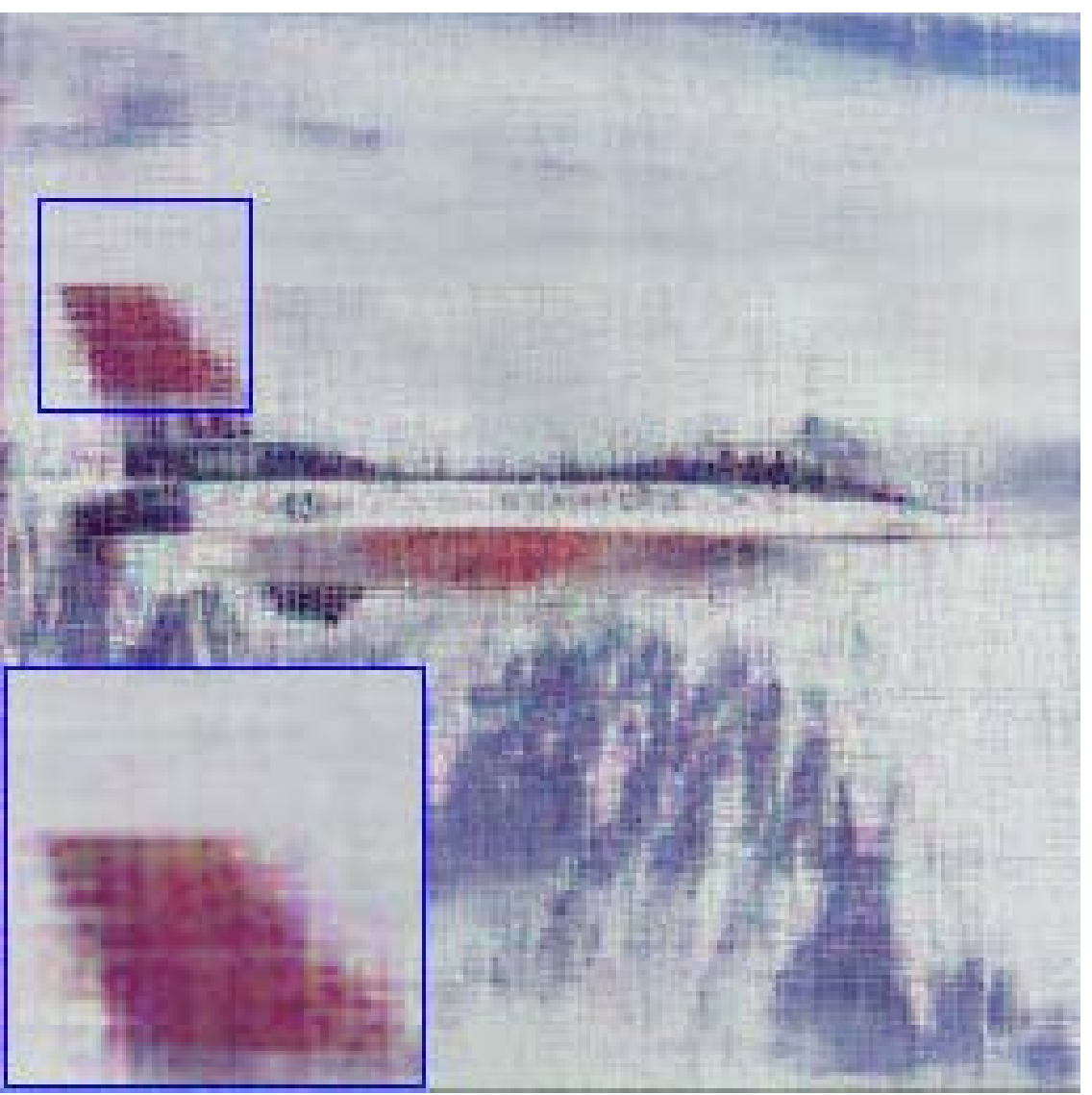}&
\includegraphics[width=0.14\textwidth]{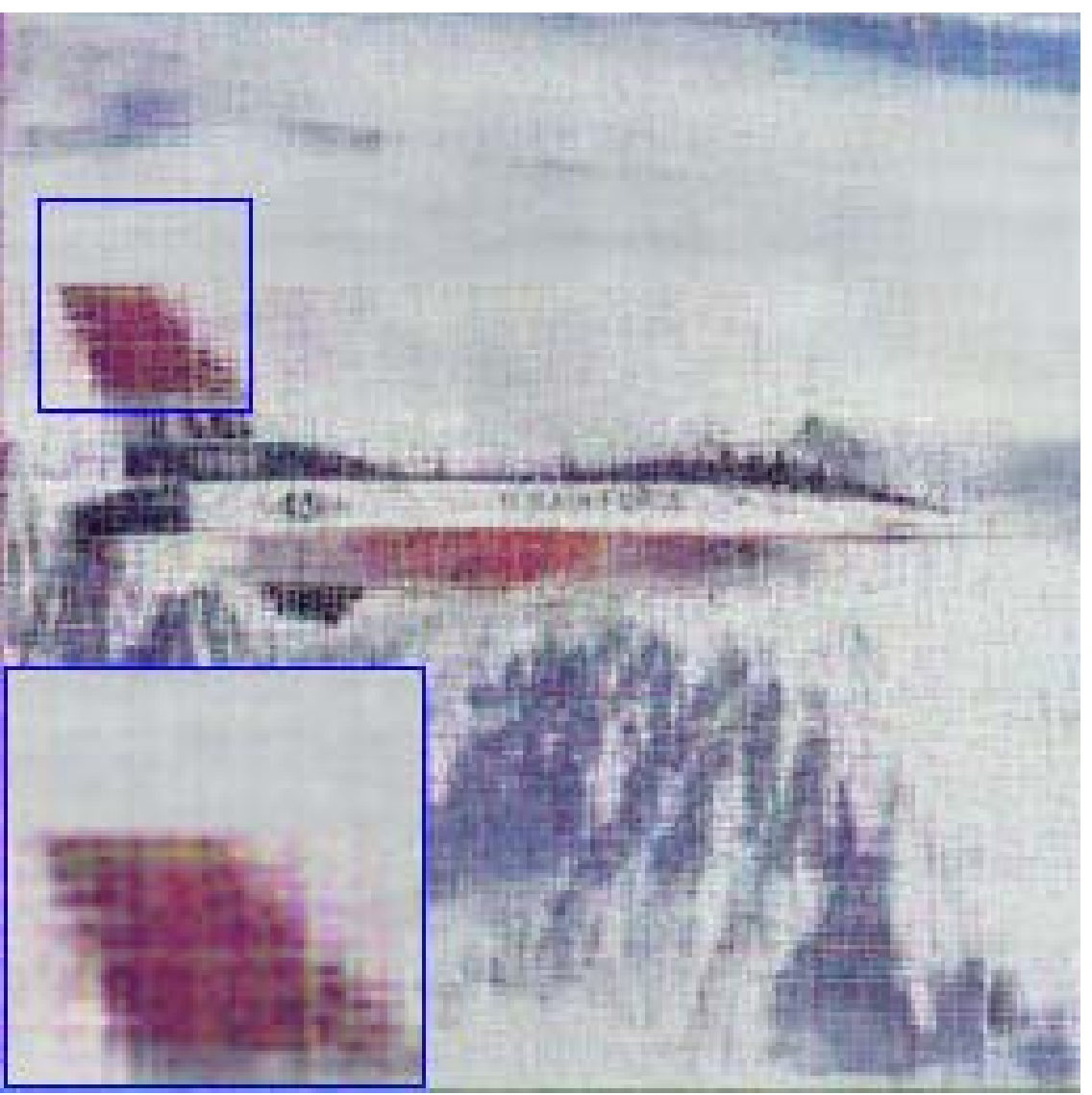}&
\includegraphics[width=0.14\textwidth]{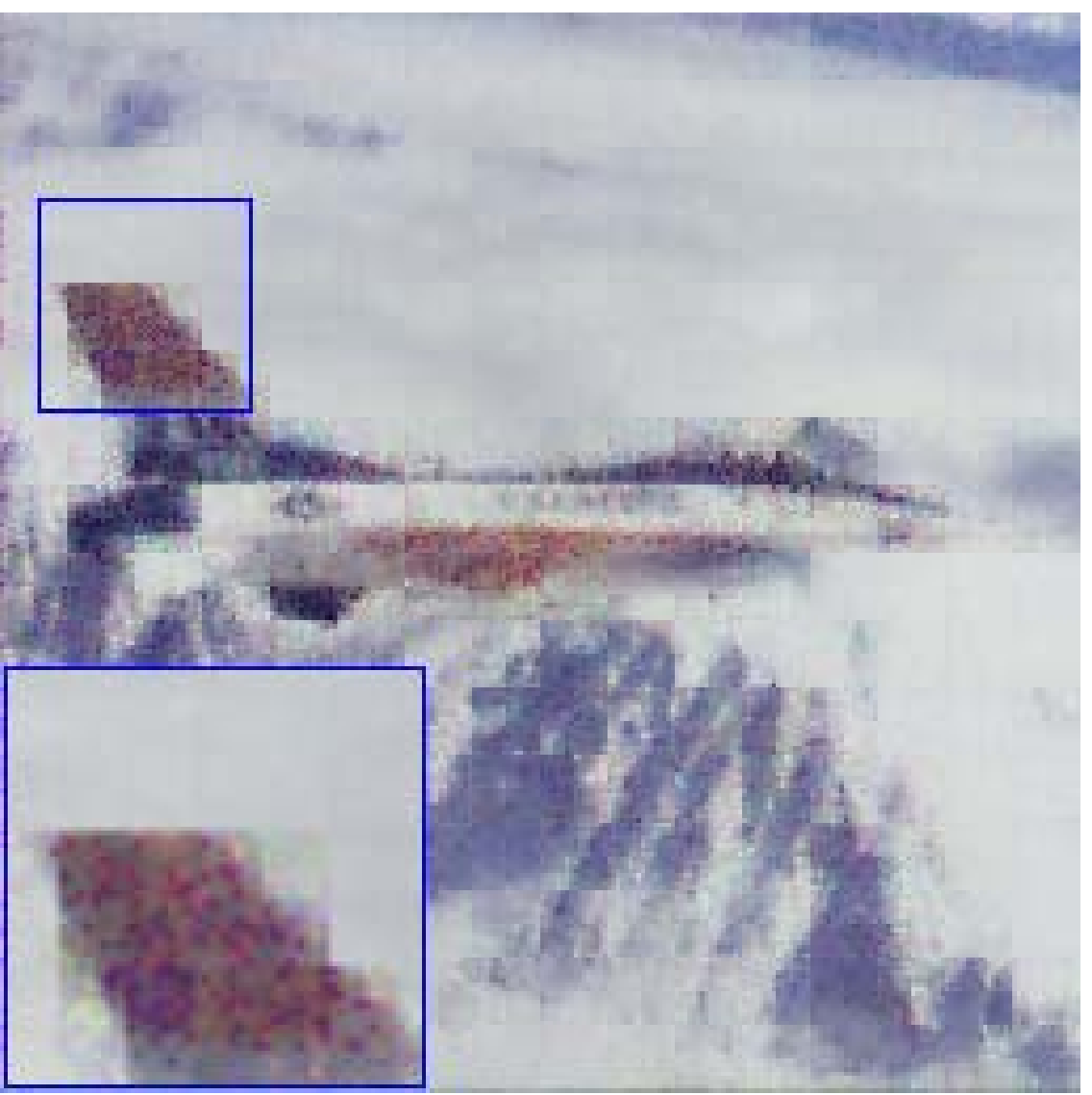}&
\includegraphics[width=0.14\textwidth]{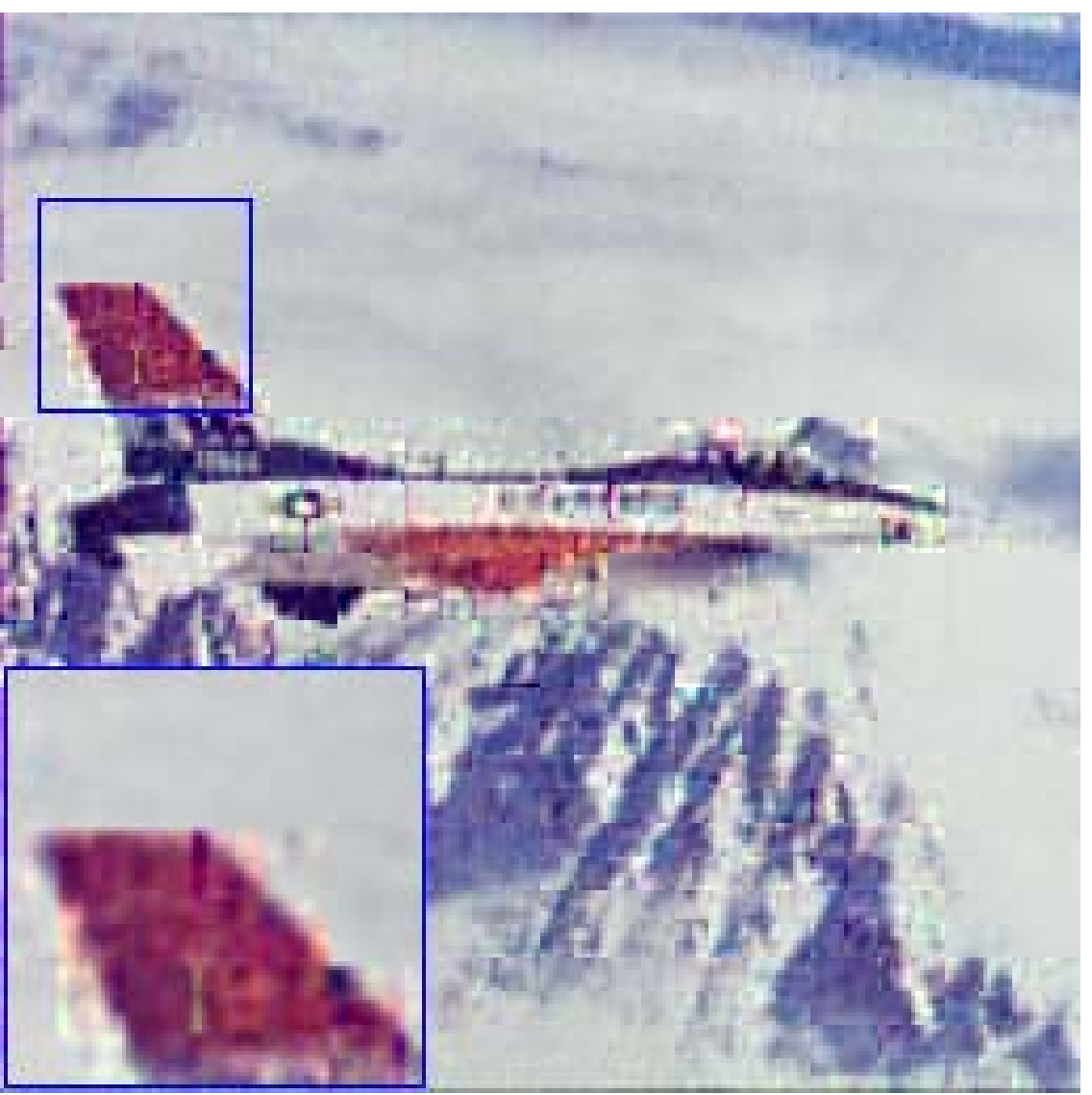}&
\includegraphics[width=0.14\textwidth]{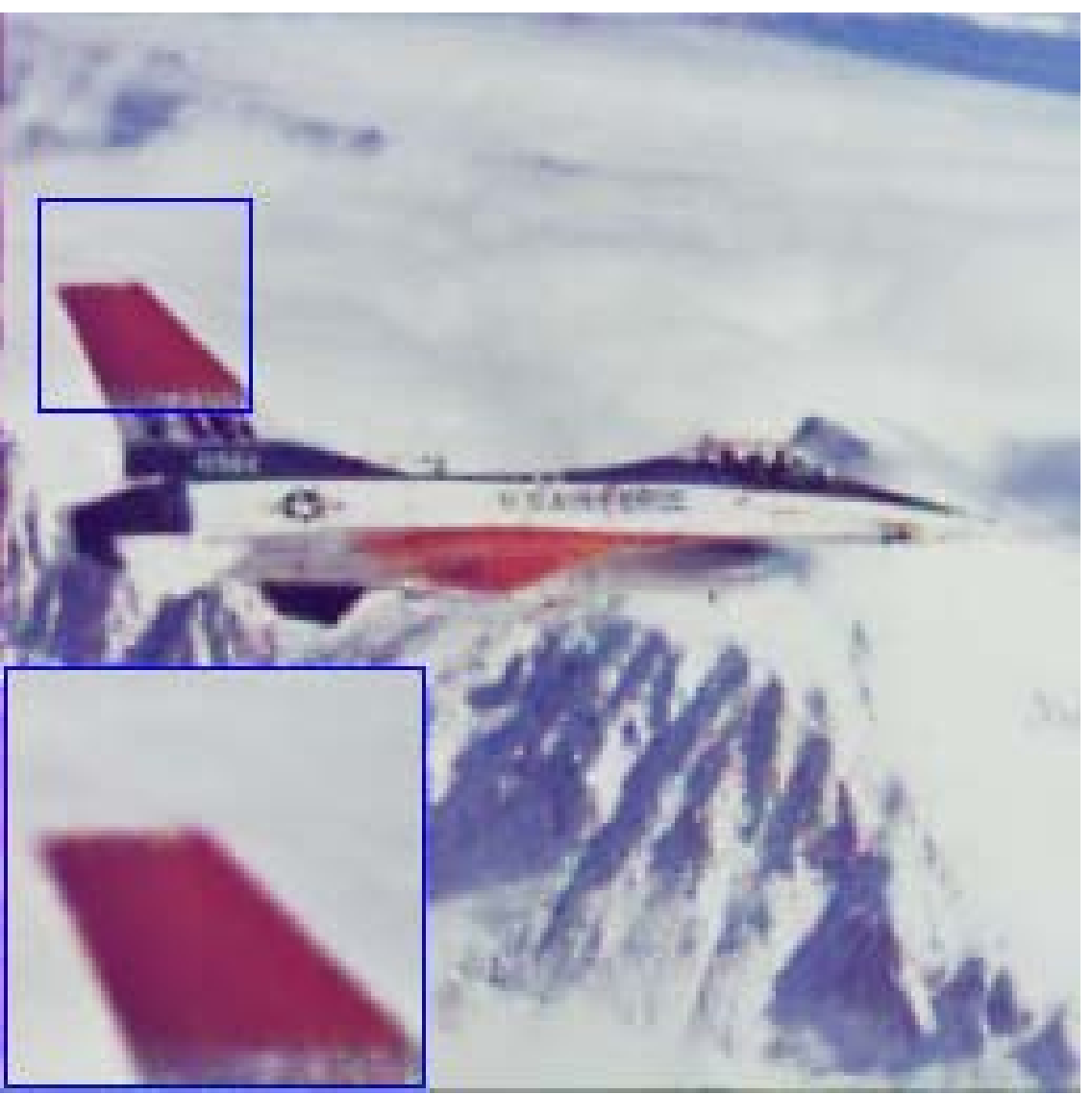}&
\includegraphics[width=0.14\textwidth]{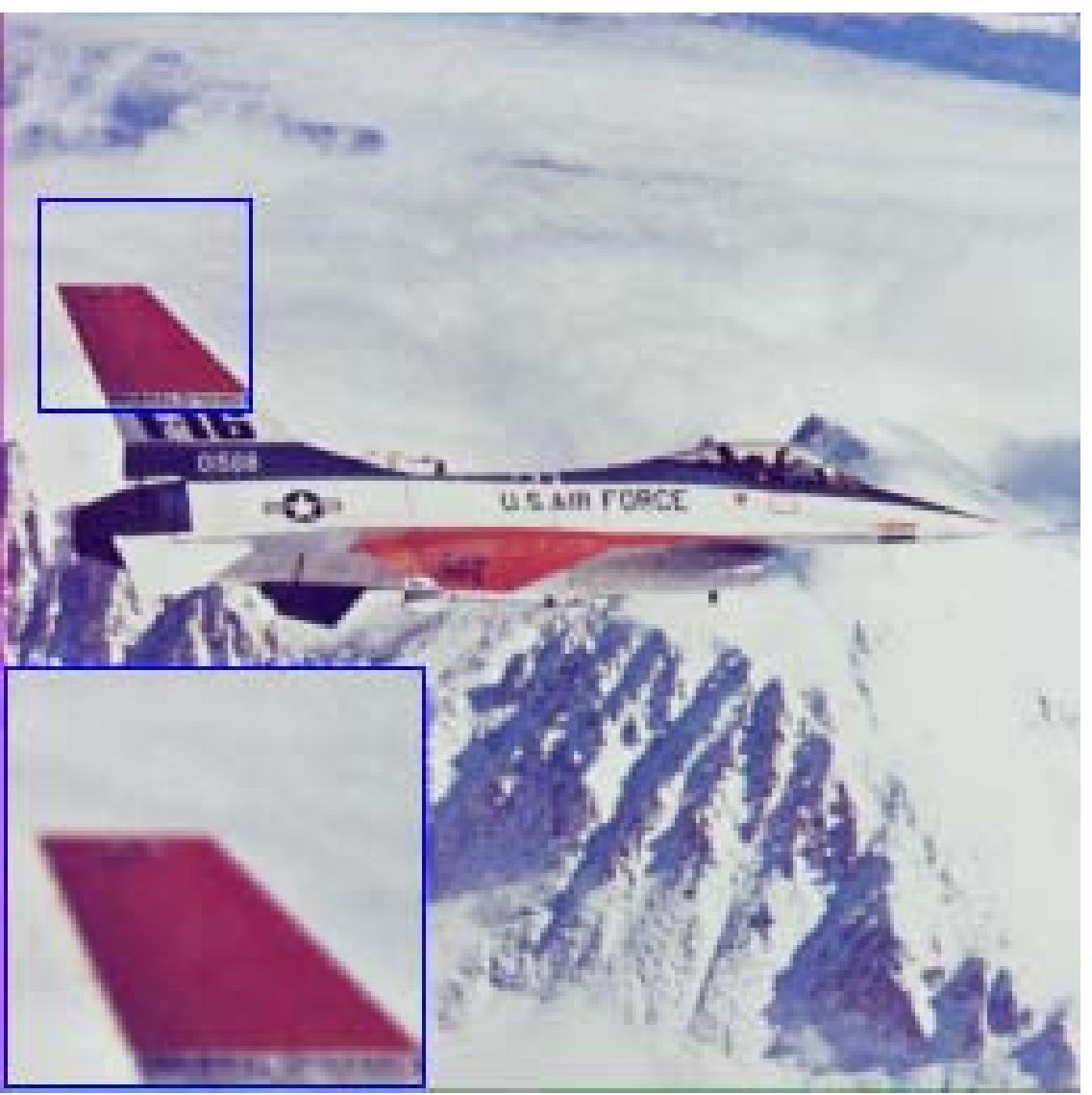}\\
\includegraphics[width=0.14\textwidth]{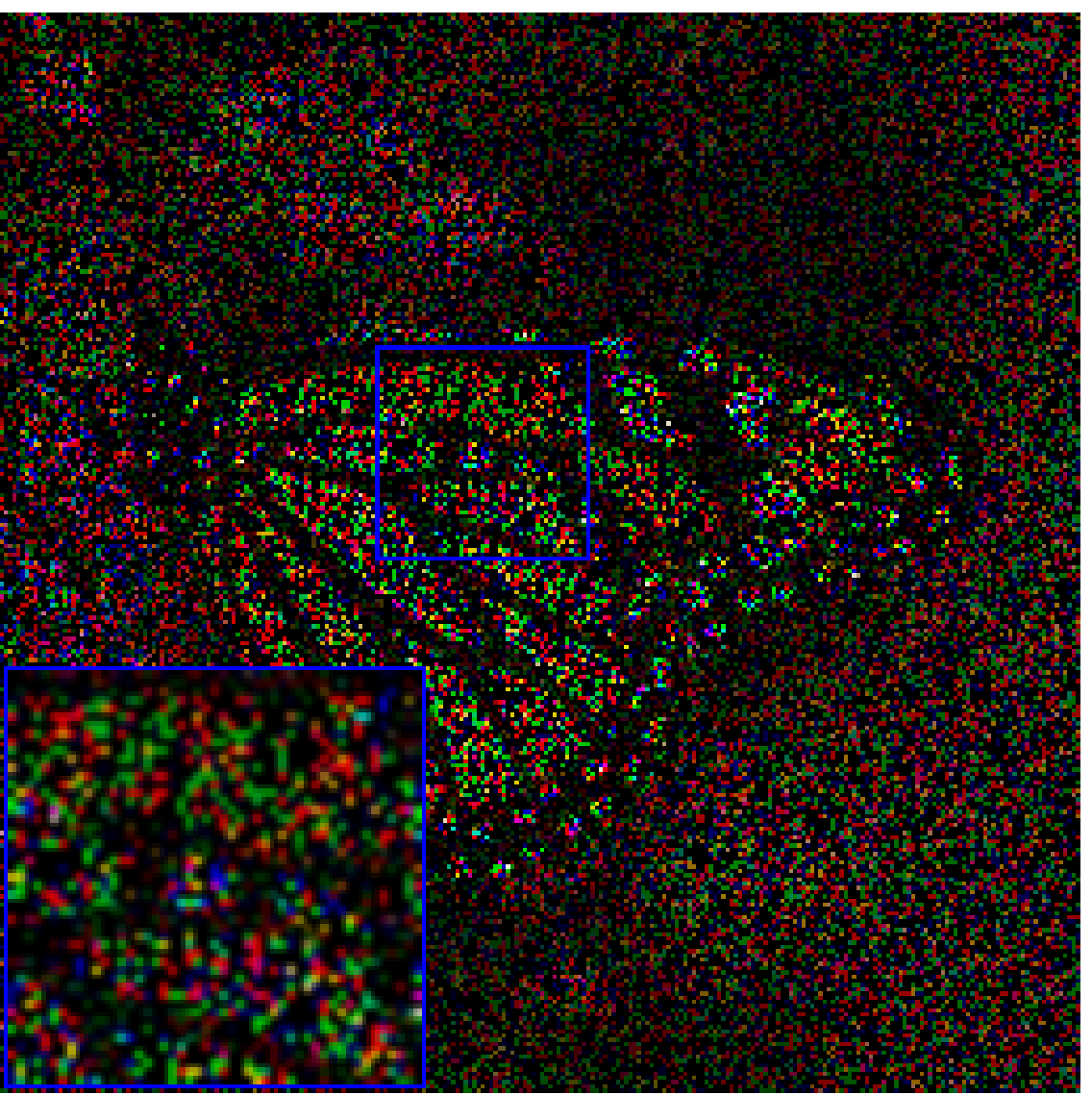}&
\includegraphics[width=0.14\textwidth]{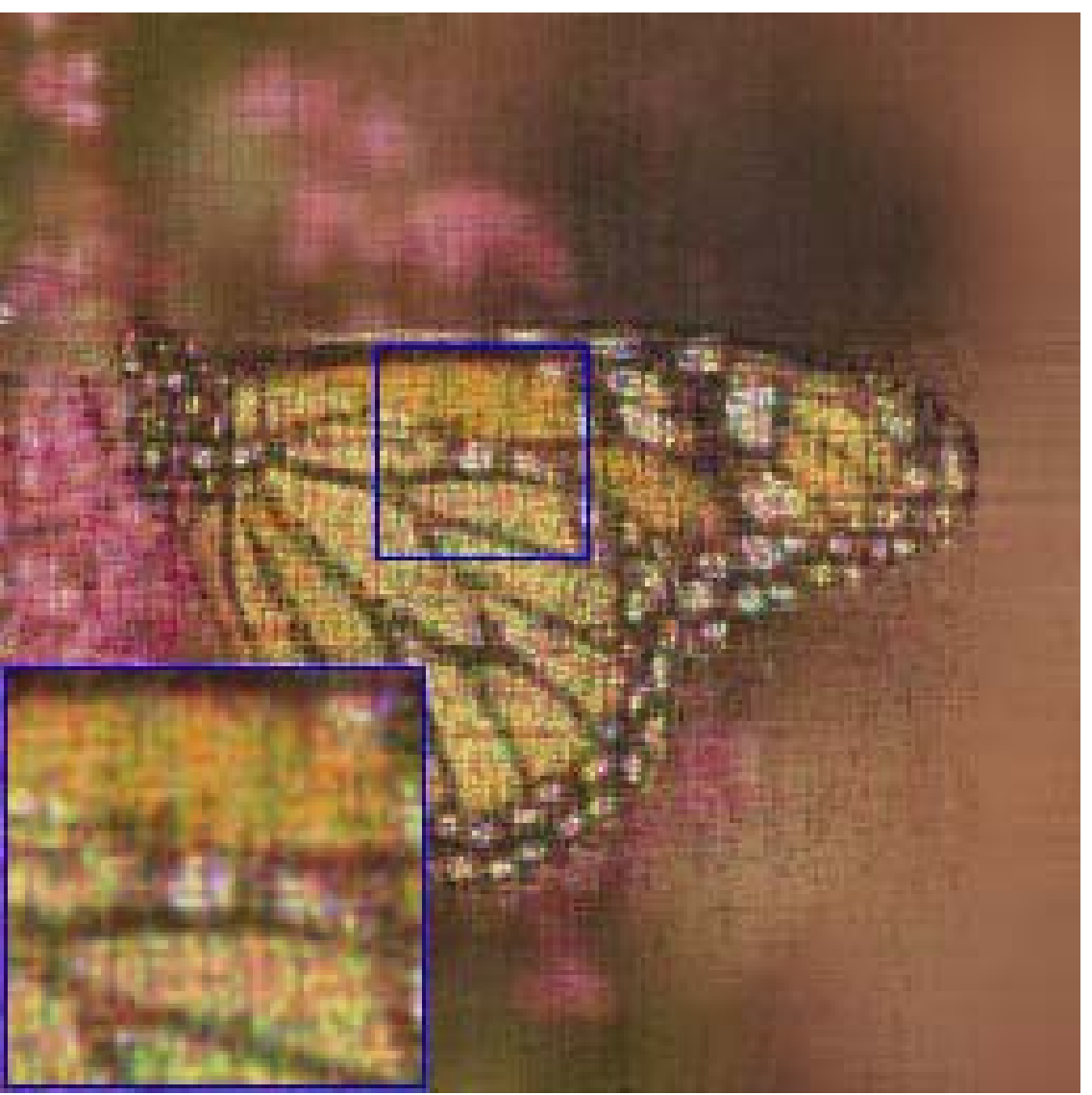}&
\includegraphics[width=0.14\textwidth]{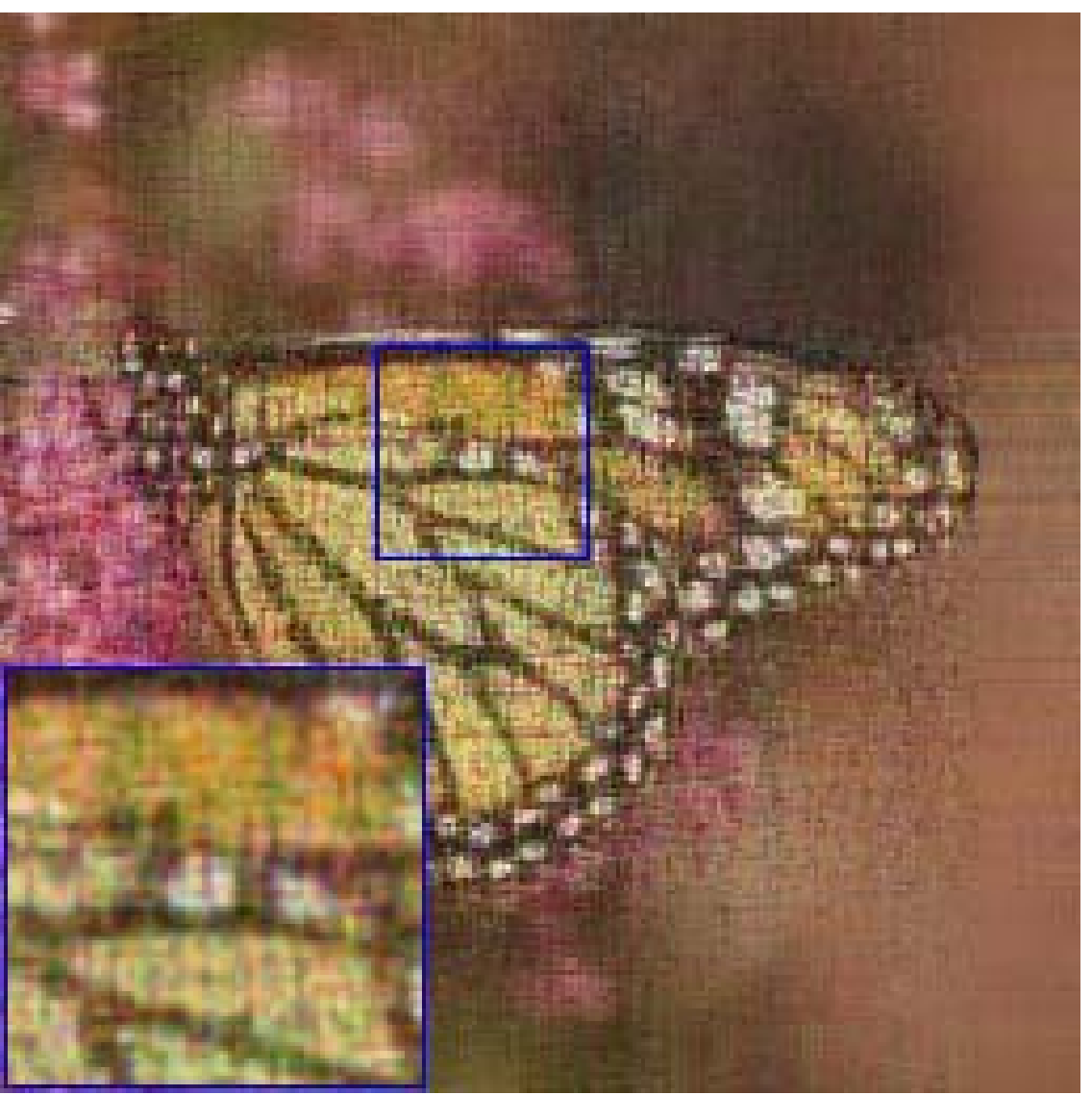}&
\includegraphics[width=0.14\textwidth]{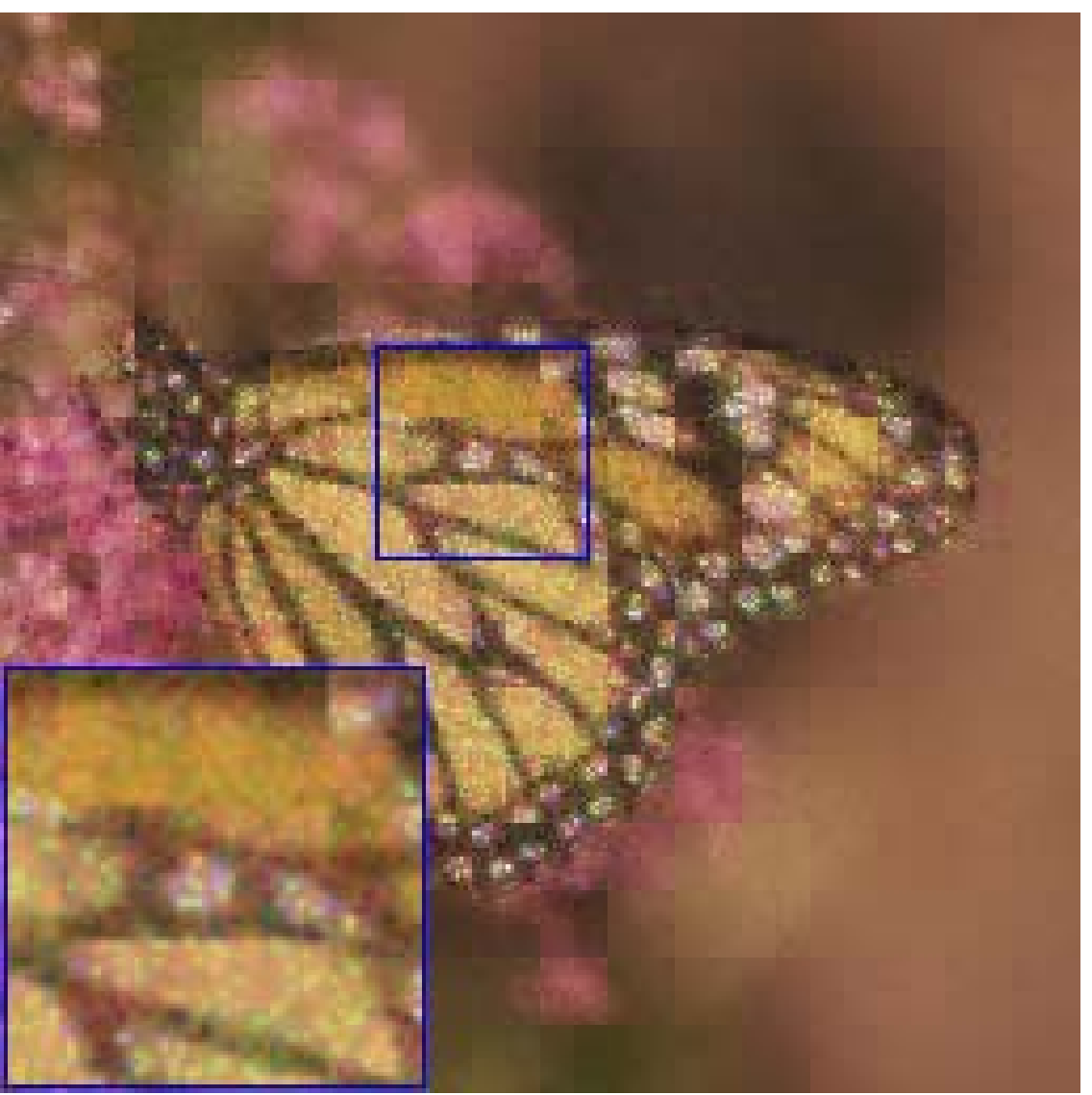}&
\includegraphics[width=0.14\textwidth]{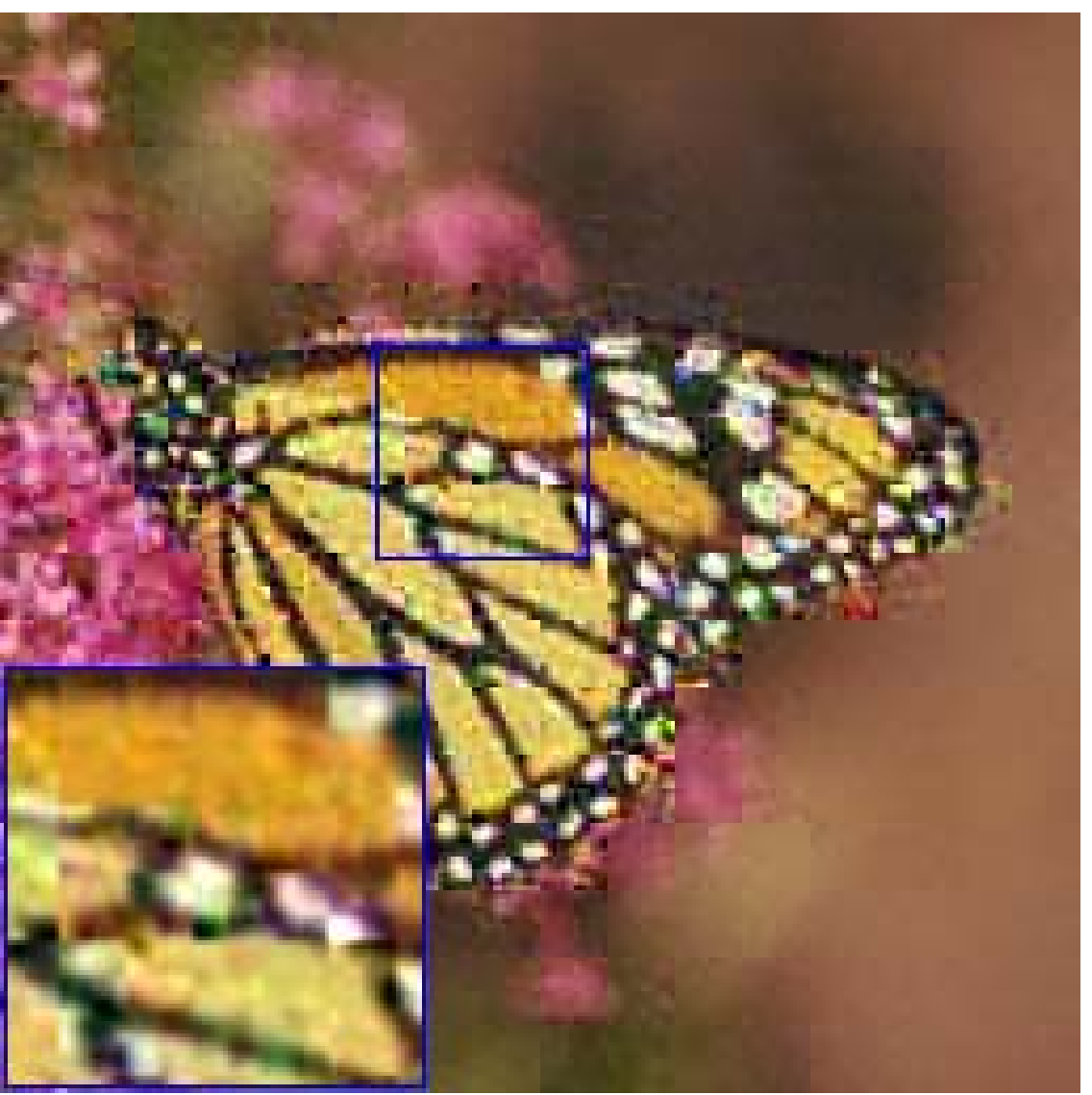}&
\includegraphics[width=0.14\textwidth]{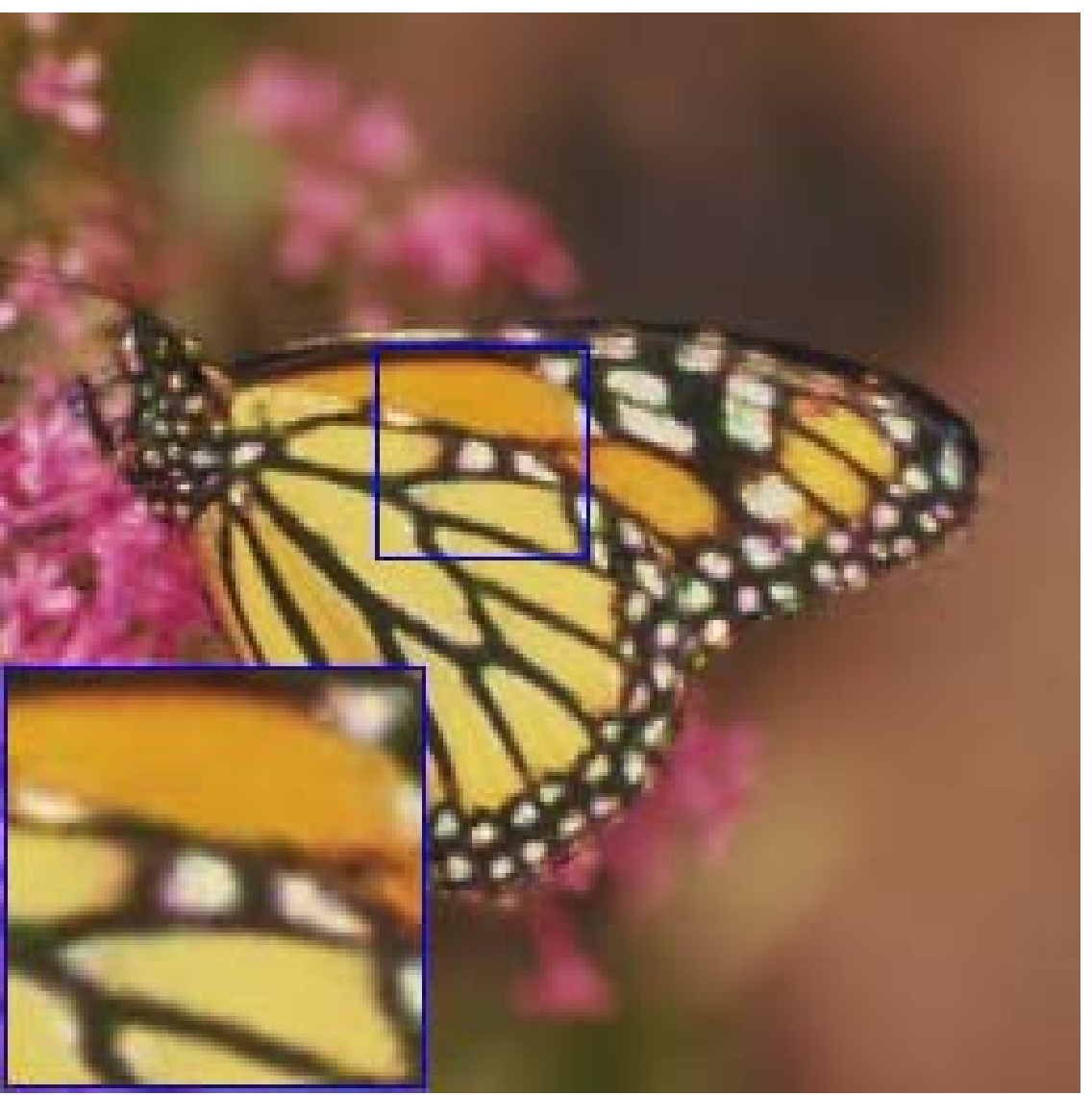}&
\includegraphics[width=0.14\textwidth]{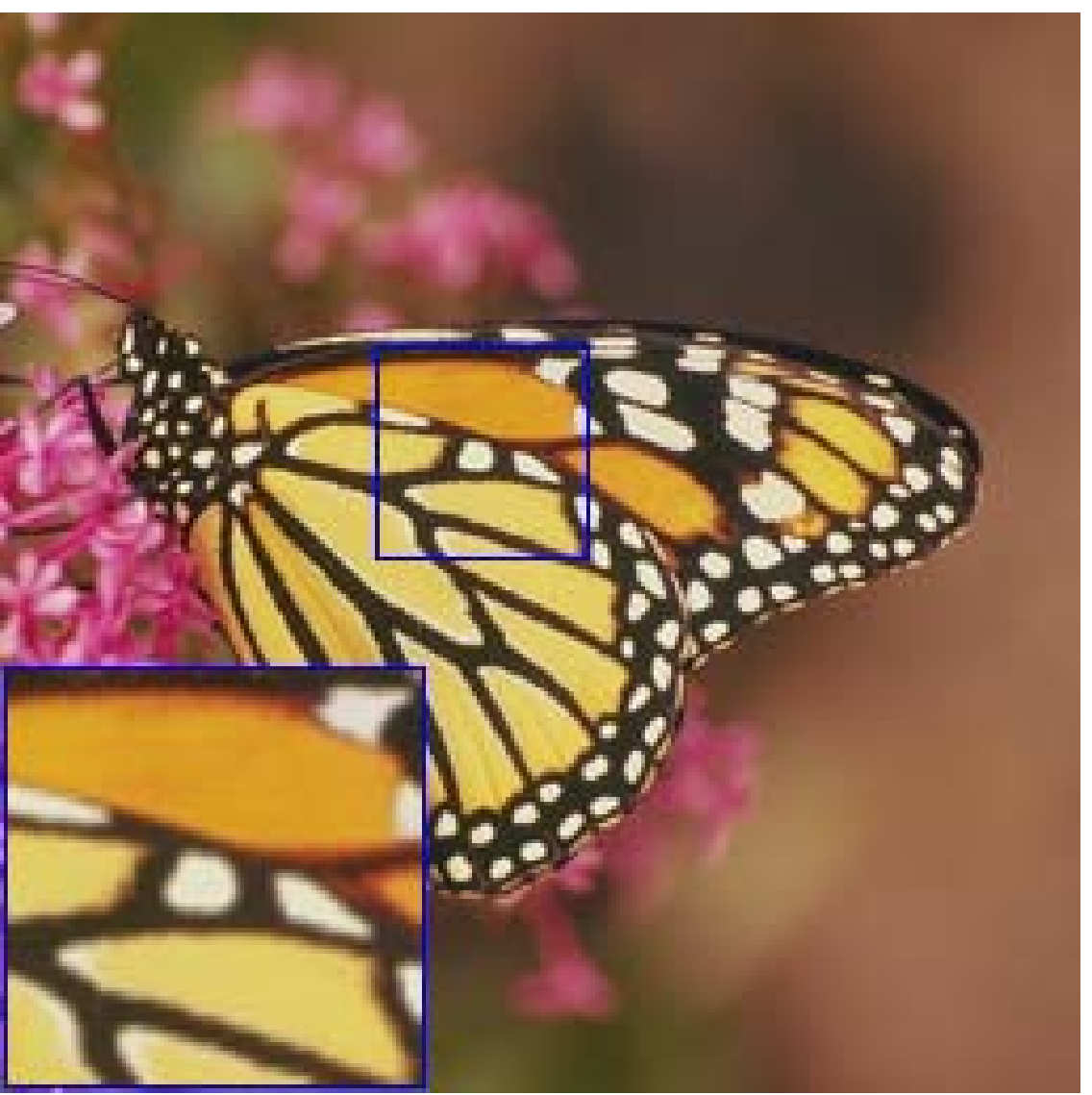}\\
\includegraphics[width=0.14\textwidth]{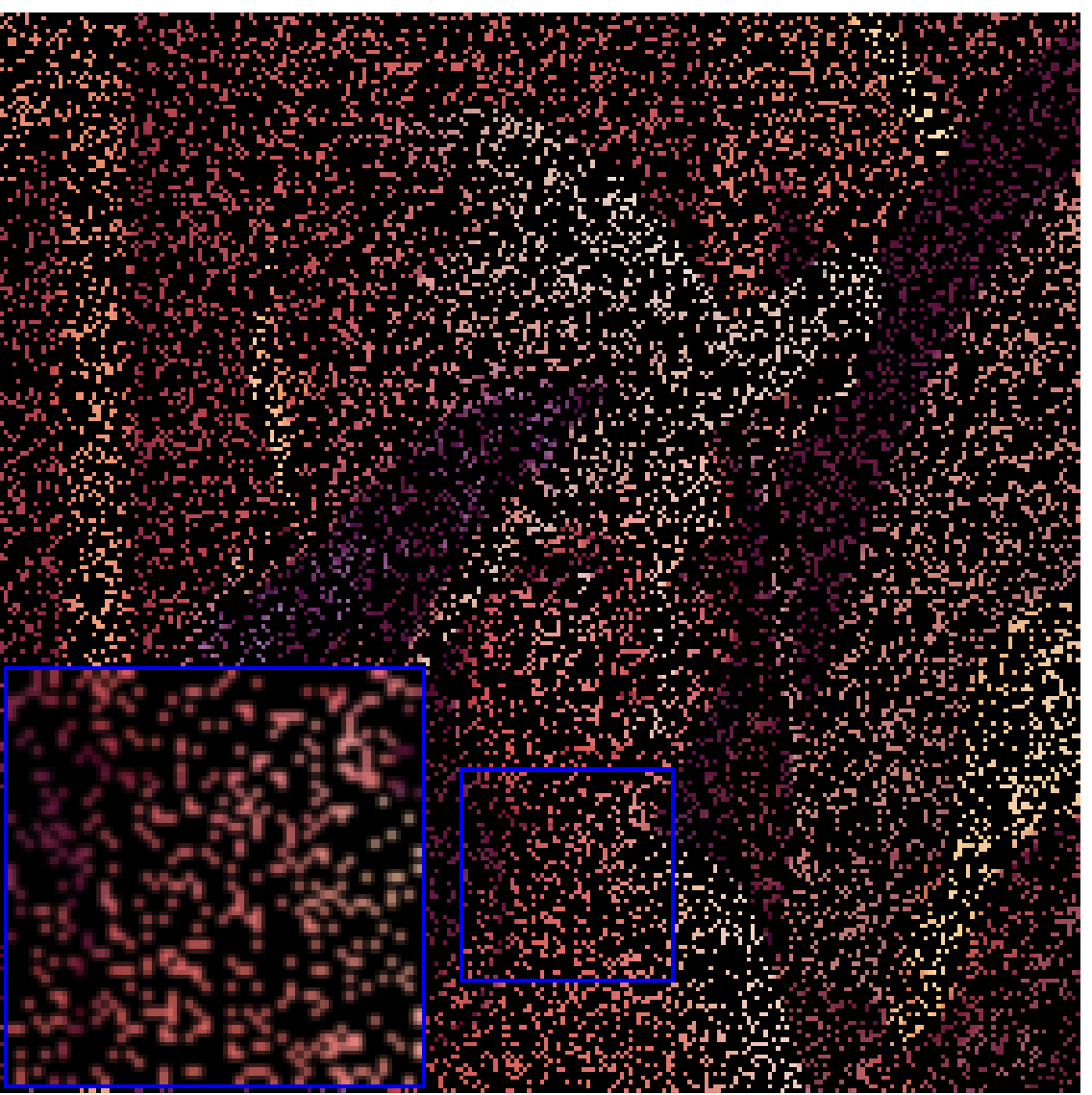}&
\includegraphics[width=0.14\textwidth]{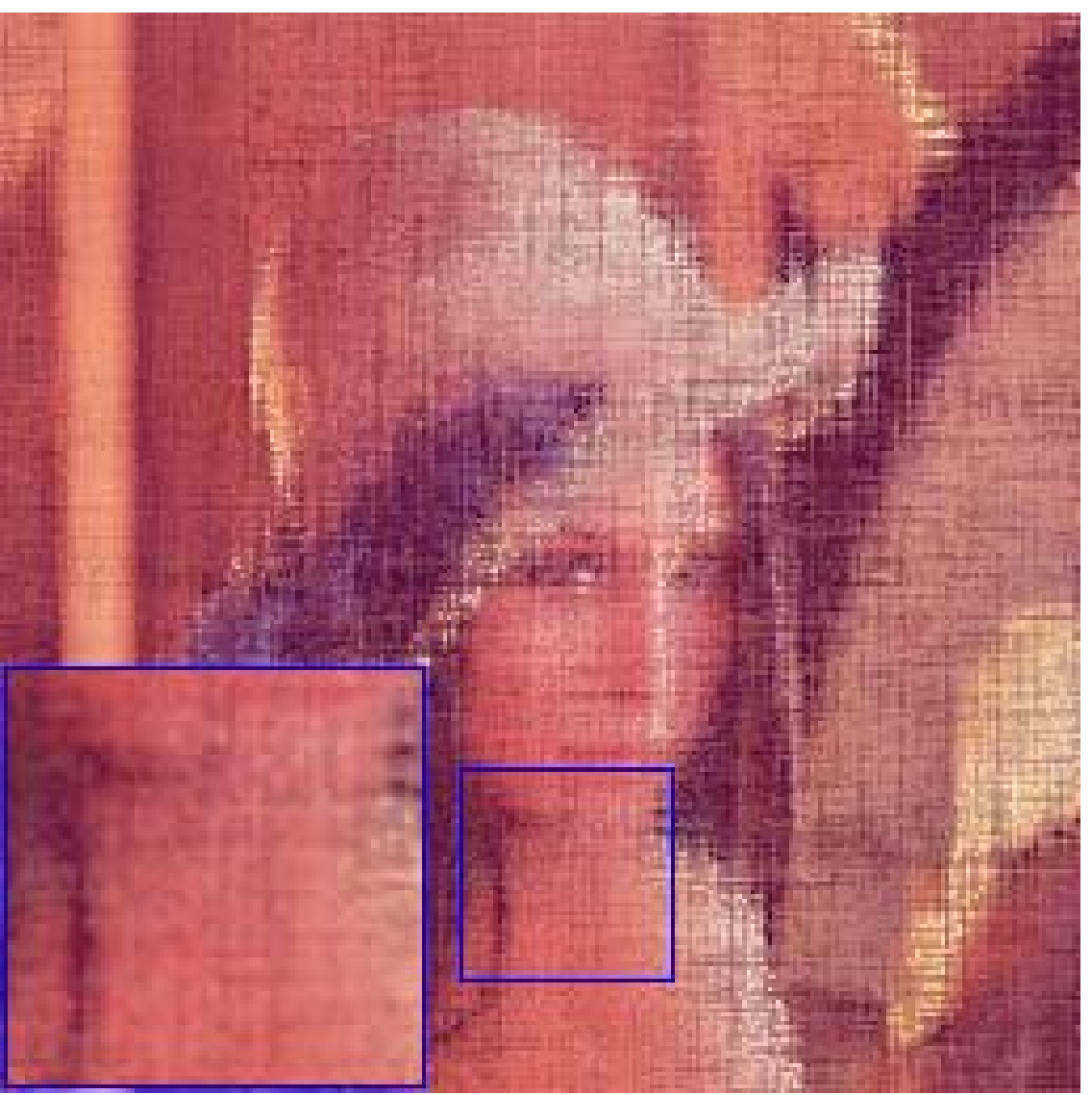}&
\includegraphics[width=0.14\textwidth]{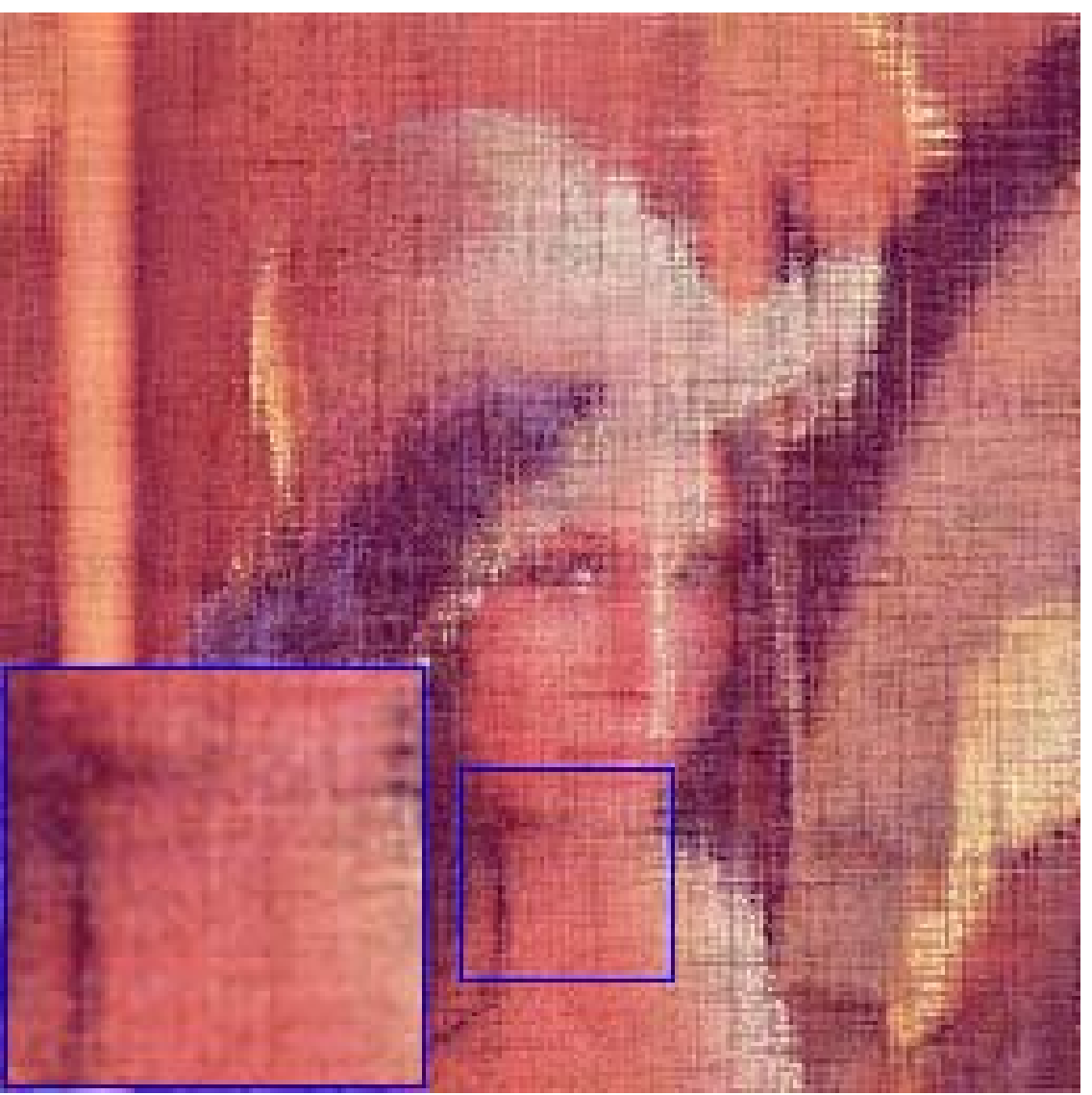}&
\includegraphics[width=0.14\textwidth]{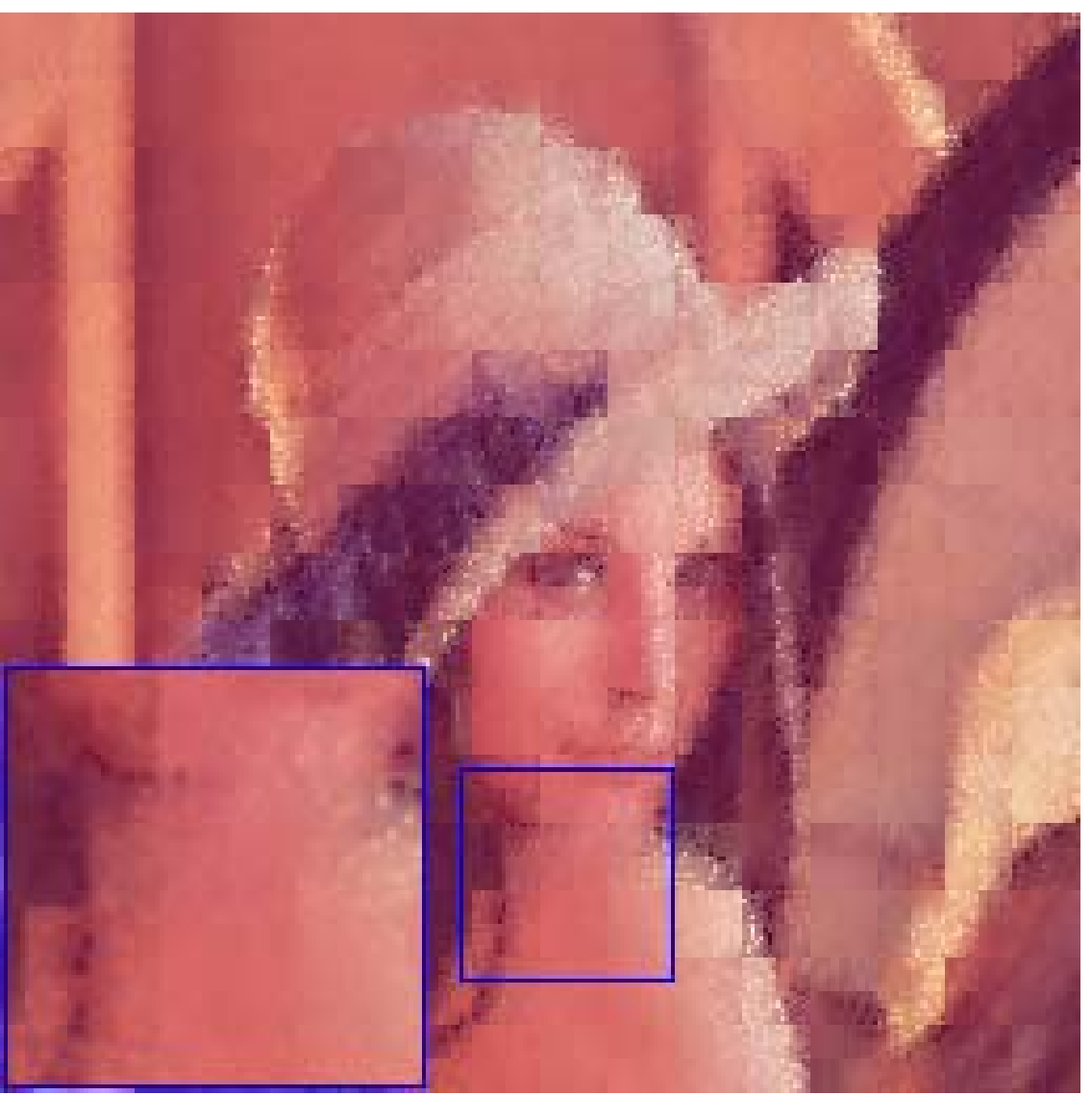}&
\includegraphics[width=0.14\textwidth]{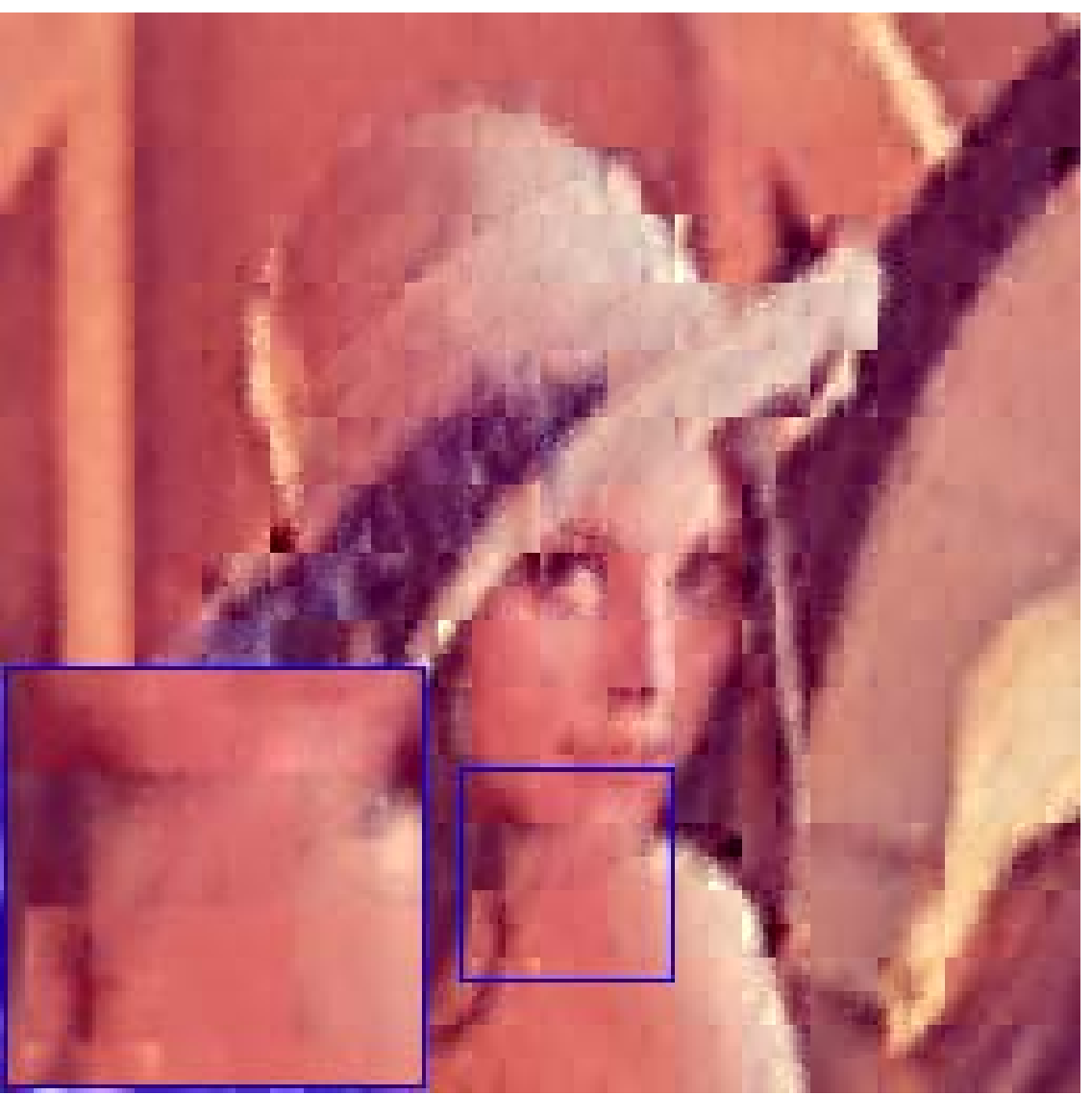}&
\includegraphics[width=0.14\textwidth]{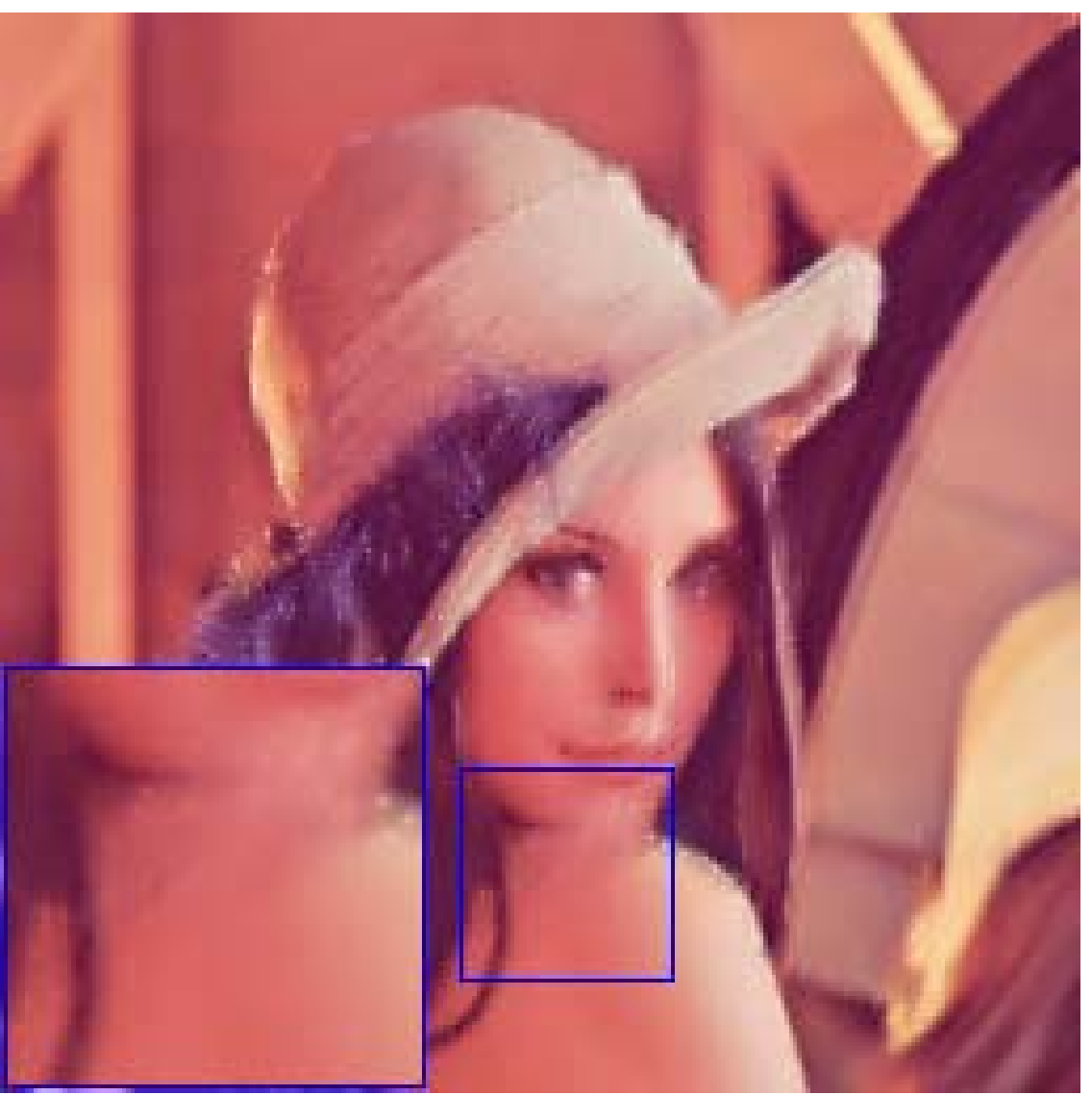}&
\includegraphics[width=0.14\textwidth]{figs/lena.pdf}\\
\includegraphics[width=0.14\textwidth]{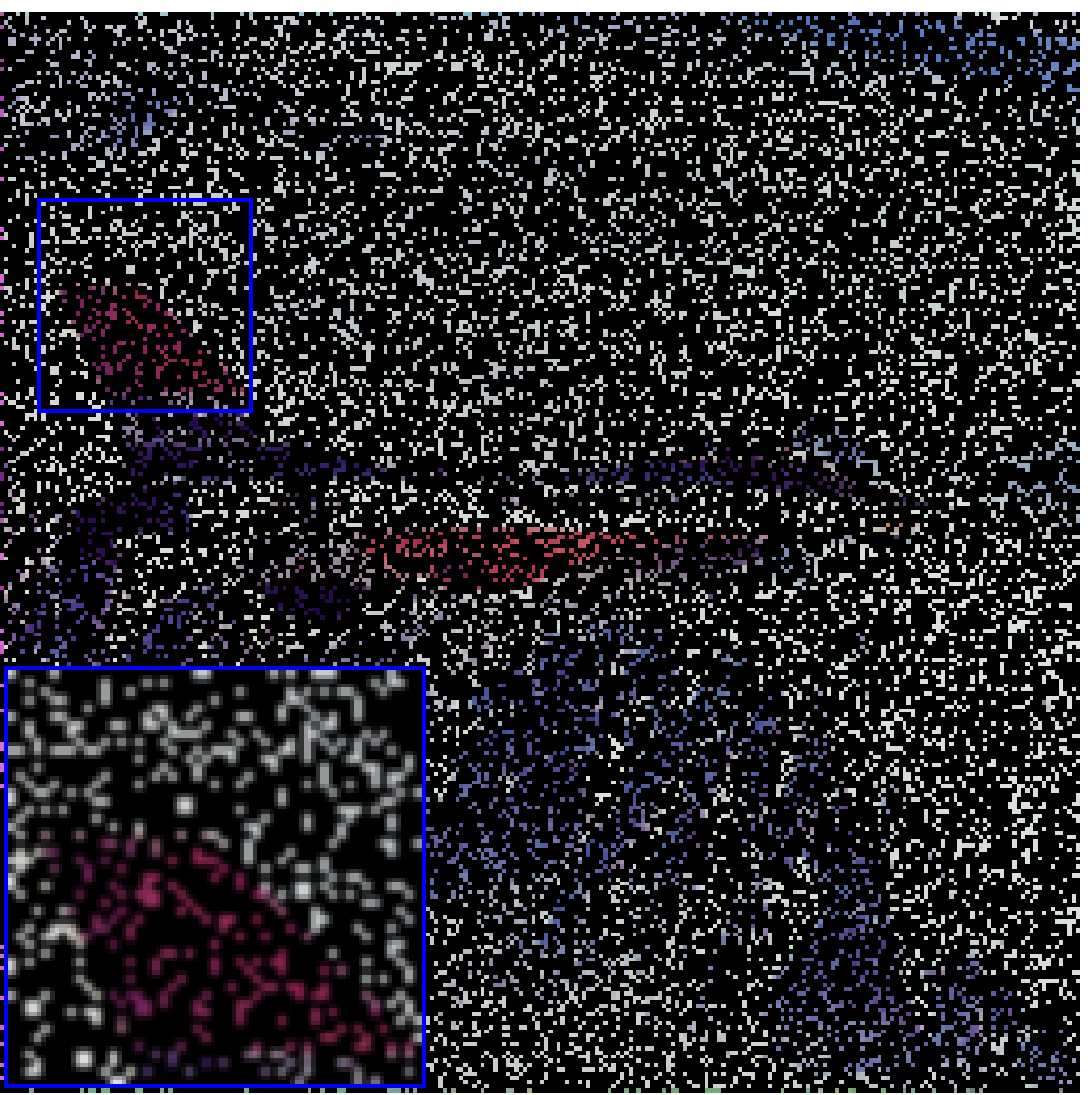}&
\includegraphics[width=0.14\textwidth]{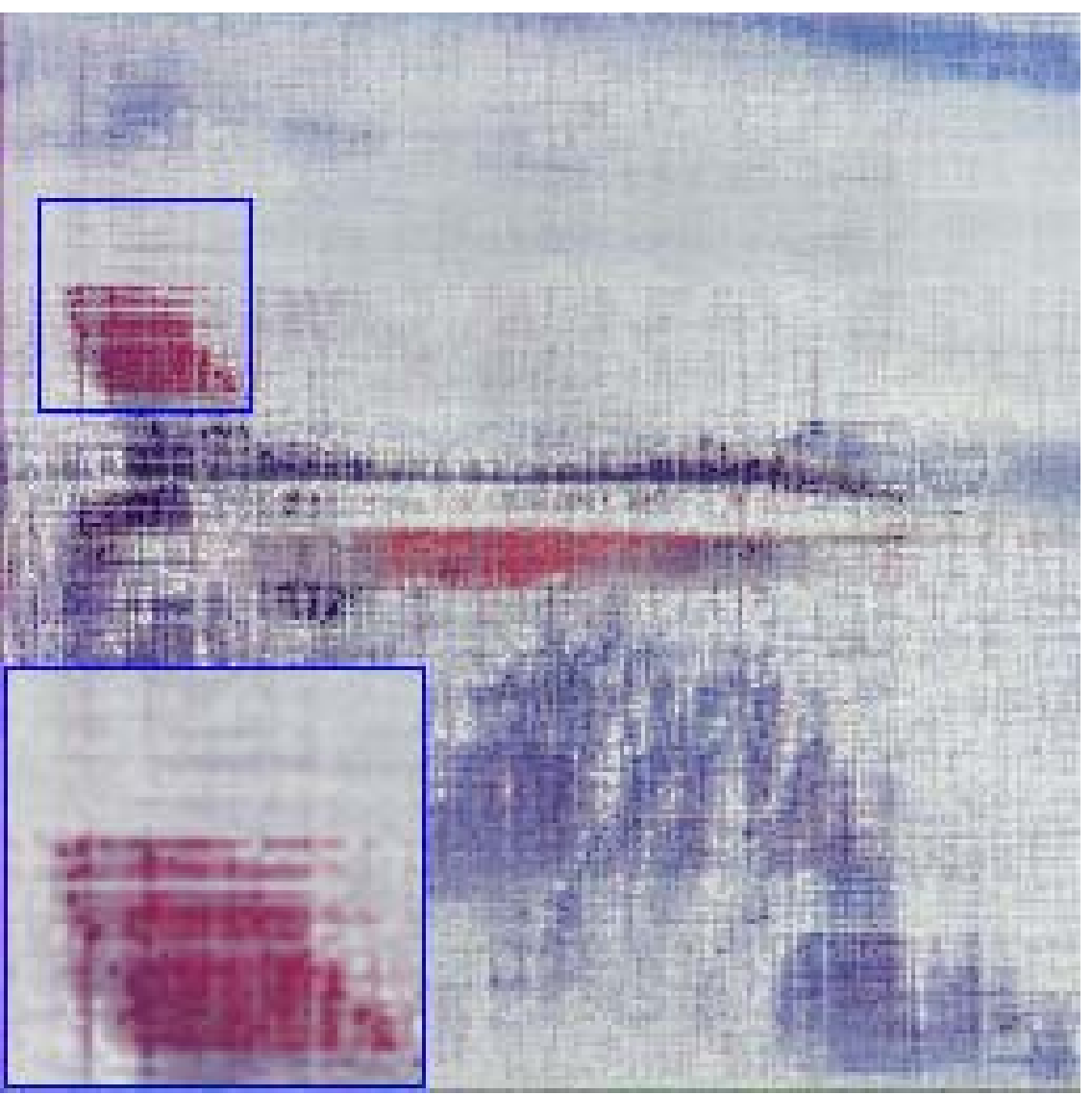}&
\includegraphics[width=0.14\textwidth]{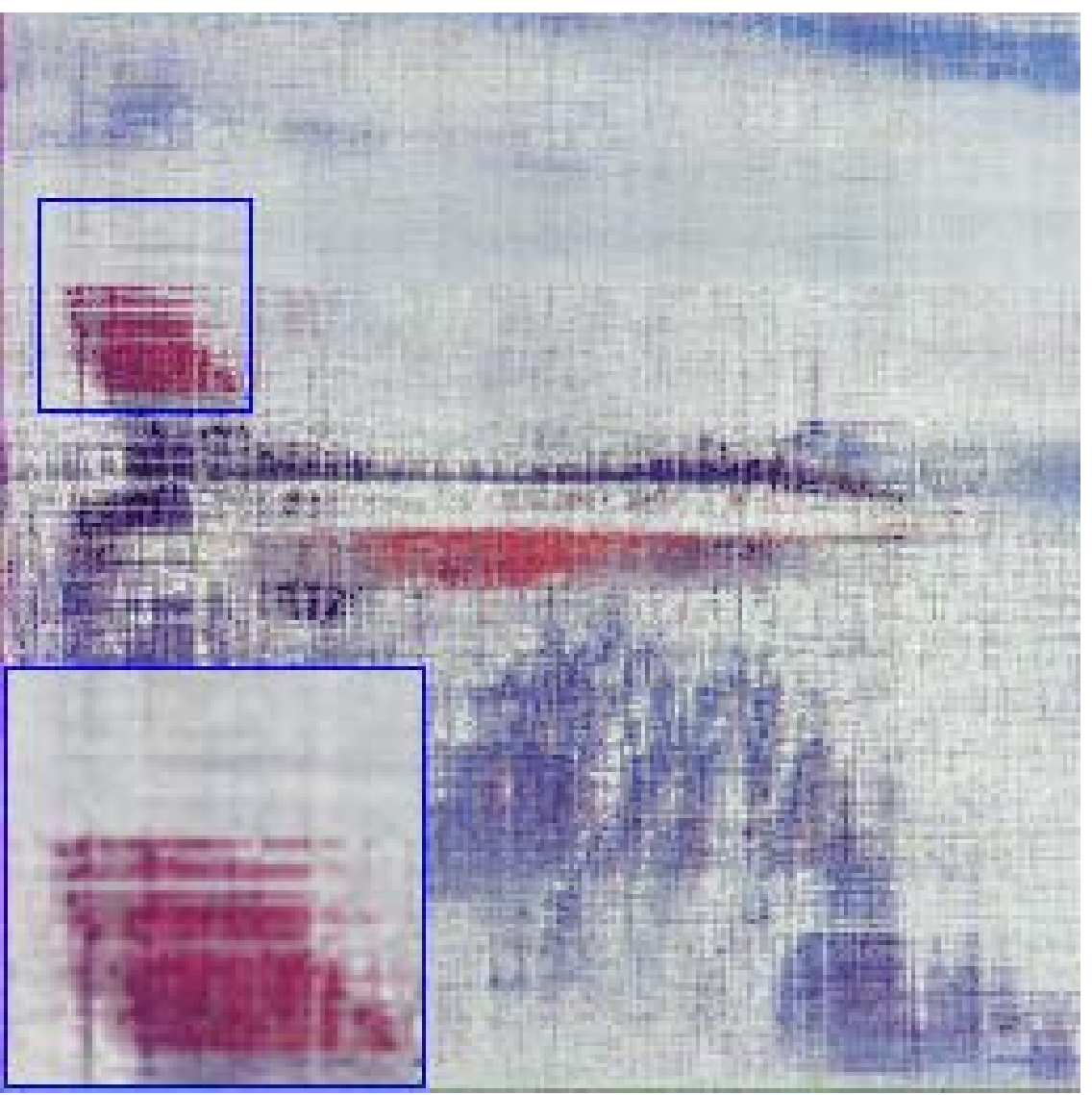}&
\includegraphics[width=0.14\textwidth]{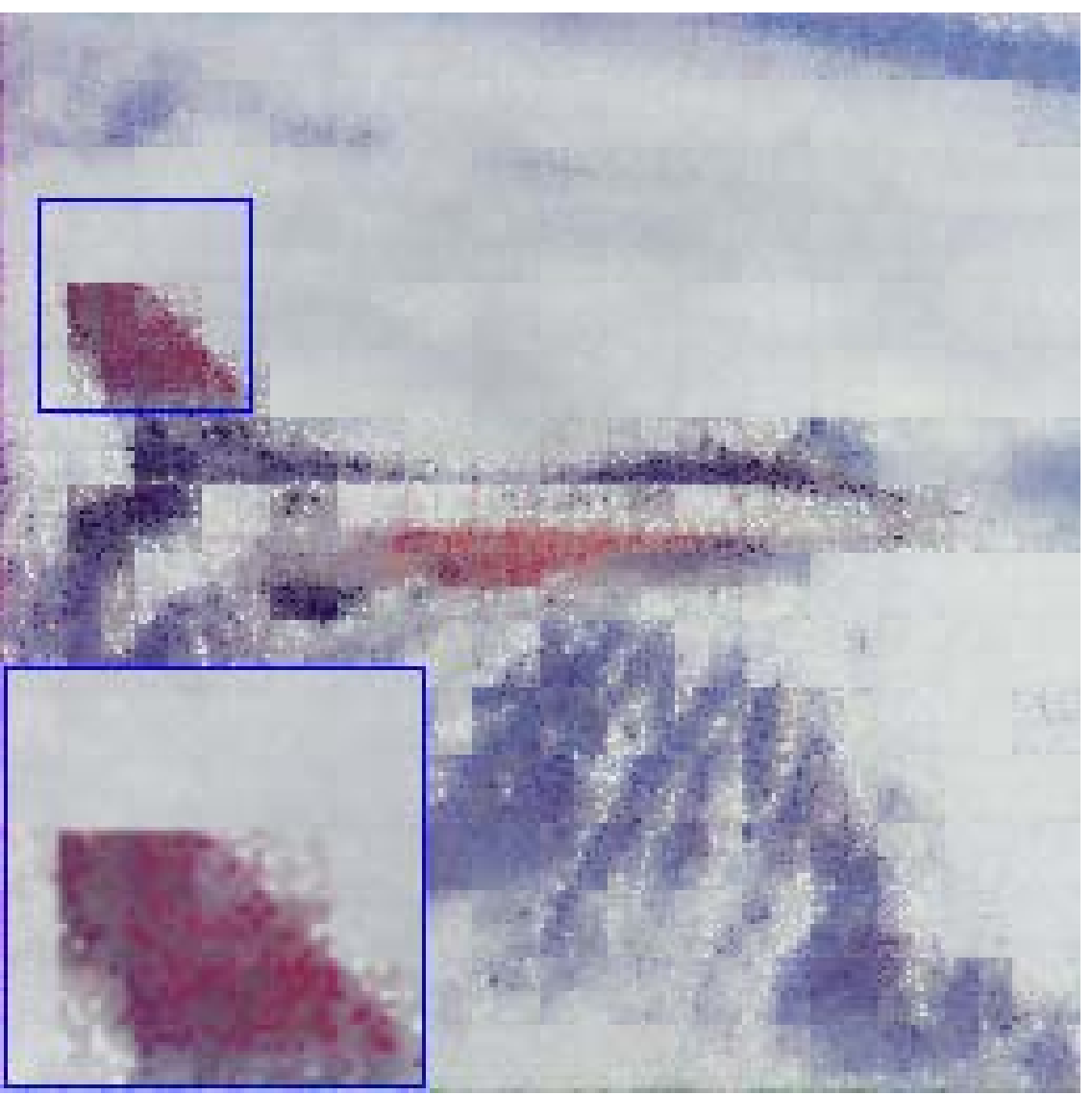}&
\includegraphics[width=0.14\textwidth]{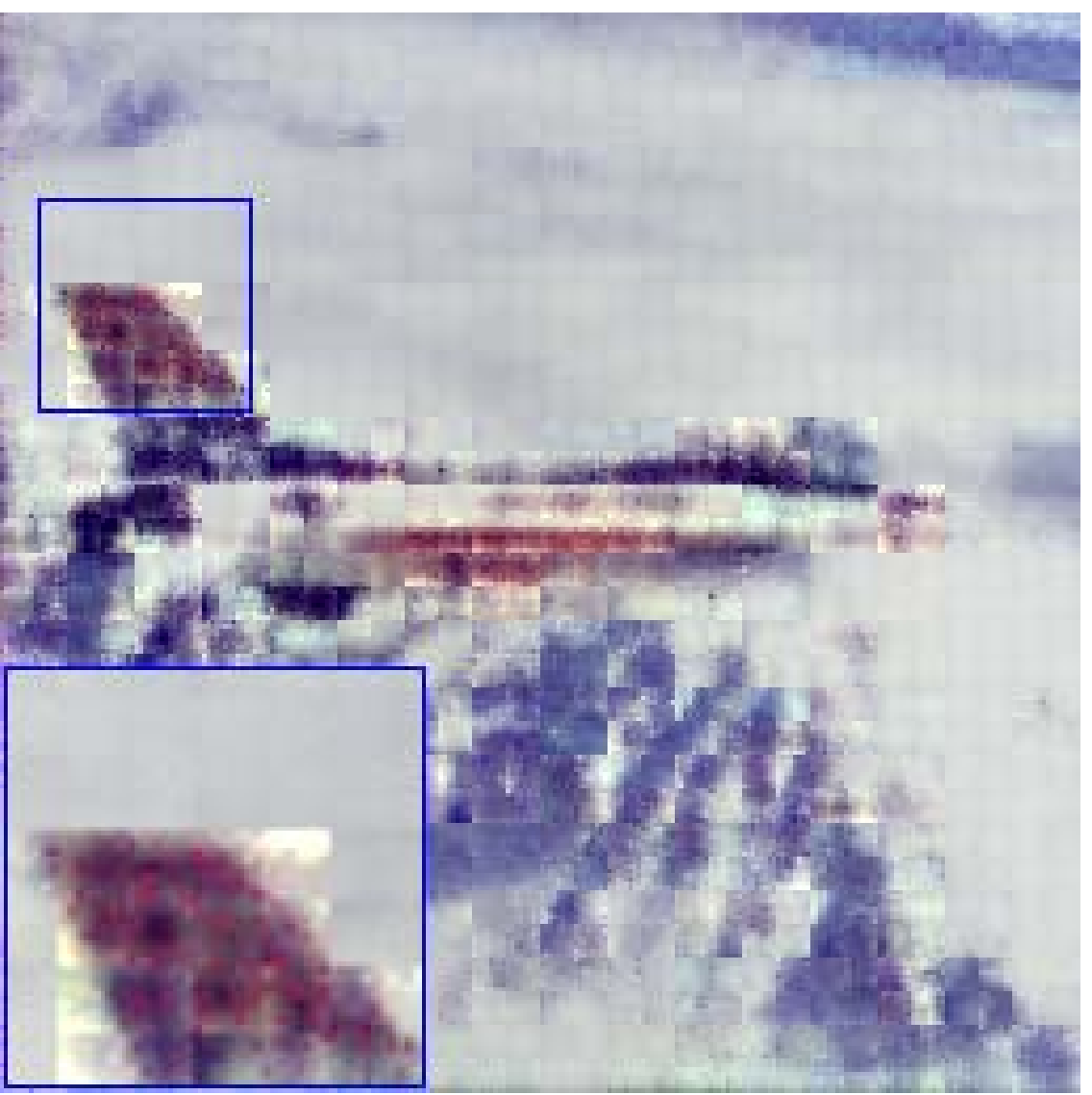}&
\includegraphics[width=0.14\textwidth]{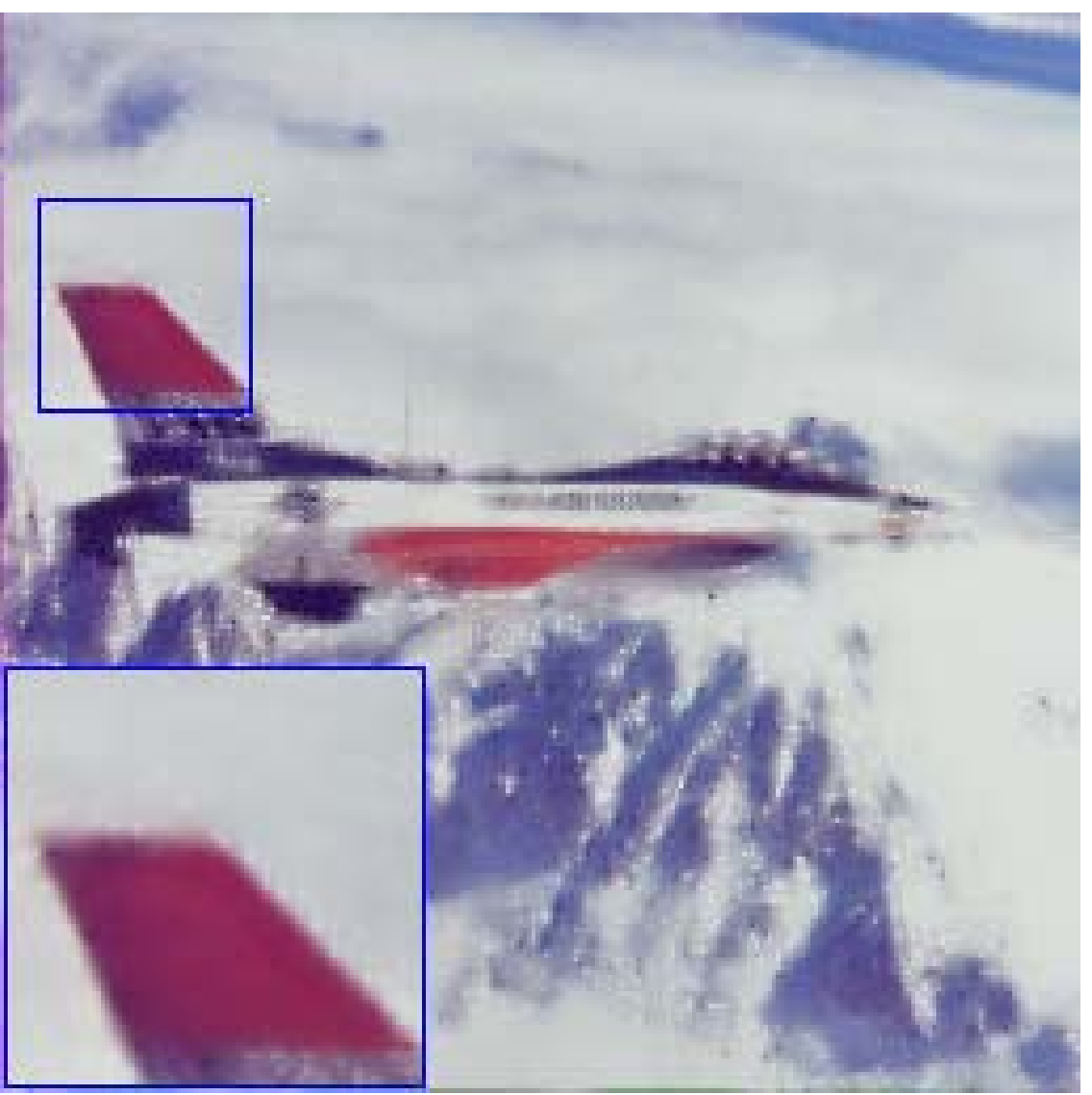}&
\includegraphics[width=0.14\textwidth]{figs/airplane.pdf}\\
\includegraphics[width=0.139\textwidth]{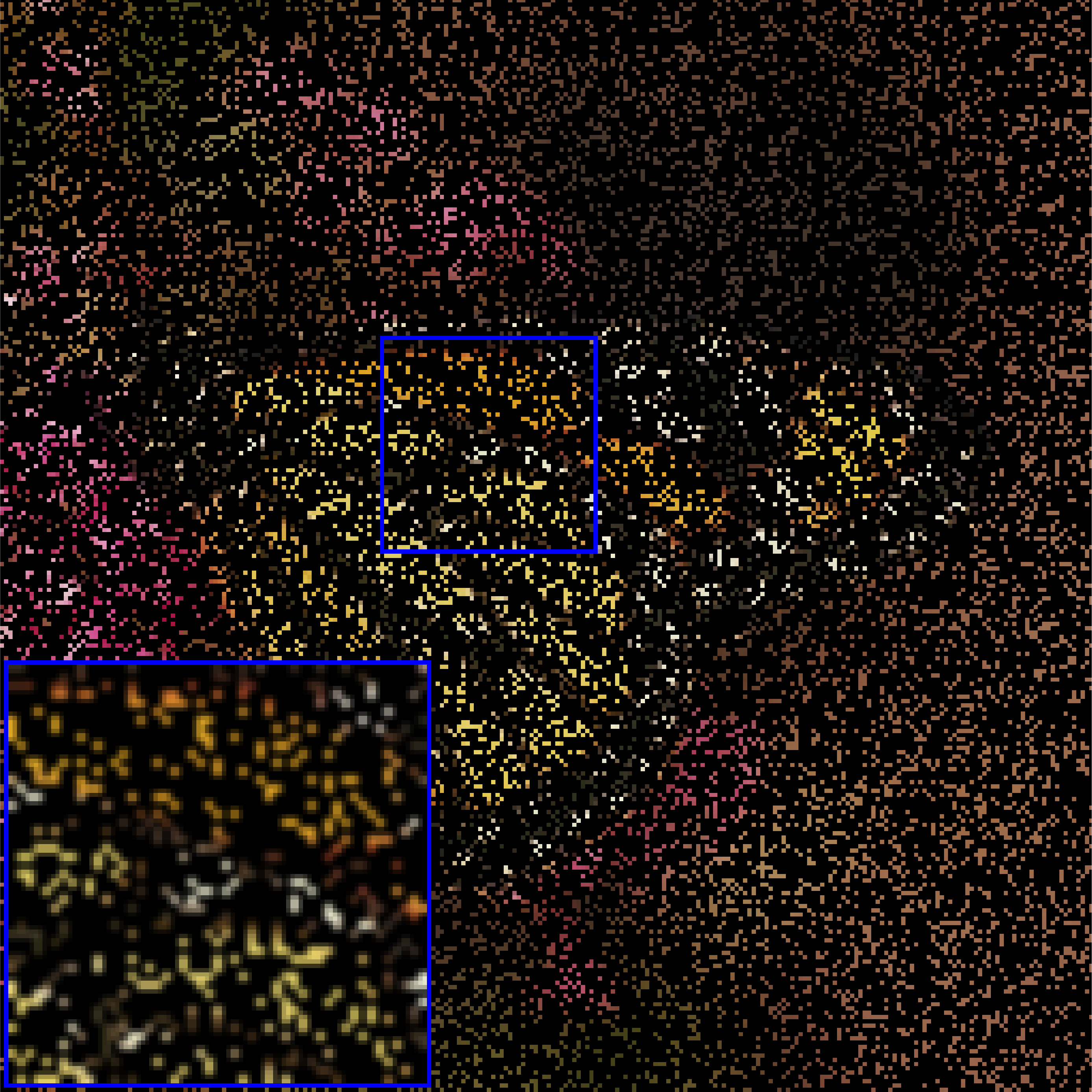}&
\includegraphics[width=0.14\textwidth]{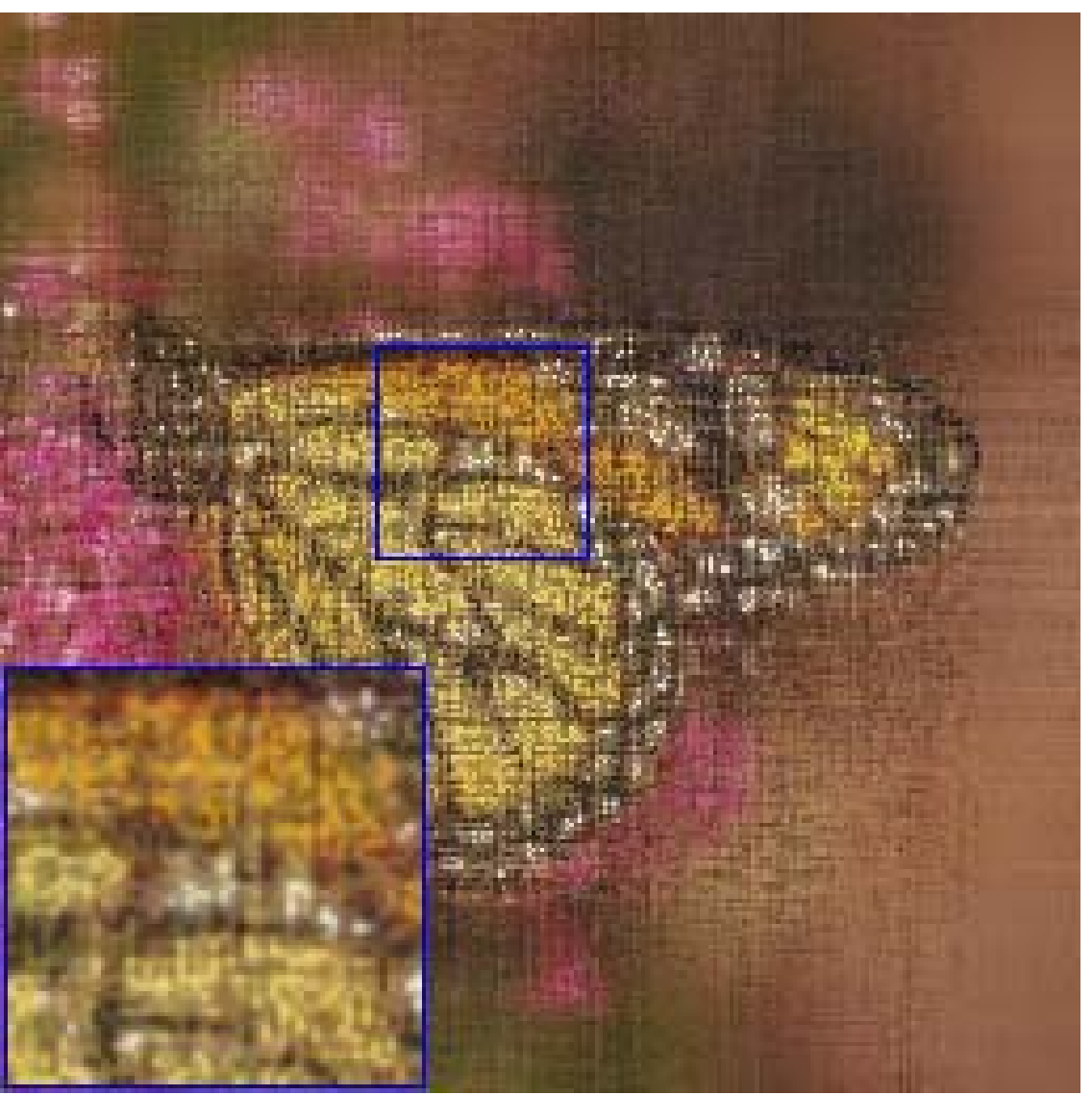}&
\includegraphics[width=0.139\textwidth]{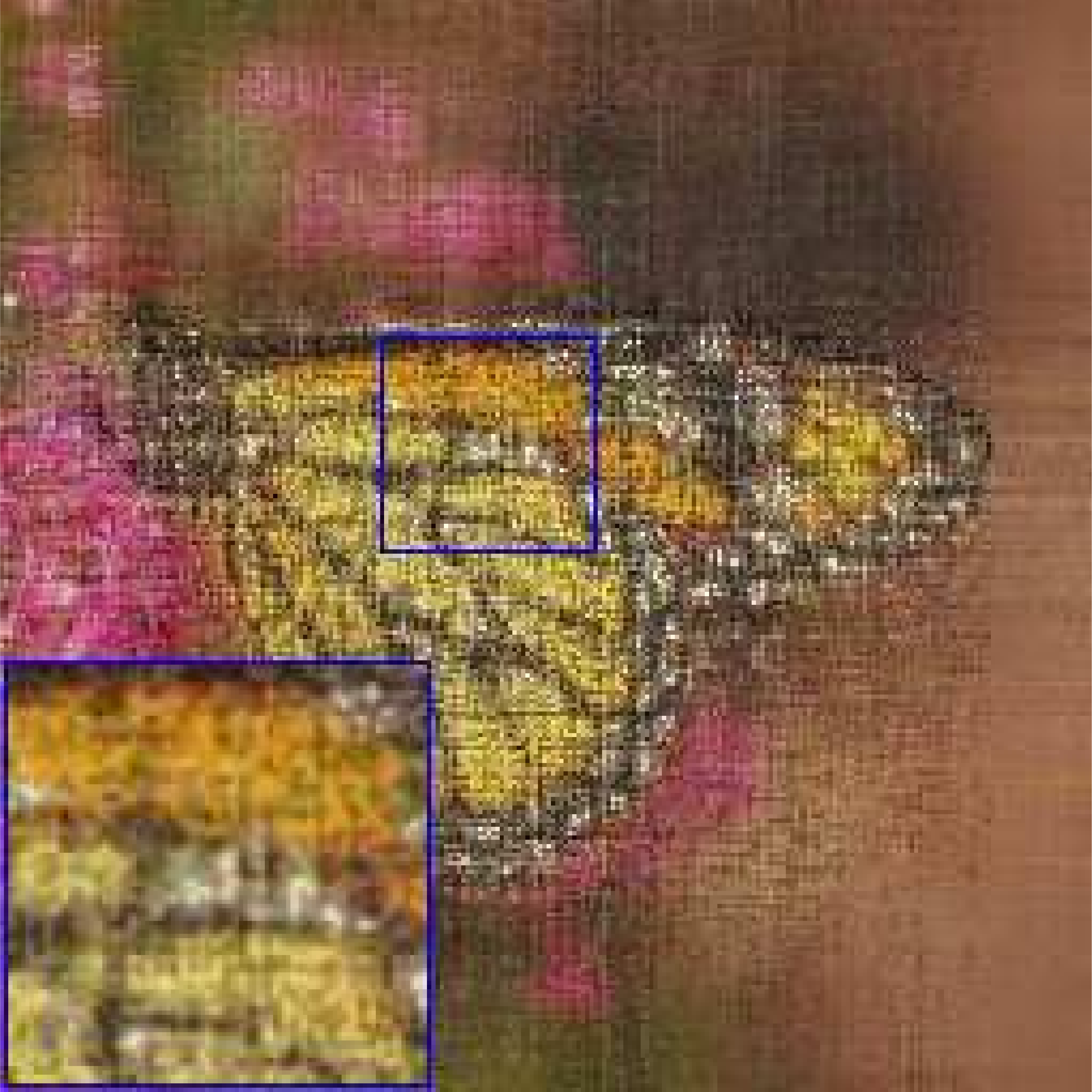}&
\includegraphics[width=0.139\textwidth]{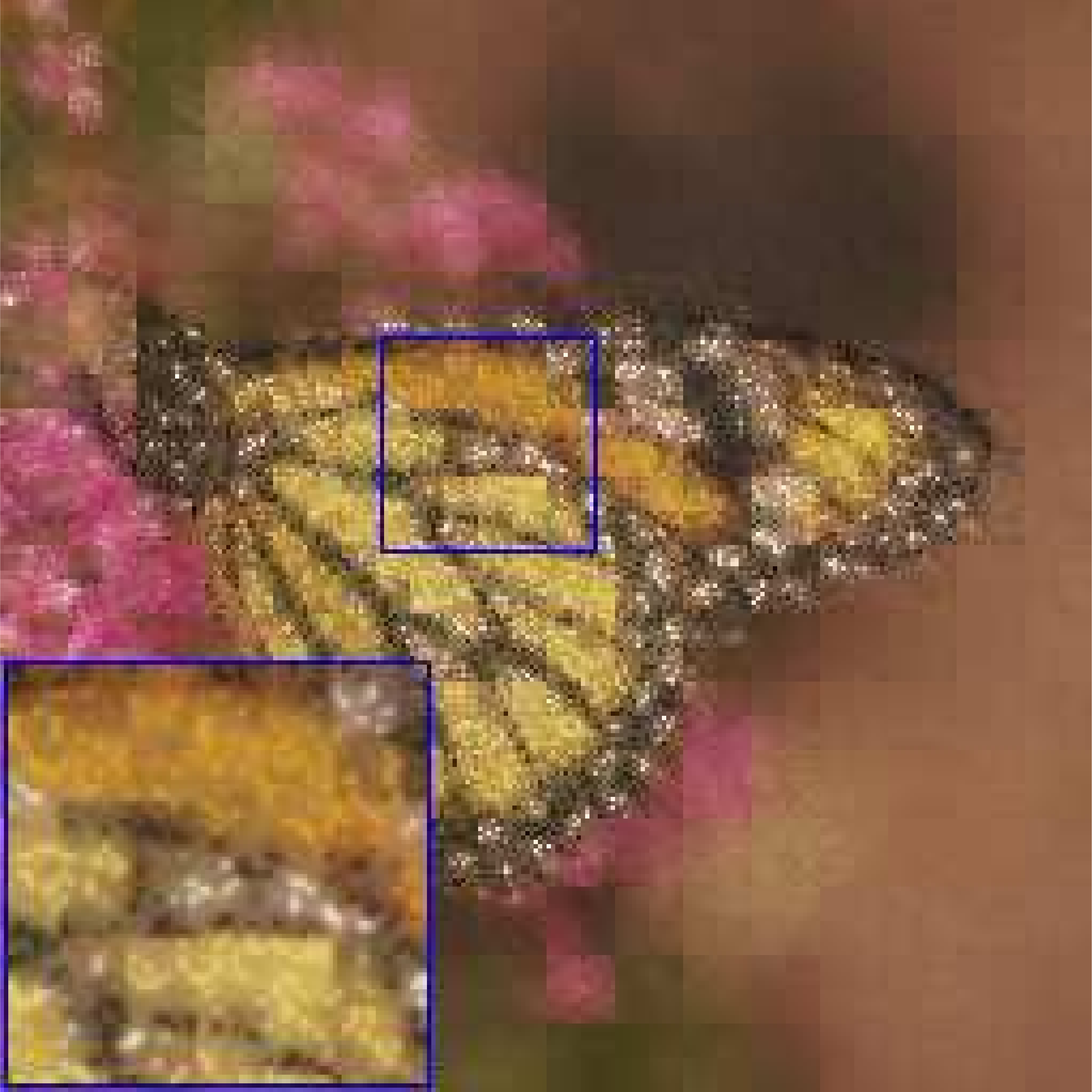}&
\includegraphics[width=0.139\textwidth]{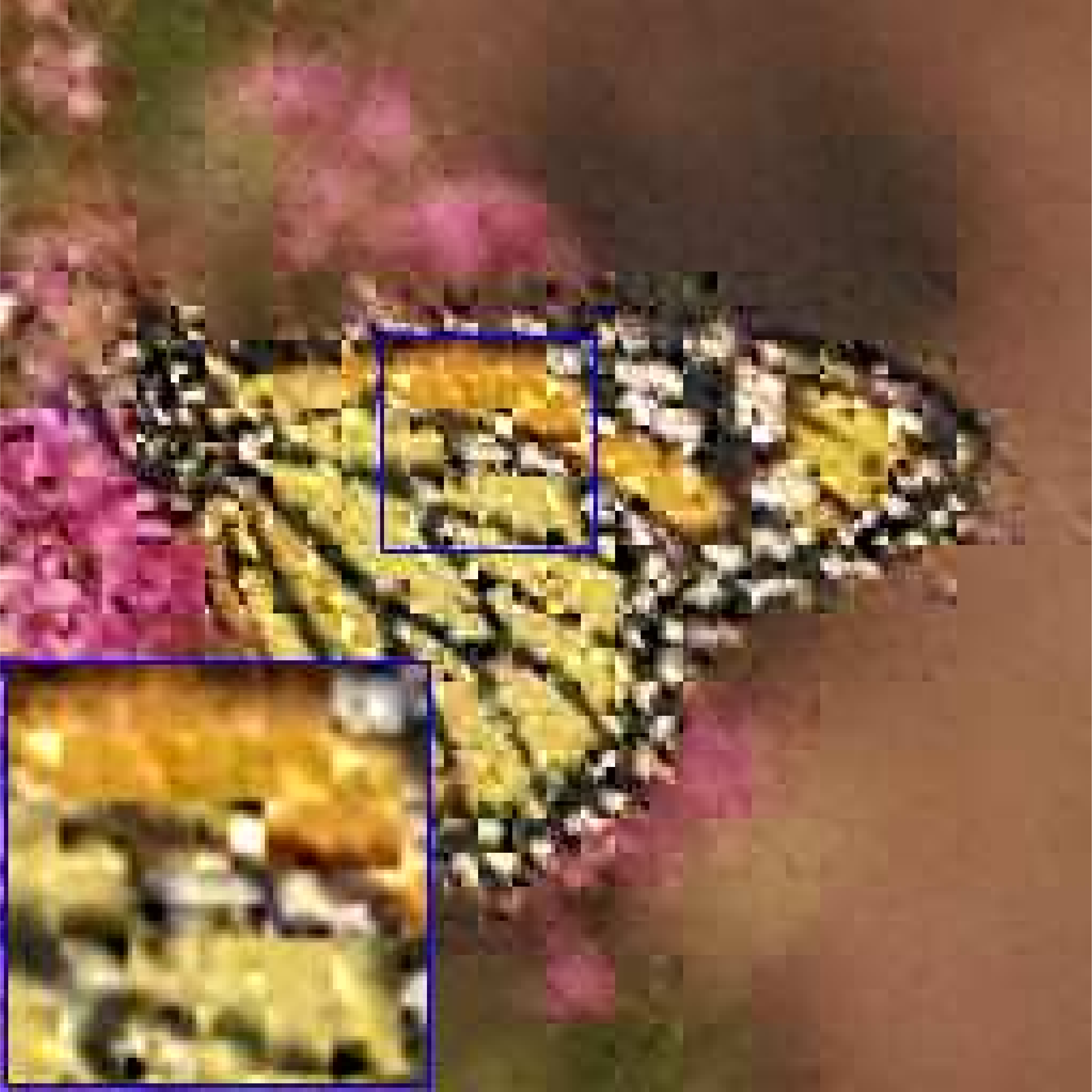}&
\includegraphics[width=0.139\textwidth]{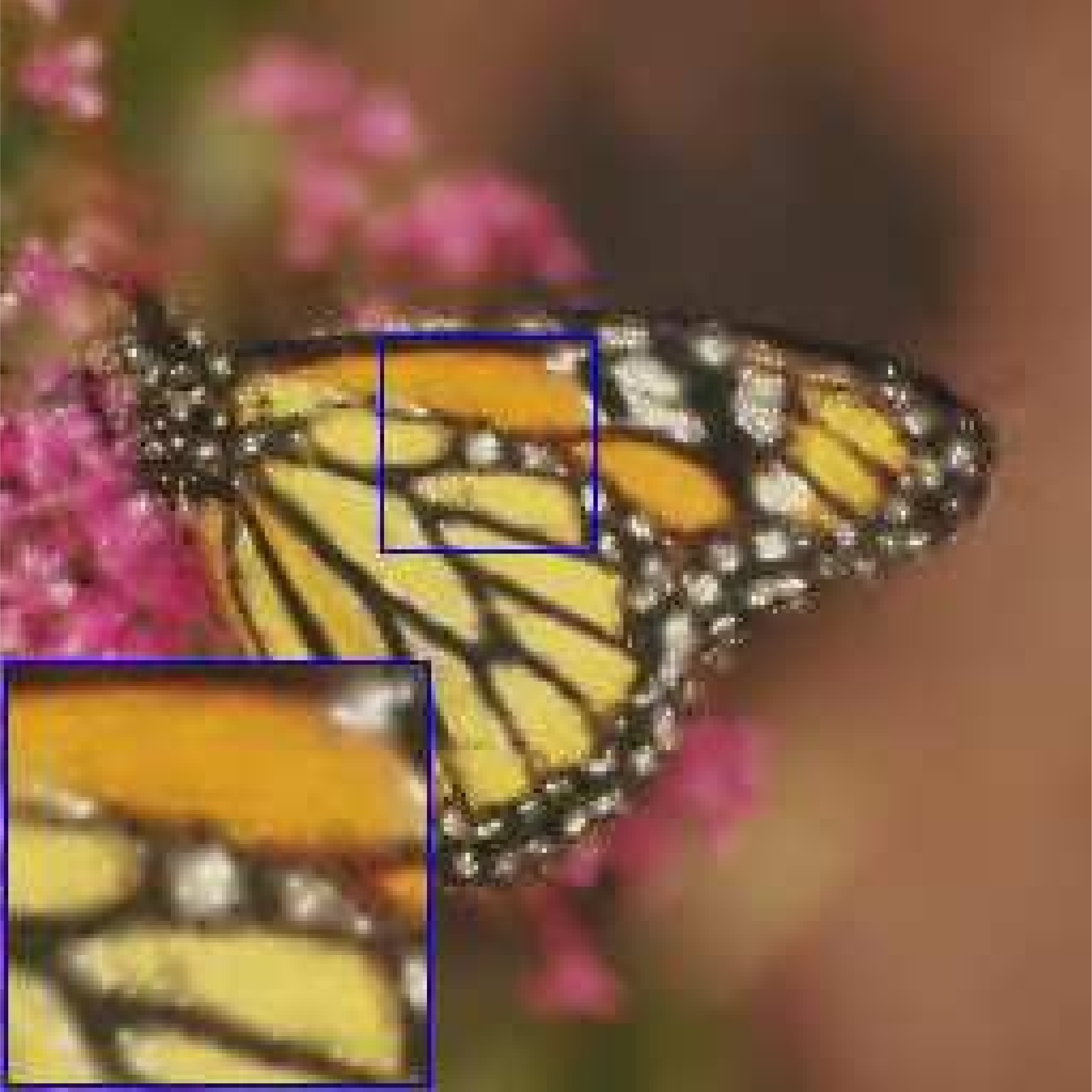}&
\includegraphics[width=0.14\textwidth]{figs/monarch.pdf}\\
 {\footnotesize\textrm{(a) Observed}} & {\footnotesize\textrm{(b) HaLRTC}} & {\footnotesize\textrm{(c) tSVD}} & {\footnotesize\textrm{(d) SiLRTC-TT}} & {\footnotesize\textrm{(e) TMac-TT}} & {\footnotesize\textrm{(f) NL-TT}}& {\footnotesize\textrm{(g) Original}}\\
\end{tabular}
\caption{\small{The results of testing color images with $SR = 0.2$ recovered by different methods. The first three rows and second three rows represent random sampling and tube sampling, respectively. From left to right: (a) the observed image, the results by (b) HaLRTC, (c) tSVD, (d) SiLRTC-TT, (e) TMac-TT, (f) NL-TT, and (g) the original image.}}
  \label{fig:color_entry}
  \end{center}\vspace{-0.3cm}
\end{figure*}

\textbf{Parameters setting.} In the block-matching operator, there are three important parameters: the cube size, the similar cube number, and the overlapping cube size. We set the cube size $s\in [10,20]$ with the increment 2, the number of similar cubes $h\in [30,50]$ with the increment 4, and the overlapping cube size as $o=1$. In our model \eqref{proposed model}, we assign larger weights to $\textbf{X}_{p,[k]}$ with balanced sizes, i.e.,
\begin{equation}\small
\alpha_{k}=\frac{\delta_{k}}{\sum_{k=1}^{j-1}\delta_{k}} \quad  \text{with} \quad  \delta_{k}=\min(\Pi_{d=1}^{k}n_{d}, \Pi_{d=k+1}^{j}n_{d}),
\end{equation}
where $n_{d}$ is the $d$-th order of $\mathcal{X}_{p}$ and $k=1, \ldots, j-1$. In the ADMM solver, we empirically select the penalty parameter $\beta$ in \eqref{Lagrangian function} from the candidate set: \ \{0.05, 0.08, 0.1, 0.3, 0.5\}, to attain the highest PSNR value. For each completed groups, we stop the proposed algorithm according to the relative error of the tensor $\mathcal{X}_{p}$ between two successive iterations as follows:
\begin{equation}\small
\frac{\|\mathcal{X}^{l+1}_{p}-\mathcal{X}^{l}_{p}\|_{F}}{\|\mathcal{X}^{l}_{p}\|_{F}}\leq 10^{-4}.
\end{equation}
In addition, we set the maximum inner iterations $l_{max}=500$. We try our best to tune the parameters involved in the competing algorithms according to the reference papers' suggestion.

\begin{table}[!ht]
\renewcommand\arraystretch{1.2}
  \centering
  \caption{The PSNR and SSIM values obtained by HaLRTC, tSVD, SiLRTC-TT, TMac-TT and NL-TT for color image data with different sampling rates (SRs). The first three rows and second three rows represent random sampling and tube sampling, respectively.}
\begin{tabular}{c|c|cc|cc|cc|cc}
\hline

\hline
\multirow{2}{*}{Image} &SR &\multicolumn{2}{c|}{0.1} &\multicolumn{2}{c|}{0.2} &\multicolumn{2}{c|}{0.3} & \multicolumn{2}{c}{0.4}  \\
\cline{3-10}
    & Method     & PSNR  & SSIM    & PSNR   & SSIM    & PSNR   & SSIM    & PSNR   & SSIM  \\

    \hline
    \multirow{5}[0]{*}{\emph{lena}}
    & HaLRTC    & 19.29  & 0.4151  & 23.10  & 0.6047  & 25.68  & 0.7311  & 28.00  & 0.8205 \\
    & tSVD      & 19.55  & 0.3500  & 23.33  & 0.5572  & 26.08  & 0.7033  & 28.60  & 0.8066 \\
    & SiLRTC-TT & 21.67  & 0.5954  & 24.80  & 0.7366  & 27.01  & 0.8226  & 28.90  & 0.8782 \\
    & TMac-TT   & 24.25  & 0.6829  & 27.22  & 0.8097  & 28.87  & 0.8584  & 30.22  & 0.8902 \\
    & NL-TT     & \textbf{26.46}   & \textbf{0.8110}  & \textbf{30.09}   & \textbf{0.8970} & \textbf{32.02} & \textbf{0.9309} & \textbf{33.87} & \textbf{0.9528}\\
    \hline
    \multirow{5}[0]{*}{\hspace{-0.5cm} \emph{airplane}} \hspace{-0.5cm}
    & HaLRTC    & \hspace{-0.23cm} 19.80 \hspace{-0.3cm} & 0.4621 \hspace{-0.23cm} & \hspace{-0.23cm} 23.18 \hspace{-0.3cm} & 0.6437 \hspace{-0.23cm} & \hspace{-0.23cm} 25.62 \hspace{-0.3cm} & 0.7614 \hspace{-0.23cm} & \hspace{-0.23cm} 27.97 \hspace{-0.3cm} & 0.8399 \hspace{-0.23cm} \\
    & tSVD      & \hspace{-0.23cm} 19.87 \hspace{-0.3cm} & 0.4196 \hspace{-0.23cm} & \hspace{-0.23cm} 23.30 \hspace{-0.3cm} & 0.6139 \hspace{-0.23cm} & \hspace{-0.23cm} 25.86 \hspace{-0.3cm} & 0.7387 \hspace{-0.23cm} & \hspace{-0.23cm} 28.25 \hspace{-0.3cm} & 0.8258 \hspace{-0.23cm} \\
    & \hspace{-0.4cm} SiLRTC-TT \hspace{-0.3cm} & \hspace{-0.23cm} 20.81 \hspace{-0.4cm} & 0.6072 \hspace{-0.23cm} & \hspace{-0.23cm} 23.42 \hspace{-0.3cm} & 0.7361 \hspace{-0.23cm} & \hspace{-0.23cm} 25.62 \hspace{-0.3cm} & 0.8213 \hspace{-0.23cm} & \hspace{-0.23cm} 27.55 \hspace{-0.3cm} & 0.8768 \hspace{-0.23cm} \\
    & TMac-TT   & \hspace{-0.23cm} 22.46 \hspace{-0.3cm} & 0.6766 \hspace{-0.23cm} & \hspace{-0.23cm} 25.81 \hspace{-0.3cm} & 0.8105 \hspace{-0.23cm} & \hspace{-0.23cm} 27.67 \hspace{-0.3cm} & 0.8622 \hspace{-0.23cm} & \hspace{-0.23cm} 28.97 \hspace{-0.3cm} & 0.8915 \hspace{-0.23cm} \\
    & NL-TT     & \hspace{-0.23cm} \textbf{24.33} \hspace{-0.3cm} & \textbf{0.7840} \hspace{-0.23cm} & \hspace{-0.23cm} \textbf{28.33} \hspace{-0.3cm} & \textbf{0.8929} \hspace{-0.23cm} & \hspace{-0.23cm} \textbf{30.29} \hspace{-0.3cm} & \textbf{0.9268} \hspace{-0.23cm} & \hspace{-0.23cm} \textbf{31.99} \hspace{-0.3cm} & \textbf{0.9489} \hspace{-0.23cm} \\
    \hline
    \multirow{5}[0]{*}{\hspace{-0.5cm} \emph{monarch}} \hspace{-0.5cm}
    & HaLRTC    & \hspace{-0.23cm} 17.12 \hspace{-0.3cm} & 0.4381 \hspace{-0.23cm} & \hspace{-0.23cm} 19.59 \hspace{-0.3cm} & 0.6069 \hspace{-0.23cm} & \hspace{-0.23cm} 21.89 \hspace{-0.3cm} & 0.7404 \hspace{-0.23cm} & \hspace{-0.23cm} 24.20 \hspace{-0.3cm} & 0.8271 \hspace{-0.23cm} \\
    & tSVD      & \hspace{-0.23cm} 17.14 \hspace{-0.3cm} & 0.3372 \hspace{-0.23cm} & \hspace{-0.23cm} 19.98 \hspace{-0.3cm} & 0.5462 \hspace{-0.23cm} & \hspace{-0.23cm} 22.60 \hspace{-0.3cm} & 0.6980 \hspace{-0.23cm} & \hspace{-0.23cm} 25.23 \hspace{-0.3cm} & 0.8023 \hspace{-0.23cm} \\
    & \hspace{-0.4cm} SiLRTC-TT \hspace{-0.3cm} & \hspace{-0.23cm} 17.95 \hspace{-0.4cm} & 0.5784 \hspace{-0.23cm} & \hspace{-0.23cm} 20.32 \hspace{-0.3cm} & 0.7196 \hspace{-0.23cm} & \hspace{-0.23cm} 22.38 \hspace{-0.3cm} & 0.8100 \hspace{-0.23cm} & \hspace{-0.23cm} 24.39 \hspace{-0.3cm} & 0.8702 \hspace{-0.23cm} \\
    & TMac-TT   & \hspace{-0.23cm} 19.21 \hspace{-0.3cm} & 0.6621 \hspace{-0.23cm} & \hspace{-0.23cm} 22.45 \hspace{-0.3cm} & 0.7912 \hspace{-0.23cm} & \hspace{-0.23cm} 24.86 \hspace{-0.3cm} & 0.8505 \hspace{-0.23cm} & \hspace{-0.23cm} 27.24 \hspace{-0.3cm} & 0.9046 \hspace{-0.23cm} \\
    & NL-TT     & \hspace{-0.23cm} \textbf{22.22} \hspace{-0.3cm} & \textbf{0.8307} \hspace{-0.23cm} & \hspace{-0.23cm} \textbf{25.42} \hspace{-0.3cm} & \textbf{0.9140} \hspace{-0.23cm} & \hspace{-0.23cm} \textbf{27.95} \hspace{-0.3cm} & \textbf{0.9496} \hspace{-0.23cm} & \hspace{-0.23cm} \textbf{30.74} \hspace{-0.3cm} & \textbf{0.9729} \hspace{-0.23cm} \\
    \hline
    \multirow{5}[0]{*}{\hspace{-0.5cm} \emph{lena}} \hspace{-0.5cm}
    & HaLRTC    & \hspace{-0.23cm} 17.54 \hspace{-0.3cm} & 0.2942 \hspace{-0.23cm} & \hspace{-0.23cm} 20.97 \hspace{-0.3cm} & 0.4651 \hspace{-0.23cm} & \hspace{-0.23cm} 23.59 \hspace{-0.3cm} & 0.6144 \hspace{-0.23cm} & \hspace{-0.23cm} 25.88 \hspace{-0.3cm} & 0.7272 \hspace{-0.23cm} \\
    & tSVD      & \hspace{-0.23cm} 17.88 \hspace{-0.3cm} & 0.2570 \hspace{-0.23cm} & \hspace{-0.23cm} 20.85 \hspace{-0.3cm} & 0.4186 \hspace{-0.23cm} & \hspace{-0.23cm} 23.29 \hspace{-0.3cm} & 0.5676 \hspace{-0.23cm} & \hspace{-0.23cm} 25.50 \hspace{-0.3cm} & 0.6857 \hspace{-0.23cm} \\
    & \hspace{-0.4cm} SiLRTC-TT \hspace{-0.3cm} & \hspace{-0.23cm} 20.90 \hspace{-0.4cm} & 0.5462 \hspace{-0.23cm} & \hspace{-0.23cm} 23.61 \hspace{-0.3cm} & 0.6830 \hspace{-0.23cm} & \hspace{-0.23cm} 25.69 \hspace{-0.3cm} & 0.7732 \hspace{-0.23cm} & \hspace{-0.23cm} 27.35 \hspace{-0.3cm} & 0.8353 \hspace{-0.23cm} \\
    & TMac-TT   & \hspace{-0.23cm} 21.62 \hspace{-0.3cm} & 0.5629 \hspace{-0.23cm} & \hspace{-0.23cm} 24.60 \hspace{-0.3cm} & 0.7193 \hspace{-0.23cm} & \hspace{-0.23cm} 26.22 \hspace{-0.3cm} & 0.7764 \hspace{-0.23cm} & \hspace{-0.23cm} 27.55 \hspace{-0.3cm} & 0.8392 \hspace{-0.23cm} \\
    & NL-TT     & \hspace{-0.23cm} \textbf{23.94} \hspace{-0.3cm} & \textbf{0.7351} \hspace{-0.23cm} & \hspace{-0.23cm} \textbf{27.45} \hspace{-0.3cm} & \textbf{0.8459} \hspace{-0.23cm} & \hspace{-0.23cm} \textbf{29.33} \hspace{-0.3cm} & \textbf{0.8928} \hspace{-0.23cm} & \hspace{-0.23cm} \textbf{31.38} \hspace{-0.3cm} & \textbf{0.9259} \hspace{-0.23cm} \\
    \hline
    \multirow{5}[0]{*}{\hspace{-0.5cm} \emph{airplane}} \hspace{-0.5cm}
    & HaLRTC    & \hspace{-0.23cm} 17.81 \hspace{-0.3cm} & 0.3050 \hspace{-0.23cm} & \hspace{-0.23cm} 20.77 \hspace{-0.3cm} & 0.4847 \hspace{-0.23cm} & \hspace{-0.23cm} 23.15 \hspace{-0.3cm} & 0.6214 \hspace{-0.23cm} & \hspace{-0.23cm} 25.29 \hspace{-0.3cm} & 0.7289 \hspace{-0.23cm} \\
    & tSVD      & \hspace{-0.23cm} 17.97 \hspace{-0.3cm} & 0.2900 \hspace{-0.23cm} & \hspace{-0.23cm} 20.66 \hspace{-0.3cm} & 0.4588 \hspace{-0.23cm} & \hspace{-0.23cm} 22.97 \hspace{-0.3cm} & 0.5926 \hspace{-0.23cm} & \hspace{-0.23cm} 25.06 \hspace{-0.3cm} & 0.7029 \hspace{-0.23cm} \\
    & \hspace{-0.4cm} SiLRTC-TT \hspace{-0.3cm} & \hspace{-0.23cm} 20.20 \hspace{-0.4cm} & 0.5570 \hspace{-0.23cm} & \hspace{-0.23cm} 22.49 \hspace{-0.3cm} & 0.6809 \hspace{-0.23cm} & \hspace{-0.23cm} 24.33 \hspace{-0.3cm} & 0.7661 \hspace{-0.23cm} & \hspace{-0.23cm} 26.09 \hspace{-0.3cm} & 0.8298 \hspace{-0.23cm} \\
    & TMac-TT   & \hspace{-0.23cm} 21.06 \hspace{-0.3cm} & 0.6169 \hspace{-0.23cm} & \hspace{-0.23cm} 23.15 \hspace{-0.3cm} & 0.7114 \hspace{-0.23cm} & \hspace{-0.23cm} 24.41 \hspace{-0.3cm} & 0.7729 \hspace{-0.23cm} & \hspace{-0.23cm} 26.17 \hspace{-0.3cm} & 0.8416 \hspace{-0.23cm} \\
    & NL-TT     & \hspace{-0.23cm} \textbf{22.45} \hspace{-0.3cm} & \textbf{0.7255} \hspace{-0.23cm} & \hspace{-0.23cm} \textbf{25.25} \hspace{-0.3cm} & \textbf{0.8210} \hspace{-0.23cm} & \hspace{-0.23cm} \textbf{27.29} \hspace{-0.3cm} & \textbf{0.8749} \hspace{-0.23cm} & \hspace{-0.23cm} \textbf{29.24} \hspace{-0.3cm} & \textbf{0.9149} \hspace{-0.23cm} \\
    \hline
    \multirow{5}[0]{*}{\hspace{-0.5cm} \emph{monarch}} \hspace{-0.5cm}
    & HaLRTC    & \hspace{-0.23cm} 16.04 \hspace{-0.3cm} & 0.3424 \hspace{-0.23cm} & \hspace{-0.23cm} 18.28 \hspace{-0.3cm} & 0.5031 \hspace{-0.23cm} & \hspace{-0.23cm} 20.12 \hspace{-0.3cm} & 0.6363 \hspace{-0.23cm} & \hspace{-0.23cm} 21.93 \hspace{-0.3cm} & 0.7401 \hspace{-0.23cm} \\
    & tSVD      & \hspace{-0.23cm} 16.33 \hspace{-0.3cm} & 0.2786 \hspace{-0.23cm} & \hspace{-0.23cm} 18.21 \hspace{-0.3cm} & 0.4312 \hspace{-0.23cm} & \hspace{-0.23cm} 19.90 \hspace{-0.3cm} & 0.5620 \hspace{-0.23cm} & \hspace{-0.23cm} 21.65 \hspace{-0.3cm} & 0.6791 \hspace{-0.23cm} \\
    & \hspace{-0.4cm} SiLRTC-TT \hspace{-0.4cm} & \hspace{-0.23cm} 17.46 \hspace{-0.4cm} & 0.5472 \hspace{-0.23cm} & \hspace{-0.23cm} 19.48 \hspace{-0.3cm} & 0.6695 \hspace{-0.23cm} & \hspace{-0.23cm} 21.19 \hspace{-0.3cm} & 0.7606 \hspace{-0.23cm} & \hspace{-0.23cm} 22.83 \hspace{-0.3cm} & 0.8290 \hspace{-0.23cm} \\
    & TMac-TT   & \hspace{-0.23cm} 15.12 \hspace{-0.3cm} & 0.3466 \hspace{-0.23cm} & \hspace{-0.23cm} 18.66 \hspace{-0.3cm} & 0.6710 \hspace{-0.23cm} & \hspace{-0.23cm} 21.74 \hspace{-0.3cm} & 0.7739 \hspace{-0.23cm} & \hspace{-0.23cm} 23.49 \hspace{-0.3cm} & 0.8282 \hspace{-0.23cm} \\
    & NL-TT     & \hspace{-0.23cm} \textbf{18.07} \hspace{-0.3cm} & \textbf{0.6564} \hspace{-0.23cm} & \hspace{-0.23cm} \textbf{22.33} \hspace{-0.3cm} & \textbf{0.8462} \hspace{-0.23cm} & \hspace{-0.23cm} \textbf{24.53} \hspace{-0.3cm} & \textbf{0.9086} \hspace{-0.23cm} & \hspace{-0.23cm} \textbf{26.25} \hspace{-0.3cm} & \textbf{0.9391} \hspace{-0.23cm} \\
    \hline

    \hline
    \end{tabular}%
  \label{tab:color_random}%
\end{table}

\subsection{Color images}
We evaluate the performance of NL-TT on color images. The test images of size $256\times 256\times 3$ are shown in Fig. \ref{fig:ori_image}. For color images, we test two kinds of missing entries: (1) random missing entries, including random sampling and tube sampling; and (2) structural missing entries, including missing curves, missing slices, missing texts, and missing blocks. The sampling rate (SR) is tested from 0.05 to 0.6.

\begin{figure}[!ht]
\scriptsize\setlength{\tabcolsep}{0.9pt}
\begin{center}
\begin{tabular}{cccc}
\includegraphics[width=0.32\textwidth]{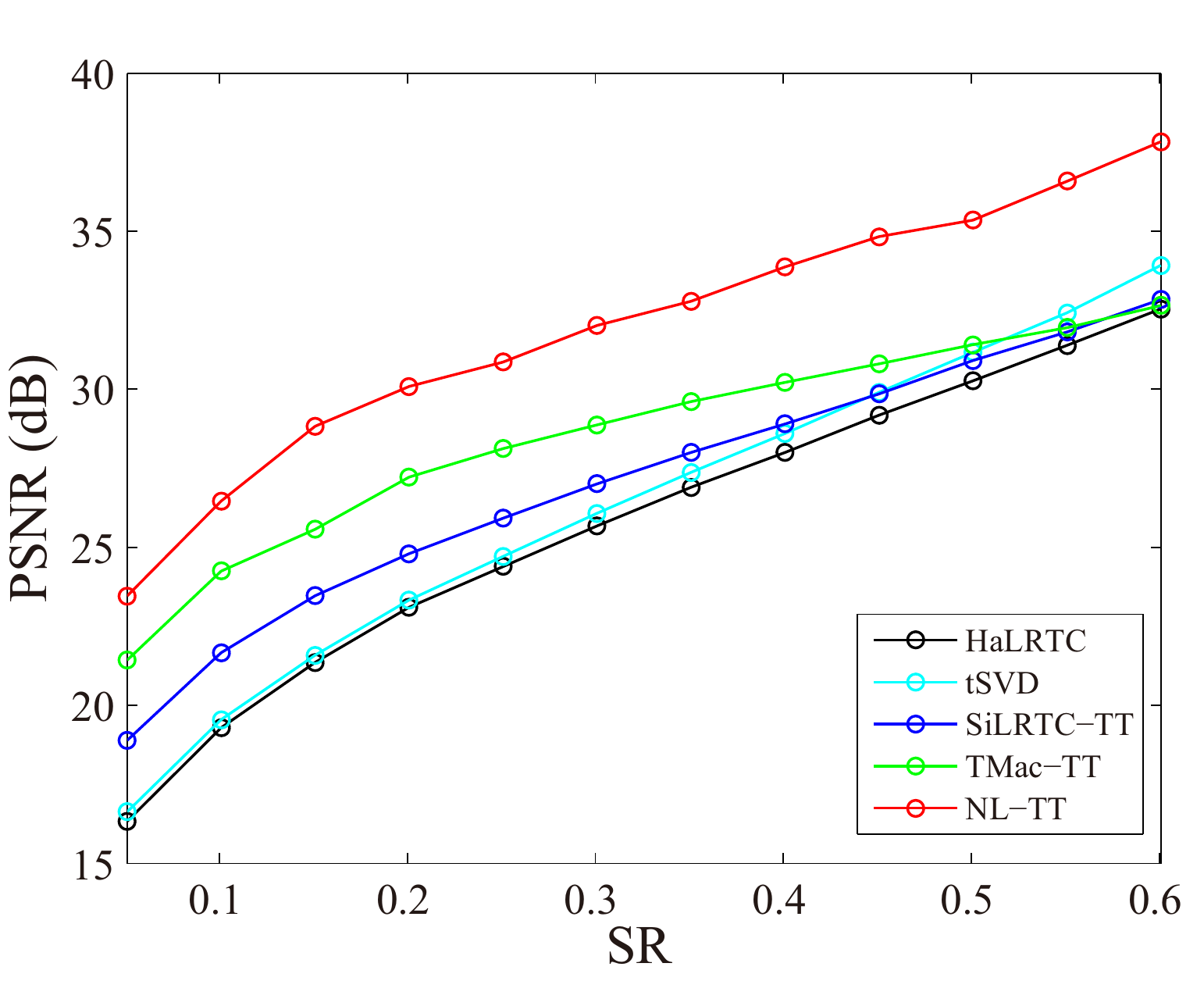}&
\includegraphics[width=0.32\textwidth]{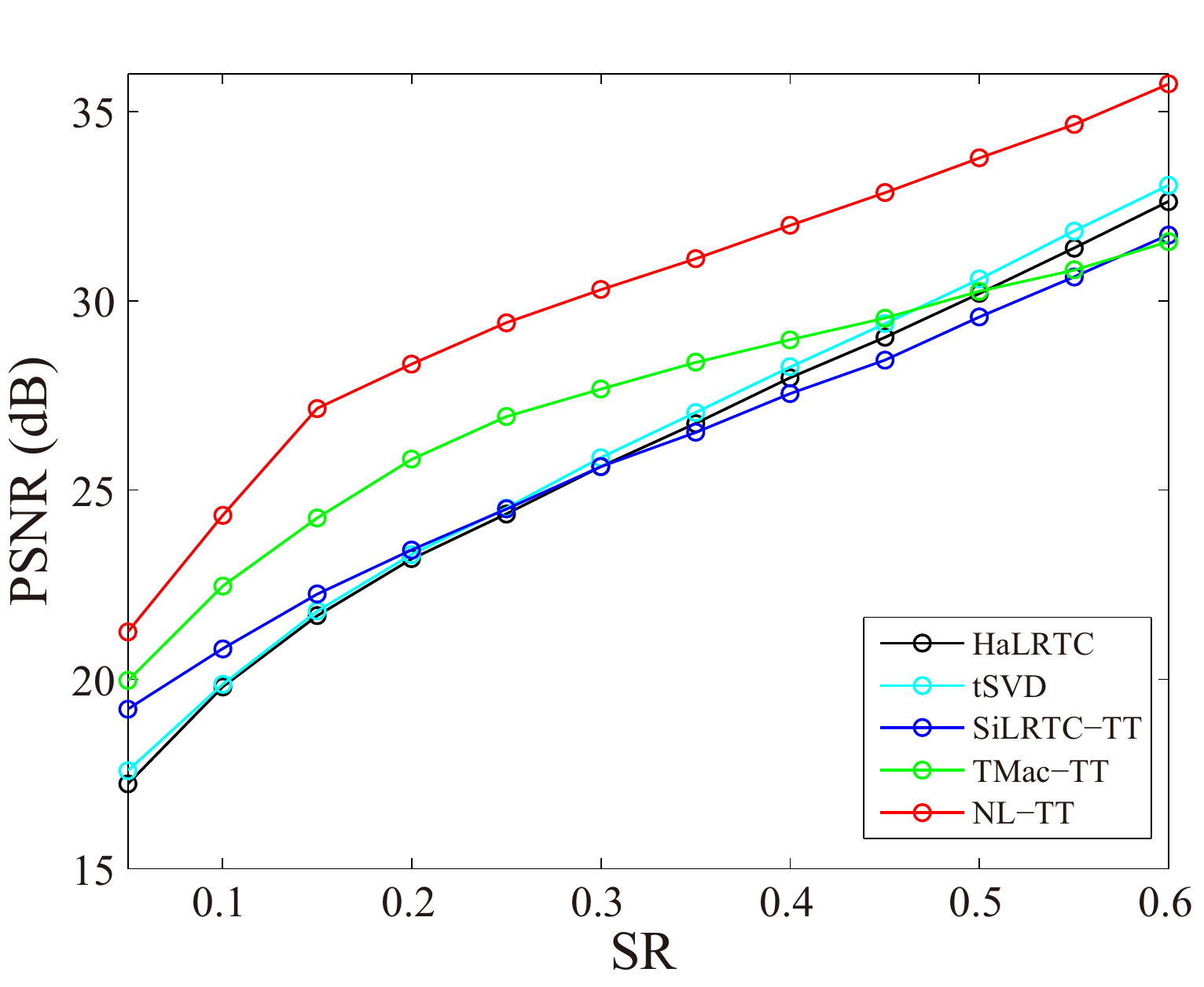}&
\includegraphics[width=0.32\textwidth]{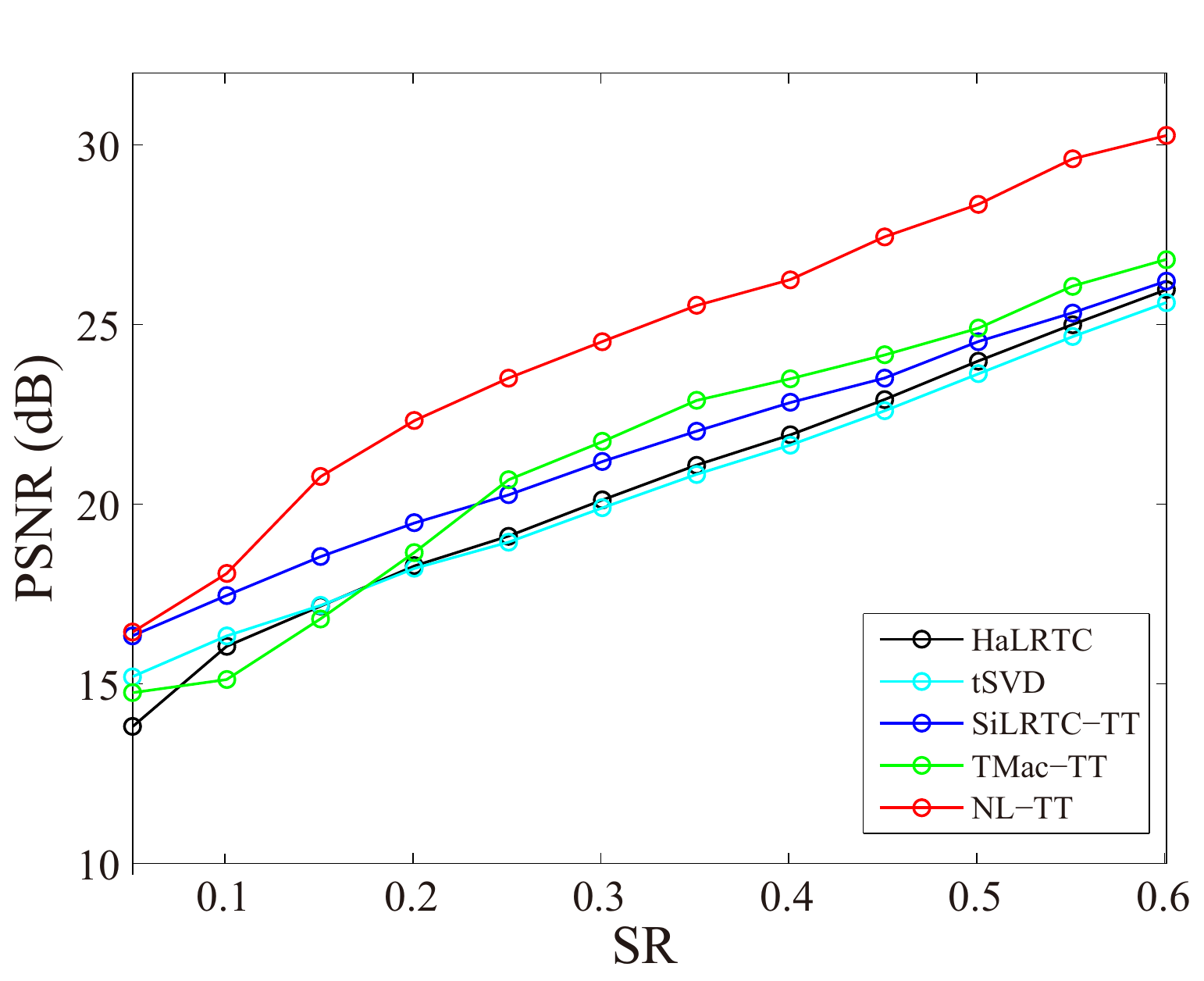}&\\
\includegraphics[width=0.32\textwidth]{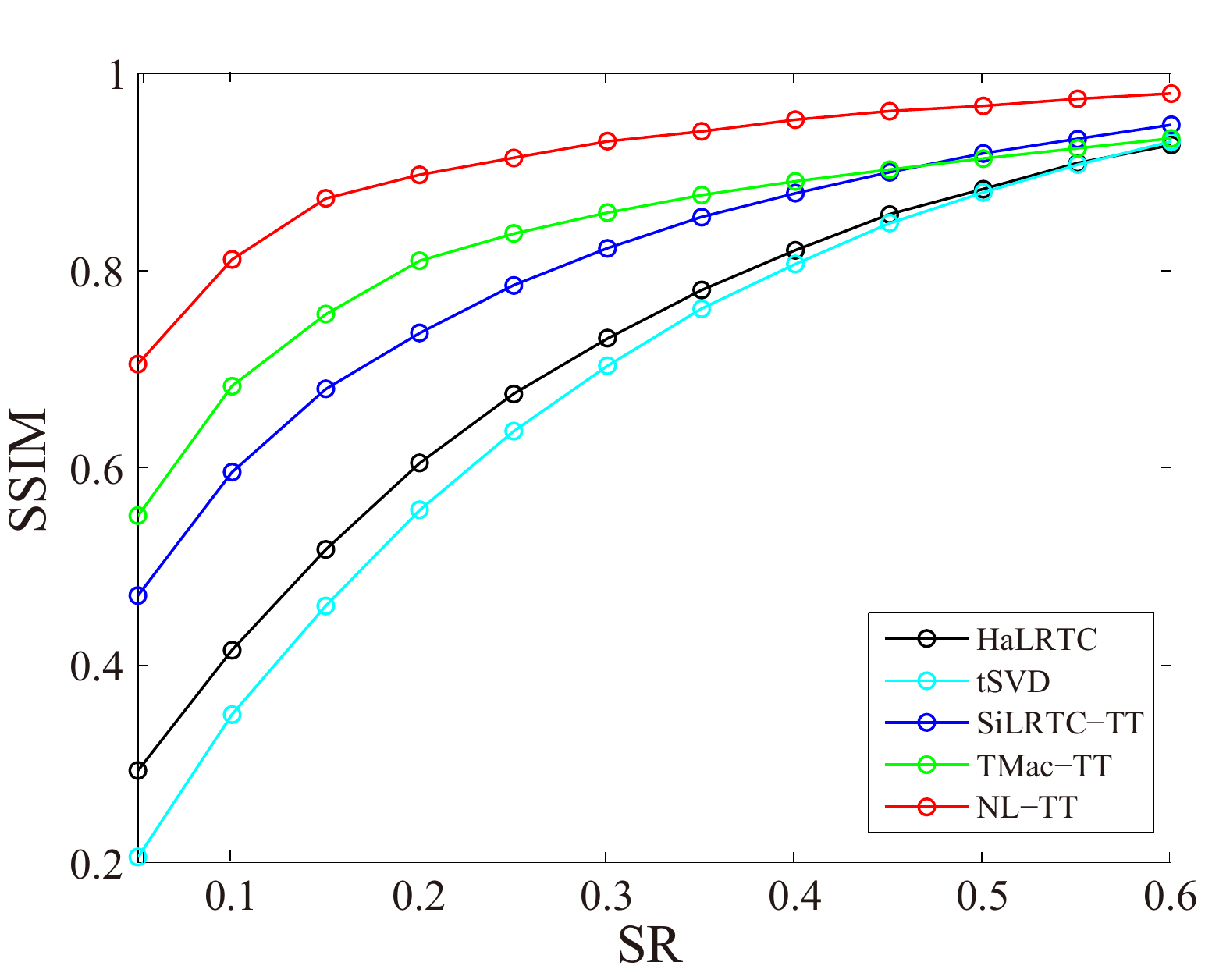}&
\includegraphics[width=0.32\textwidth]{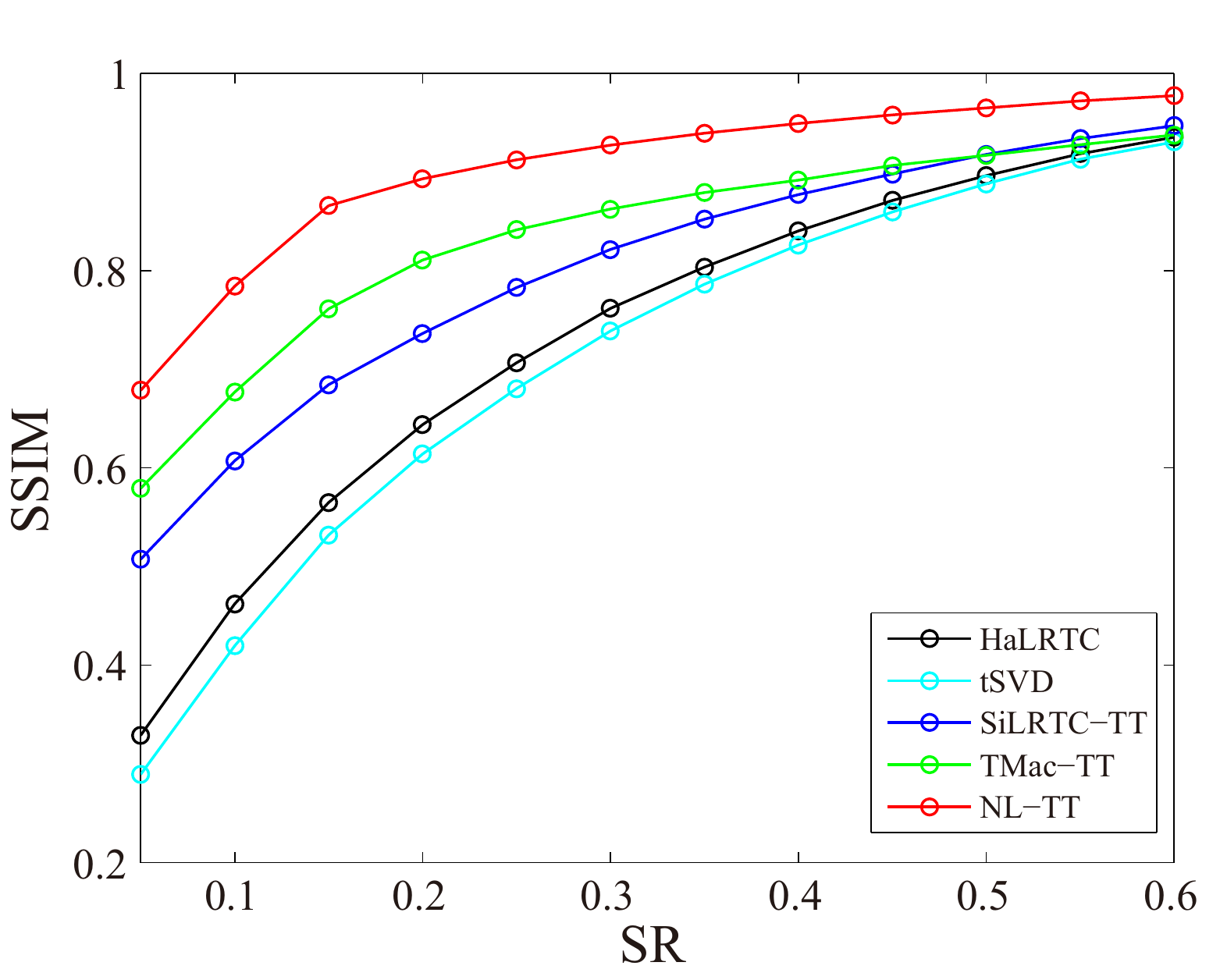}&
\includegraphics[width=0.32\textwidth]{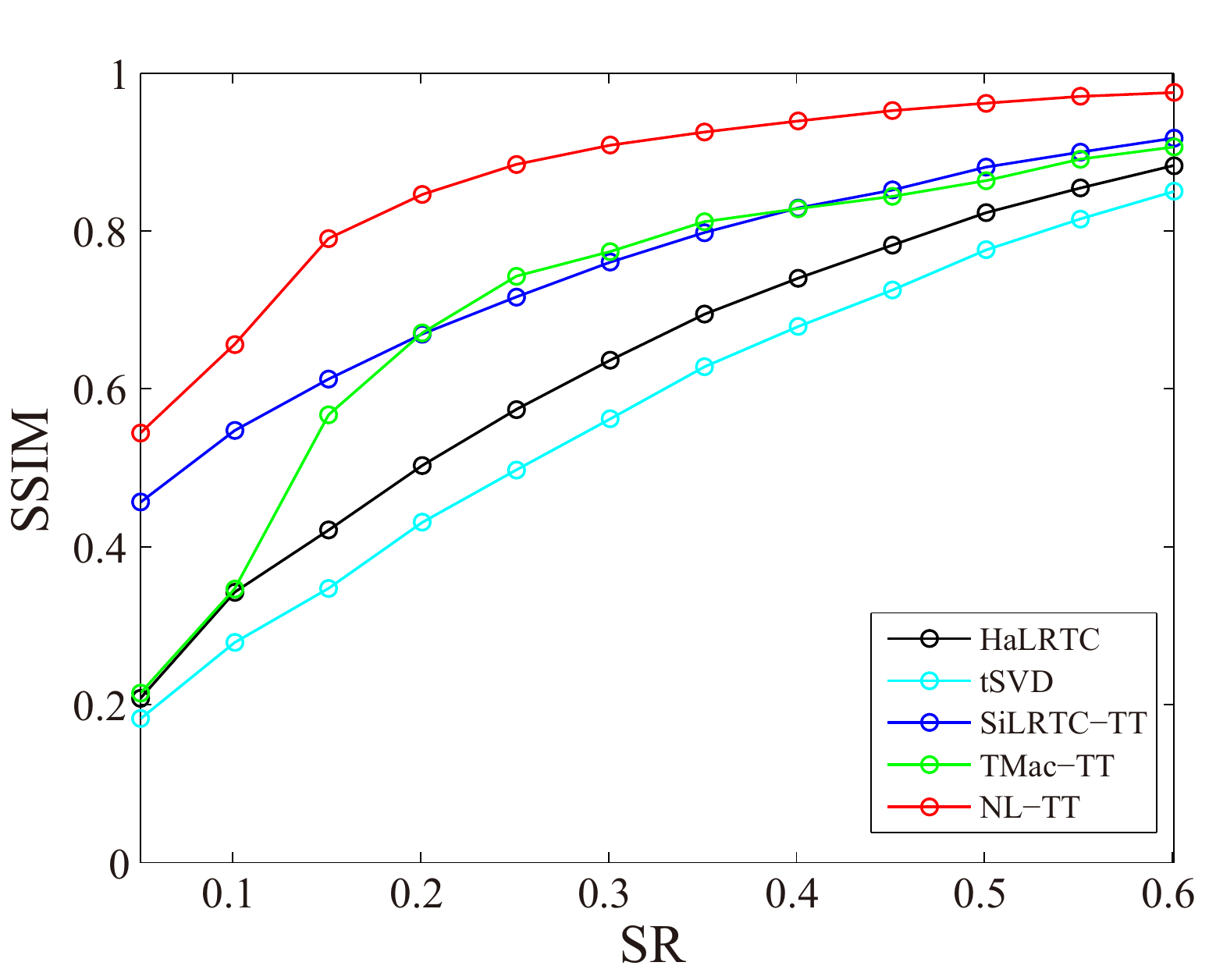} \vspace{0.1cm}\\
{\footnotesize(a)} {\small \emph{lena}} & {\footnotesize(b)} {\small \emph{airplane}} & {\footnotesize(c)} {\small \emph{monarch}}\\
\end{tabular}
\caption{\small{The PSNR and SSIM values of the reconstructed color image results for random missing entries by different methods.}}
  \label{fig:color_psnr}
  \end{center}\vspace{-0.3cm}
\end{figure}

\textbf{Random sampling and tube sampling}. The random sampling denotes that the entries in R, G, and B channels are randomly and independently lost in the color image, see the first three rows in Fig. \ref{fig:color_entry} (a). The tube sampling means that the entries are randomly lost at the same location in R, G, and B channels, see the second three rows in Fig. \ref{fig:color_entry} (a). The task for recovering the tube sampling is harder than the random sampling. Fig. \ref{fig:color_entry} shows the visually restored results recovered by HaLRTC, tSVD, SiLRTC-TT, TMac-TT, and NL-TT. The first three lines are random sampling with $SR=0.2$. The second three lines are tube sampling with $SR=0.2$. We observe that the results recovered by both HaLRTC and tSVD have undesired thorns. Although SiLRTC-TT and TMac-TT obtain much better results than HaLRTC and tSVD, some block-artifacts are created on the restored images. As a comparison, the recovered results by the proposed method are visually better than those of the compared methods. From the zoom-in regions of recovered images, we observe that NL-TT can efficiently keep the details and smoothness of images and reduce the block-artifacts compared with SiLRTC-TT and TMac-TT.

Table \ref{tab:color_random} lists the PSNR and SSIM values of the restored images by all compared methods on different SRs. The highest results for each quality index are labeled in bold. Fig. \ref{fig:color_psnr} shows the recovery PSNR and SSIM curves by all compared methods with SRs tested from 0.05 to 0.6. It is observed that for different SRs, the proposed method achieves the highest PSNR and SSIM values.

\textbf{Structural missing pixels}. We test five kinds of structural missing pixels, i.e., missing random curves for image \emph{house}, missing random vertical and horizontal slices for image \emph{facade}, missing texts for image \emph{sailboat}, and missing regular and random blacks for images \emph{barbara} and \emph{peppers}, respectively.

\begin{figure*}[!ht]
\scriptsize\setlength{\tabcolsep}{0.9pt}
\begin{center}
\begin{tabular}{cccccccc}
\includegraphics[width=0.14\textwidth]{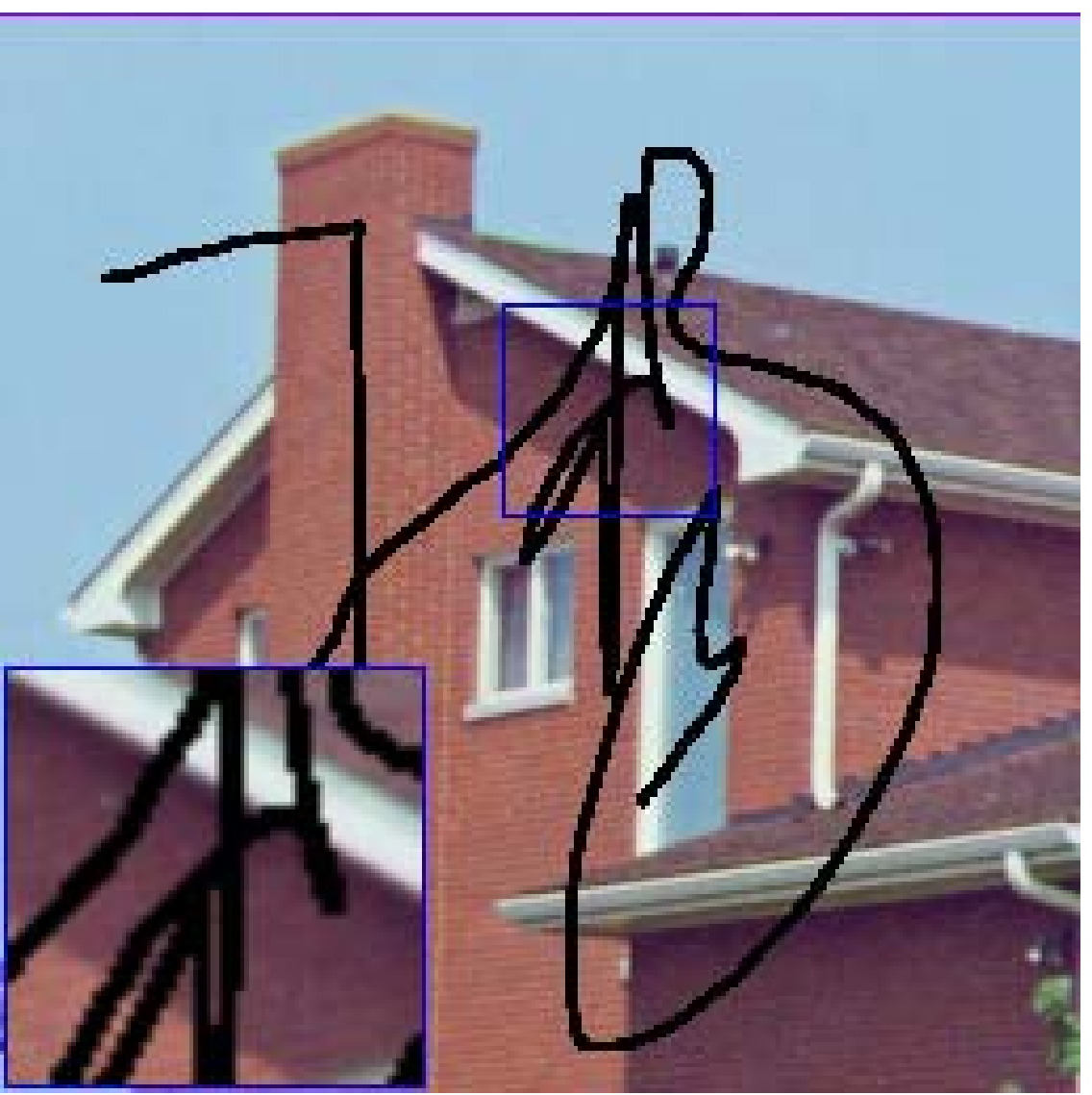}&
\includegraphics[width=0.14\textwidth]{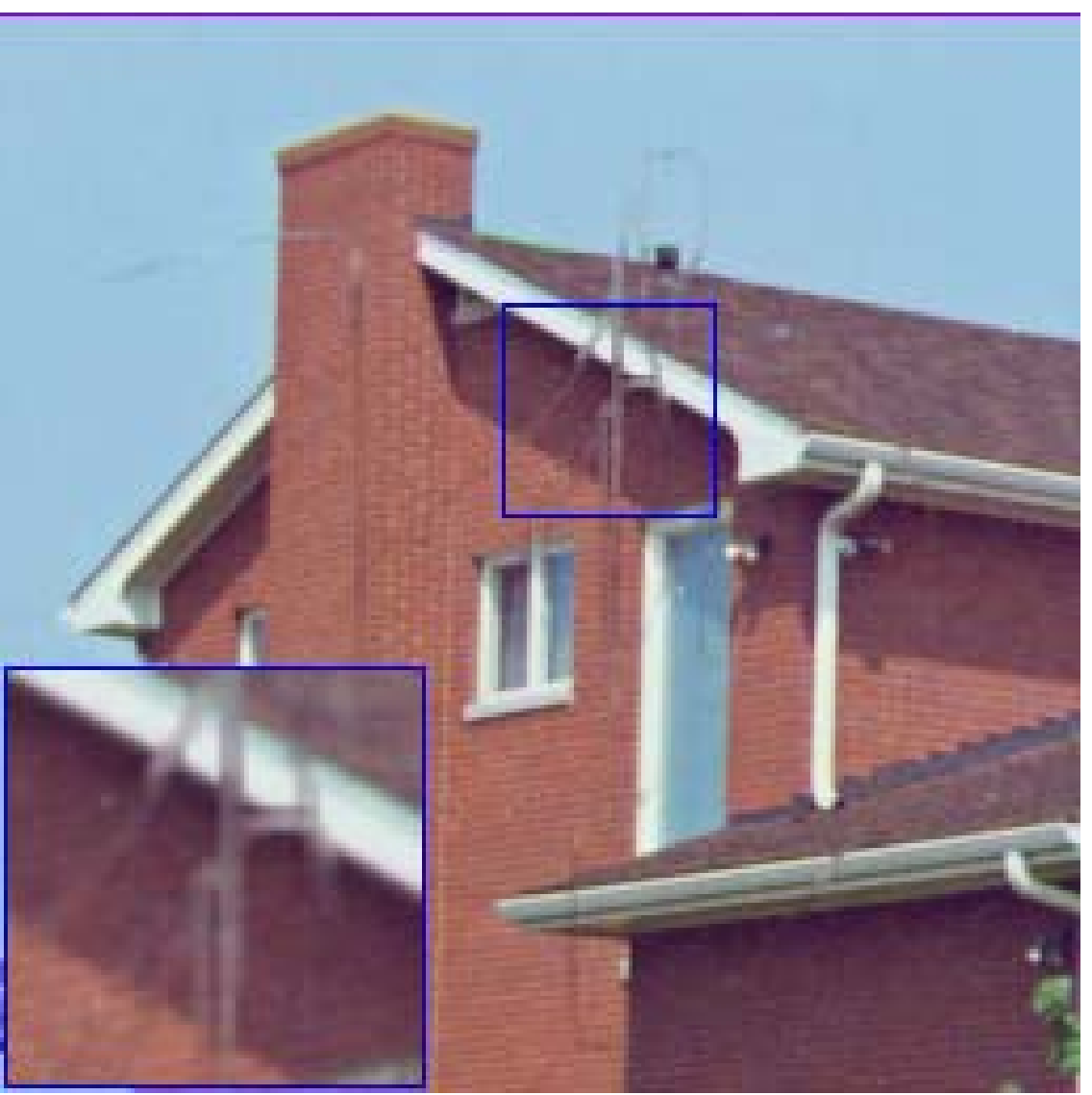}&
\includegraphics[width=0.14\textwidth]{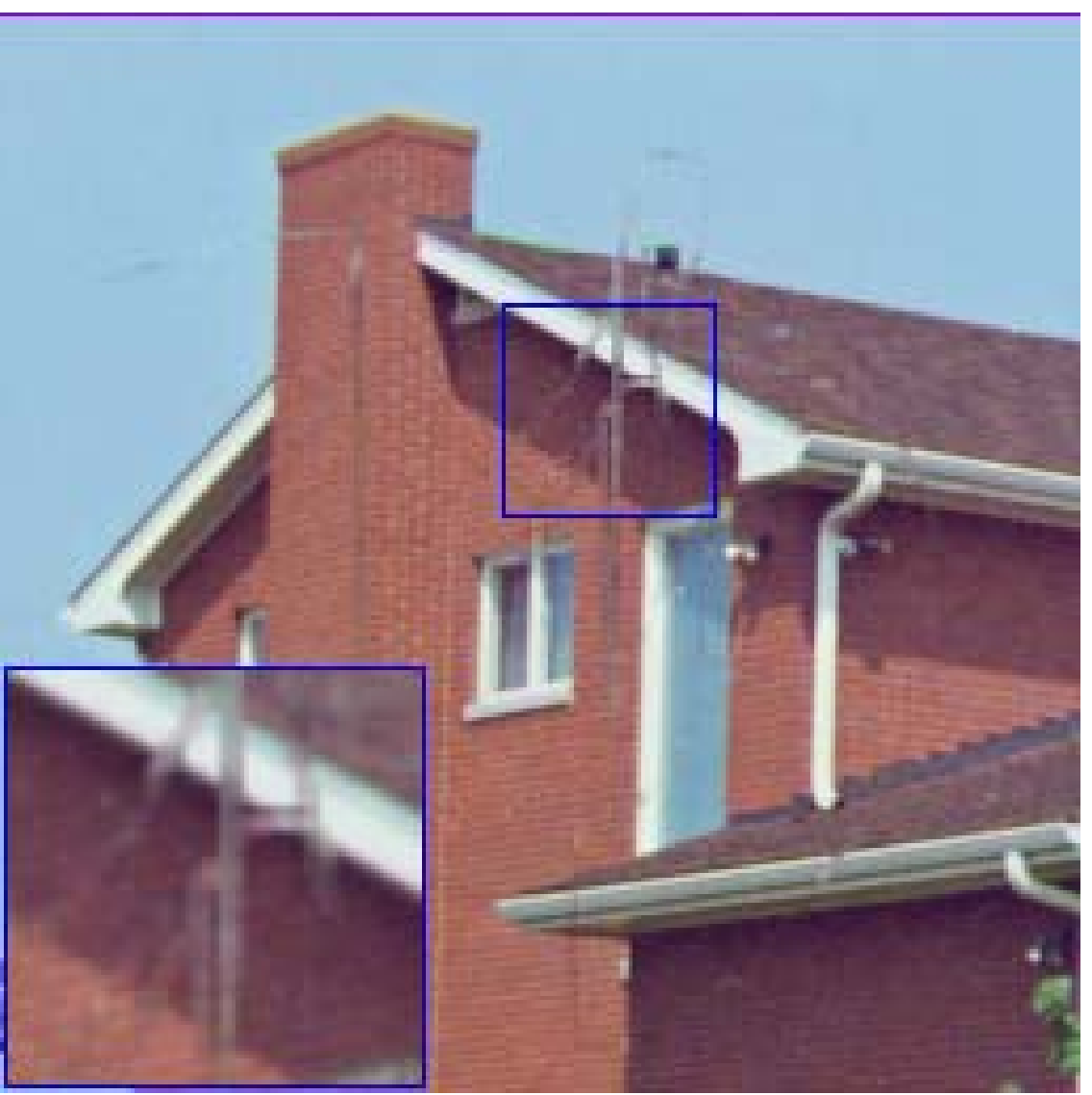}&
\includegraphics[width=0.14\textwidth]{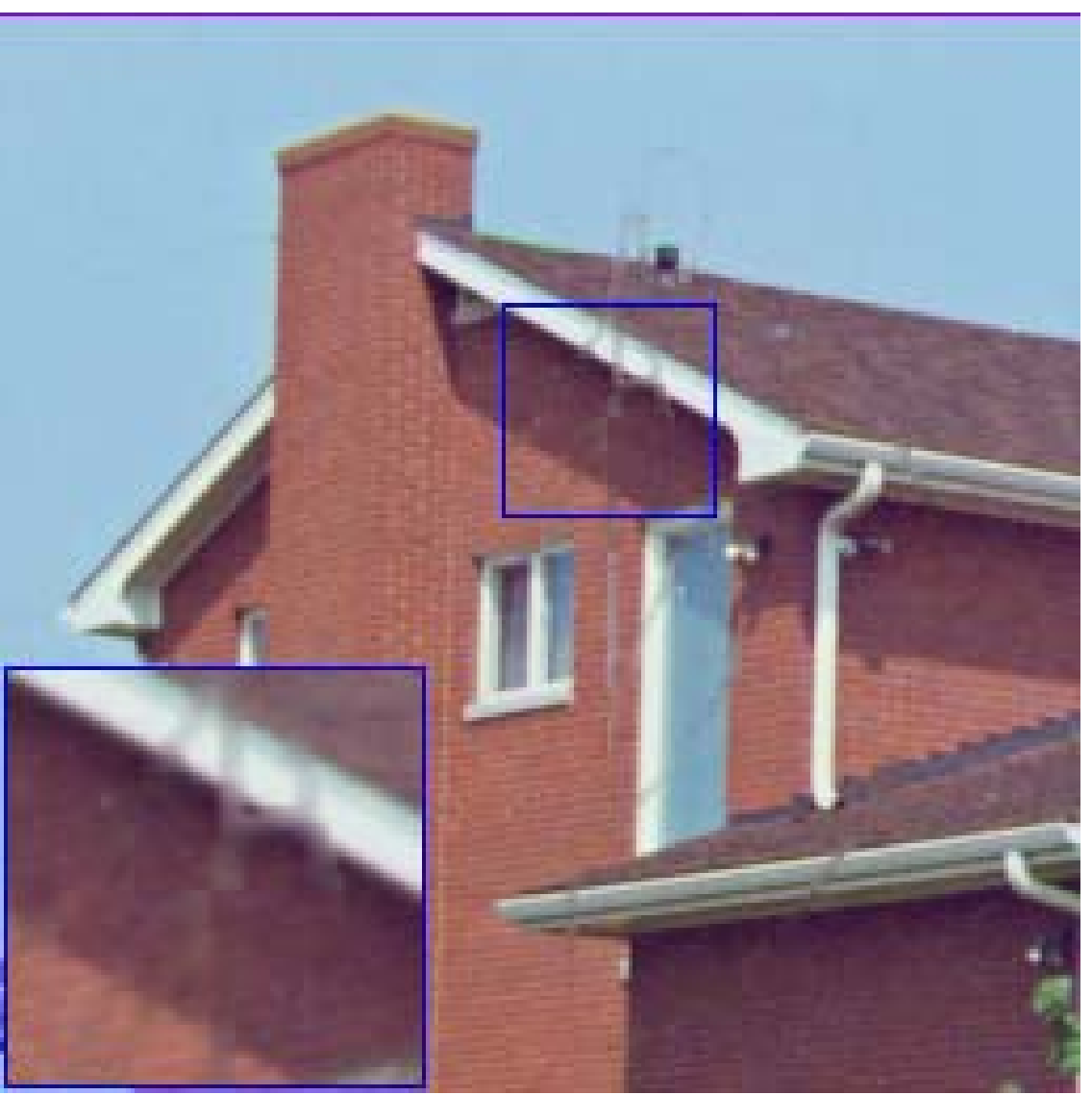}&
\includegraphics[width=0.14\textwidth]{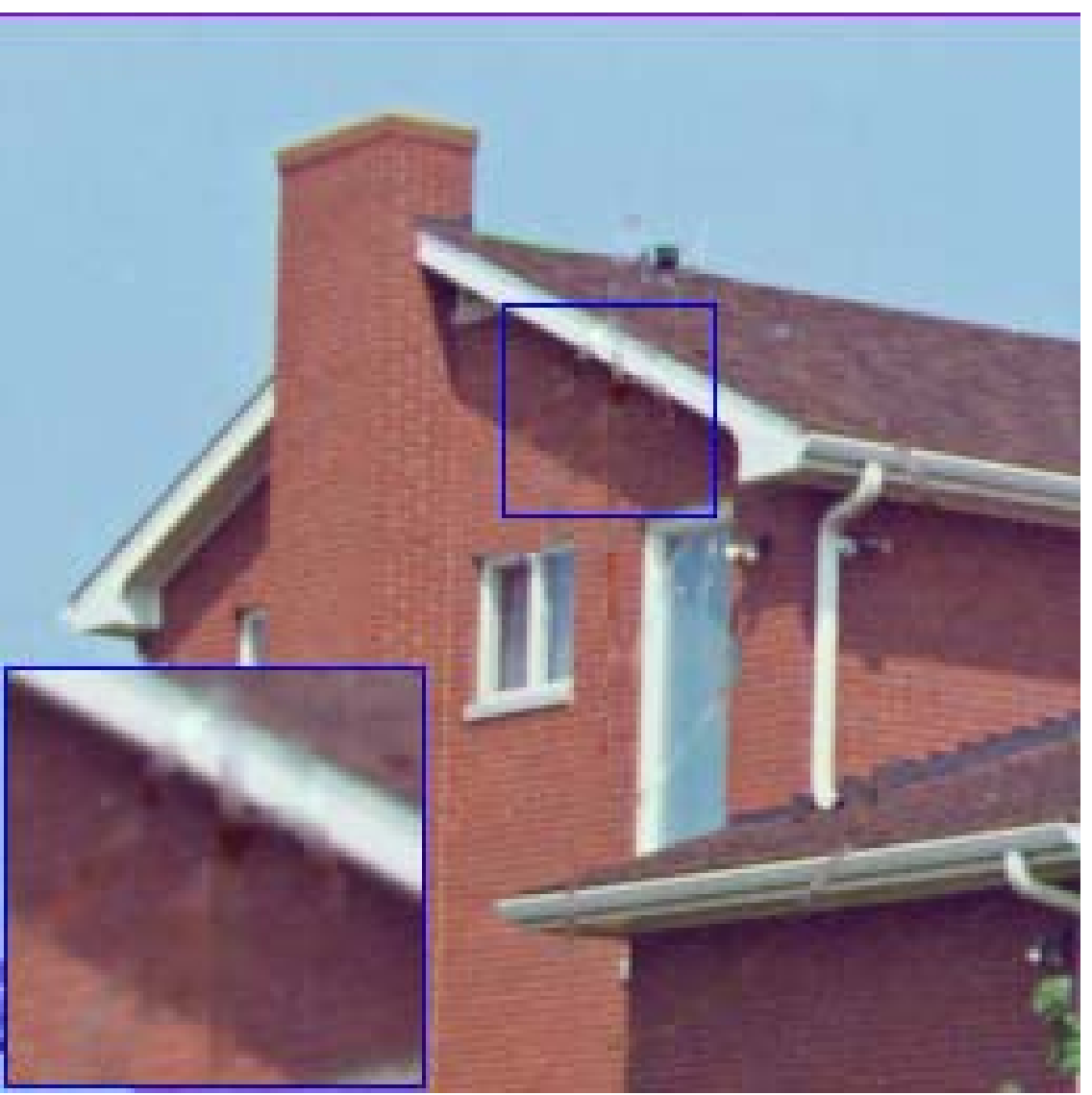}&
\includegraphics[width=0.14\textwidth]{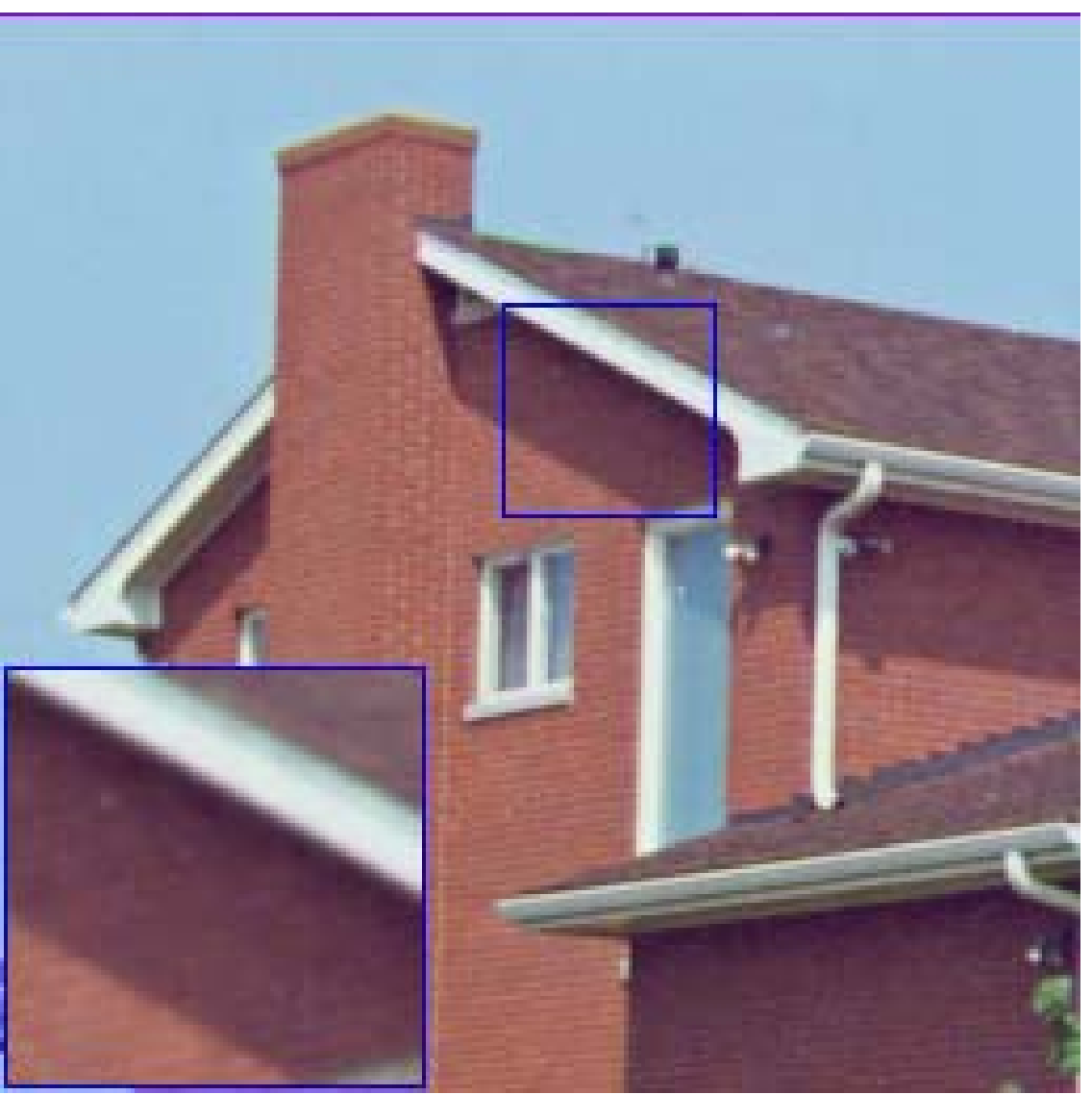}&
\includegraphics[width=0.14\textwidth]{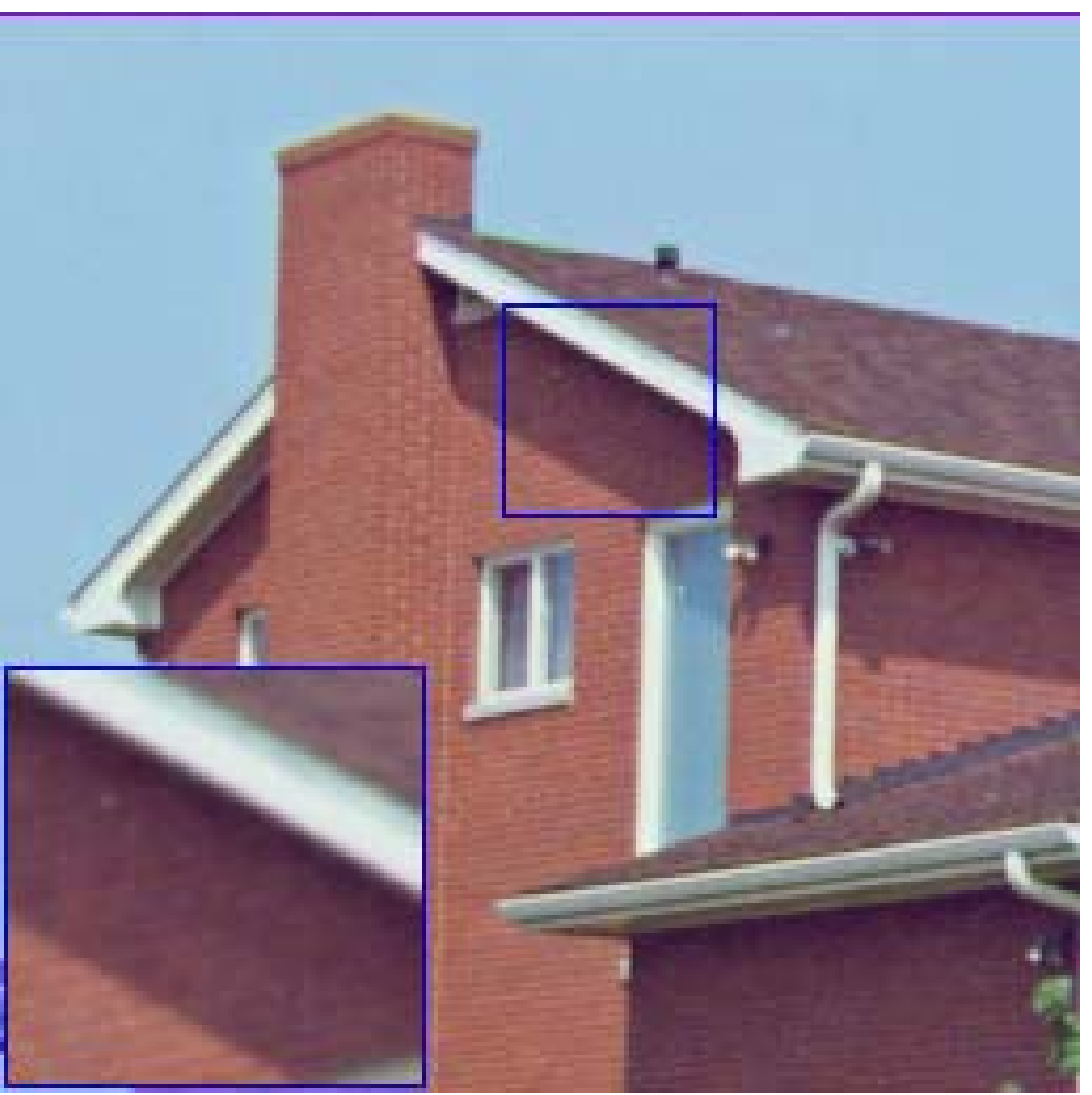}\\
\includegraphics[width=0.14\textwidth]{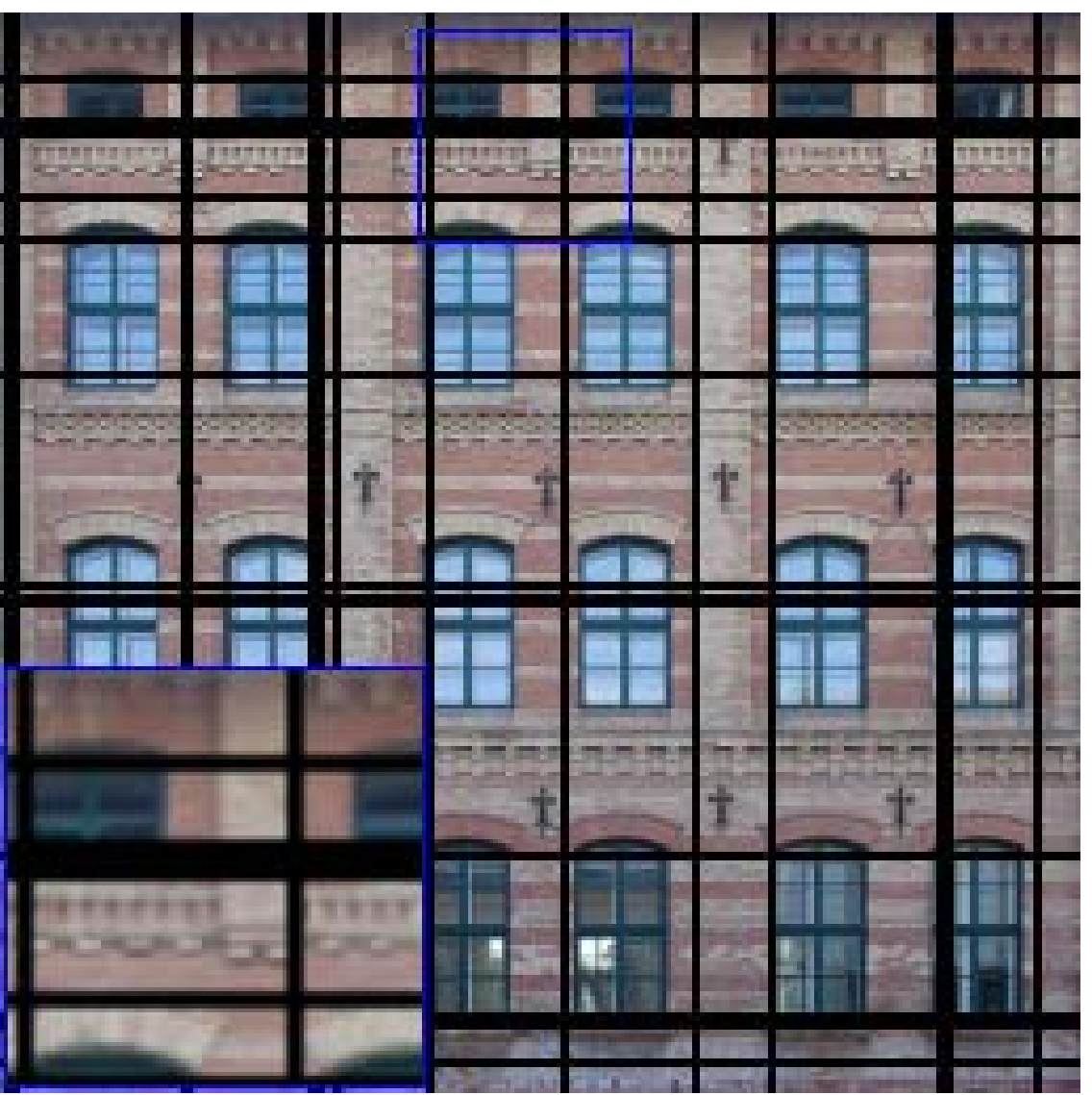}&
\includegraphics[width=0.14\textwidth]{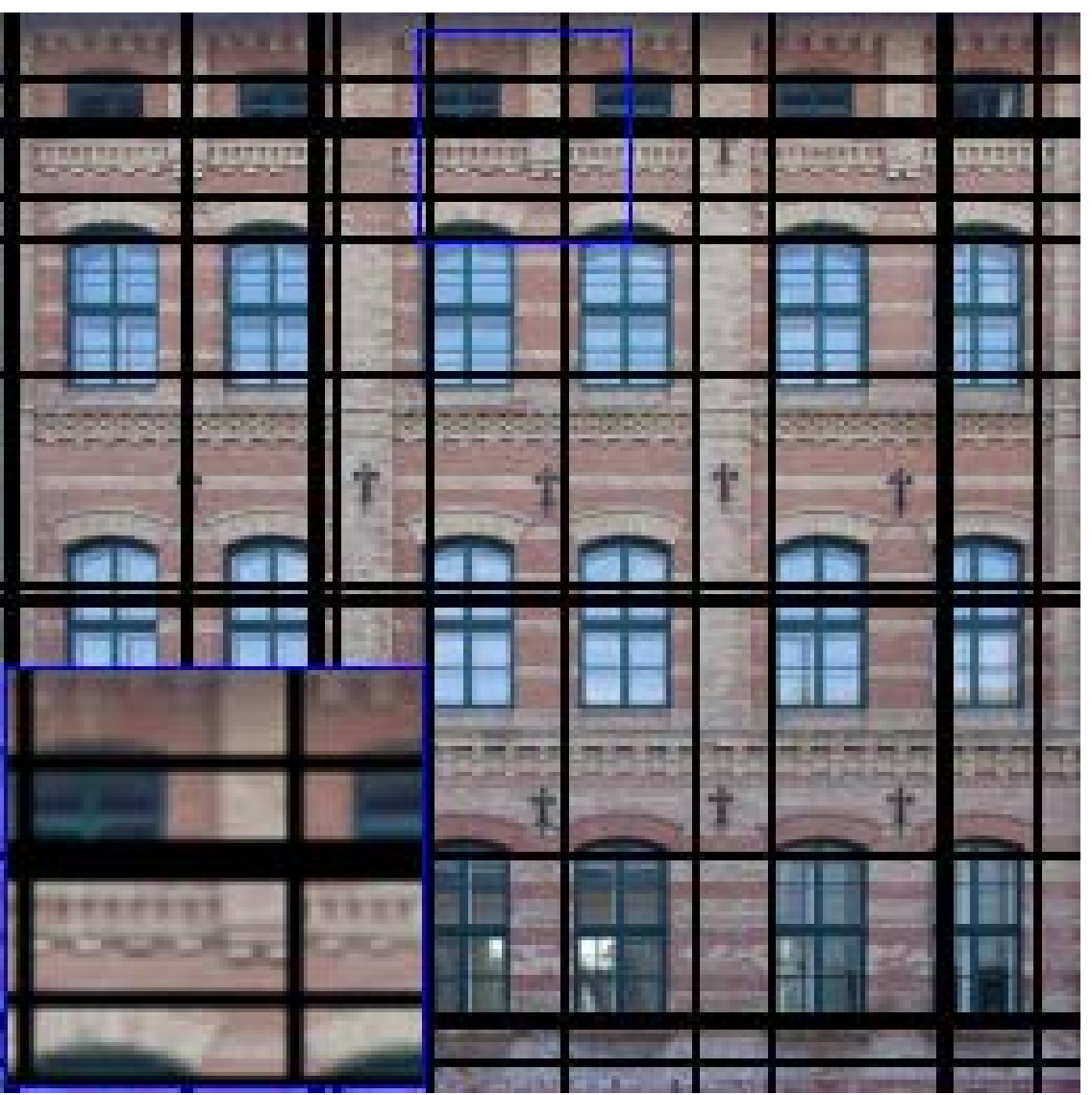}&
\includegraphics[width=0.14\textwidth]{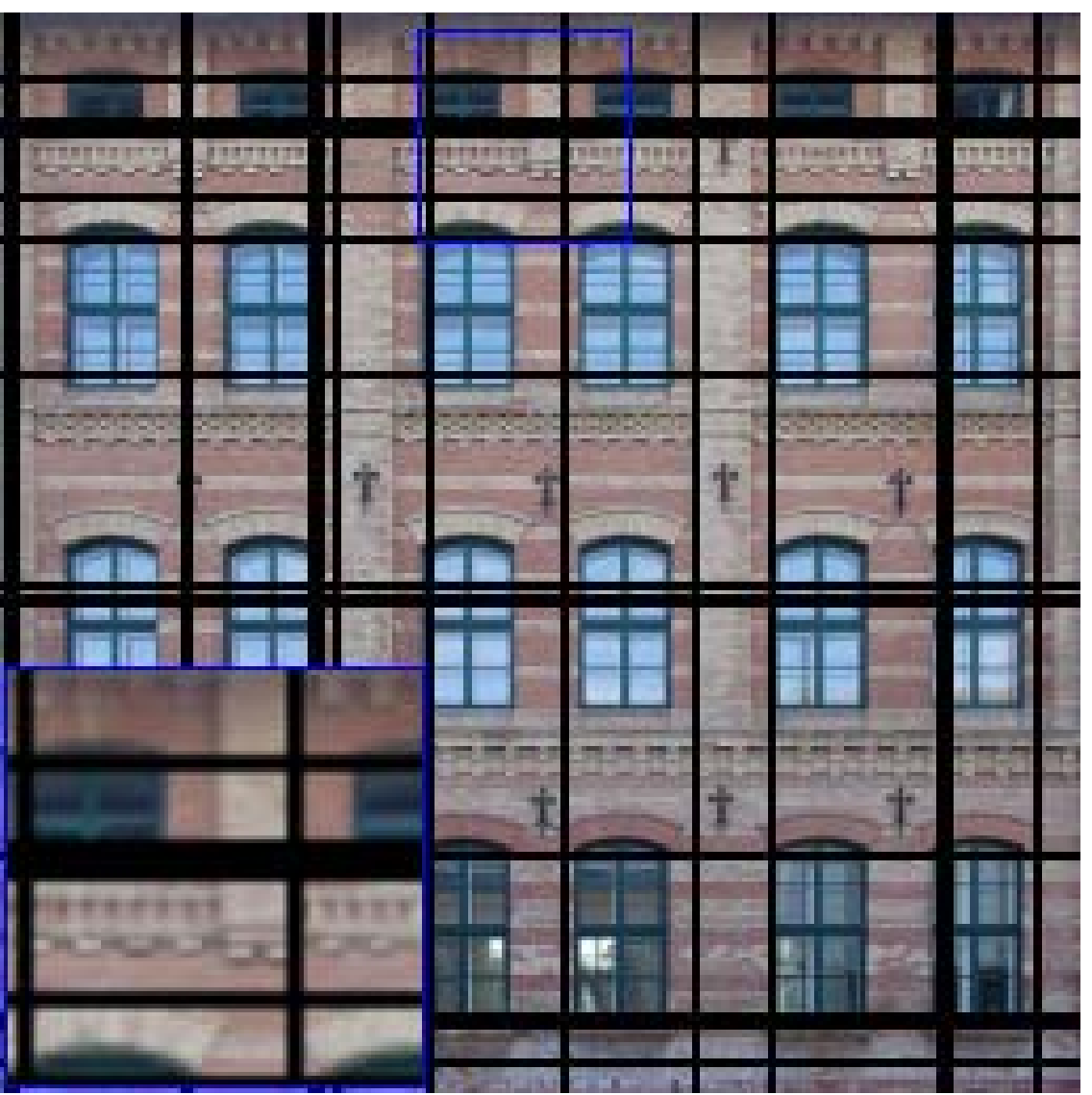}&
\includegraphics[width=0.14\textwidth]{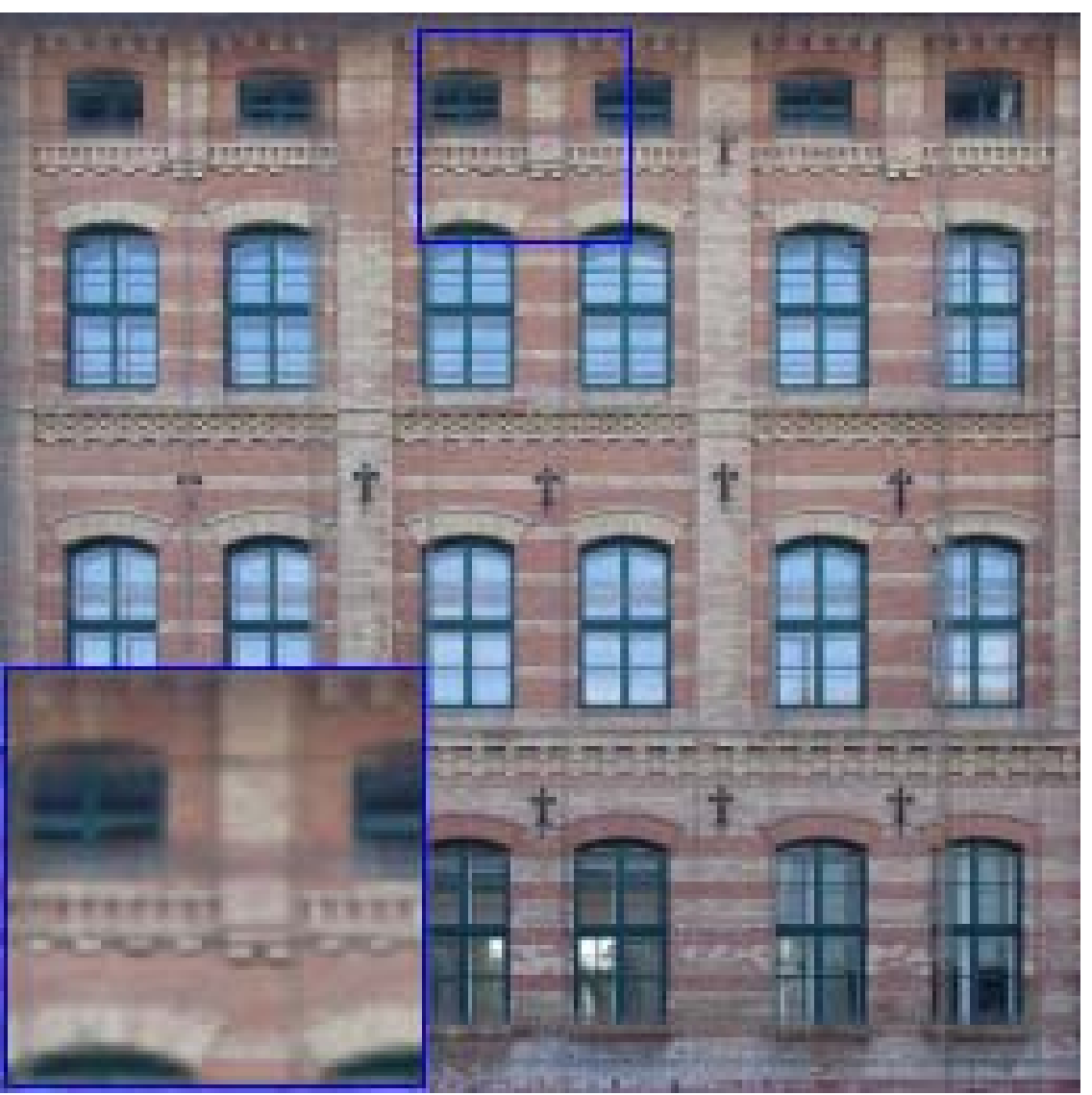}&
\includegraphics[width=0.14\textwidth]{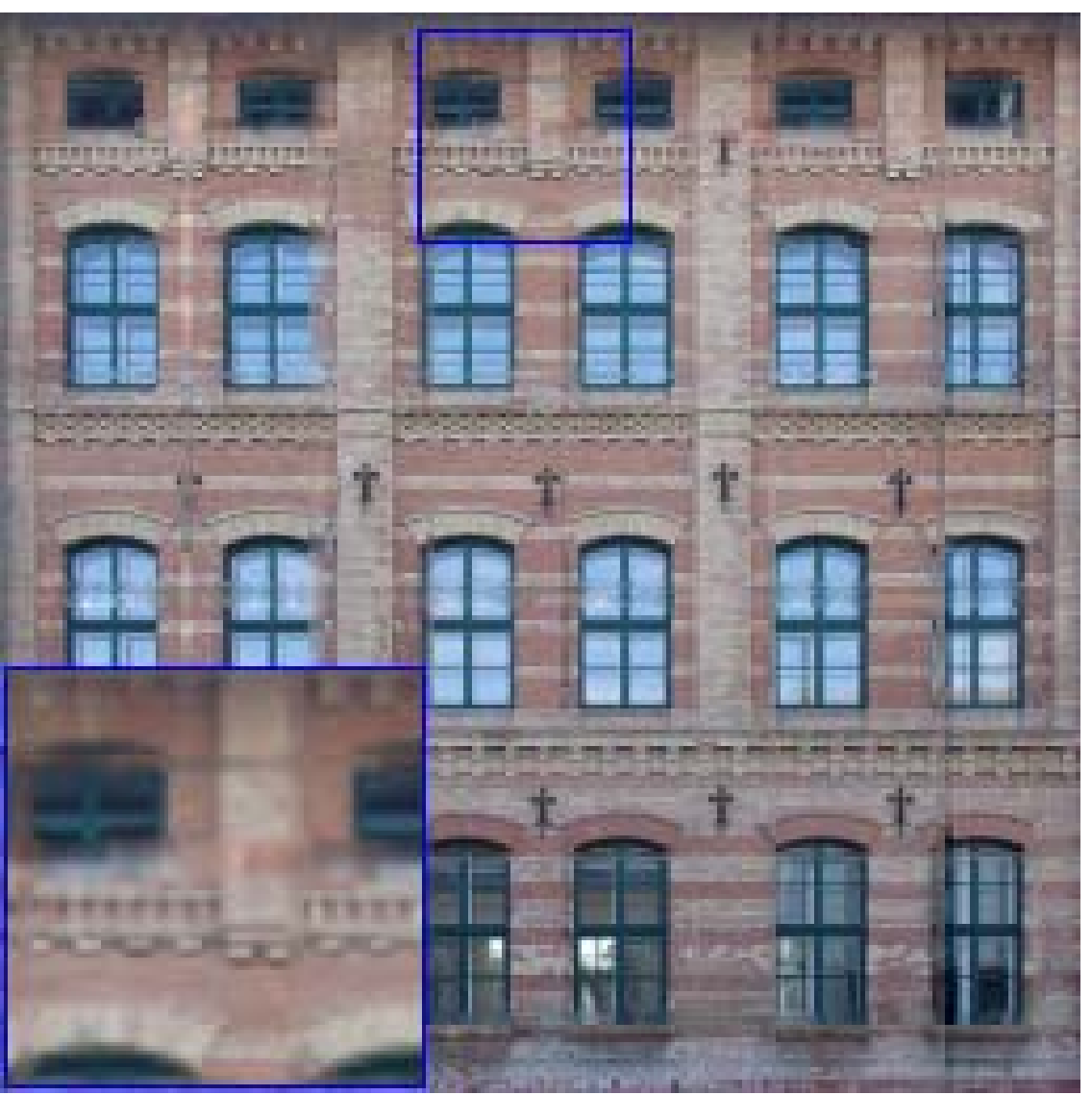}&
\includegraphics[width=0.14\textwidth]{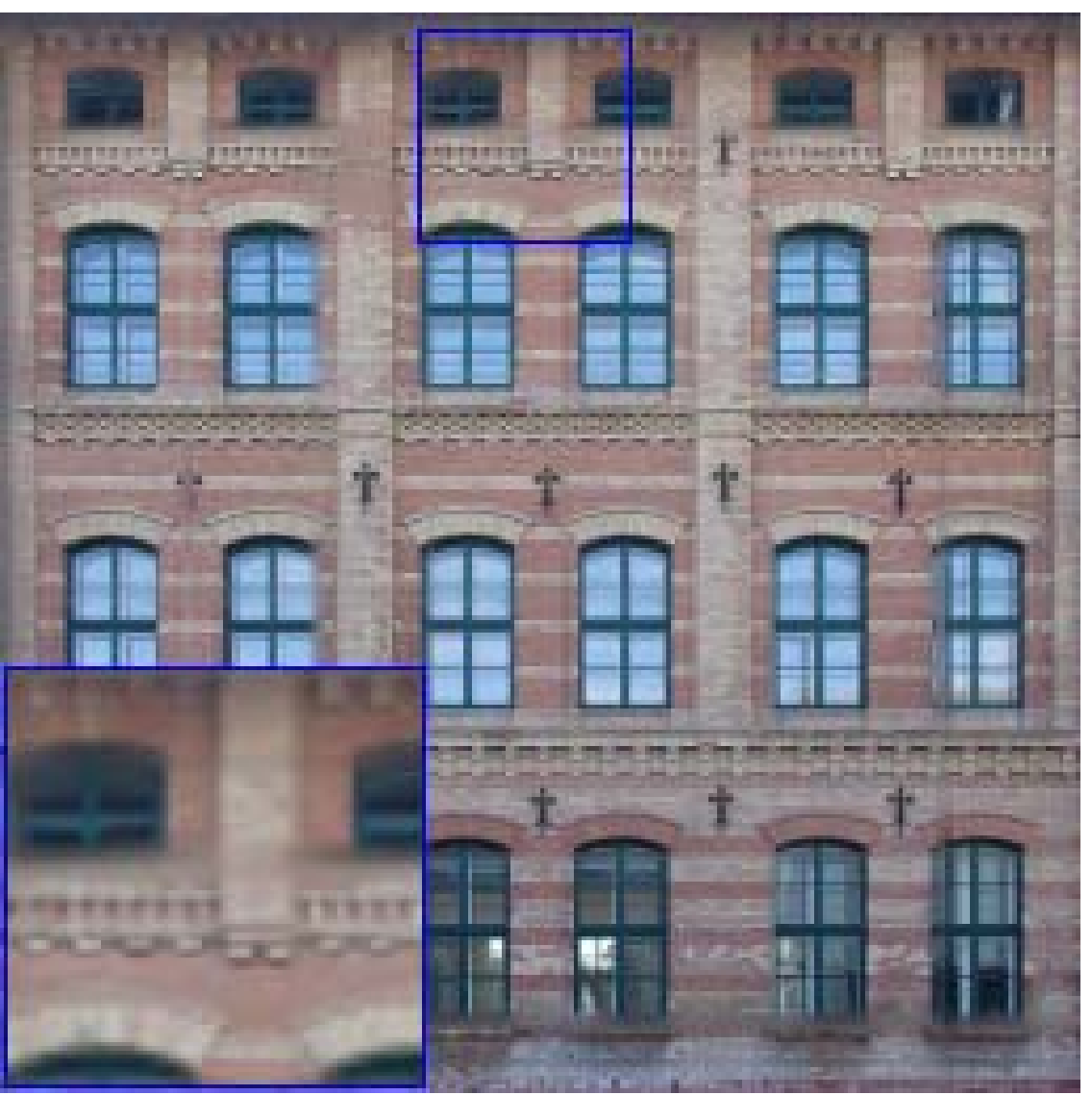}&
\includegraphics[width=0.14\textwidth]{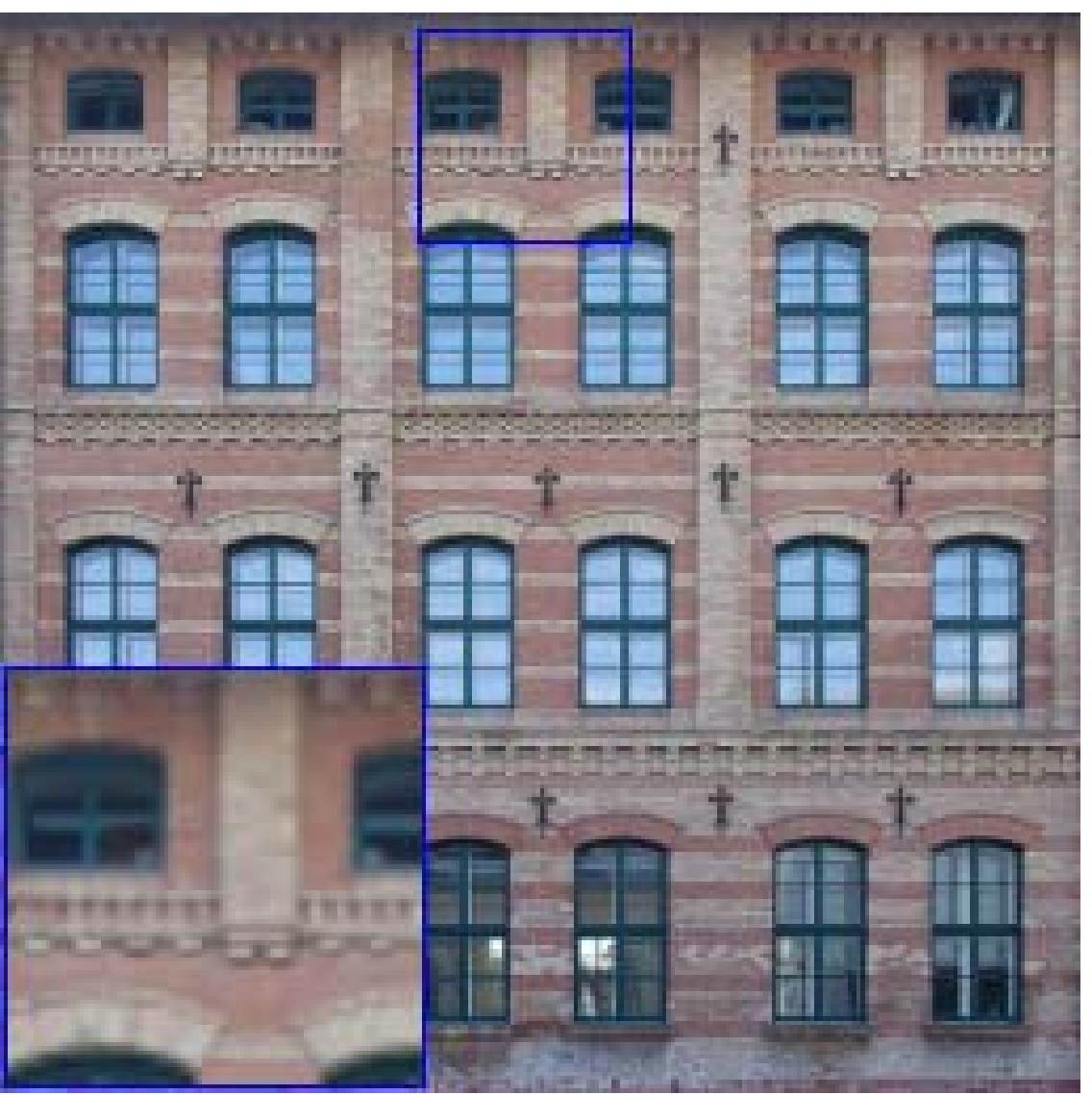}\\
\includegraphics[width=0.14\textwidth]{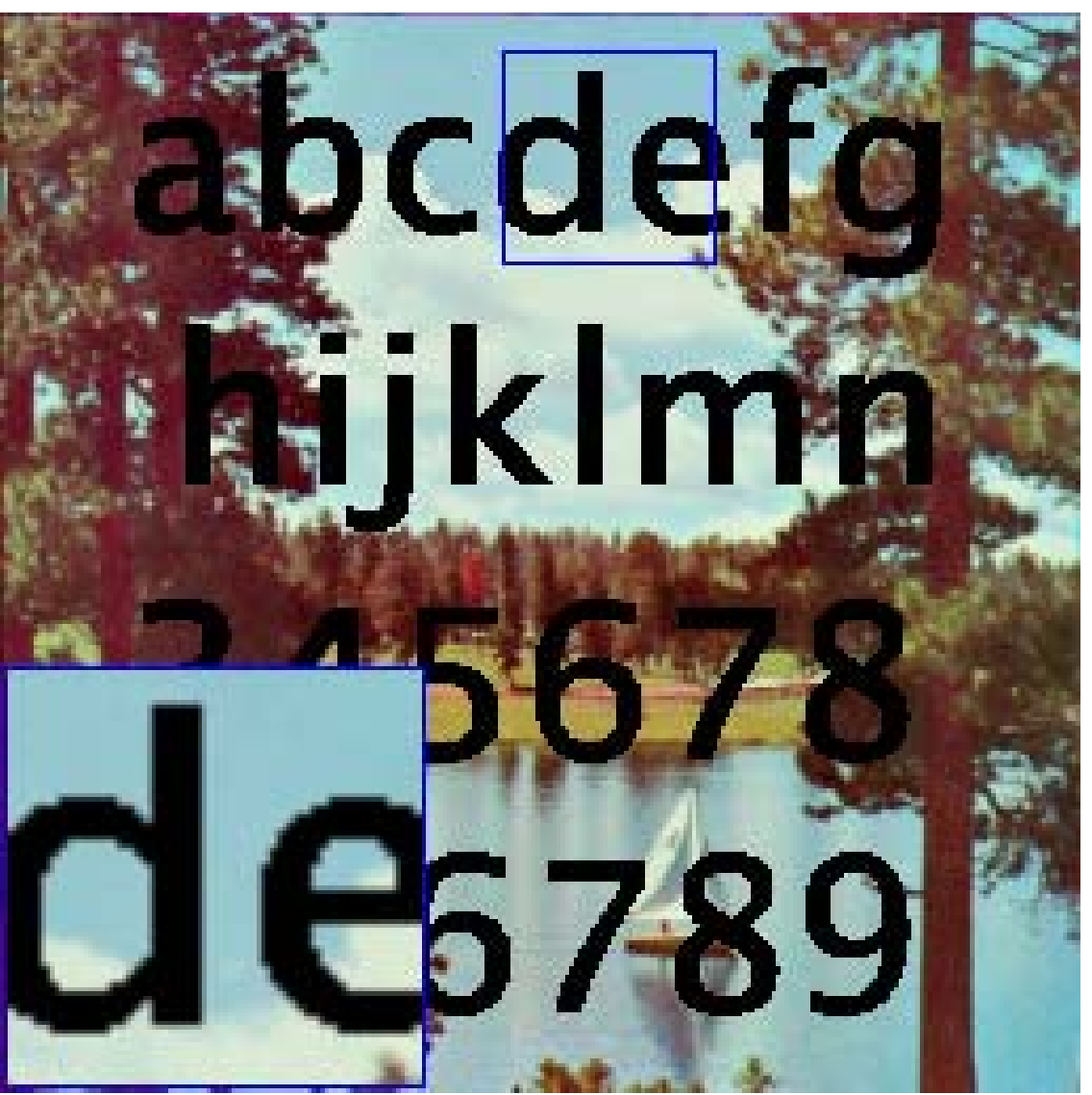}&
\includegraphics[width=0.14\textwidth]{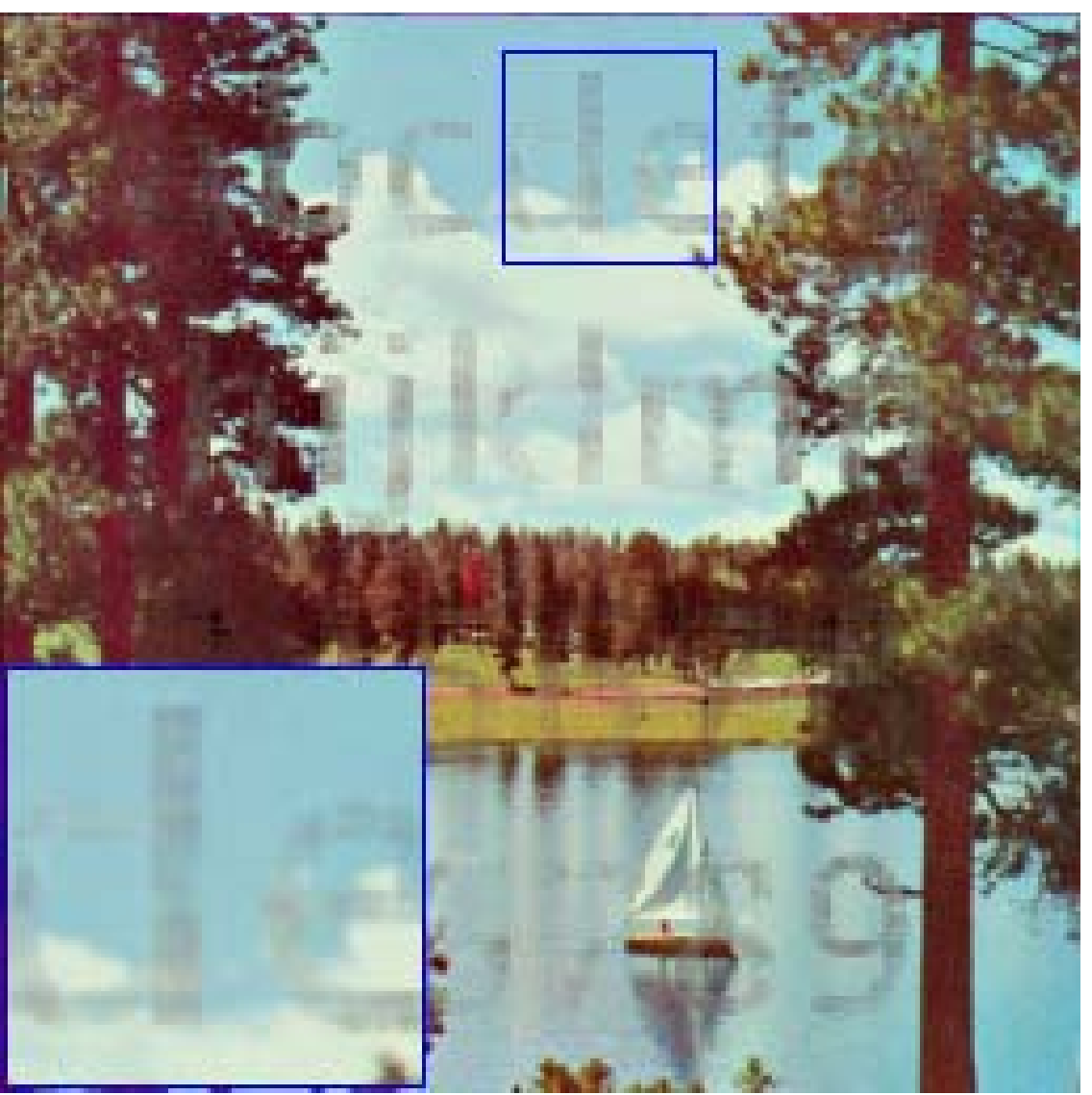}&
\includegraphics[width=0.14\textwidth]{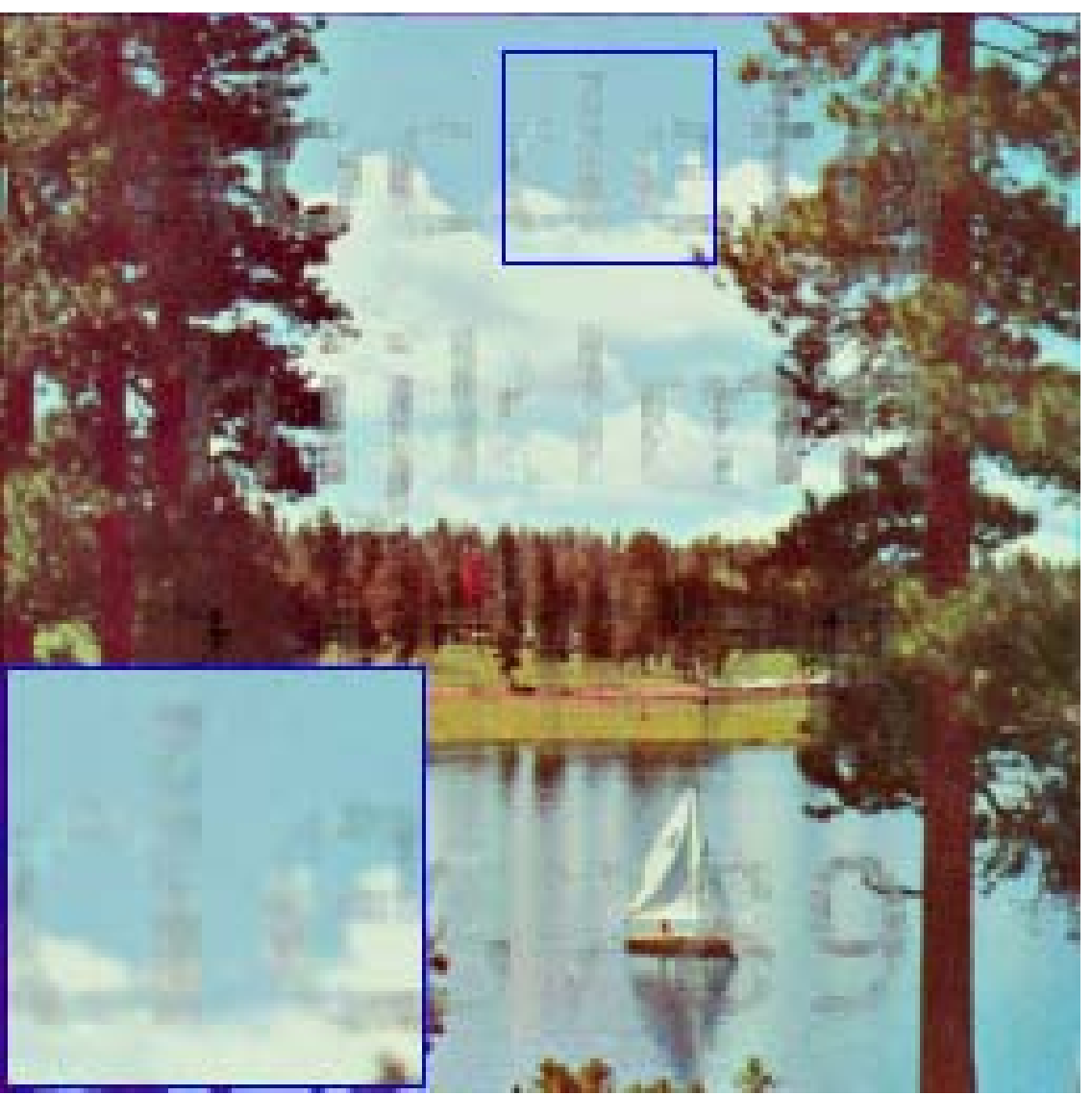}&
\includegraphics[width=0.14\textwidth]{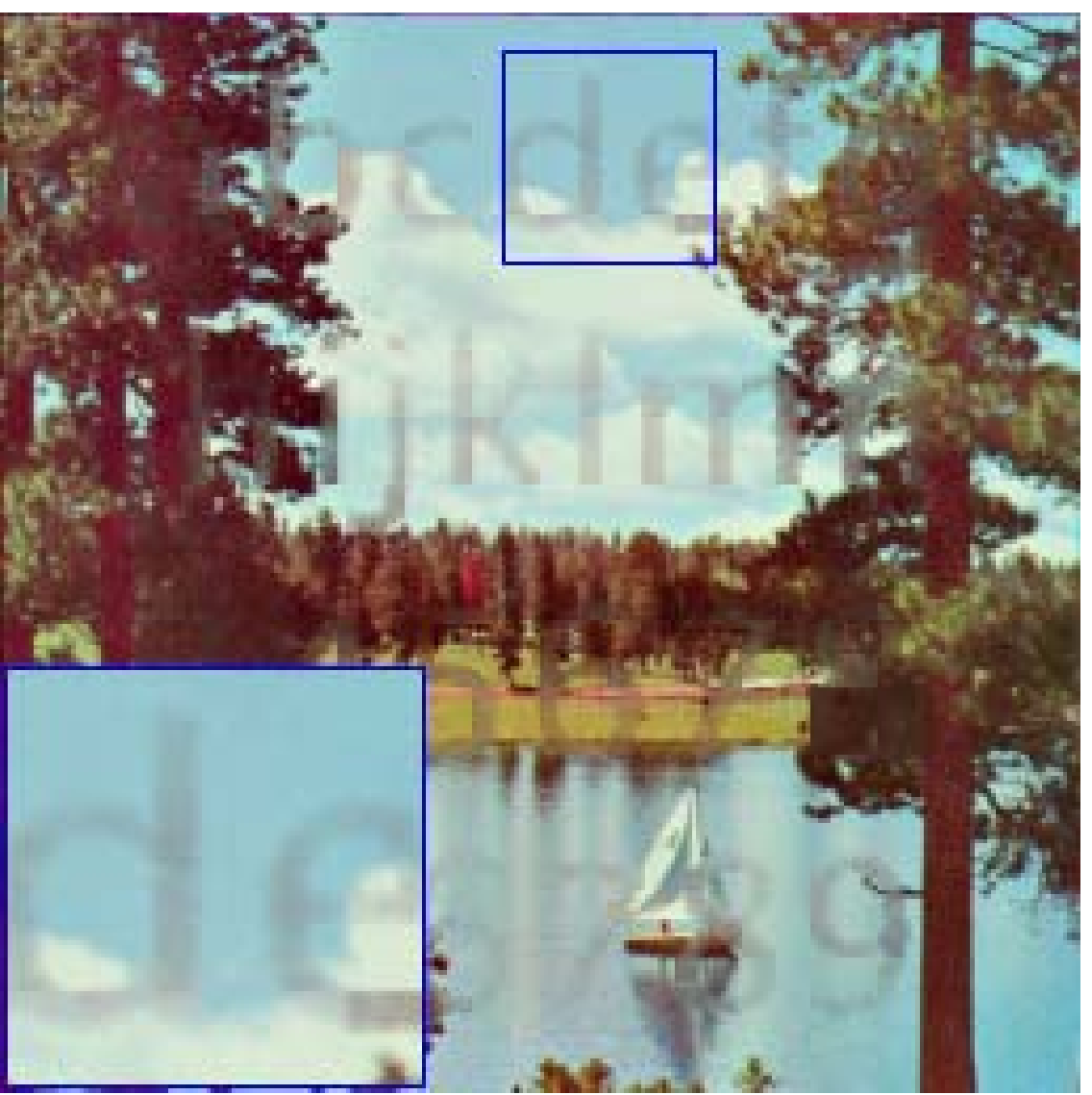}&
\includegraphics[width=0.14\textwidth]{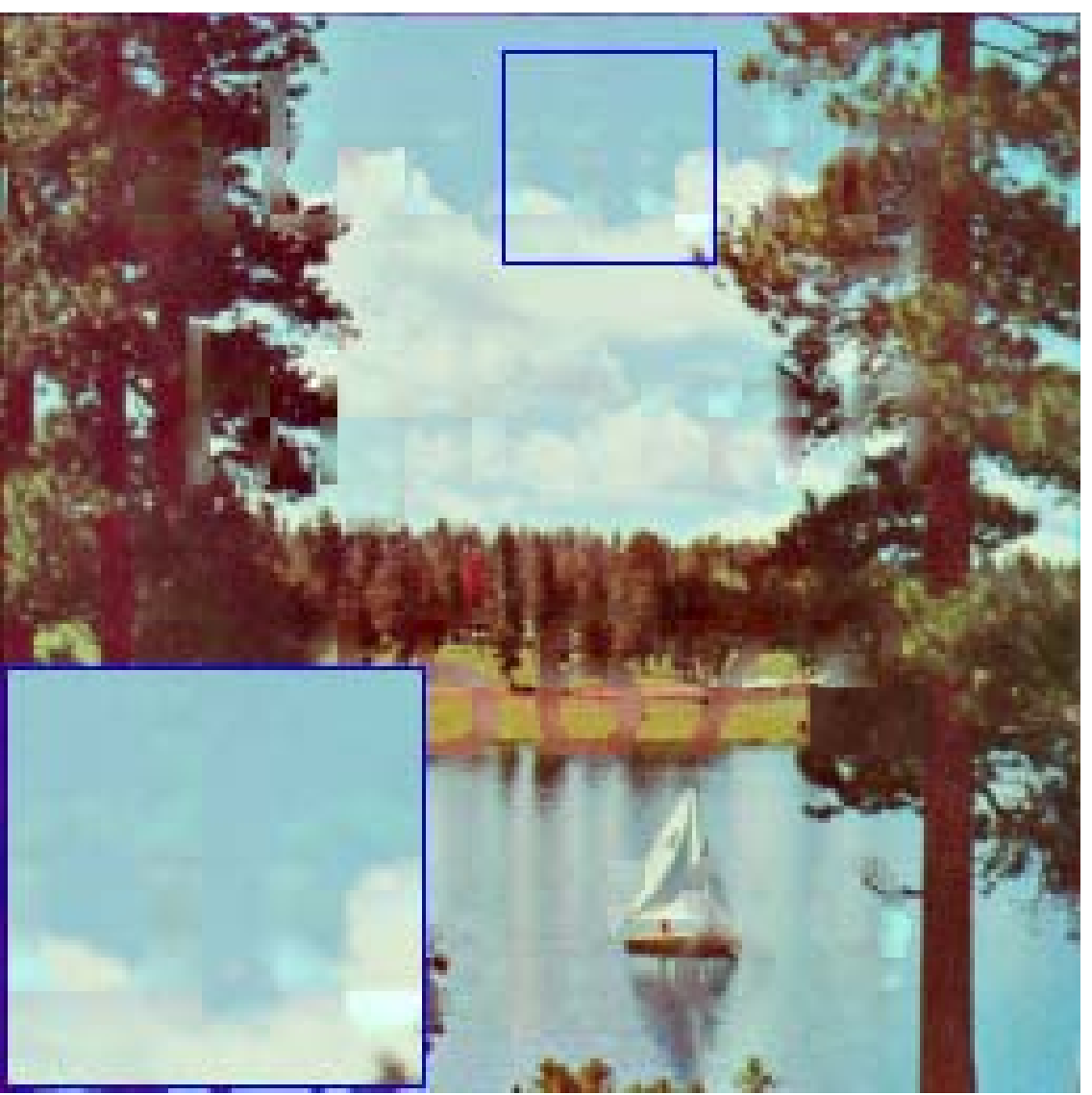}&
\includegraphics[width=0.14\textwidth]{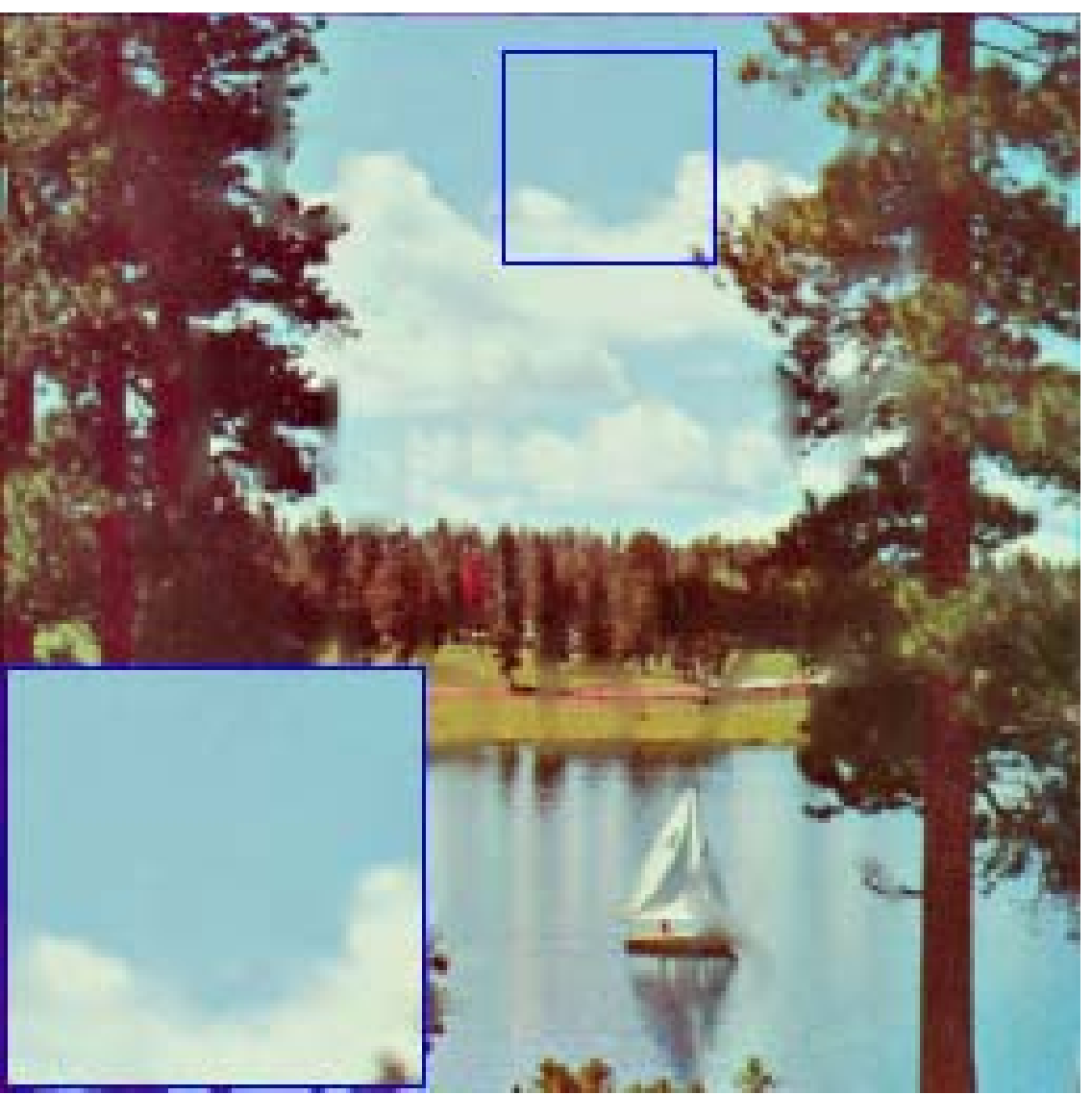}&
\includegraphics[width=0.14\textwidth]{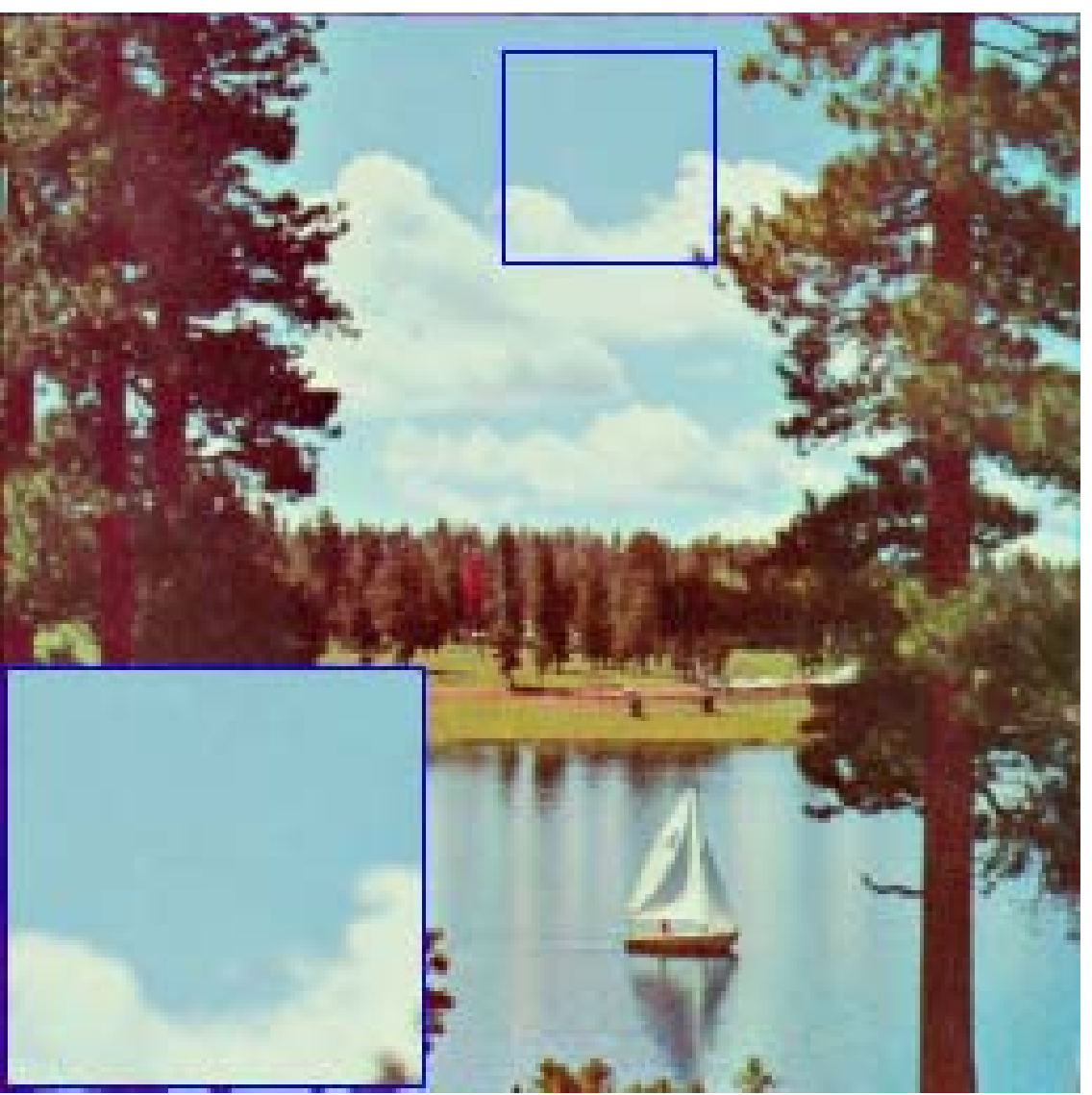}\\
\includegraphics[width=0.14\textwidth]{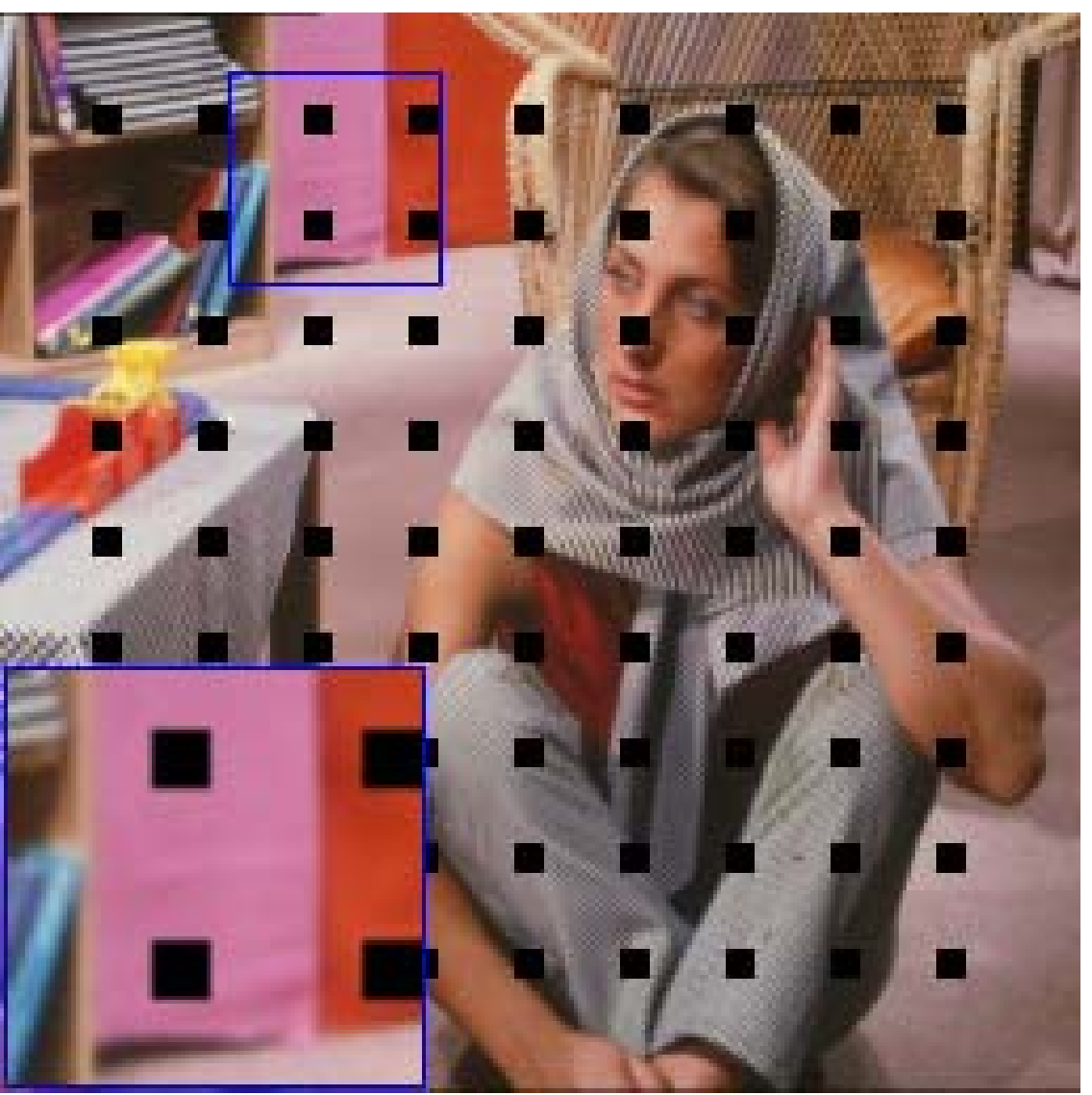}&
\includegraphics[width=0.14\textwidth]{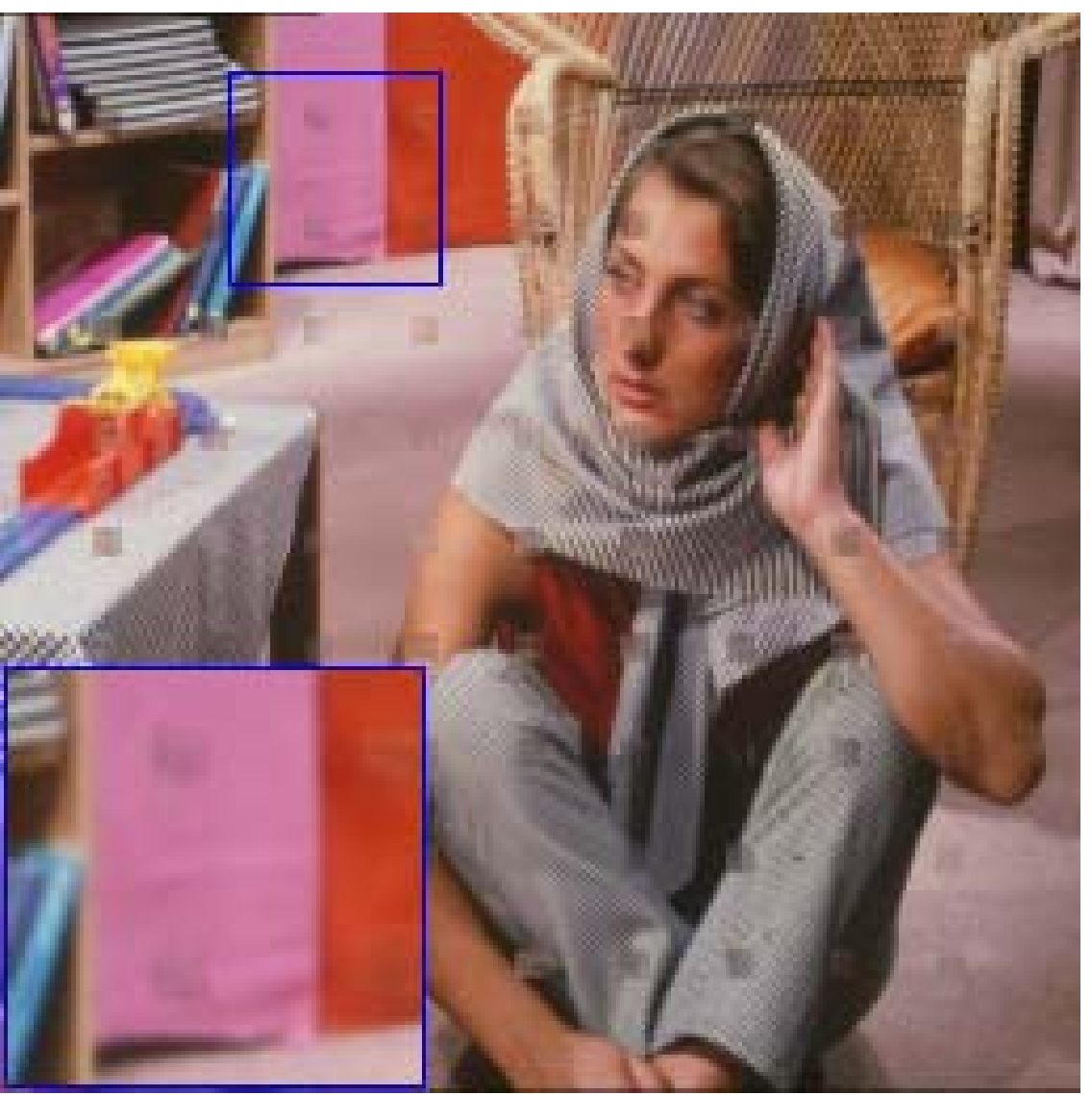}&
\includegraphics[width=0.14\textwidth]{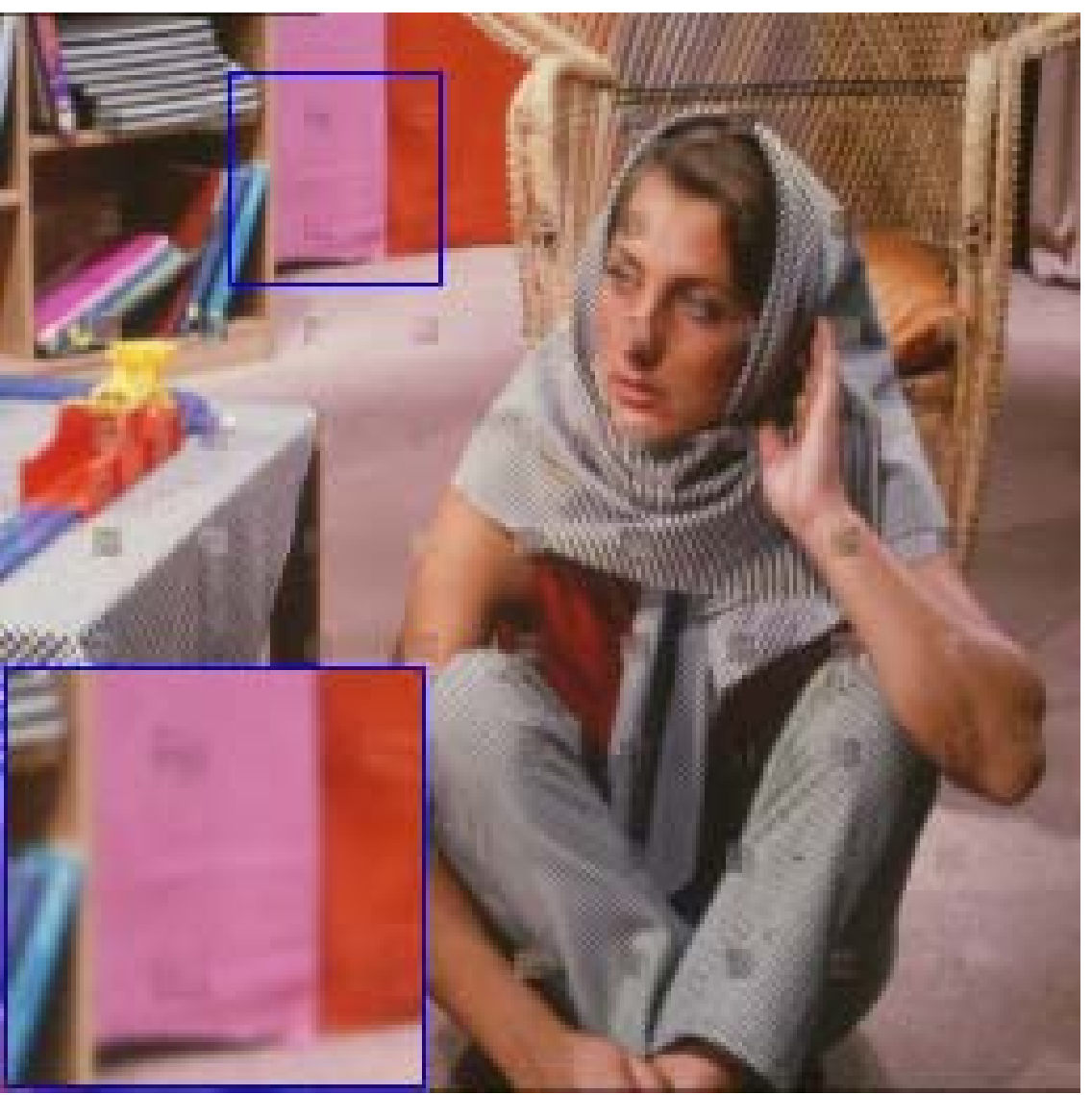}&
\includegraphics[width=0.14\textwidth]{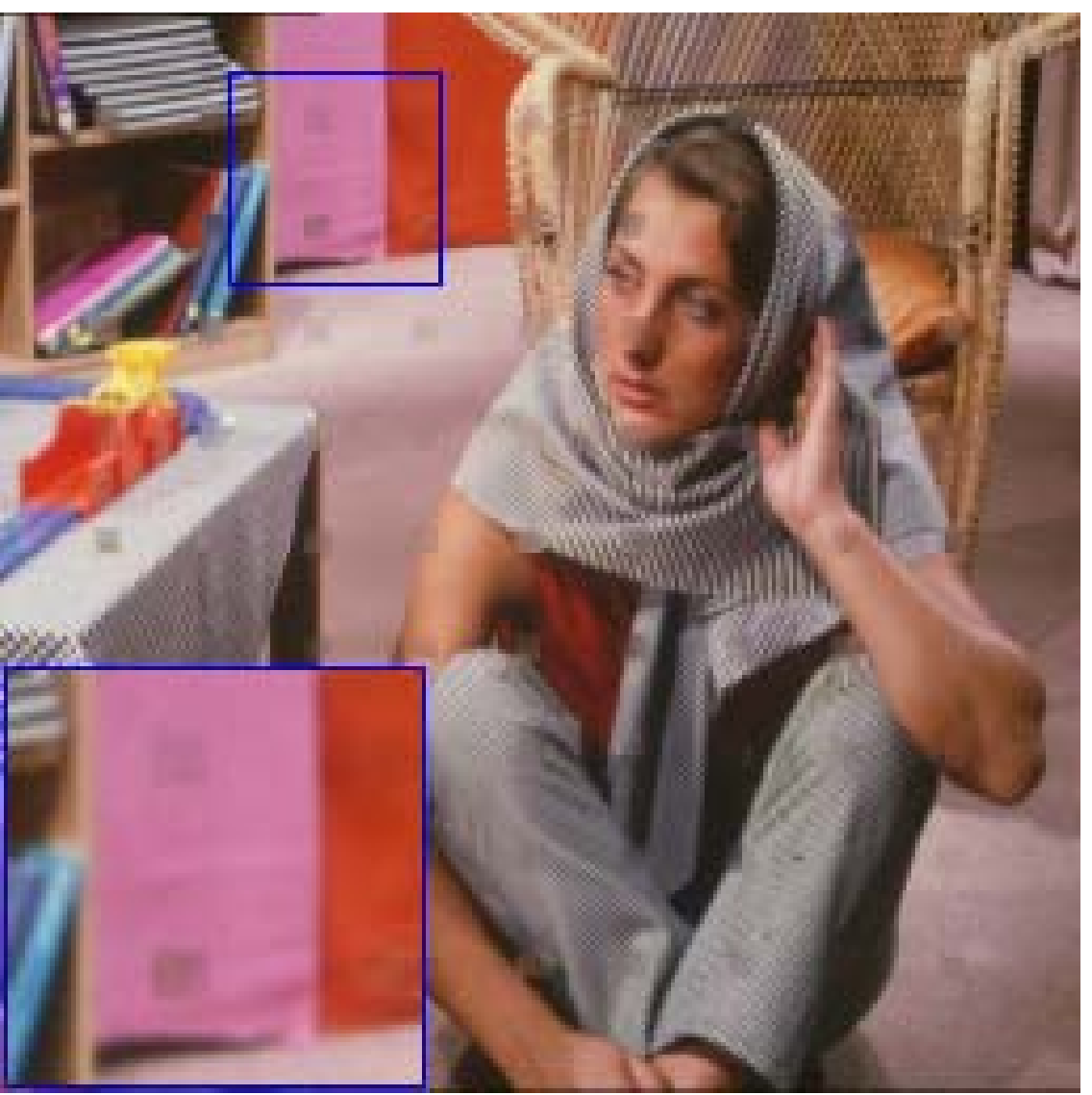}&
\includegraphics[width=0.14\textwidth]{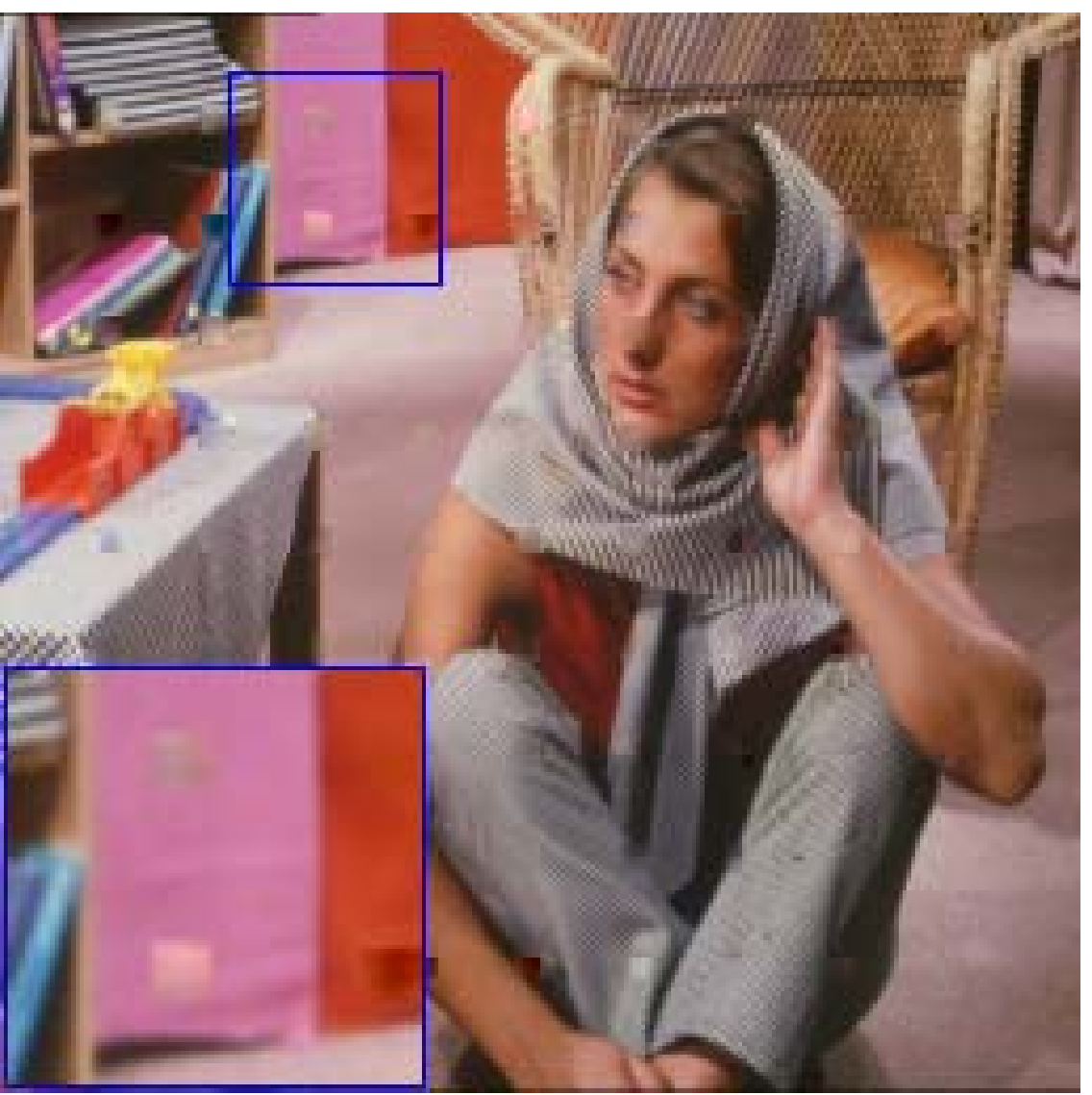}&
\includegraphics[width=0.14\textwidth]{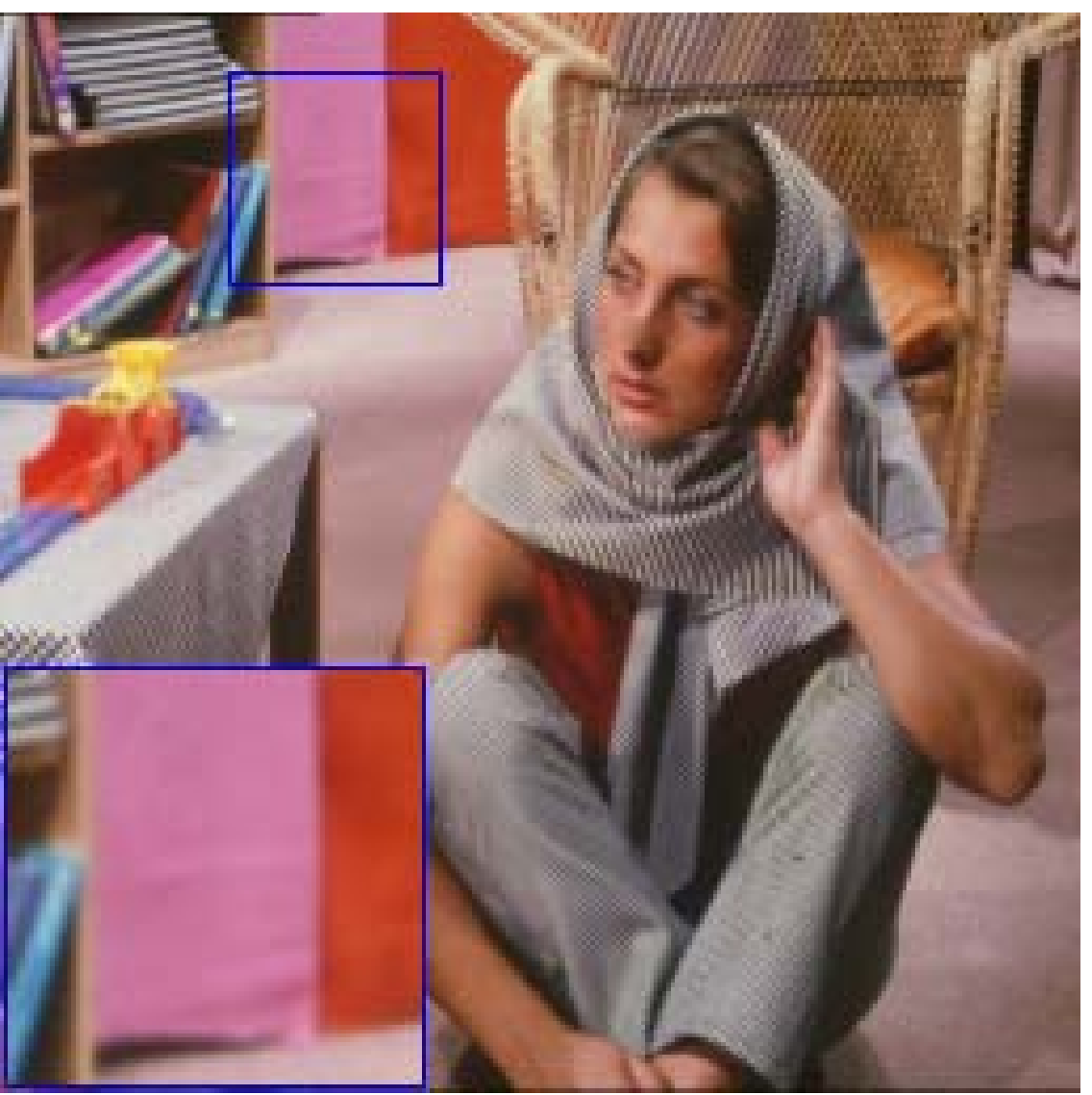}&
\includegraphics[width=0.14\textwidth]{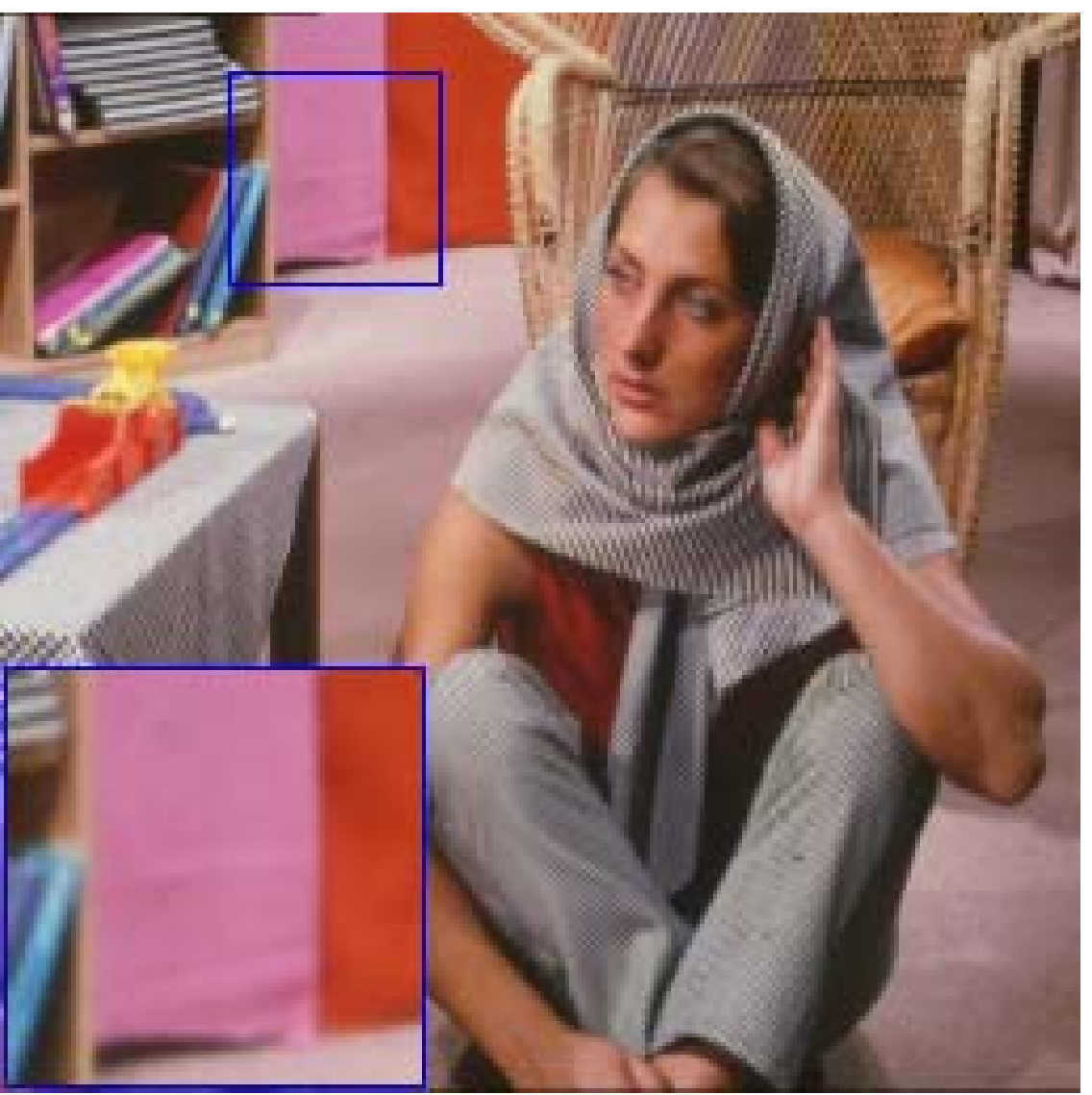}\\
\includegraphics[width=0.14\textwidth]{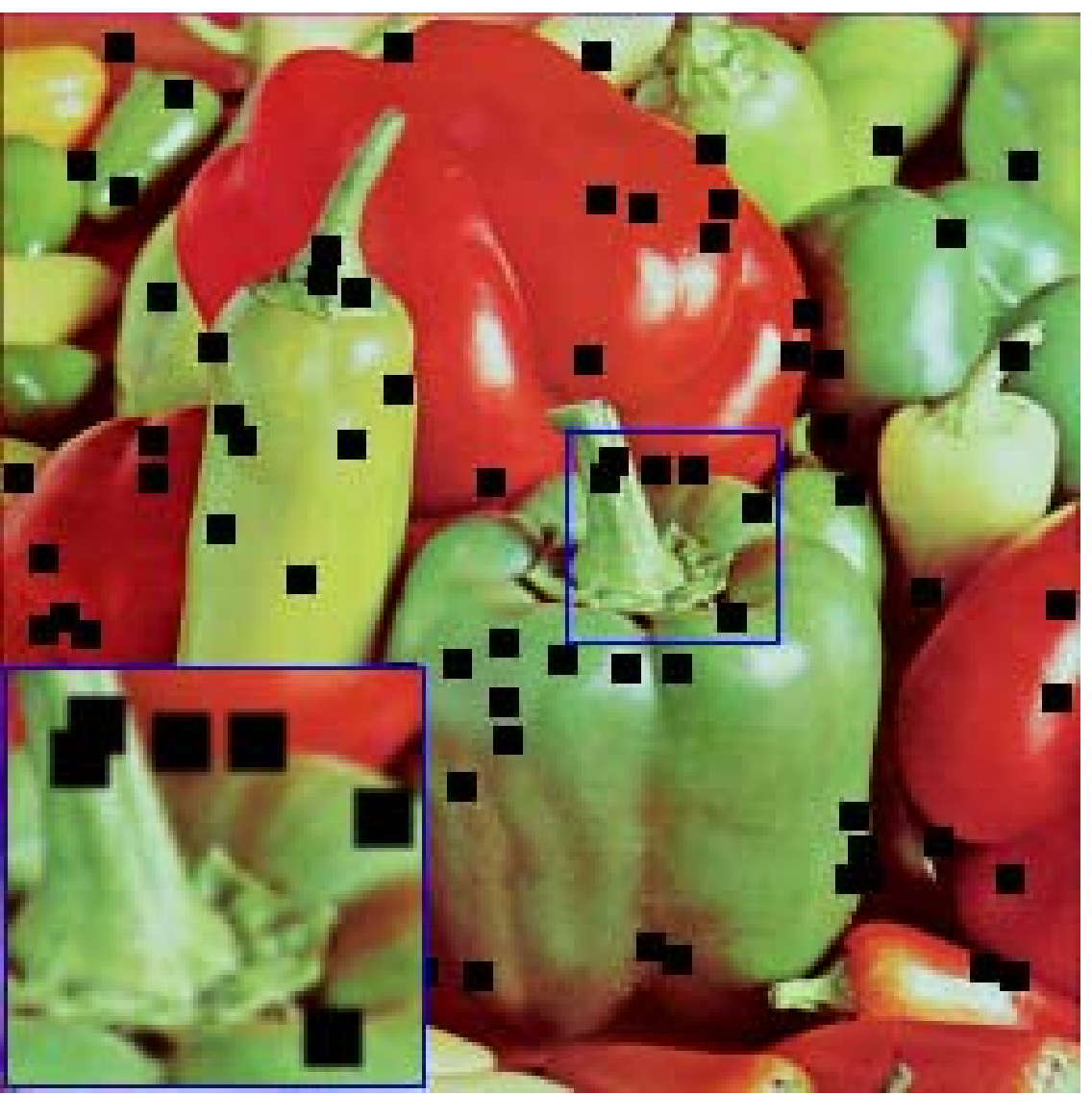}&
\includegraphics[width=0.14\textwidth]{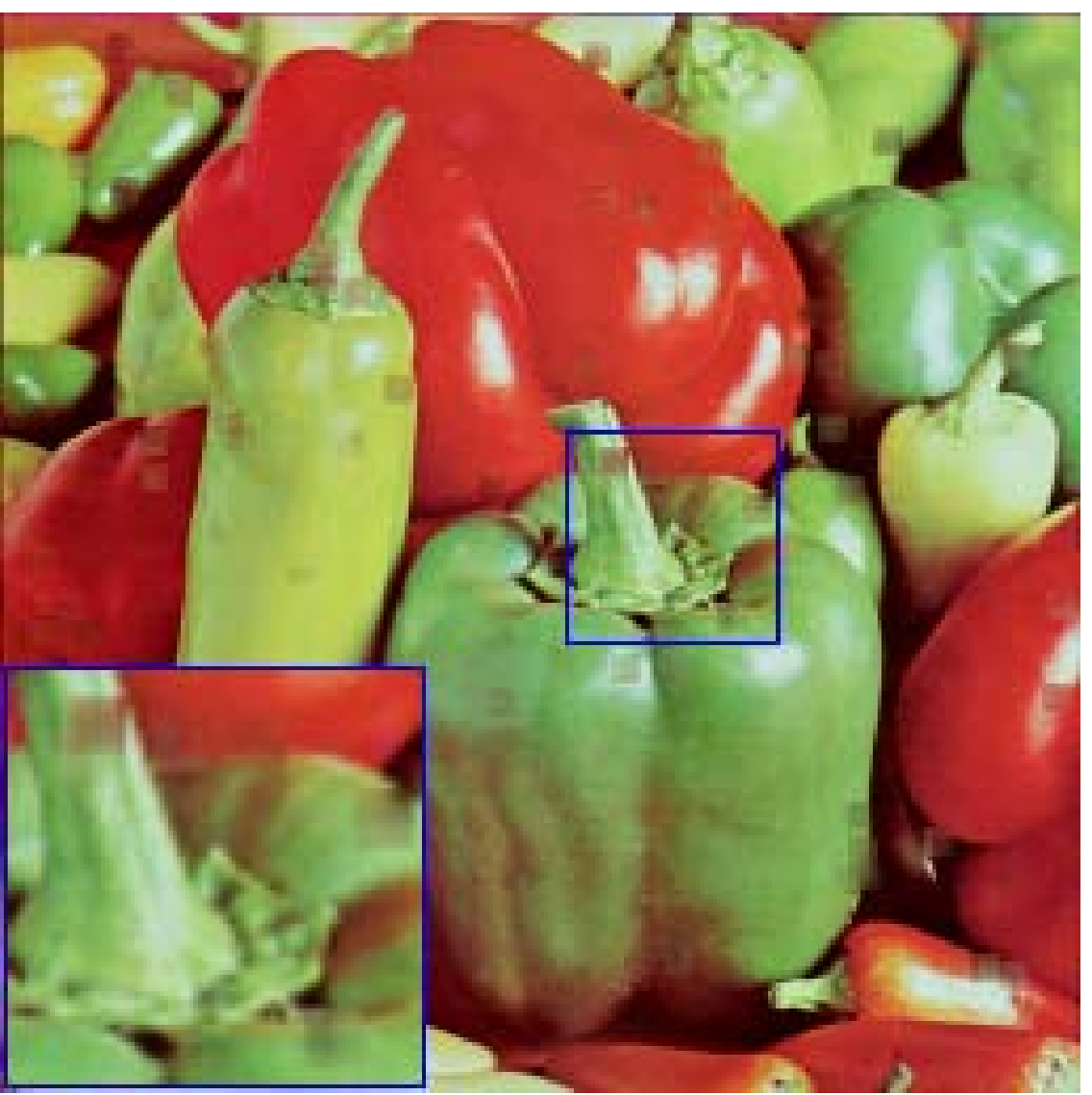}&
\includegraphics[width=0.14\textwidth]{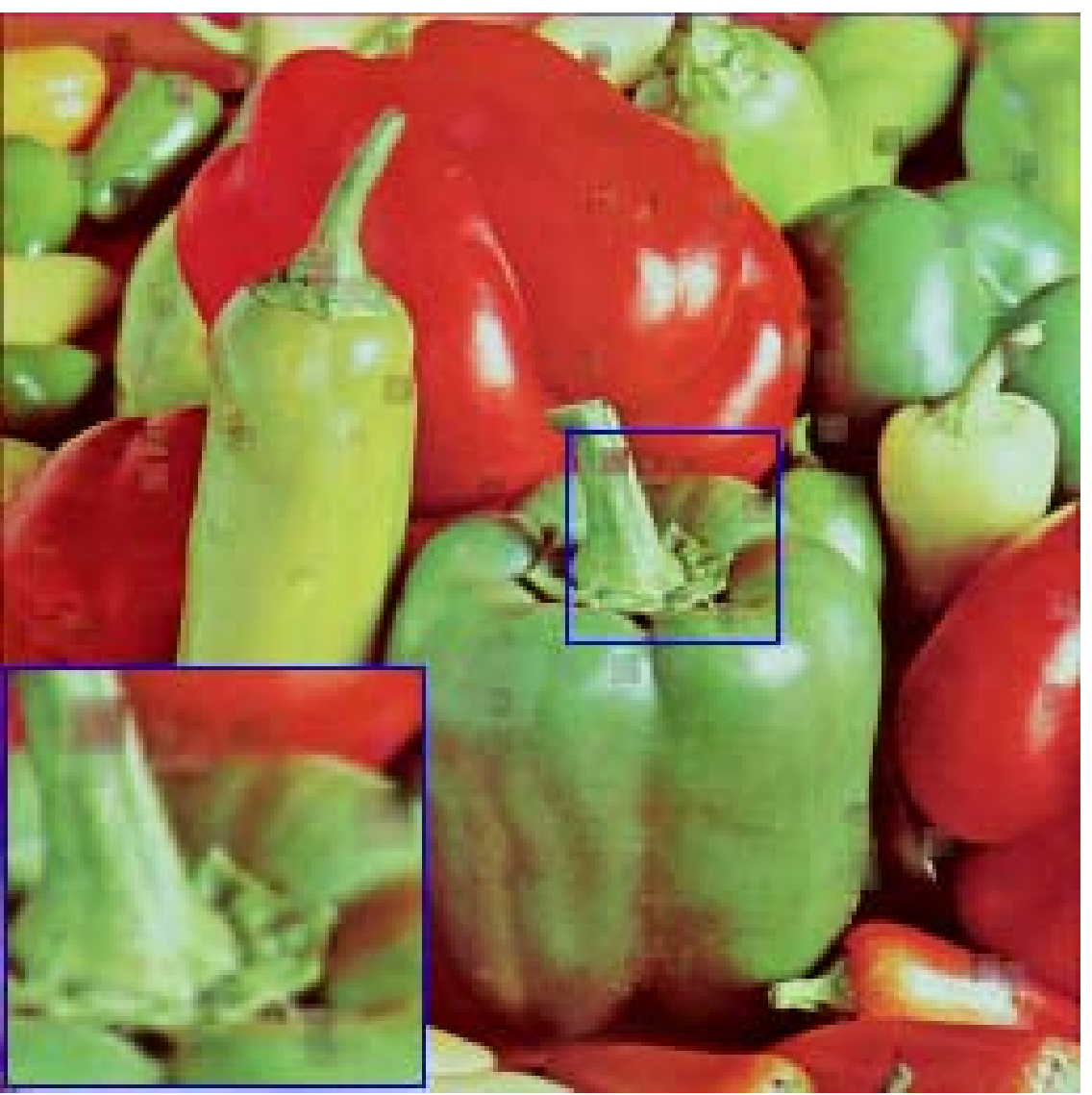}&
\includegraphics[width=0.14\textwidth]{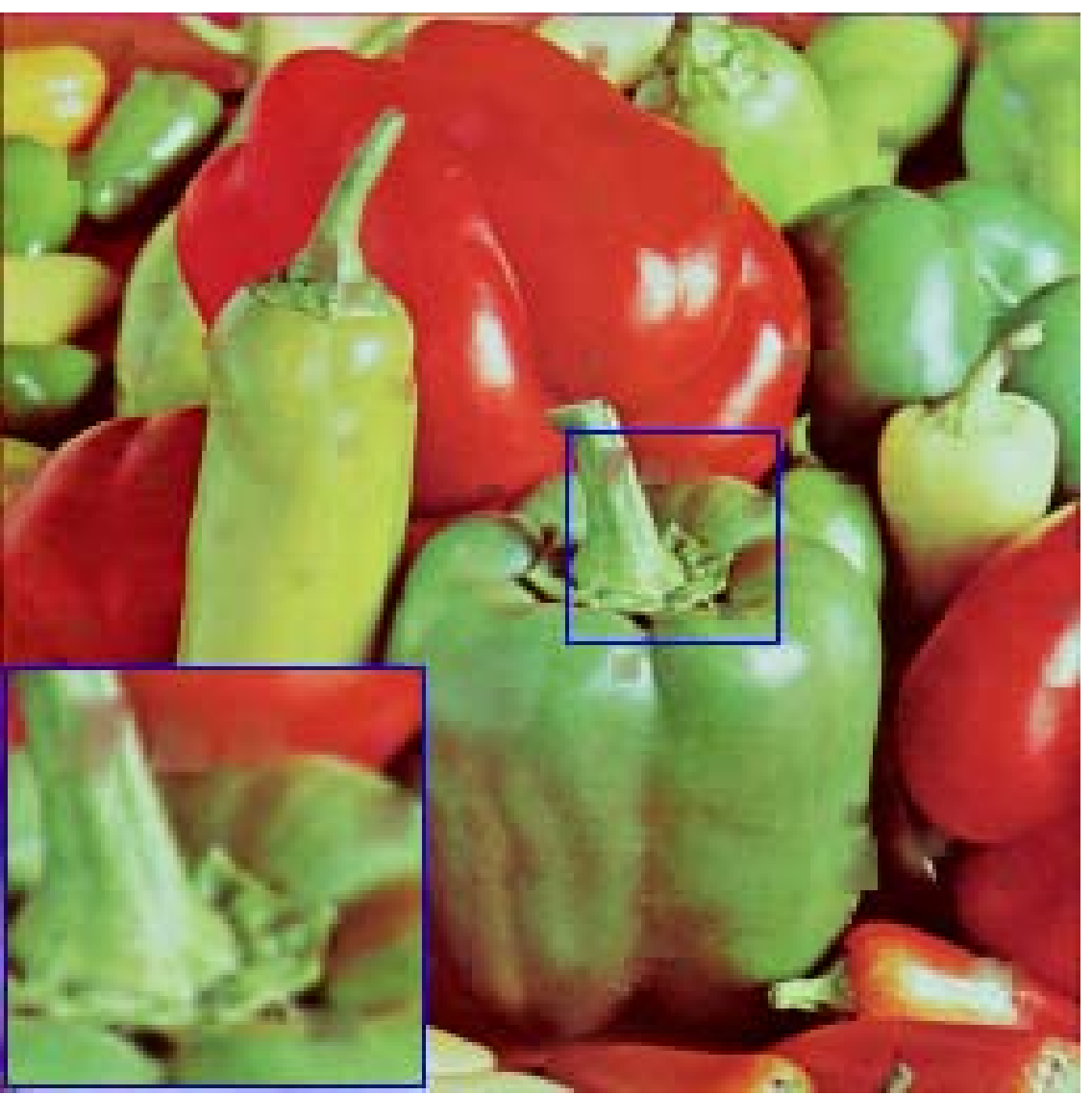}&
\includegraphics[width=0.14\textwidth]{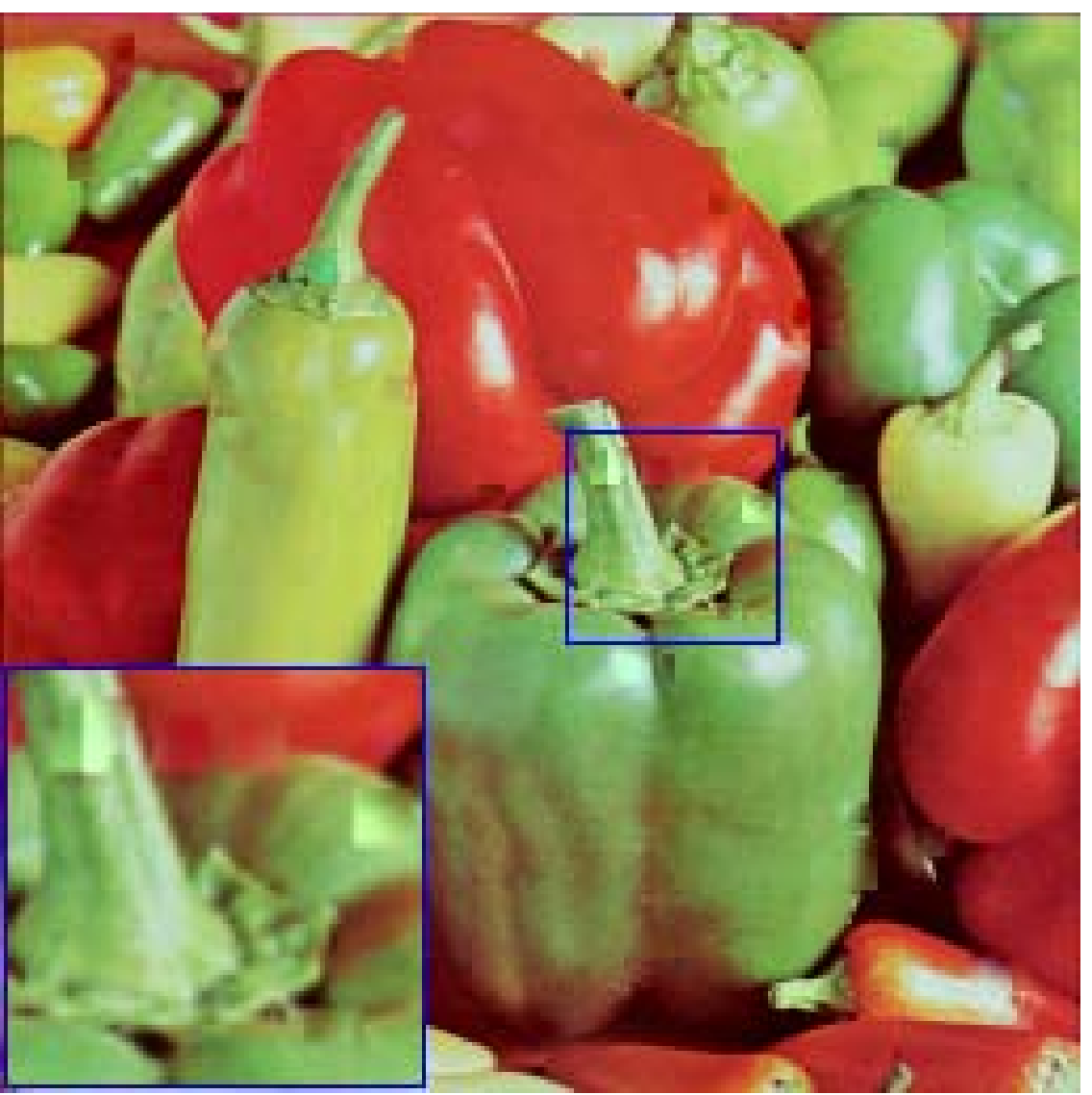}&
\includegraphics[width=0.14\textwidth]{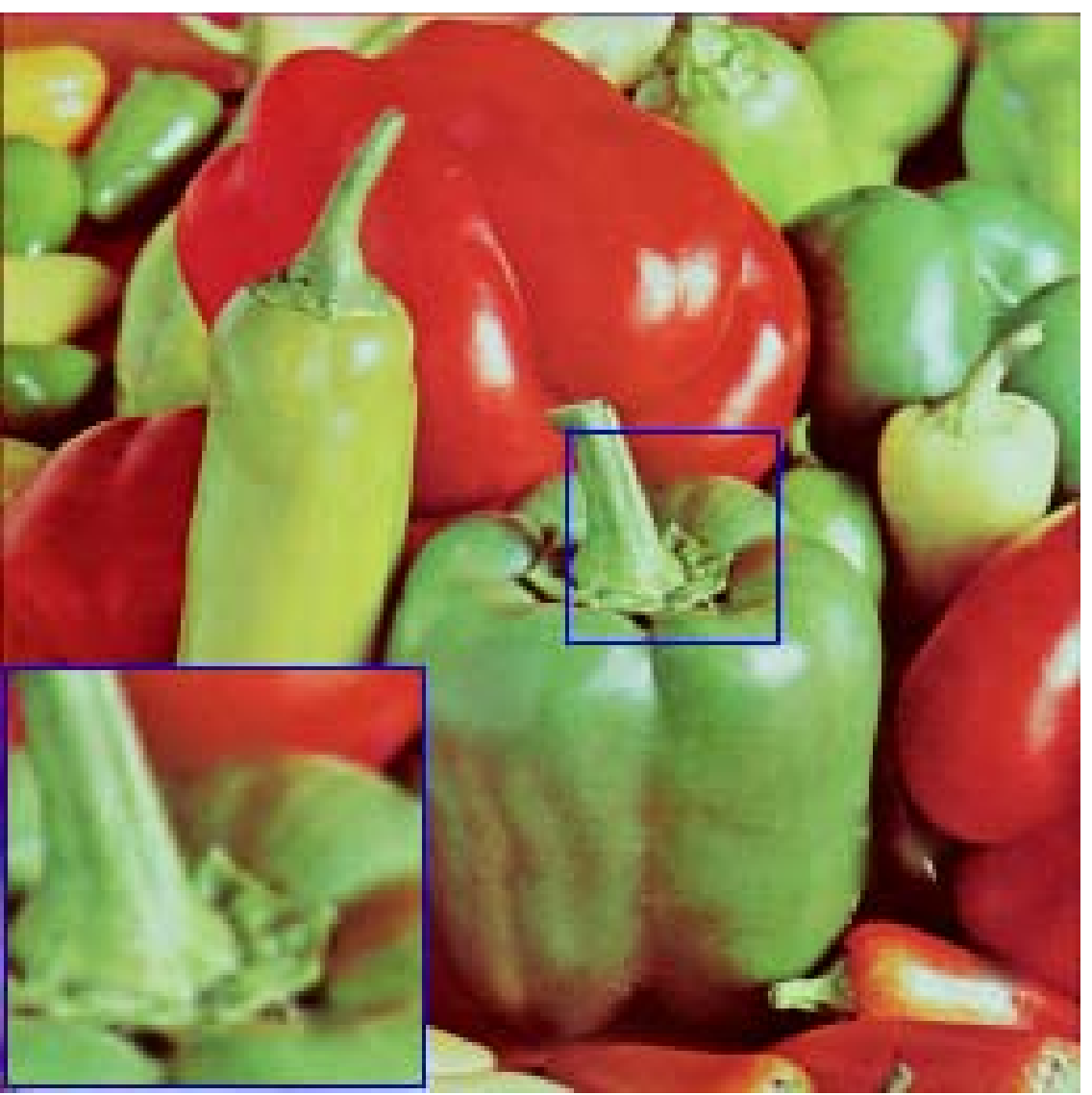}&
\includegraphics[width=0.14\textwidth]{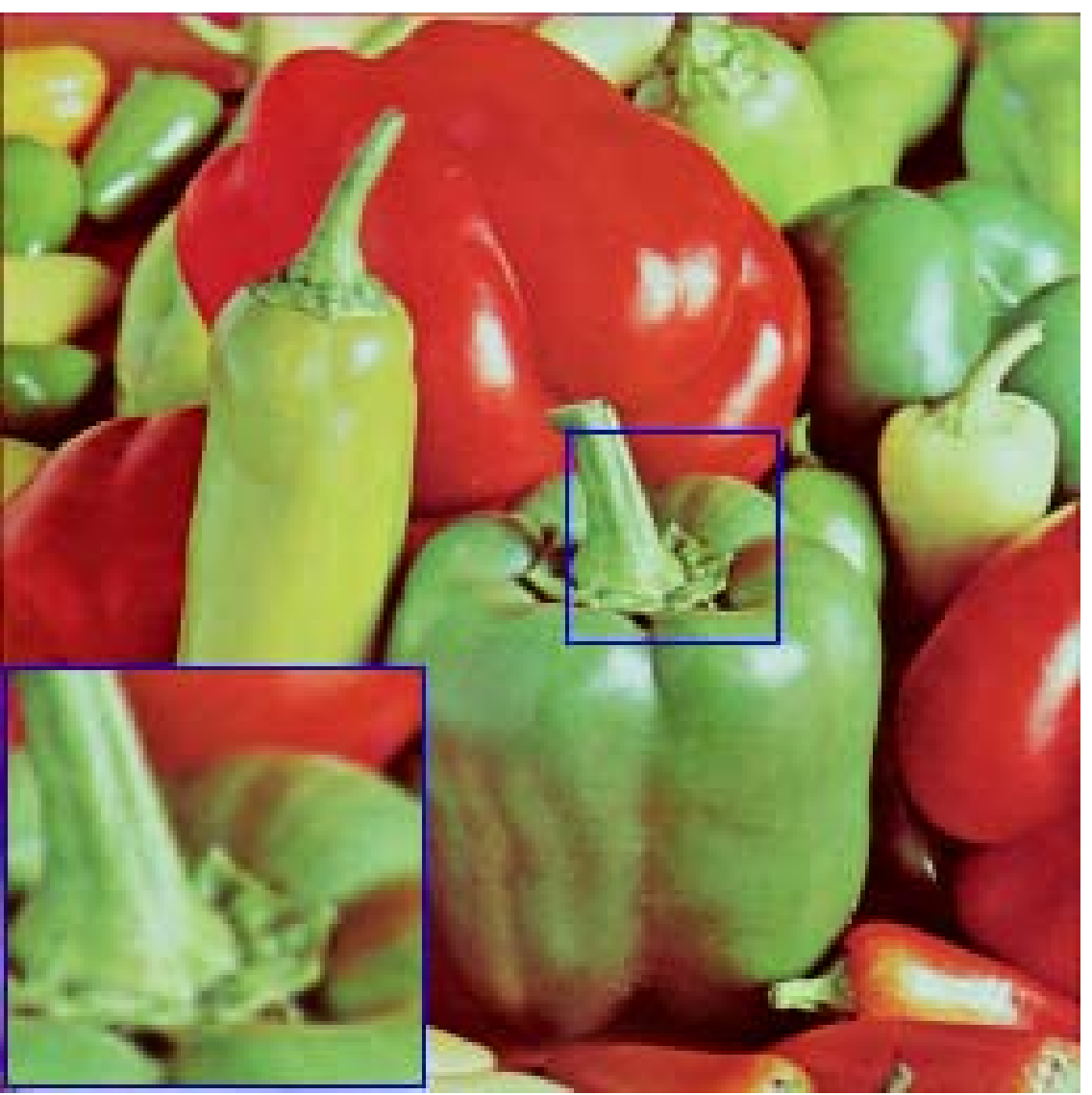}\\
 {\footnotesize\textrm{(a) Observed}} & {\footnotesize\textrm{(b) HaLRTC}} & {\footnotesize\textrm{(c) tSVD}} & {\footnotesize\textrm{(d) SiLRTC-TT}} & {\footnotesize\textrm{(e) TMac-TT}} & {\footnotesize\textrm{(f) NL-TT}}& {\footnotesize\textrm{(g) Original}}\\
\end{tabular}
\caption{\small{The results of testing color images with structural missing entries recovered by different methods. From left to right: (a) the observed image, the results by (b) HaLRTC, (c) tSVD, (d) SiLRTC-TT, (e) TMac-TT, (f) NL-TT, and (g) the original image.}}
  \label{fig:color_fix}
  \end{center}\vspace{-0.3cm}
\end{figure*}

\begin{table}[!ht]
\renewcommand\arraystretch{1.2}
  \centering
  \caption{The PSNR and SSIM values obtained by HaLRTC, tSVD, SiLRTC-TT, TMac-TT and NL-TT for color image data with structural missing entries.}
    \begin{tabular}{c|cc|cc|cc|cc|cc}    \hline

    \hline
    Method  & \multicolumn{2}{c|}{HaLRTC} & \multicolumn{2}{c|}{tSVD} & \multicolumn{2}{c|}{SiLRTC-TT} & \multicolumn{2}{c|}{TMac-TT} & \multicolumn{2}{c}{NL-TT} \\ \hline
    Image   & PSNR  & SSIM   & PSNR  & SSIM    &  PSNR  & SSIM    &  PSNR   & SSIM &  PSNR   & SSIM  \\ \hline
    \hspace{-0.23cm} \emph{house}   \hspace{-0.23cm}  & \hspace{-0.23cm} 36.44 \hspace{-0.4cm} & 0.9707 \hspace{-0.23cm} & \hspace{-0.23cm} 36.14 \hspace{-0.4cm} & 0.9681 \hspace{-0.23cm} & \hspace{-0.23cm} 38.52 \hspace{-0.4cm} & 0.9793 \hspace{-0.23cm} & \hspace{-0.23cm} 38.03 \hspace{-0.4cm} & 0.9740 \hspace{-0.23cm}
    & \hspace{-0.23cm} \textbf{45.34} \hspace{-0.4cm} & \textbf{0.9906} \hspace{-0.23cm} \\
    \hspace{-0.23cm} \emph{facade}  \hspace{-0.23cm}  & \hspace{-0.23cm} 12.95 \hspace{-0.4cm} & 0.5681 \hspace{-0.23cm} & \hspace{-0.23cm} 12.95 \hspace{-0.4cm} & 0.5681 \hspace{-0.23cm} & \hspace{-0.23cm} 28.14 \hspace{-0.4cm} & 0.9062 \hspace{-0.23cm} & \hspace{-0.23cm} 27.50 \hspace{-0.4cm} & 0.8947 \hspace{-0.23cm}
    & \hspace{-0.23cm} \textbf{29.60} \hspace{-0.4cm} & \textbf{0.9357} \hspace{-0.23cm} \\
    \hspace{-0.23cm} \emph{sailboat} \hspace{-0.23cm} & \hspace{-0.23cm} 26.49 \hspace{-0.4cm} & 0.8700 \hspace{-0.23cm} & \hspace{-0.23cm} 26.69 \hspace{-0.4cm} & 0.8696 \hspace{-0.23cm} & \hspace{-0.23cm} 26.53 \hspace{-0.4cm} & 0.8838 \hspace{-0.23cm} & \hspace{-0.23cm} 26.40 \hspace{-0.4cm} & 0.8995 \hspace{-0.23cm}
    & \hspace{-0.23cm} \textbf{27.86} \hspace{-0.4cm} & \textbf{0.9370} \hspace{-0.23cm} \\
    \hspace{-0.23cm} \emph{barbara} \hspace{-0.23cm}  & \hspace{-0.23cm} 32.44 \hspace{-0.4cm} & 0.9580 \hspace{-0.23cm} & \hspace{-0.23cm} 32.44 \hspace{-0.4cm} & 0.9579 \hspace{-0.23cm} & \hspace{-0.23cm} 33.99 \hspace{-0.4cm} & 0.9681 \hspace{-0.23cm} & \hspace{-0.23cm} 33.29 \hspace{-0.4cm} & 0.9654 \hspace{-0.23cm}
    & \hspace{-0.23cm} \textbf{37.56} \hspace{-0.4cm} & \textbf{0.9867} \hspace{-0.23cm} \\
    \hspace{-0.23cm} \emph{peppers} \hspace{-0.23cm}  & \hspace{-0.23cm} 31.64 \hspace{-0.4cm} & 0.9595 \hspace{-0.23cm} & \hspace{-0.23cm} 31.53 \hspace{-0.4cm} & 0.9551 \hspace{-0.23cm} & \hspace{-0.23cm} 32.59 \hspace{-0.4cm} & 0.9676 \hspace{-0.23cm} & \hspace{-0.23cm} 32.77 \hspace{-0.4cm} & 0.9651 \hspace{-0.23cm}
    & \hspace{-0.23cm} \textbf{36.33} \hspace{-0.4cm} & \textbf{0.9862} \hspace{-0.23cm} \\ \hline
    \hspace{-0.23cm} Average  \hspace{-0.23cm}        & \hspace{-0.23cm} 27.99 \hspace{-0.4cm} & 0.8653 \hspace{-0.23cm} & \hspace{-0.23cm} 27.95 \hspace{-0.4cm} & 0.8638 \hspace{-0.23cm} & \hspace{-0.23cm} 31.95 \hspace{-0.4cm} & 0.9410 \hspace{-0.23cm} & \hspace{-0.23cm} 31.60 \hspace{-0.4cm} & 0.9397 \hspace{-0.23cm}
    & \hspace{-0.23cm} \textbf{35.34} \hspace{-0.4cm} & \textbf{0.9672} \hspace{-0.23cm} \\ \hline

    \hline
    \end{tabular}%
  \label{table:color_fix}%
\end{table}%

Fig. \ref{fig:color_fix} shows the experimental results obtained by different methods. Enlarged subregions are marked by a blue box at the bottom left corner of each image. HaLRTC and tSVD fail to recover the missing slices. There are ``shadows" retained in the images recovered by HaLRTC, tSVD, and SiLRTC-TT. TMac-TT fills the missing areas, but causes block-artifacts  on the restored images. By contrast, the proposed method recovers most of missing areas without outlines and performs well in local details. Table \ref{table:color_fix} shows the PSNR and SSIM values obtained for all completion methods, which demonstrates that the proposed method performs better than four well-known methods in terms of the PSNR and SSIM measures. It is worth noting that our method achieves nearly 3.4 dB improvement than the second-best results in average.

\subsection{MSIs}
We test the CAVE MSI database \footnote{http://www1.cs.columbia.edu/CAVE/databases/multispectral}, which contains 32 real-world scenes, each of which has 31 spectral bands with $512\times 512$ pixels for each band. We resize the spatial resolution $512\times 512$ to $256\times 256$, and select 11 bands for our experiments. For MSIs, we only test the random sampling case. The SRs are set to be 0.05, 0.1, and 0.2, respectively.

\begin{figure*}[!ht]
\scriptsize\setlength{\tabcolsep}{0.9pt}
\begin{center}
\begin{tabular}{cccccccc}
\includegraphics[width=0.14\textwidth]{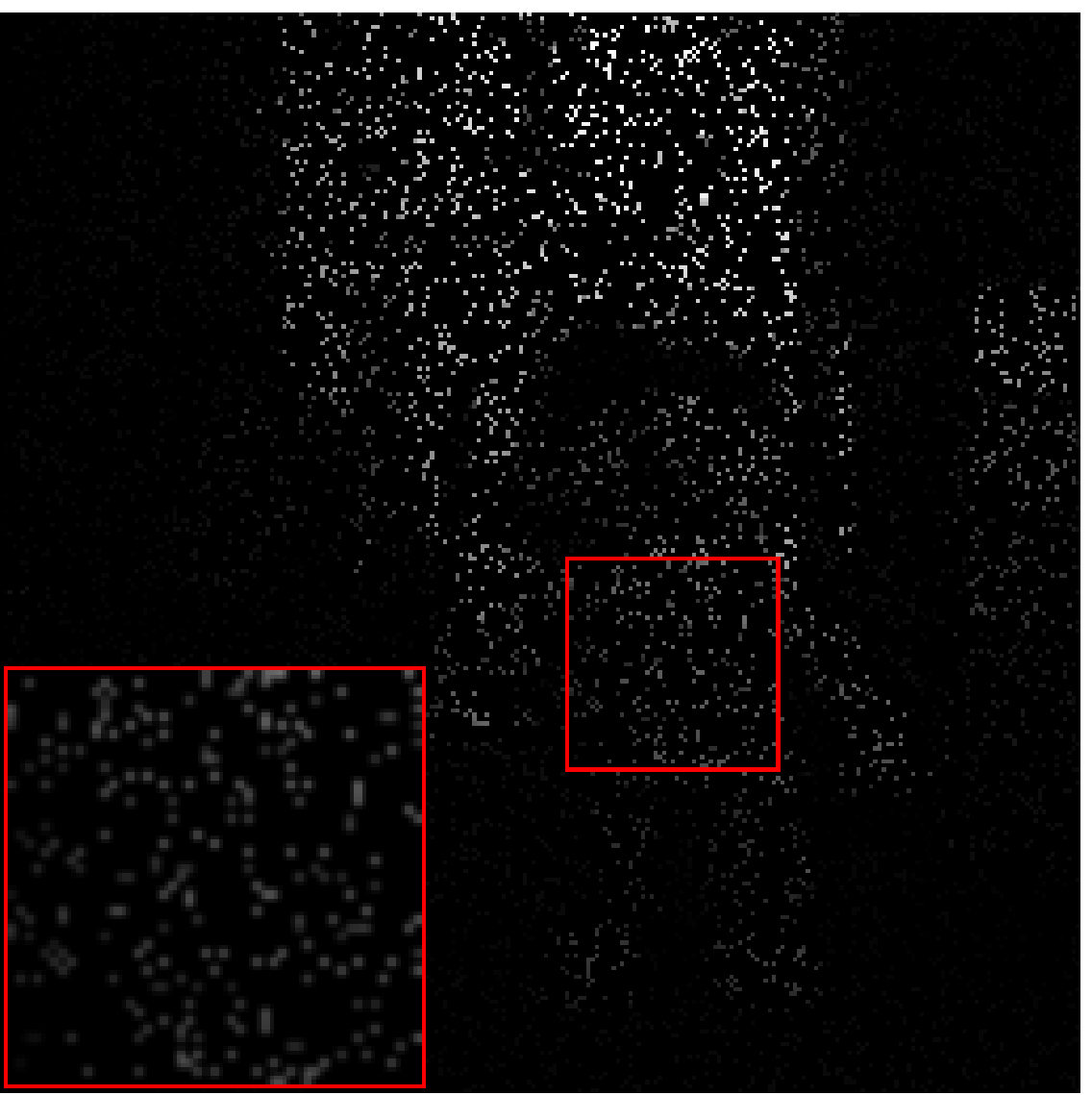}&
\includegraphics[width=0.14\textwidth]{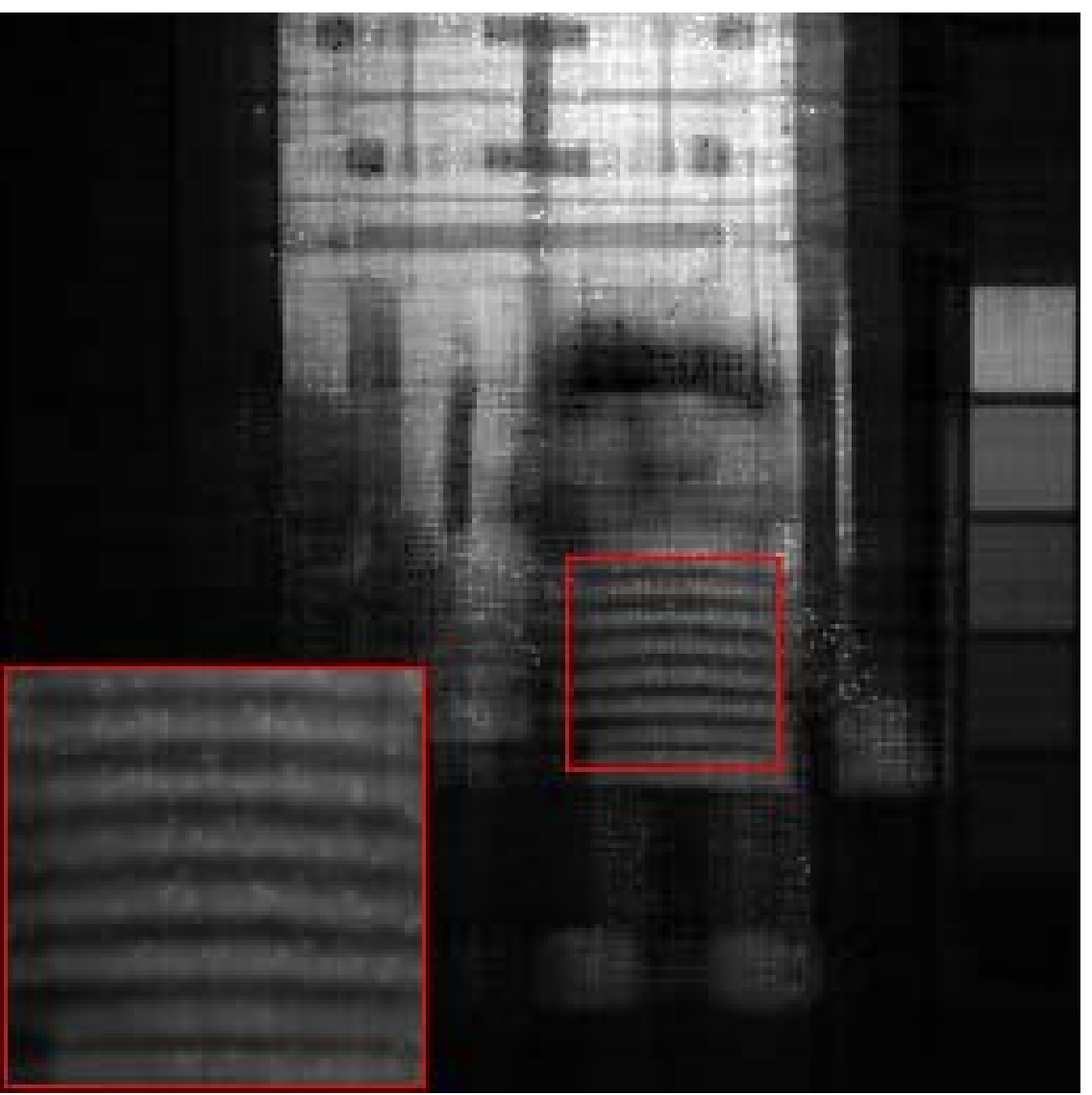}&
\includegraphics[width=0.14\textwidth]{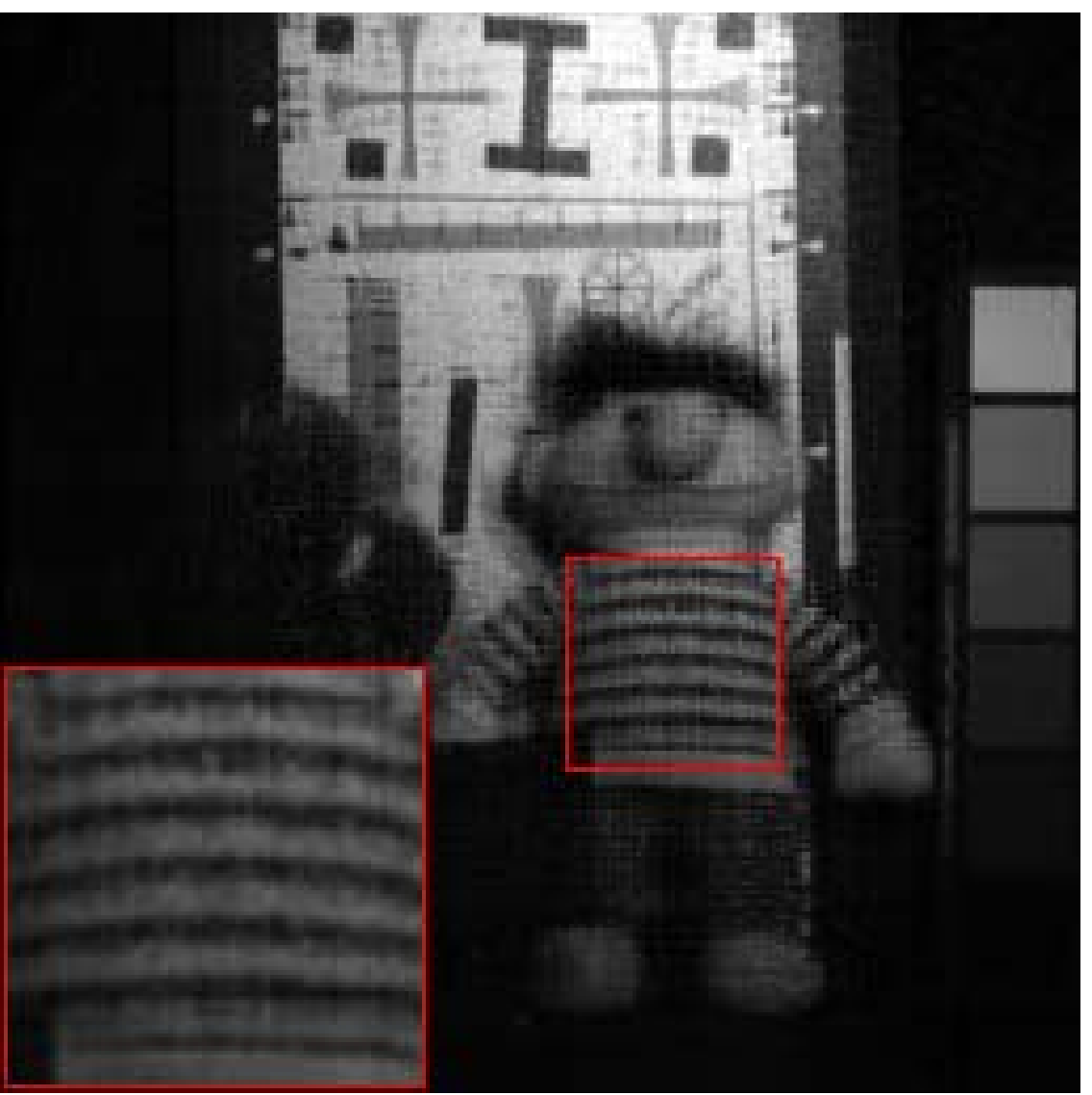}&
\includegraphics[width=0.14\textwidth]{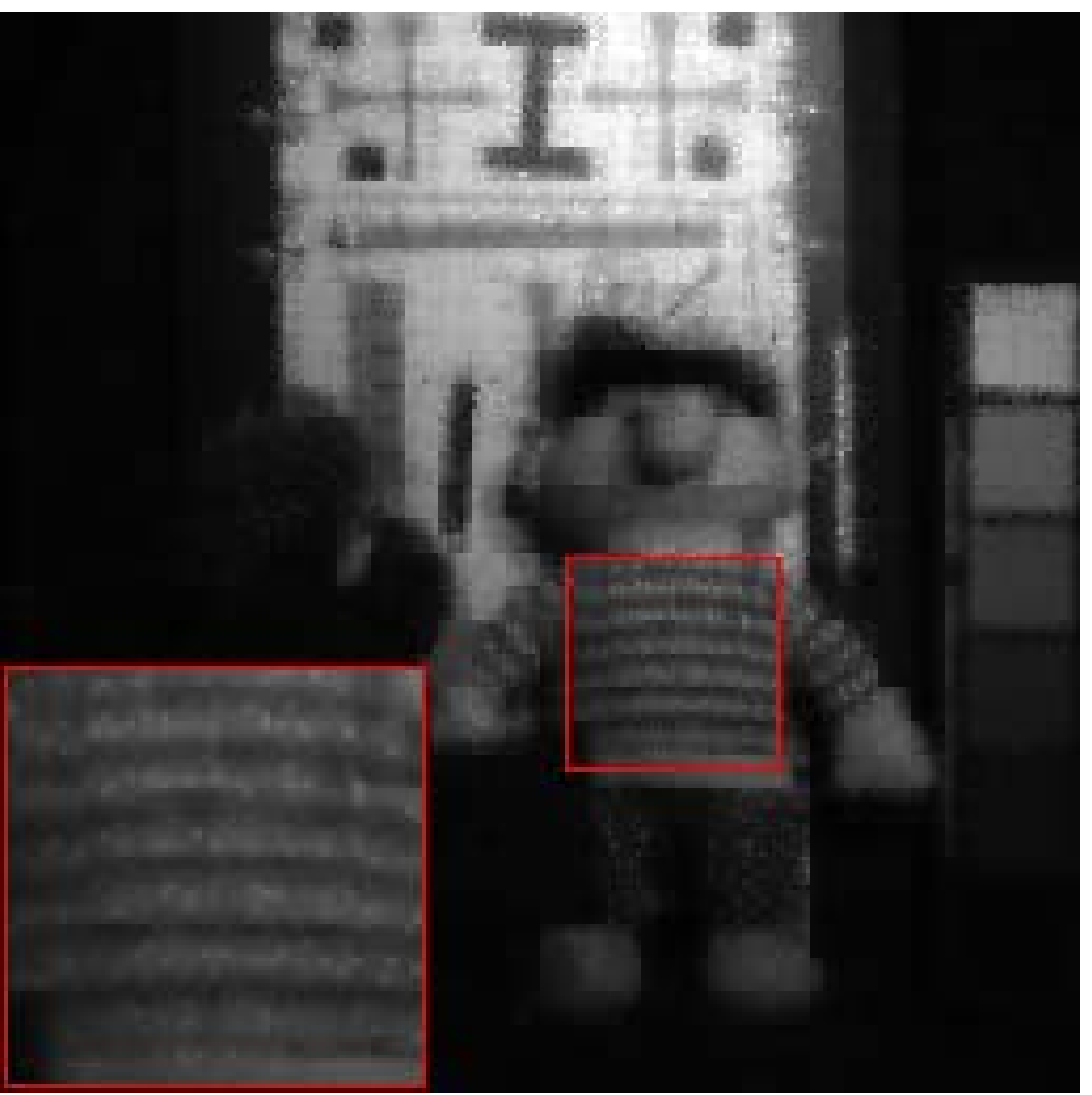}&
\includegraphics[width=0.14\textwidth]{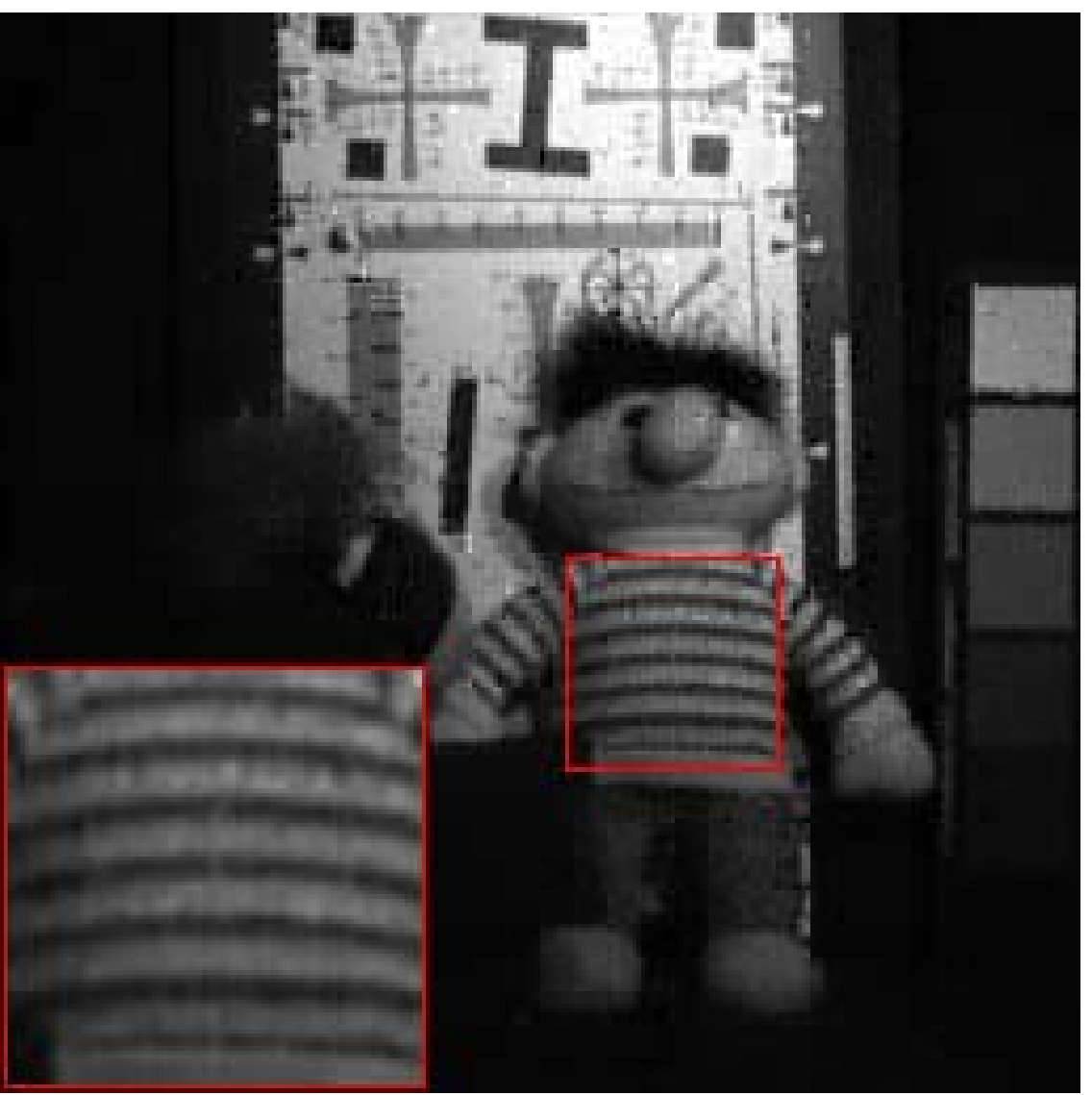}&
\includegraphics[width=0.14\textwidth]{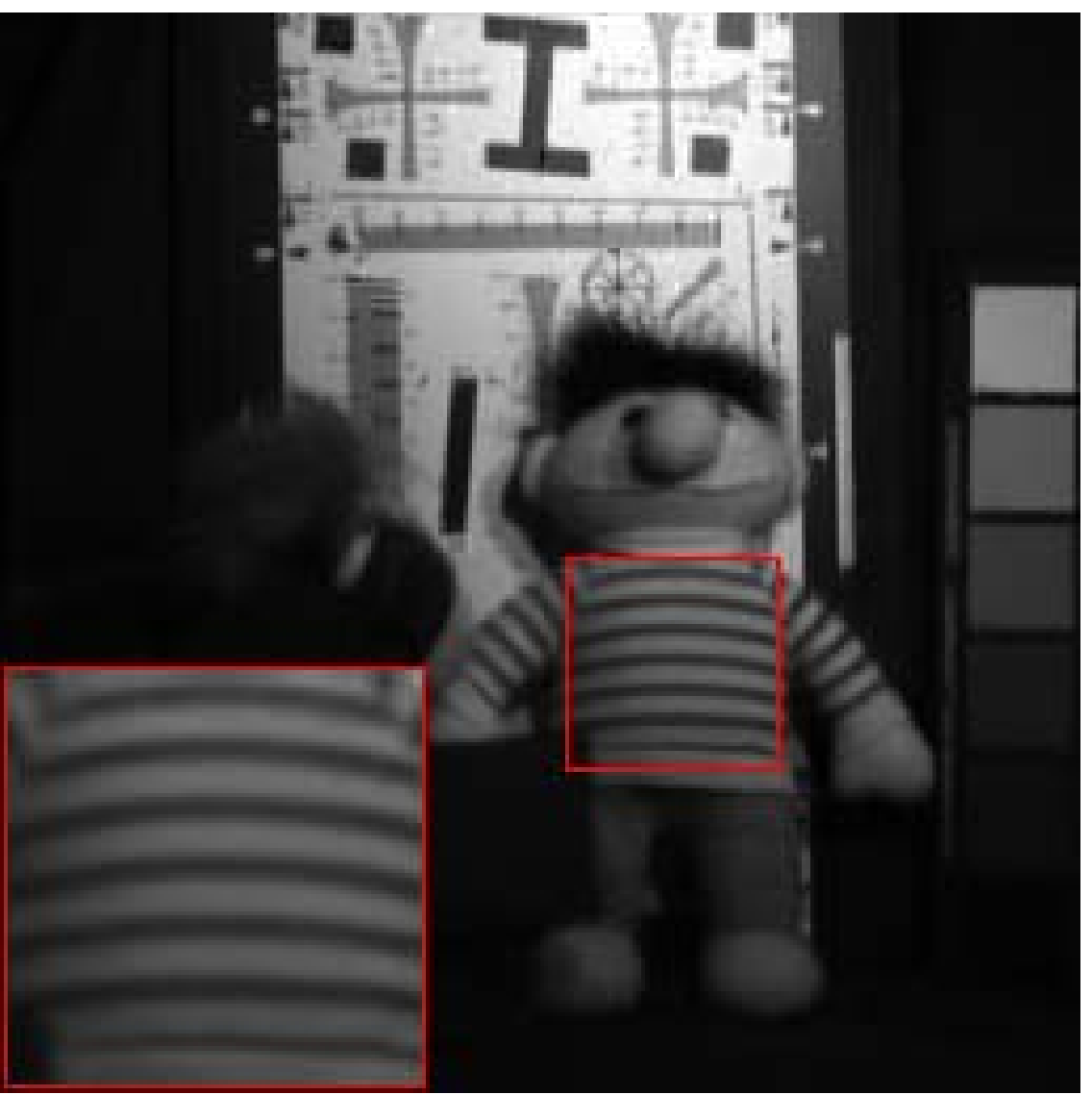}&
\includegraphics[width=0.14\textwidth]{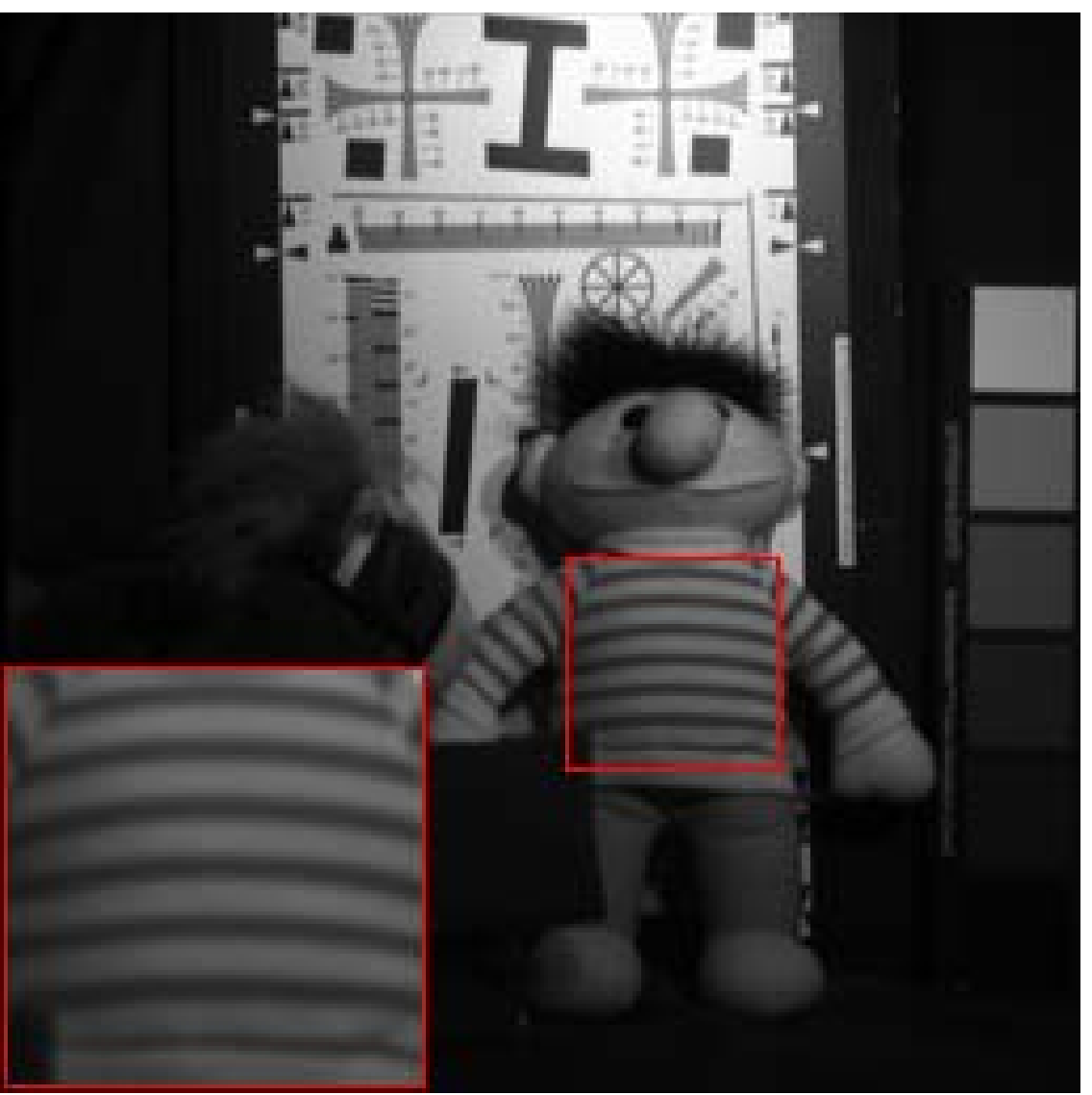}\\
\includegraphics[width=0.14\textwidth]{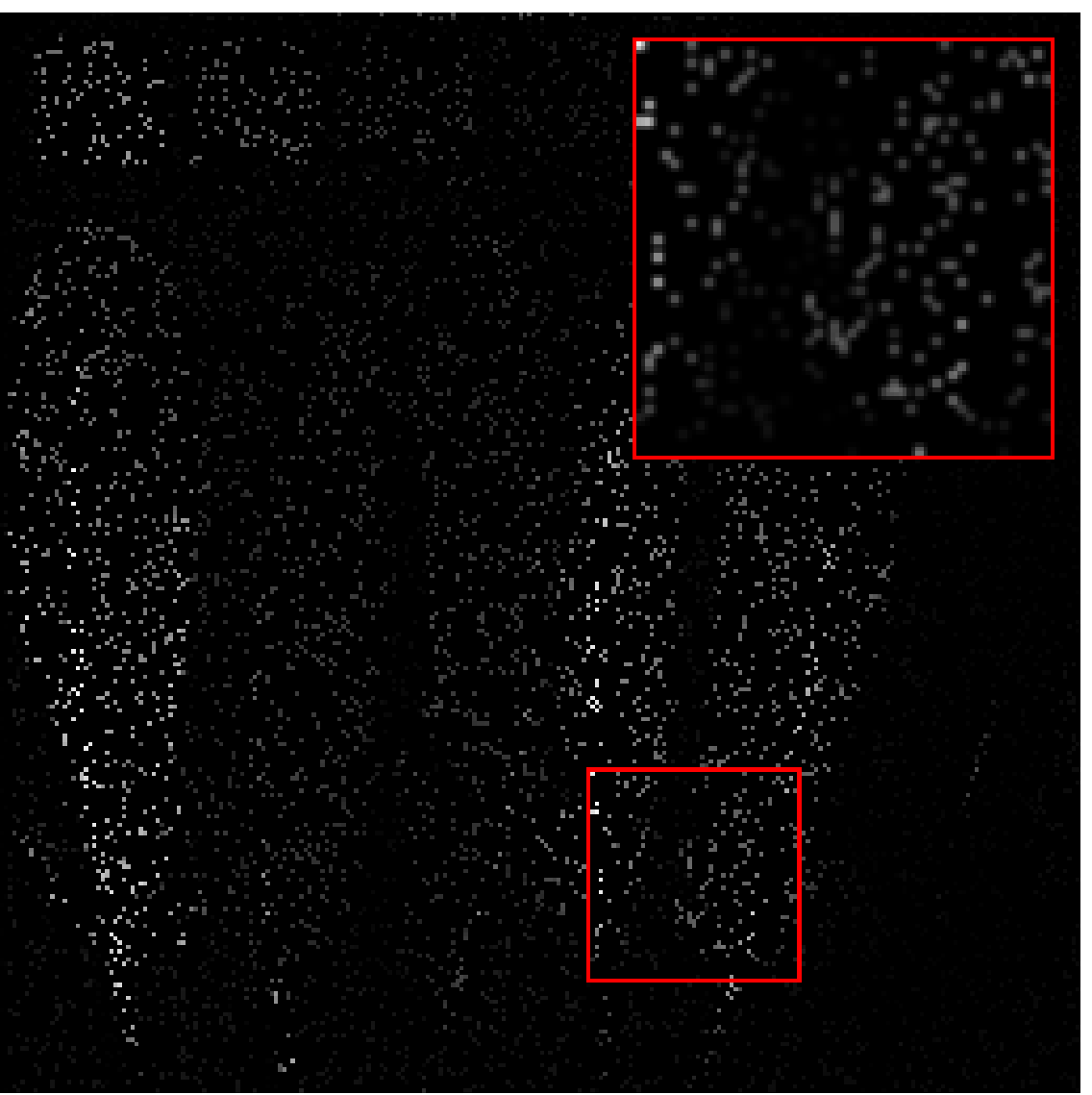}&
\includegraphics[width=0.14\textwidth]{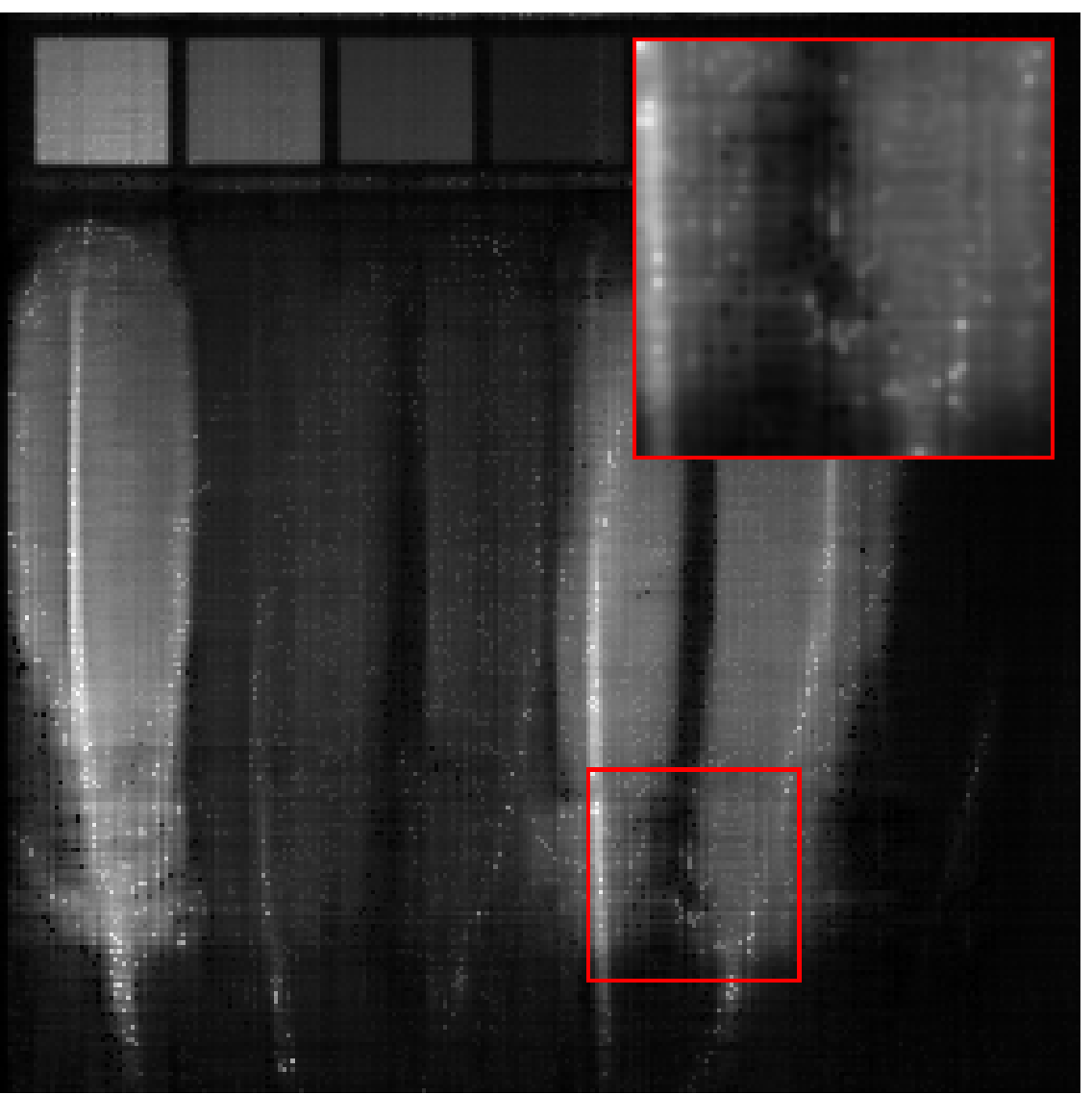}&
\includegraphics[width=0.14\textwidth]{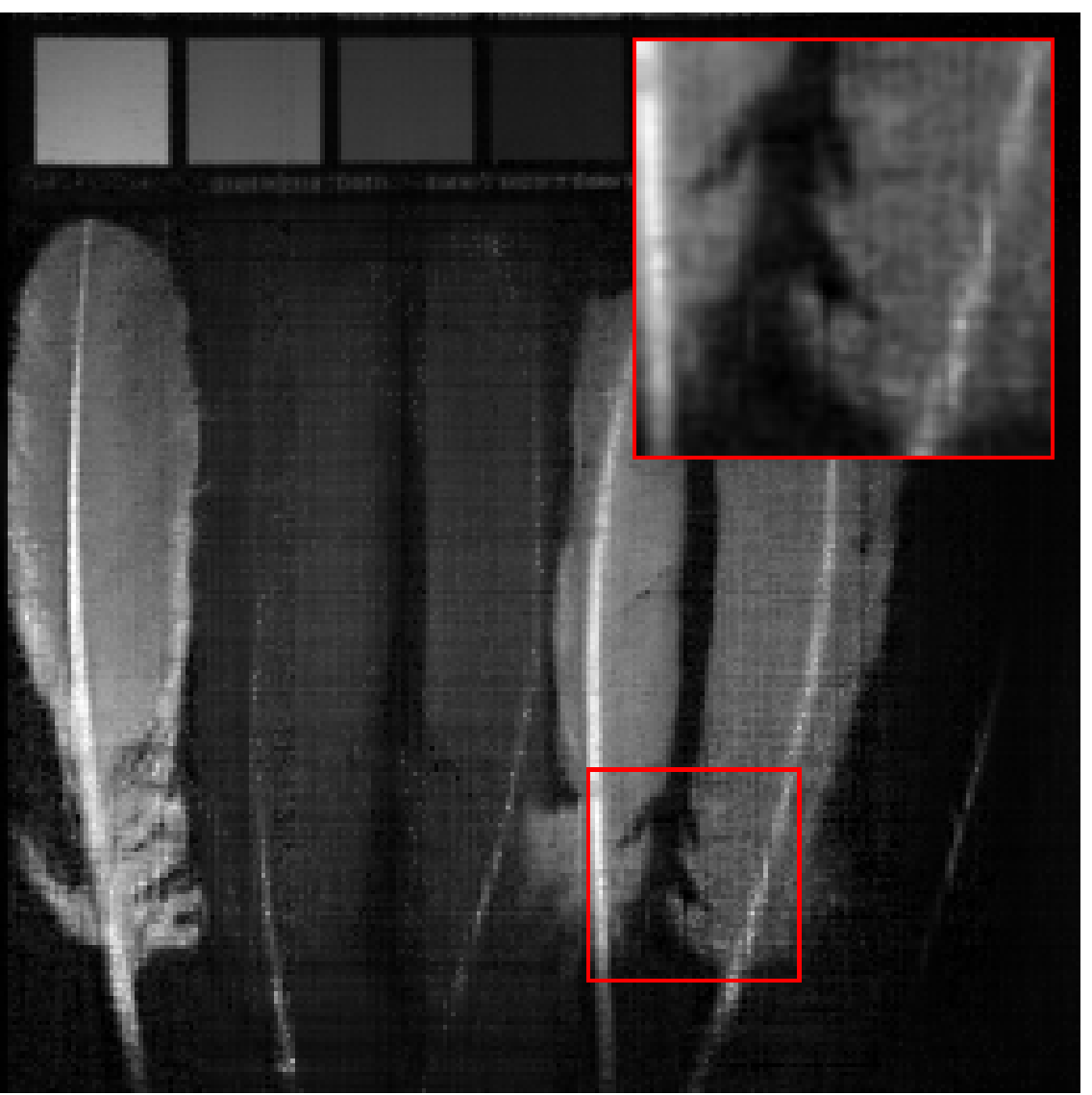}&
\includegraphics[width=0.14\textwidth]{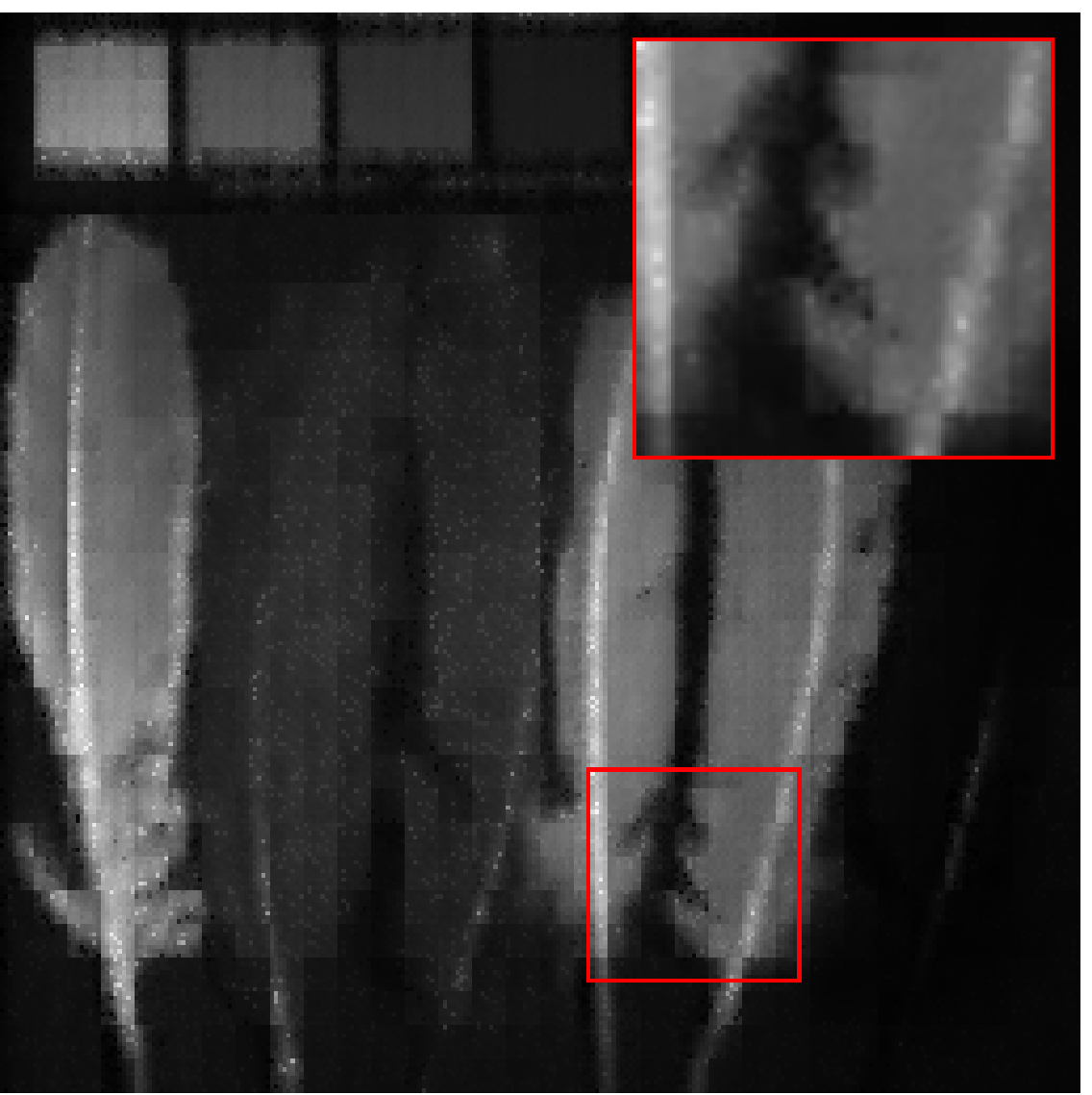}&
\includegraphics[width=0.14\textwidth]{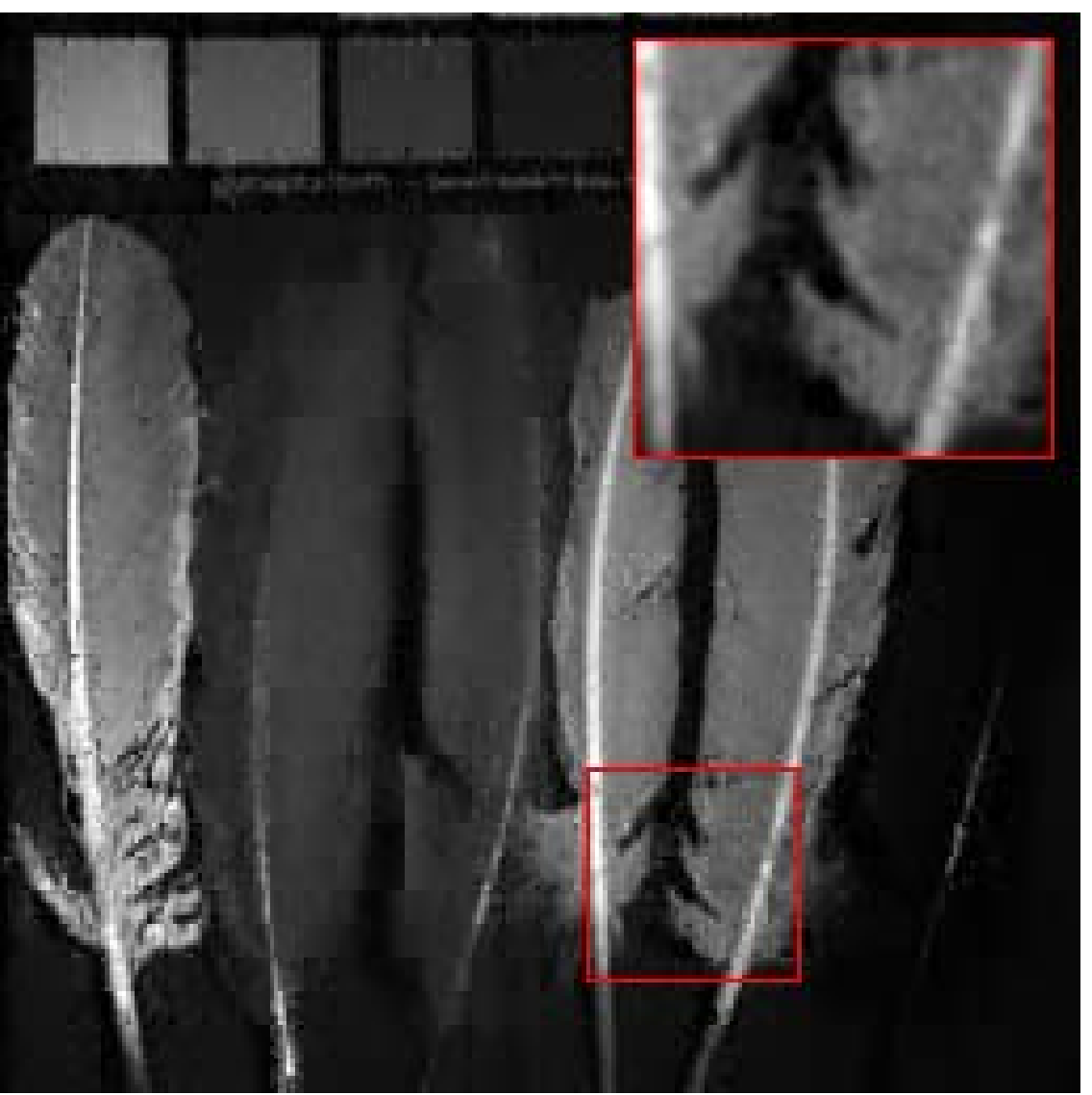}&
\includegraphics[width=0.14\textwidth]{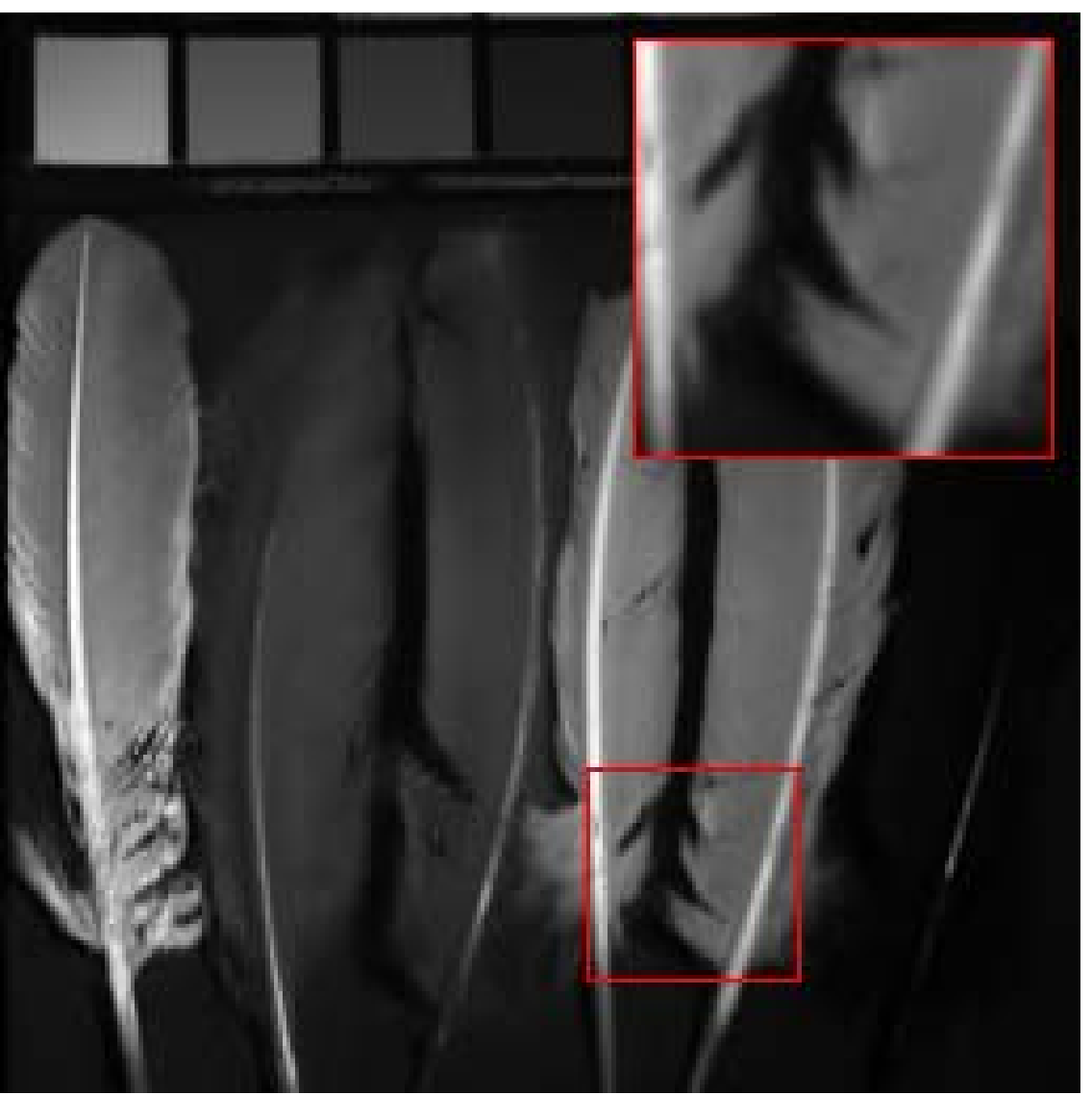}&
\includegraphics[width=0.14\textwidth]{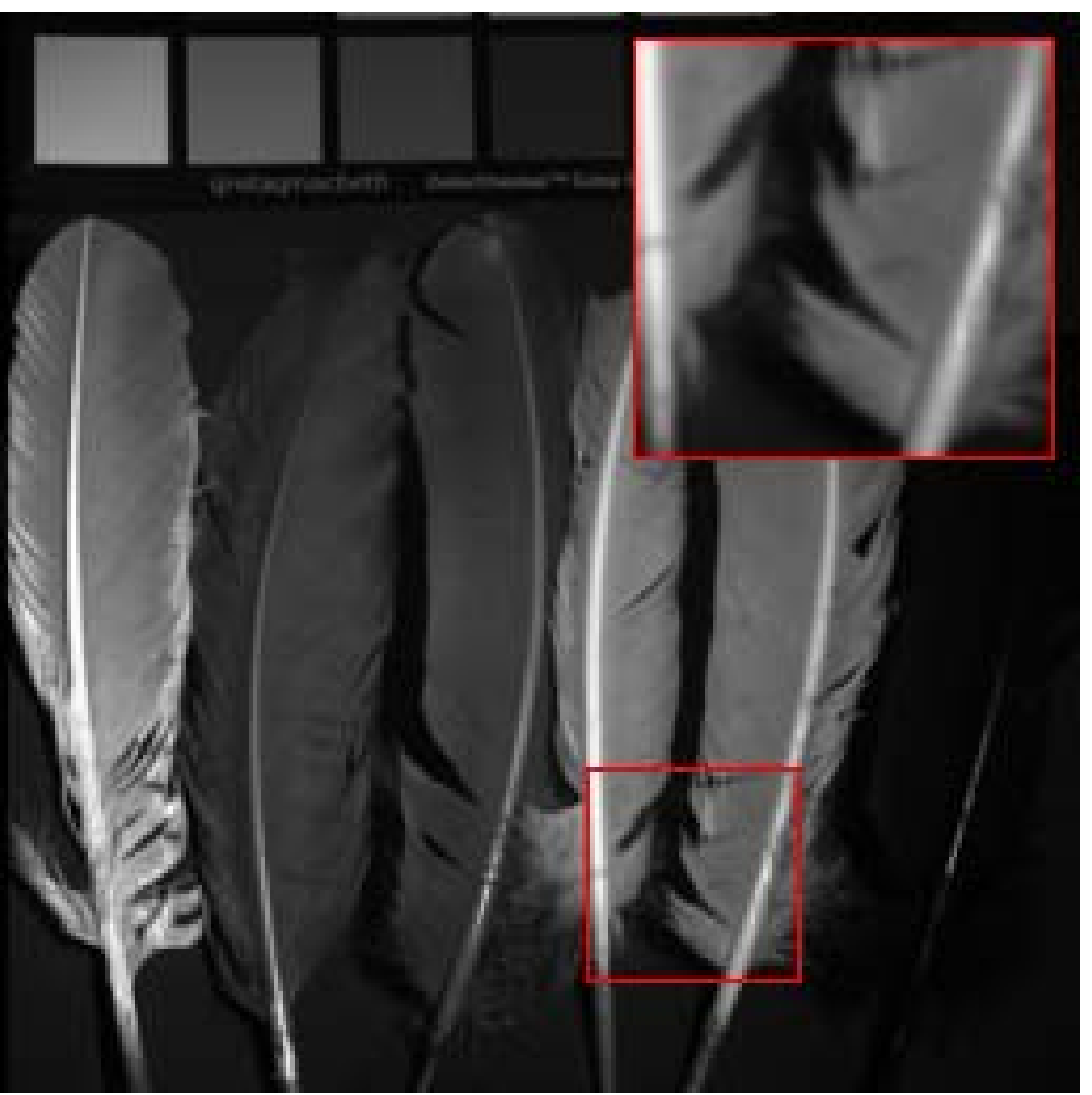}\\
\includegraphics[width=0.14\textwidth]{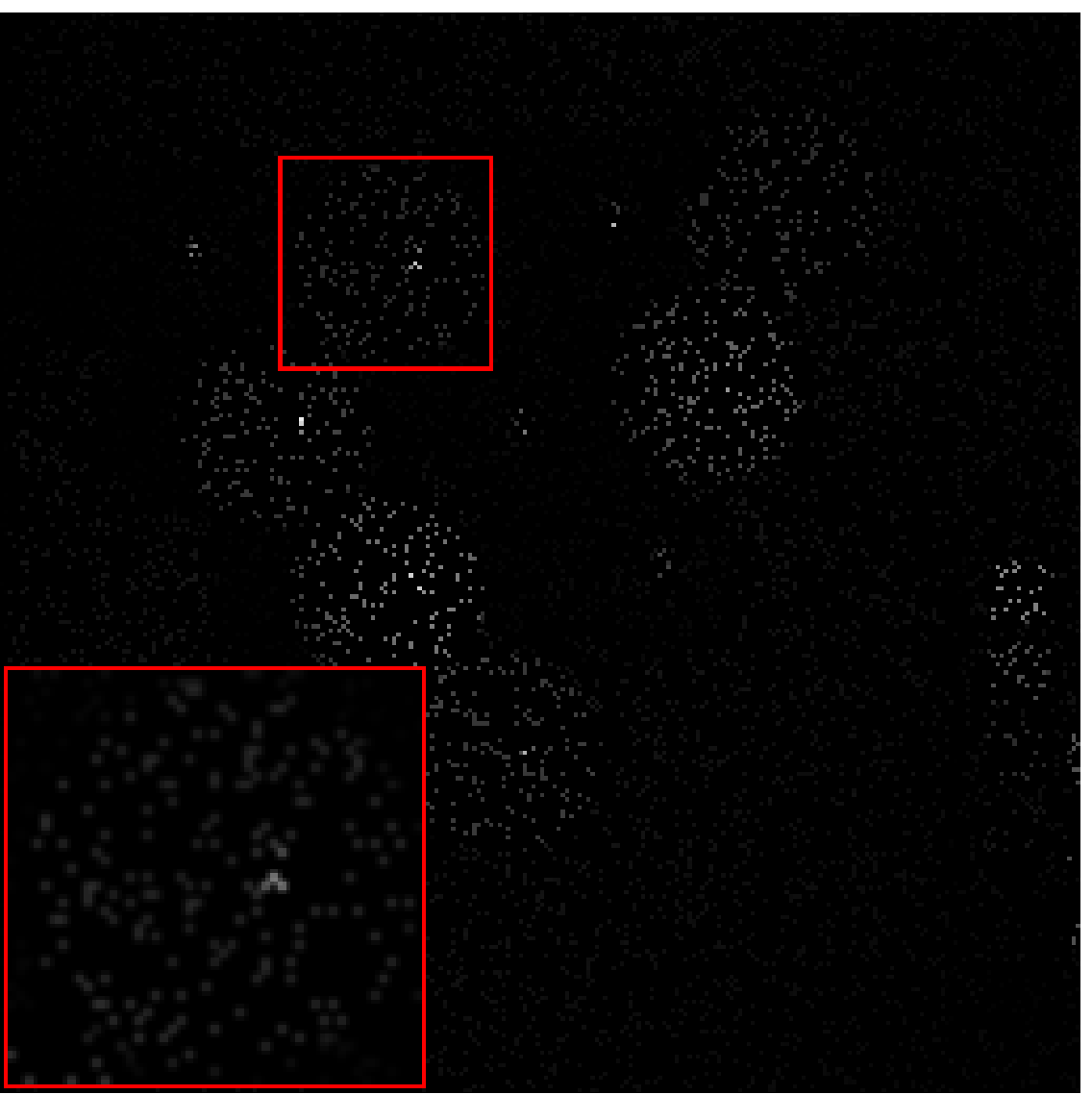}&
\includegraphics[width=0.14\textwidth]{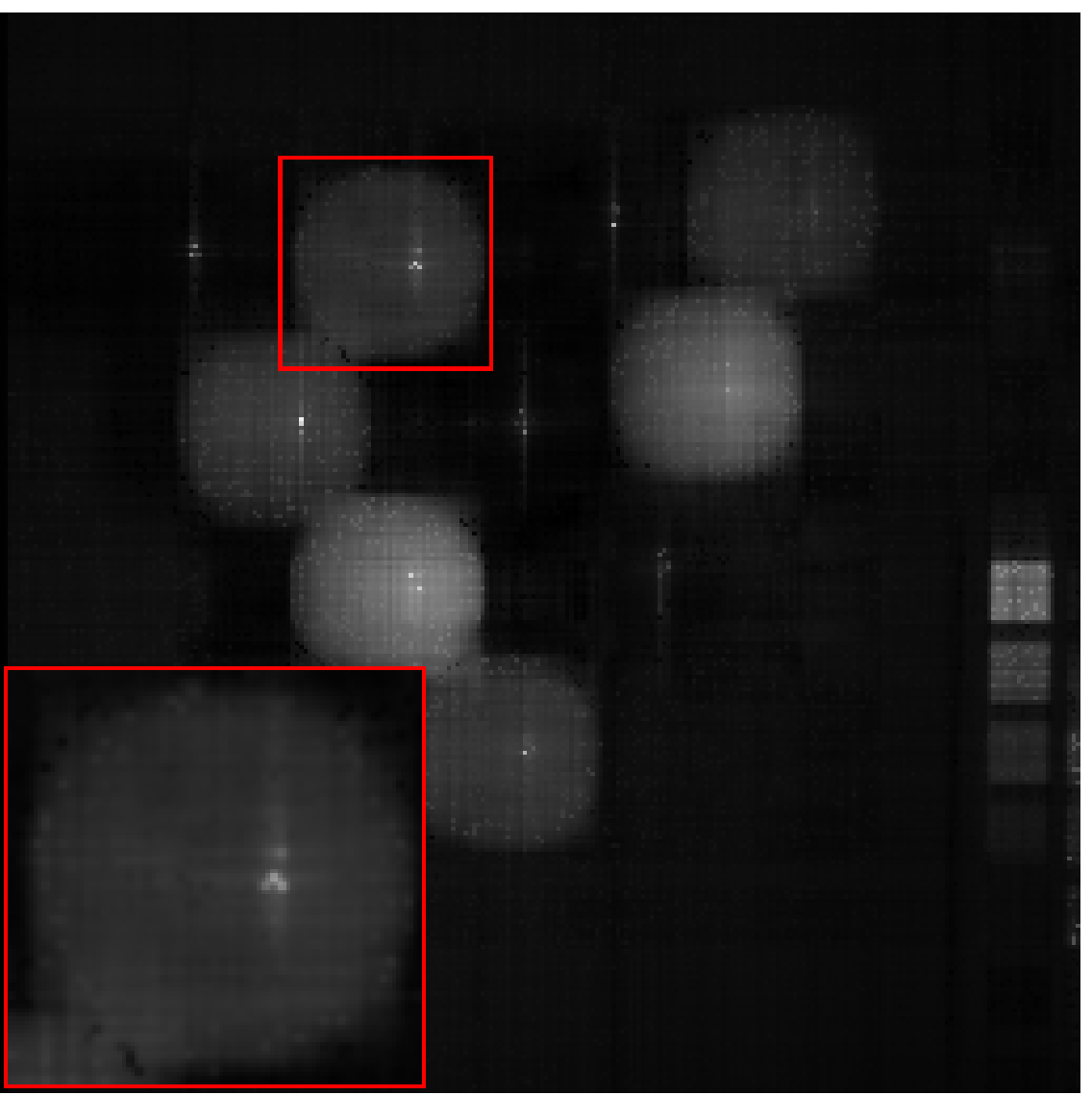}&
\includegraphics[width=0.14\textwidth]{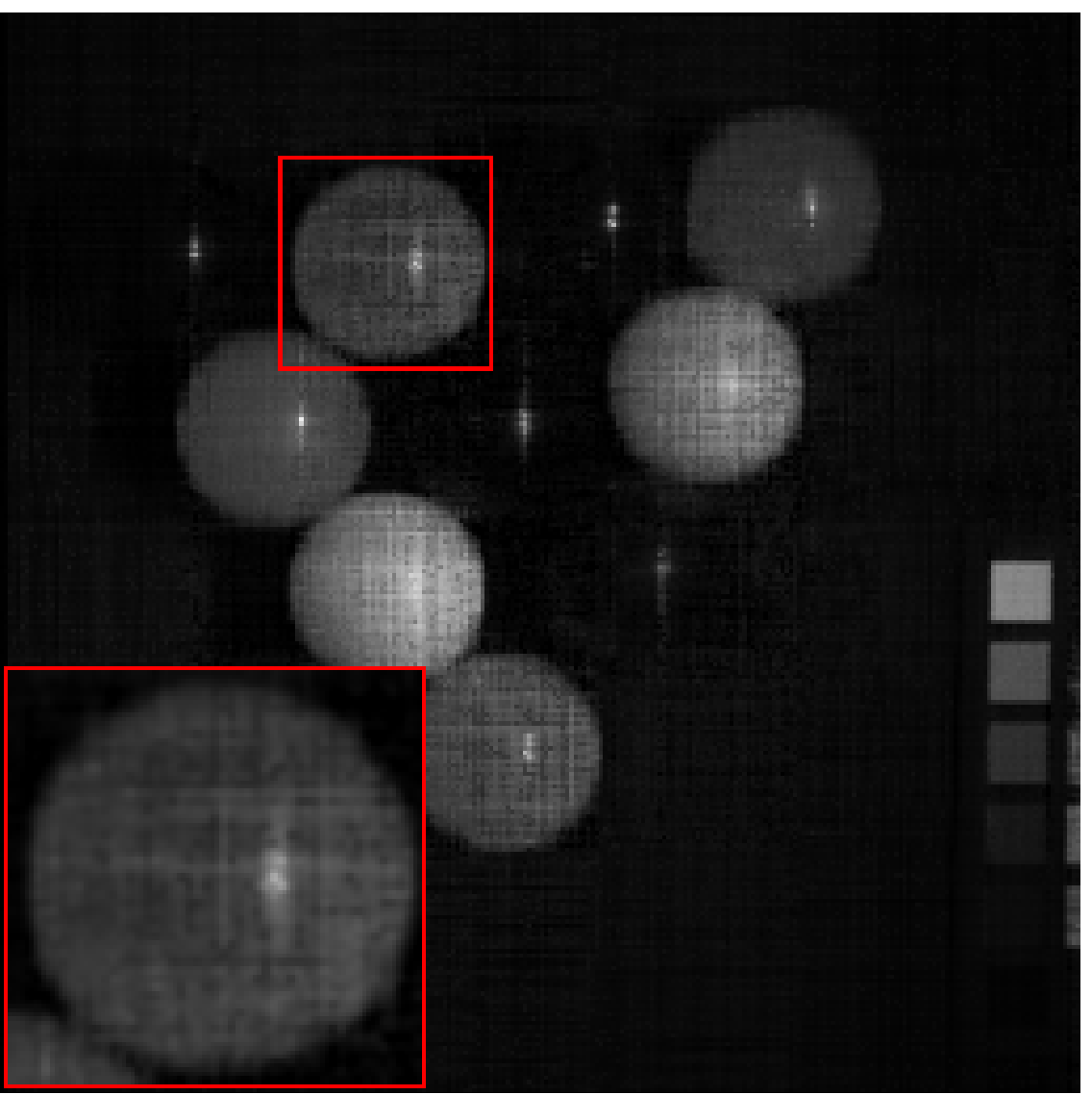}&
\includegraphics[width=0.14\textwidth]{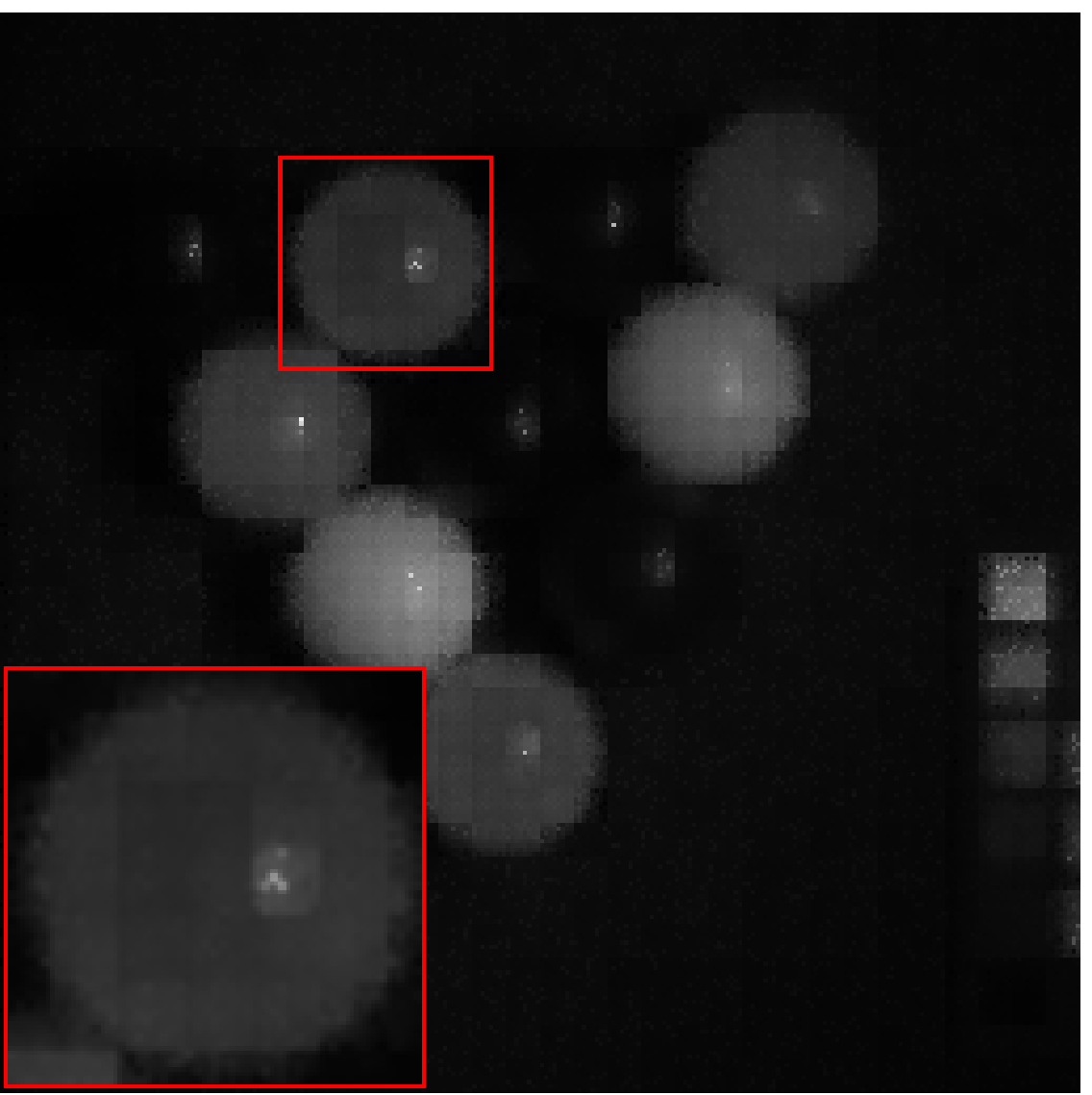}&
\includegraphics[width=0.14\textwidth]{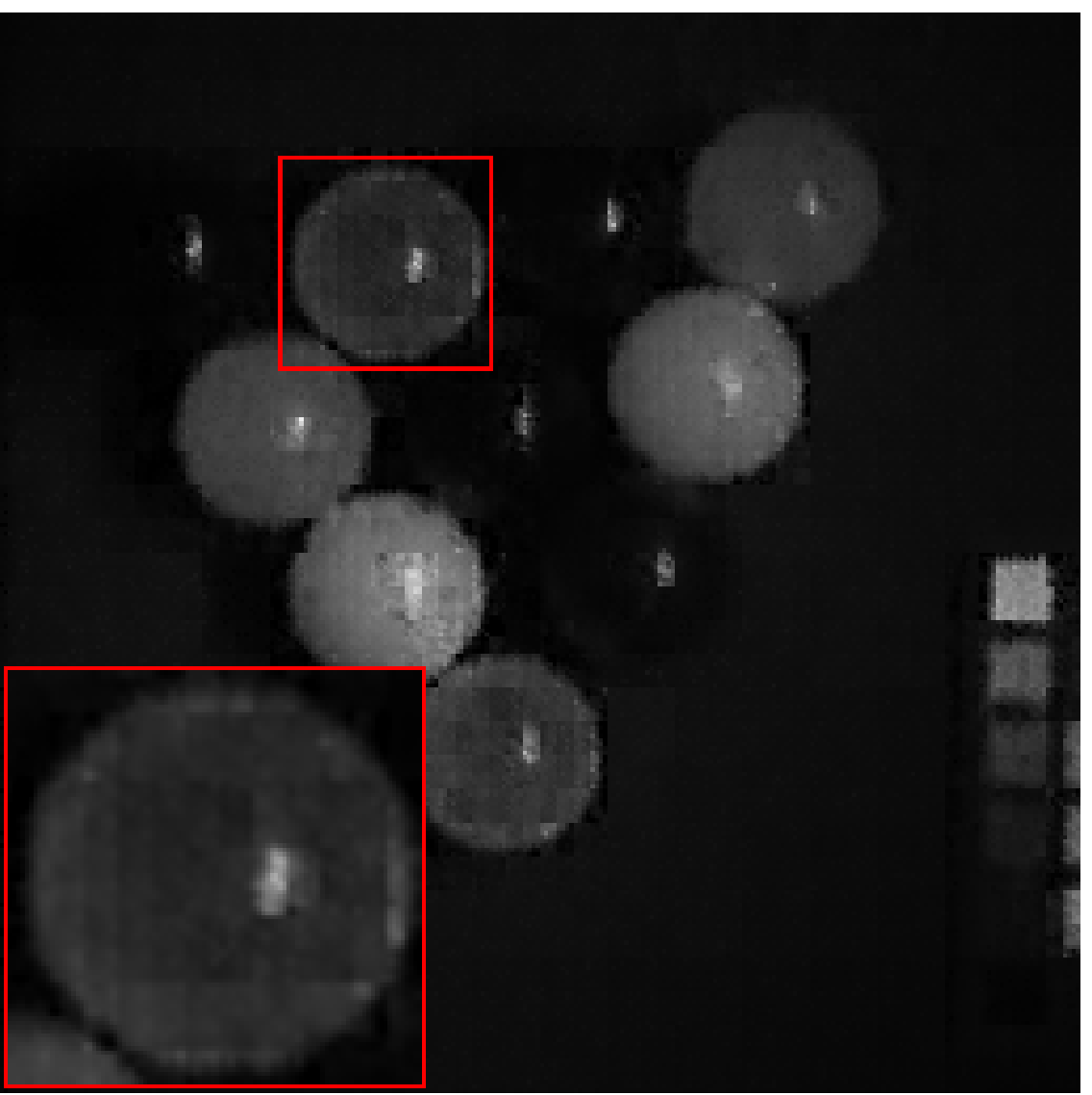}&
\includegraphics[width=0.14\textwidth]{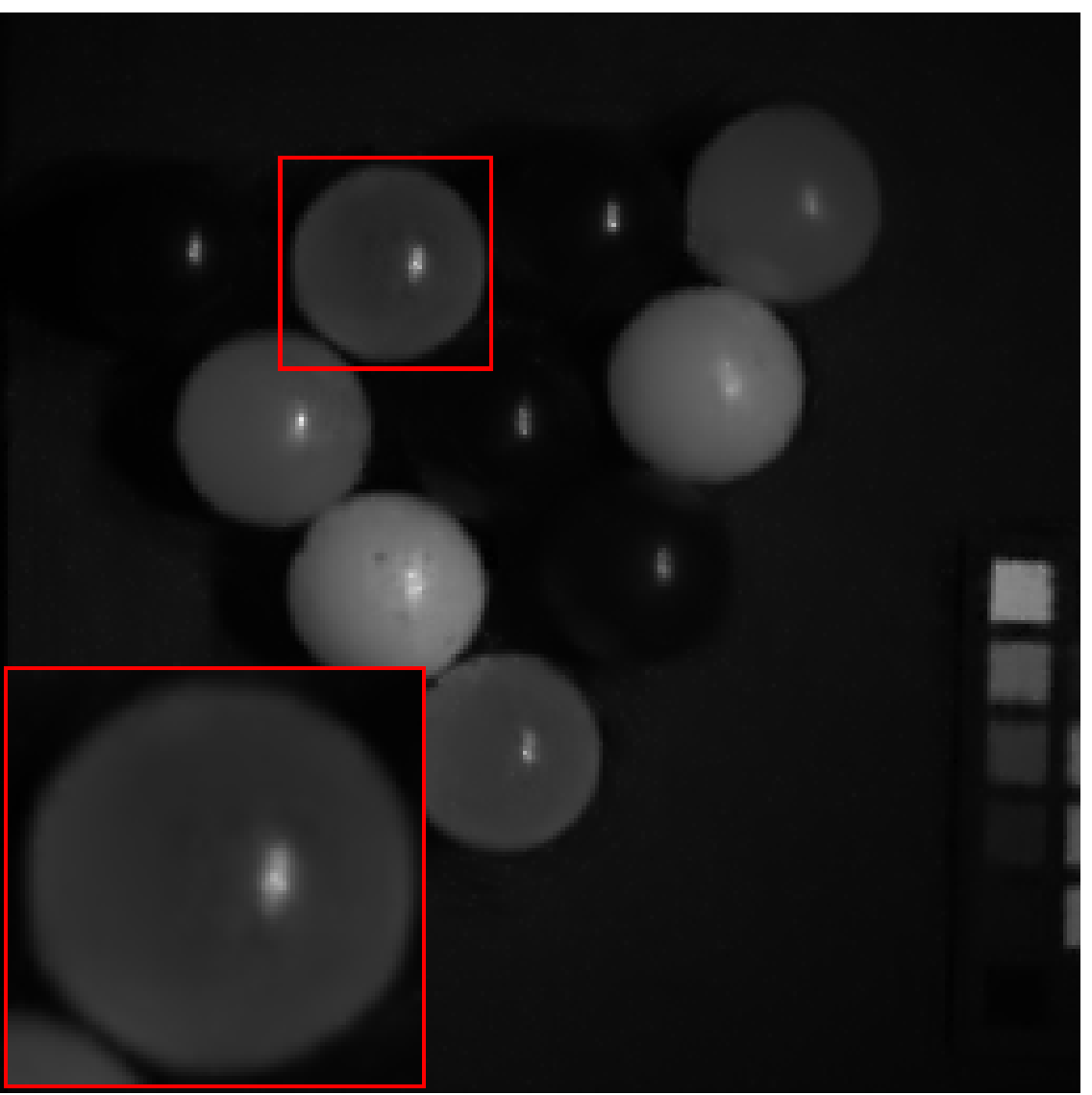}&
\includegraphics[width=0.14\textwidth]{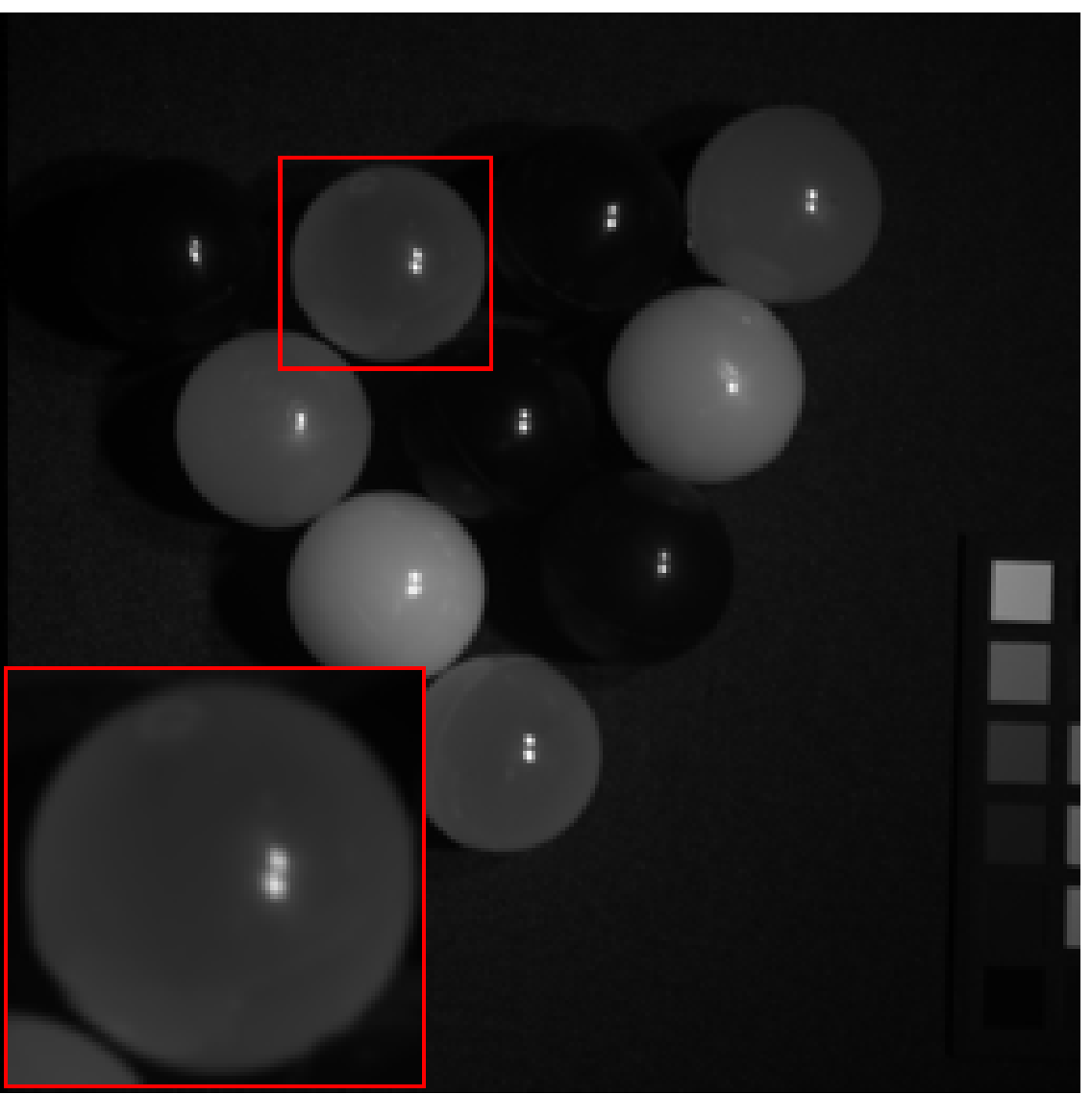}\\
 {\footnotesize\textrm{(a) Observed}} & {\footnotesize\textrm{(b) HaLRTC}} & {\footnotesize\textrm{(c) tSVD}} & {\footnotesize\textrm{(d) SiLRTC-TT}} & {\footnotesize\textrm{(e) TMac-TT}} & {\footnotesize\textrm{(f) NL-TT}}& {\footnotesize\textrm{(g) Original}}\\
\end{tabular}
\caption{\small{The results of one band of testing MSIs with $SR=0.1$ recovered by different methods. From left to right: (a) the observed image, the results by (b) HaLRTC, (c) tSVD, (d) SiLRTC-TT, (e) TMac-TT, (f) NL-TT, and (g) the original image.}}
  \label{fig:msi}
  \end{center}\vspace{-0.3cm}
\end{figure*}

\begin{figure}[!ht]
\scriptsize\setlength{\tabcolsep}{0.9pt}
\begin{center}
\begin{tabular}{ccc}
\includegraphics[width=0.32\textwidth]{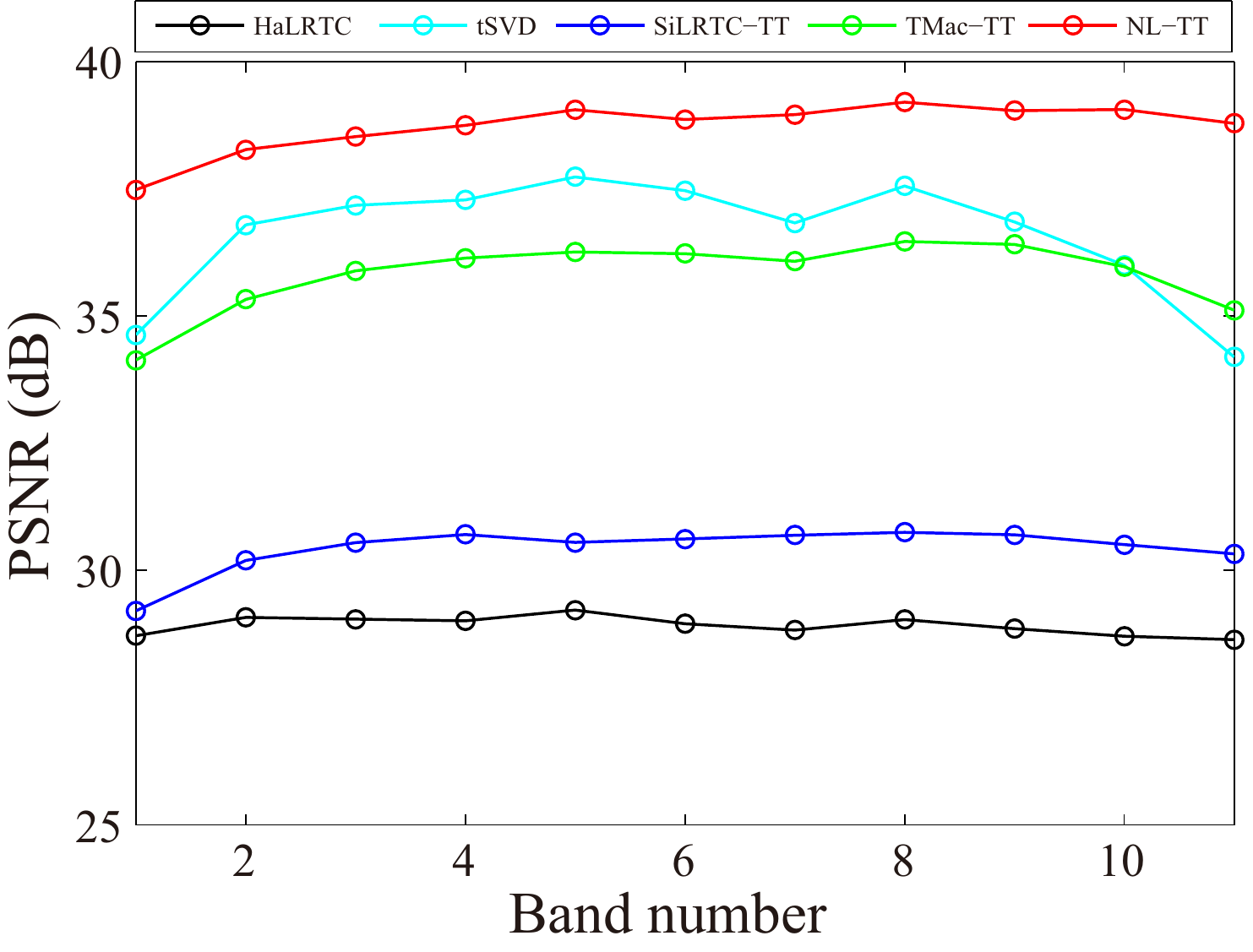}&
\includegraphics[width=0.32\textwidth]{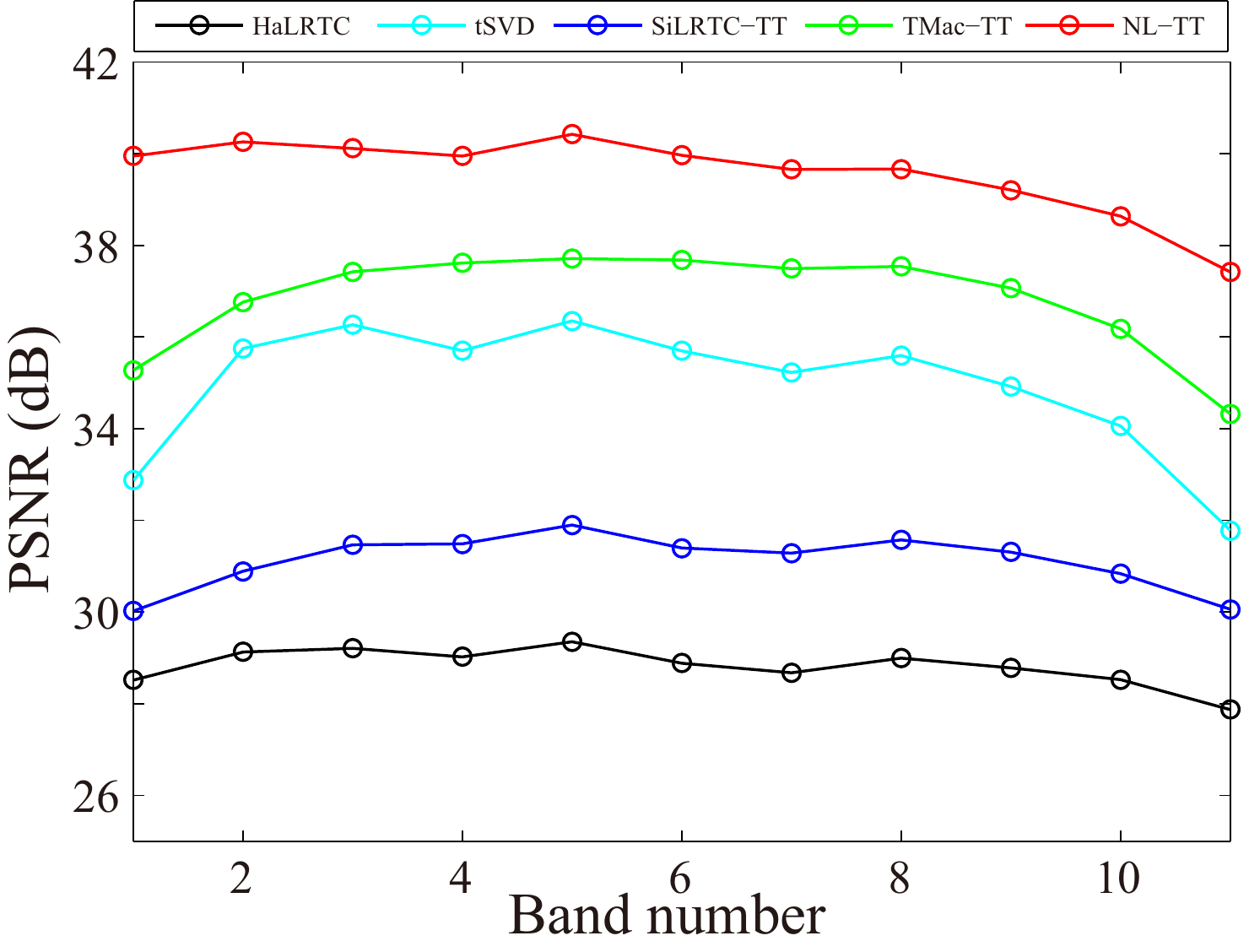}&
\includegraphics[width=0.32\textwidth]{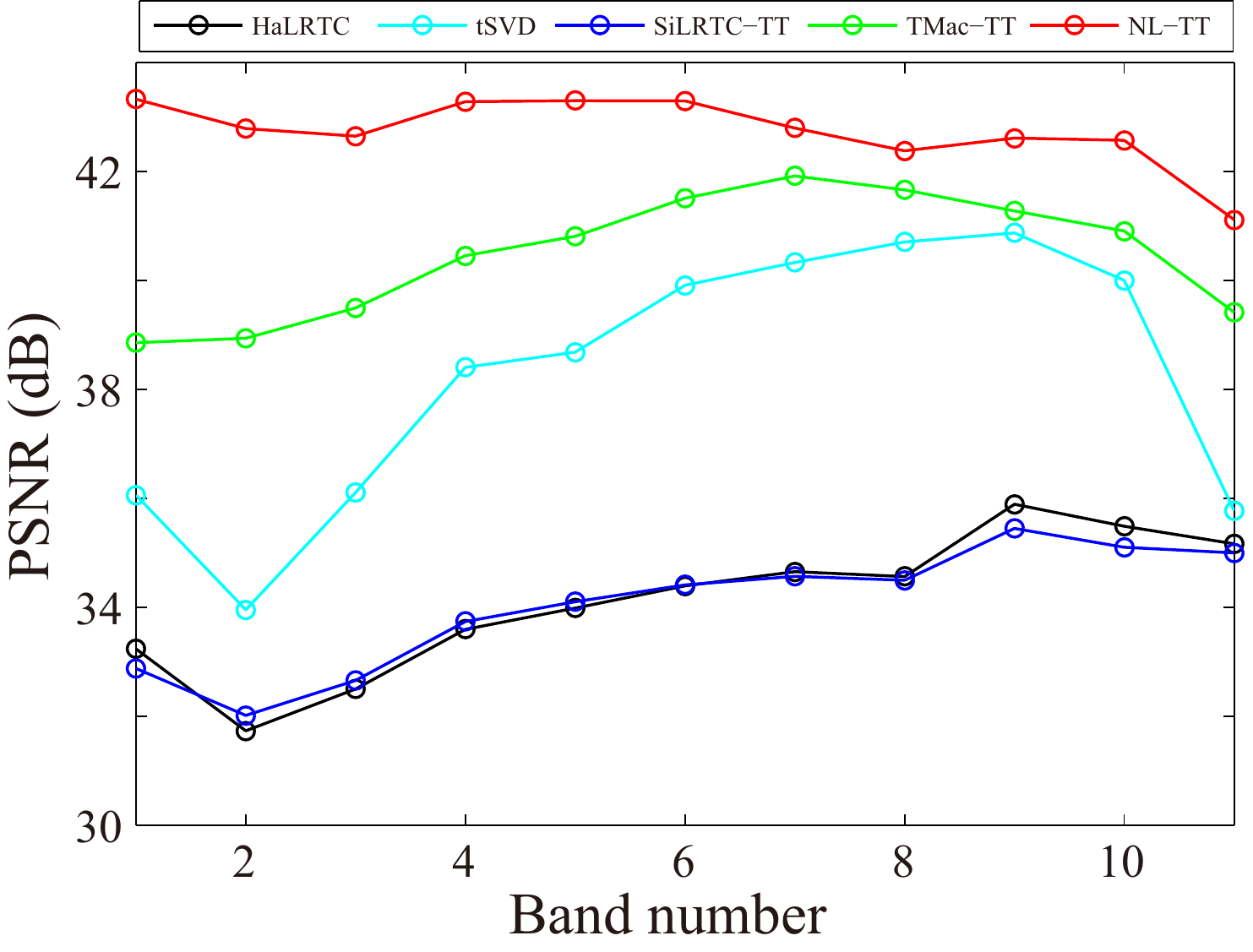}\\
\includegraphics[width=0.32\textwidth]{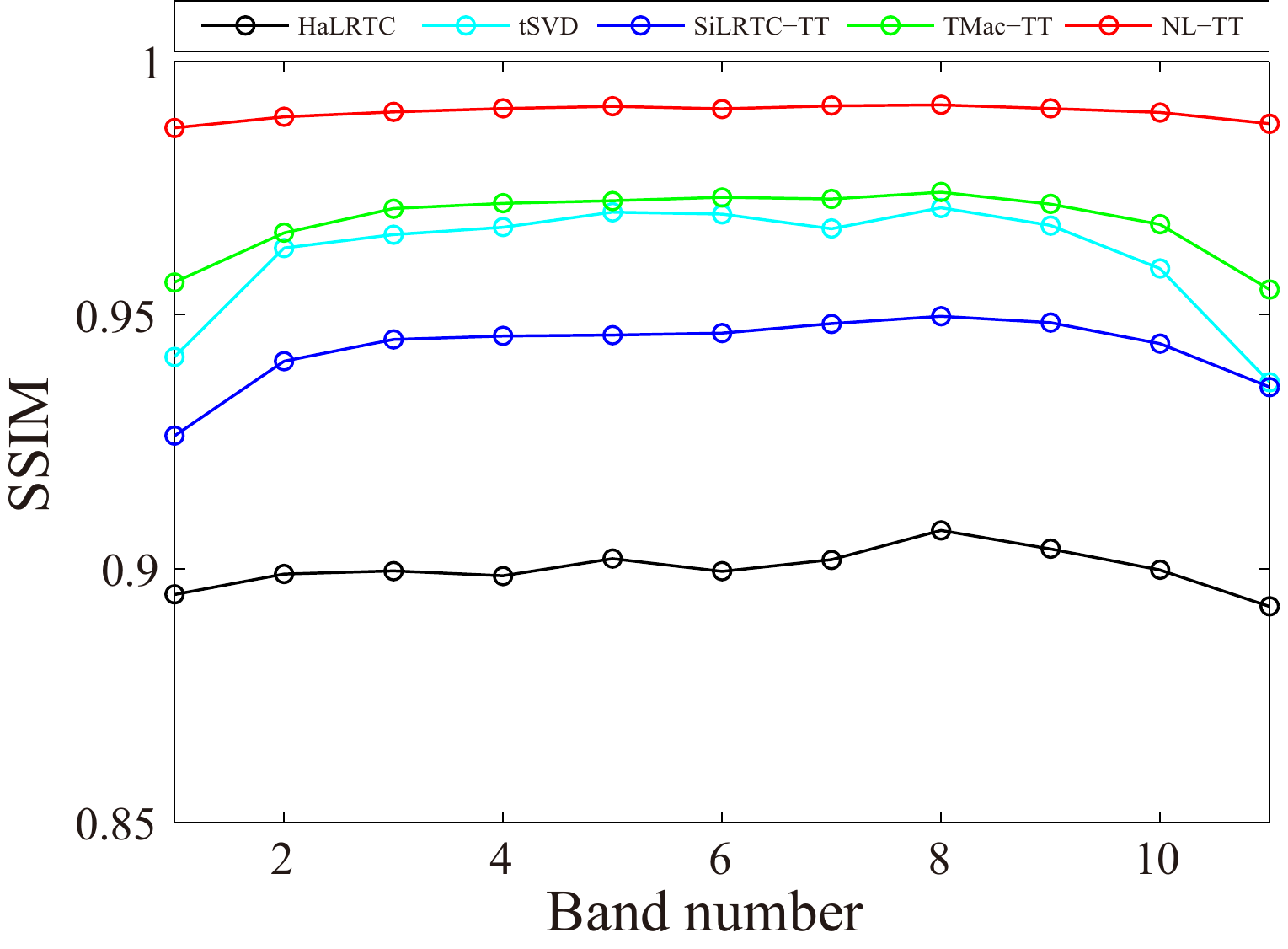}&
\includegraphics[width=0.32\textwidth]{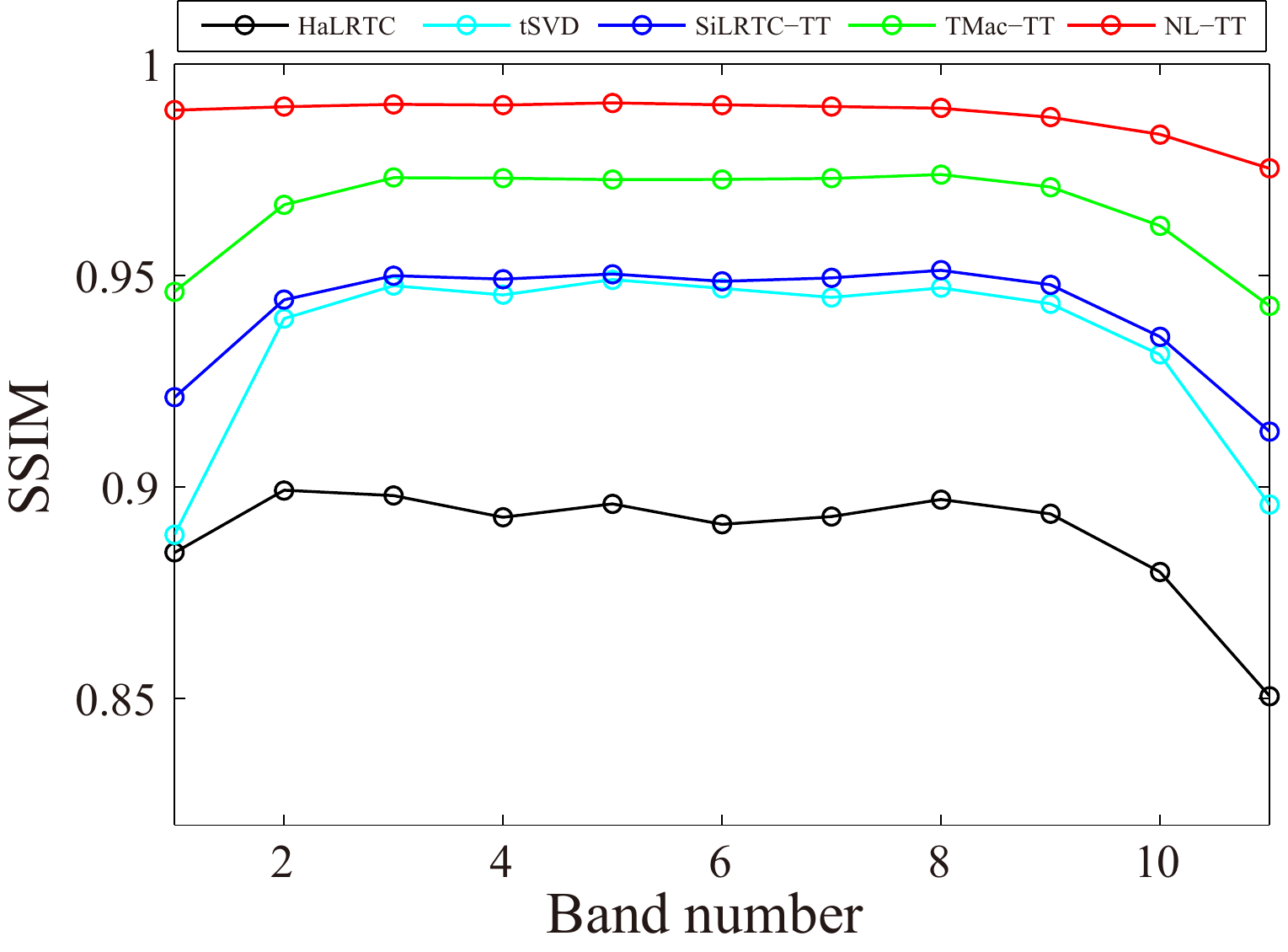}&
\includegraphics[width=0.32\textwidth]{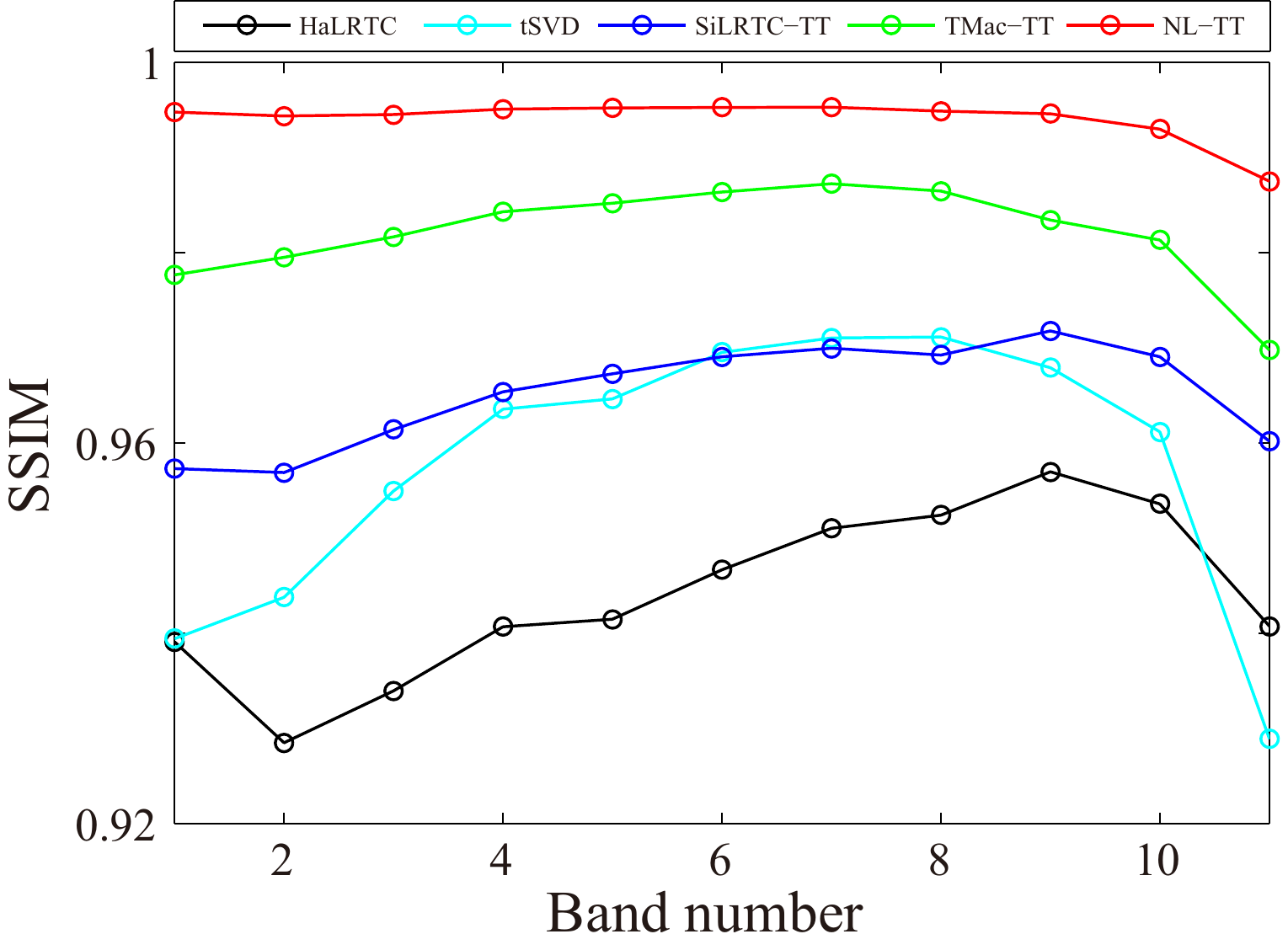} \vspace{0.1cm}\\
{\footnotesize(a)} {\small \emph{toy}} & {\footnotesize(b)} {\small \emph{feathers}} & {\footnotesize(c)} {\small \emph{superballs}}\\
\end{tabular}
\caption{\small{The PSNR and SSIM values of all bands of the reconstructed MSIs with $SR=0.2$ recovered by different methods.}}
  \label{fig:msi_all_bands_psnr}
  \end{center}\vspace{-0.3cm}
\end{figure}

\begin{figure}[!ht]
\scriptsize\setlength{\tabcolsep}{0.9pt}
\begin{center}
\begin{tabular}{cc}
\includegraphics[width=0.95\textwidth]{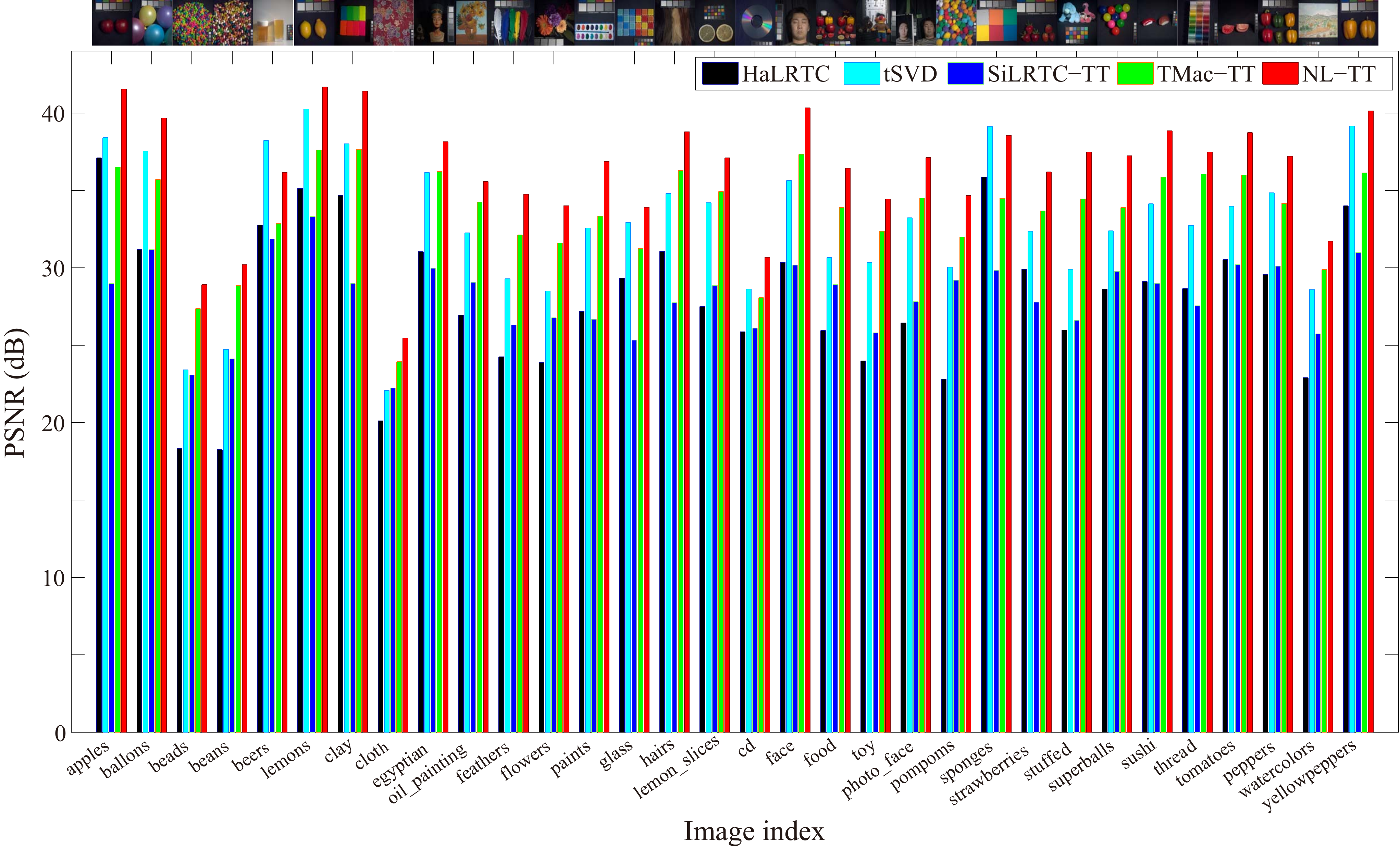} \vspace{0.1cm}\\
\end{tabular}
\caption{\small{Comparison of the PSNR values by different methods on the dataset CAVE with $SR=0.1$.}}
  \label{fig:MSI_all_psnr}
  \end{center}\vspace{-0.3cm}
\end{figure}

\begin{table}[!tb]
\renewcommand\arraystretch{1.2}
\caption{\small{The average PSNR and SSIM values obtained by HaLRTC, tSVD, SiLRTC-TT, TMac-TT and NL-TT for MSIs with different SRs.}}
\begin{center}
\begin{tabular}{c|c|cc|cc|cc}
\hline

                      \hline
\multirow{2}{*}{Image} & SR &\multicolumn{2}{c|}{0.05} &\multicolumn{2}{c|}{0.1} &\multicolumn{2}{c}{0.2} \\
\cline{3-8}
                      & Method  & PSNR  & SSIM  & PSNR  & SSIM  & PSNR  & SSIM  \\

                      \hline

\hspace{-0.2cm}\multirow{5}{*}{\emph{toy}}

                      & HaLRTC        &20.14 \hspace{-0.4cm}	&0.6519 &  23.99 \hspace{-0.4cm} &0.7790	& 28.91 \hspace{-0.4cm} &0.8994\\

                      & tSVD          &25.89 \hspace{-0.4cm} &0.7680	&30.34 \hspace{-0.4cm} 	&0.8844	&36.57 \hspace{-0.4cm}	&0.9602\\

                      &SiLRTC-TT      &22.36 \hspace{-0.4cm}	&0.7138	&25.81 \hspace{-0.4cm} 	&0.8392	&30.44 \hspace{-0.4cm}	&0.9433\\

                      & TMac-TT        &27.28 \hspace{-0.4cm}	&0.8329	&32.37 \hspace{-0.4cm}	&0.9317	&35.74 \hspace{-0.4cm}	&0.9669\\

                      & NL-TT         &  \textbf{29.58} \hspace{-0.4cm}	&\textbf{0.9243}  &\textbf{34.44} \hspace{-0.4cm}	&\textbf{0.9730}	&\textbf{38.72} \hspace{-0.4cm} &\textbf{0.9899}\\
                      \hline
\hspace{-0.2cm}\multirow{5}{*}{\emph{feathers}}
                      & HaLRTC         & 20.66 \hspace{-0.4cm}   &0.6422	&24.26 \hspace{-0.4cm} &0.7720	&28.81 \hspace{-0.4cm} 	&0.8876\\

                      & tSVD           &25.15 \hspace{-0.4cm} 	&0.6886	&29.29 \hspace{-0.4cm} &0.8266	&34.82 \hspace{-0.4cm} &0.9265\\

                      &\hspace{-0.7cm} SiLRTC-TT \hspace{-0.7cm}      & 22.86 \hspace{-0.4cm}    &0.7196	&26.32 \hspace{-0.4cm}  &0.8417	&31.11 \hspace{-0.4cm} 	&0.9411\\

                      & TMac-TT       &  27.29  \hspace{-0.4cm}	&0.7611	&32.12 \hspace{-0.4cm}  &0.9190	&36.63 \hspace{-0.4cm} 	&0.9631\\

                      & NL-TT         &  \textbf{29.61} \hspace{-0.4cm} 	&\textbf{0.9102}	&\textbf{34.76} \hspace{-0.4cm} &\textbf{0.9699}	&\textbf{39.56} \hspace{-0.4cm}	&\textbf{0.9879}\\
                      \hline
\hspace{-0.25cm}\multirow{5}{*}{\emph{superballs}} \hspace{-0.25cm}
                      & HaLRTC         & 23.28  \hspace{-0.4cm}	&0.7661	&28.63 \hspace{-0.4cm} 	&0.8621	&34.10 \hspace{-0.4cm} 	&0.9426\\

                      & tSVD          & 28.24 \hspace{-0.4cm} &0.7636	&32.39 \hspace{-0.4cm} 	&0.8663	&38.20 \hspace{-0.4cm} 	&0.9564\\

                      &\hspace{-0.7cm} SiLRTC-TT \hspace{-0.7cm}      & 26.27 \hspace{-0.4cm} 	&0.8290	&29.79 \hspace{-0.4cm} 	&0.9087	&34.03 \hspace{-0.4cm} 	&0.9651 \\

                      & TMac-TT        &  29.97 \hspace{-0.4cm} 	&0.8343	&33.90 \hspace{-0.4cm}	&0.9346	&40.19  \hspace{-0.4cm}	&0.9803\\

                      & NL-TT        & \textbf{32.93} \hspace{-0.4cm} 	&\textbf{0.9507}	&\textbf{37.25} \hspace{-0.4cm}	&\textbf{0.9812}	&\textbf{42.67} \hspace{-0.4cm}	 &\textbf{0.9939}\\
                      \hline

\hline
\end{tabular}
\end{center}
\label{table:MSI}
\end{table}

We show one band recovered results in \emph{toy}, \emph{feathers}, and \emph{superballs} by different methods in Fig. \ref{fig:msi} and display the PSNR and SSIM values of each band with $SR = 0.2$ in Fig. \ref{fig:msi_all_bands_psnr}. From Fig. \ref{fig:msi}, the proposed method achieves the best results in preserving the textures and details. From Fig. \ref{fig:msi_all_bands_psnr}, we see that our method performs higher PSNR and SSIM values than other methods for all bands. Fig. \ref{fig:MSI_all_psnr} lists the comparison of the PSNR values on all 32 MSIs. Table \ref{table:MSI} lists the average performance (over different SRs) of all methods. From these quantitative comparisons, we observe that our method outperforms other competing methods with respect to PSNR and SSIM.

\begin{figure*}[!ht]
\scriptsize\setlength{\tabcolsep}{0.9pt}
\begin{center}
\begin{tabular}{cccccccc}
\includegraphics[width=0.14\textwidth]{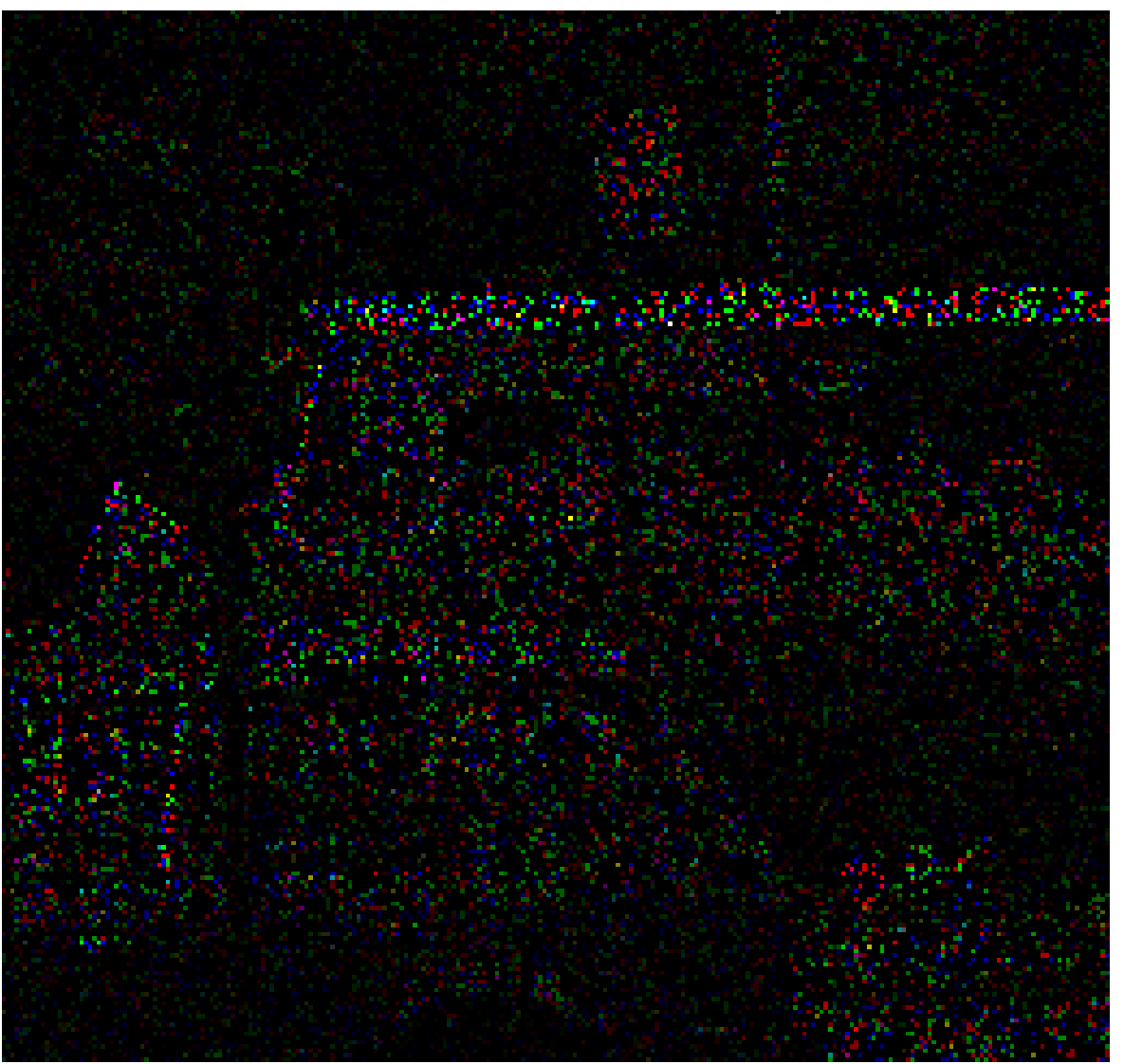}&
\includegraphics[width=0.14\textwidth]{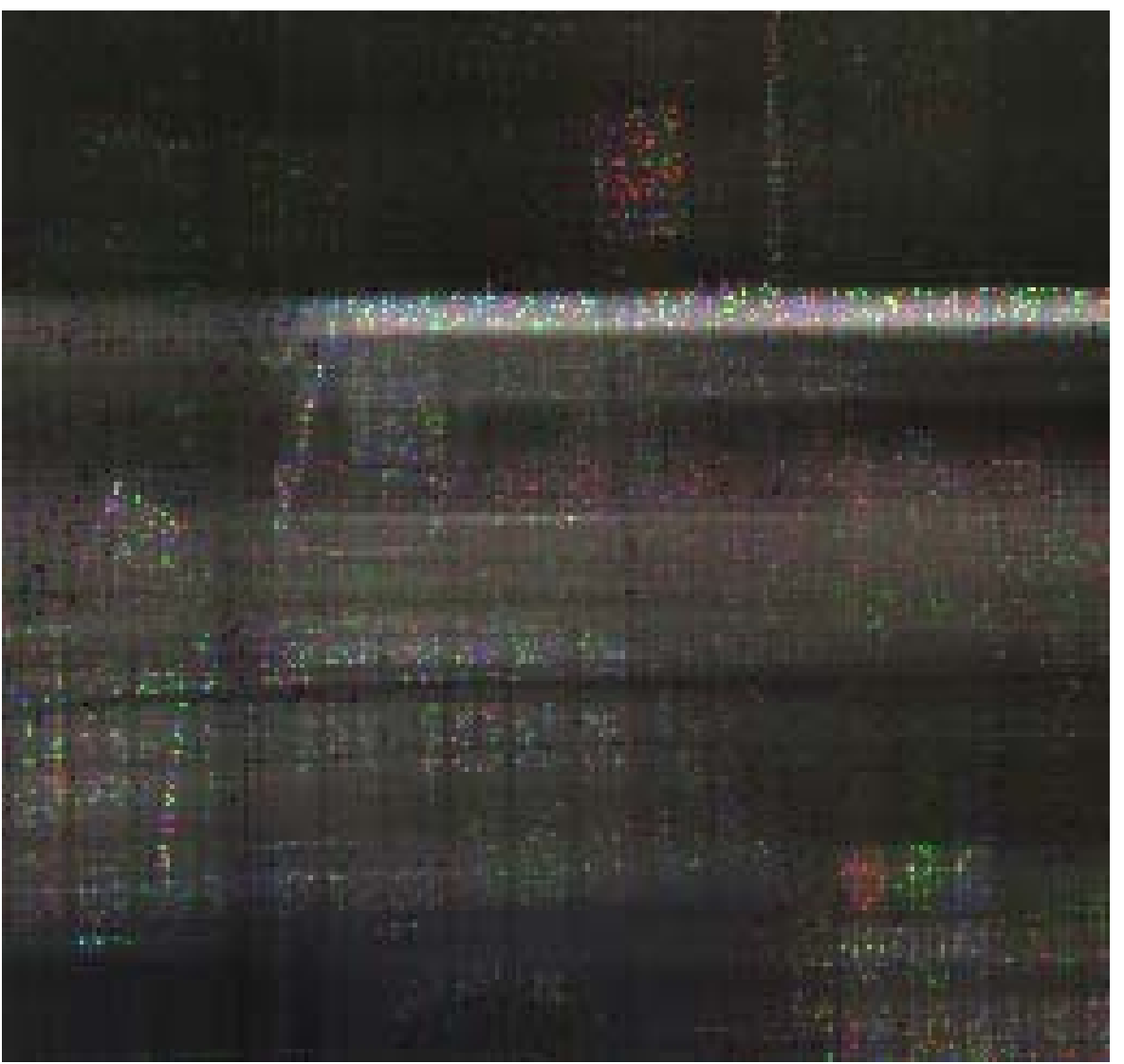}&
\includegraphics[width=0.14\textwidth]{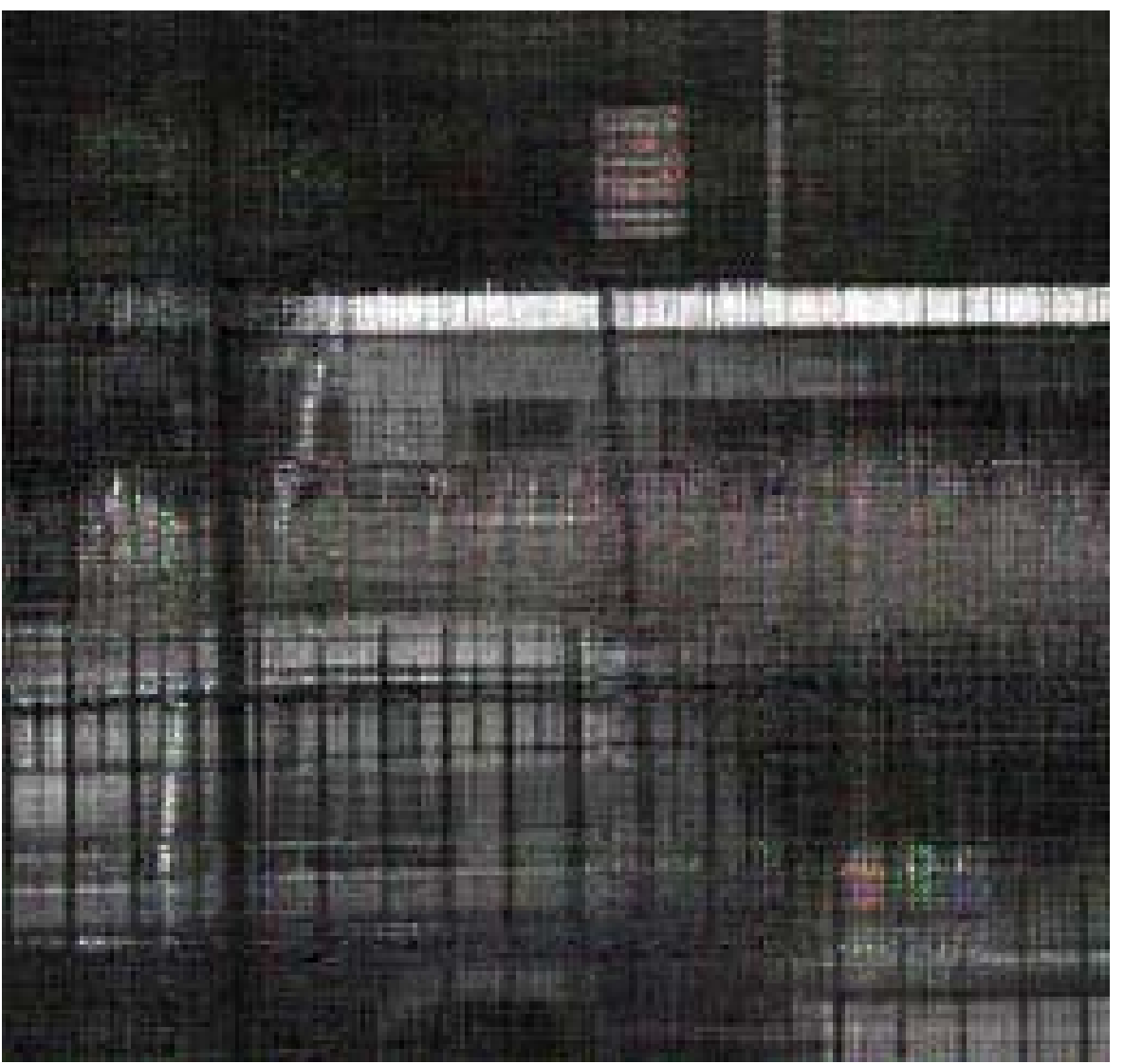}&
\includegraphics[width=0.14\textwidth]{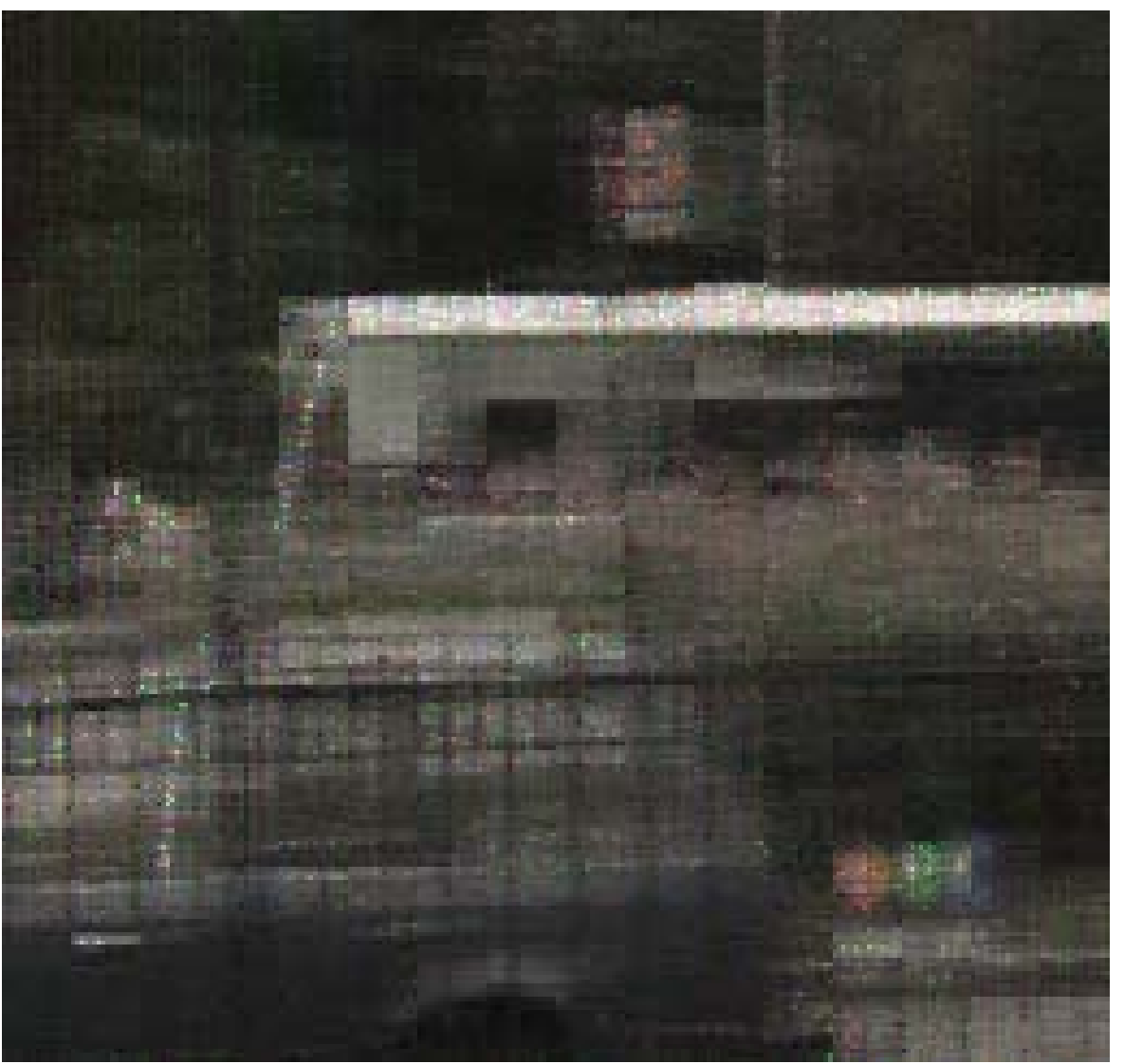}&
\includegraphics[width=0.14\textwidth]{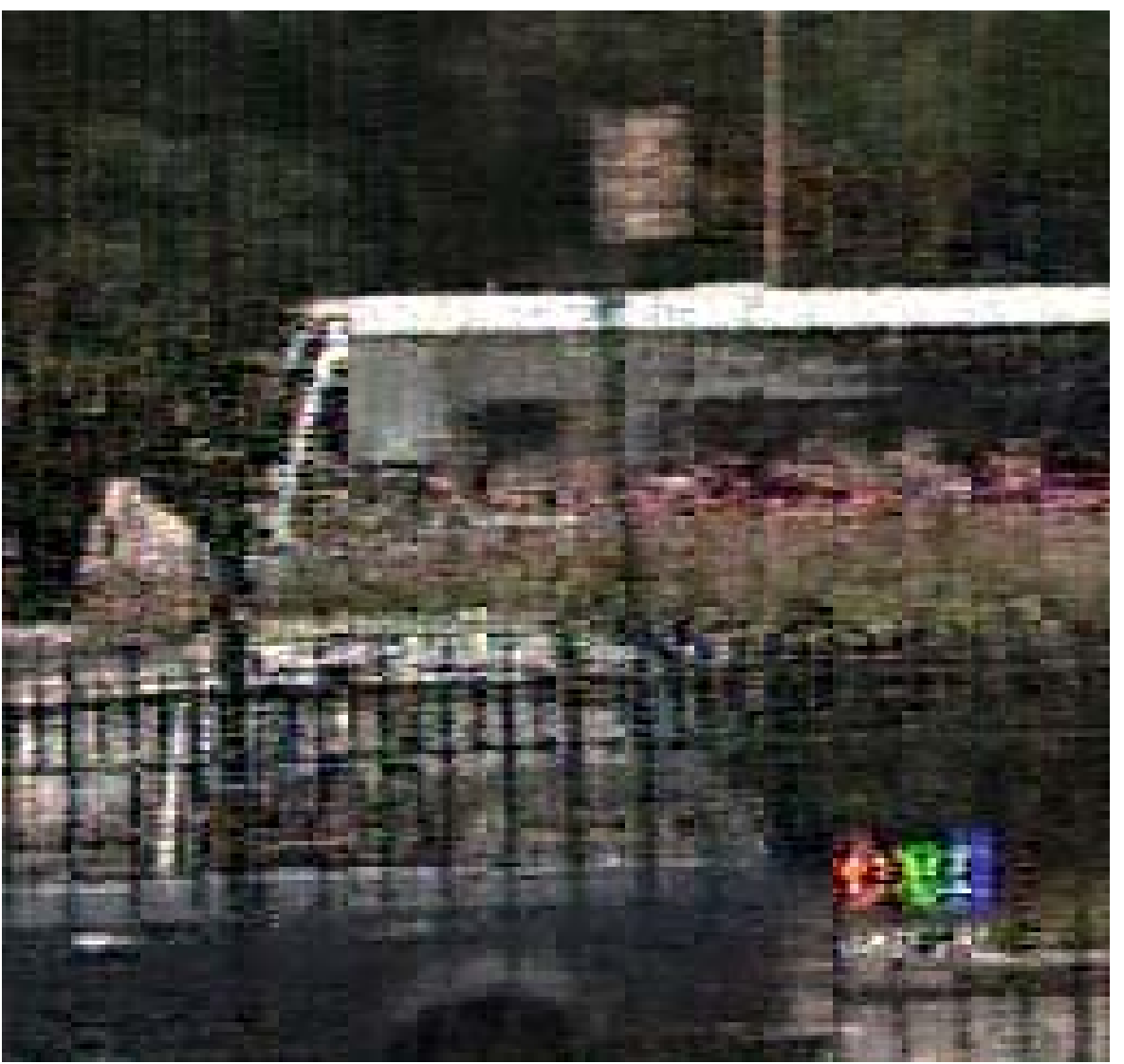}&
\includegraphics[width=0.14\textwidth]{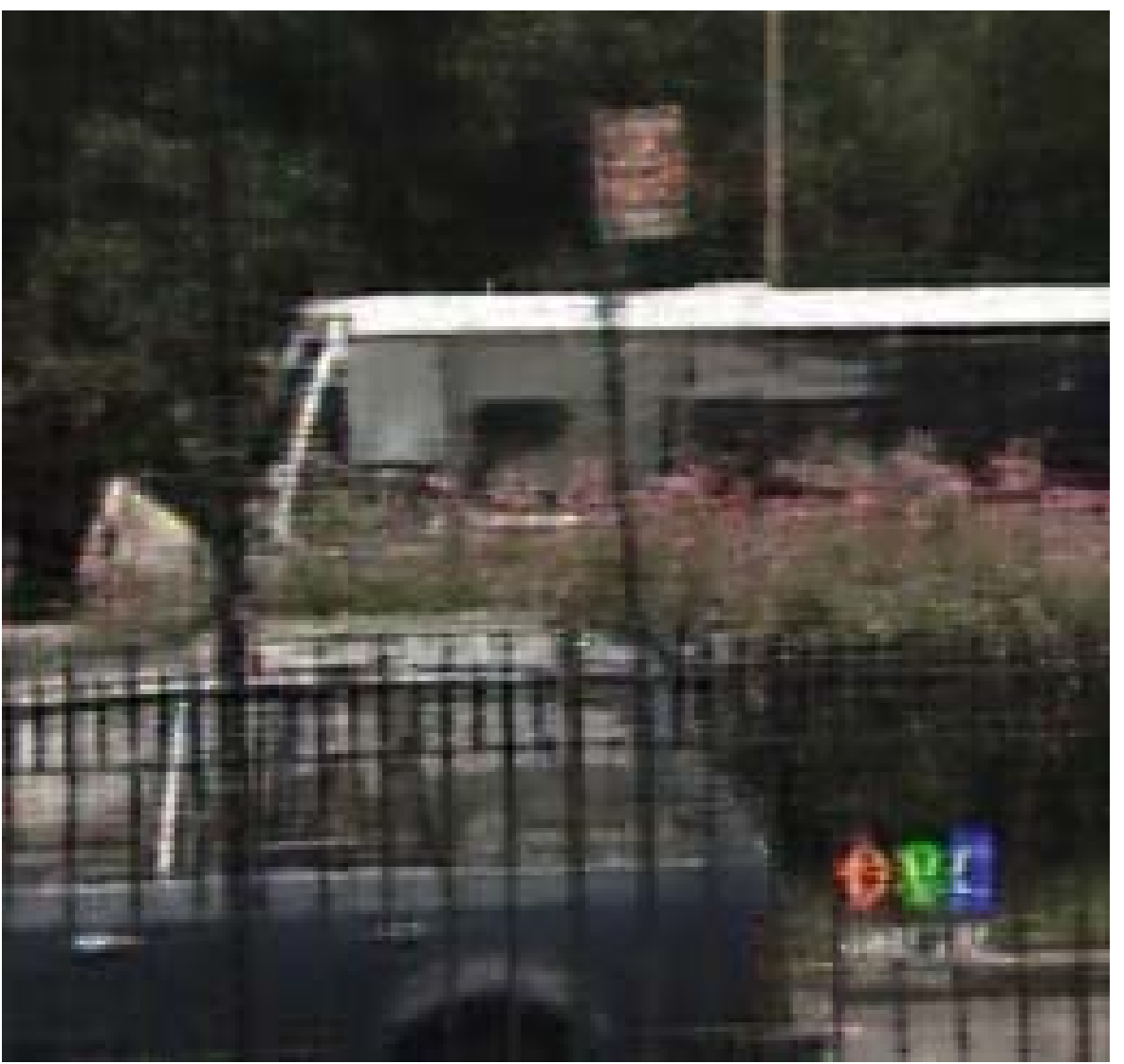}&
\includegraphics[width=0.14\textwidth]{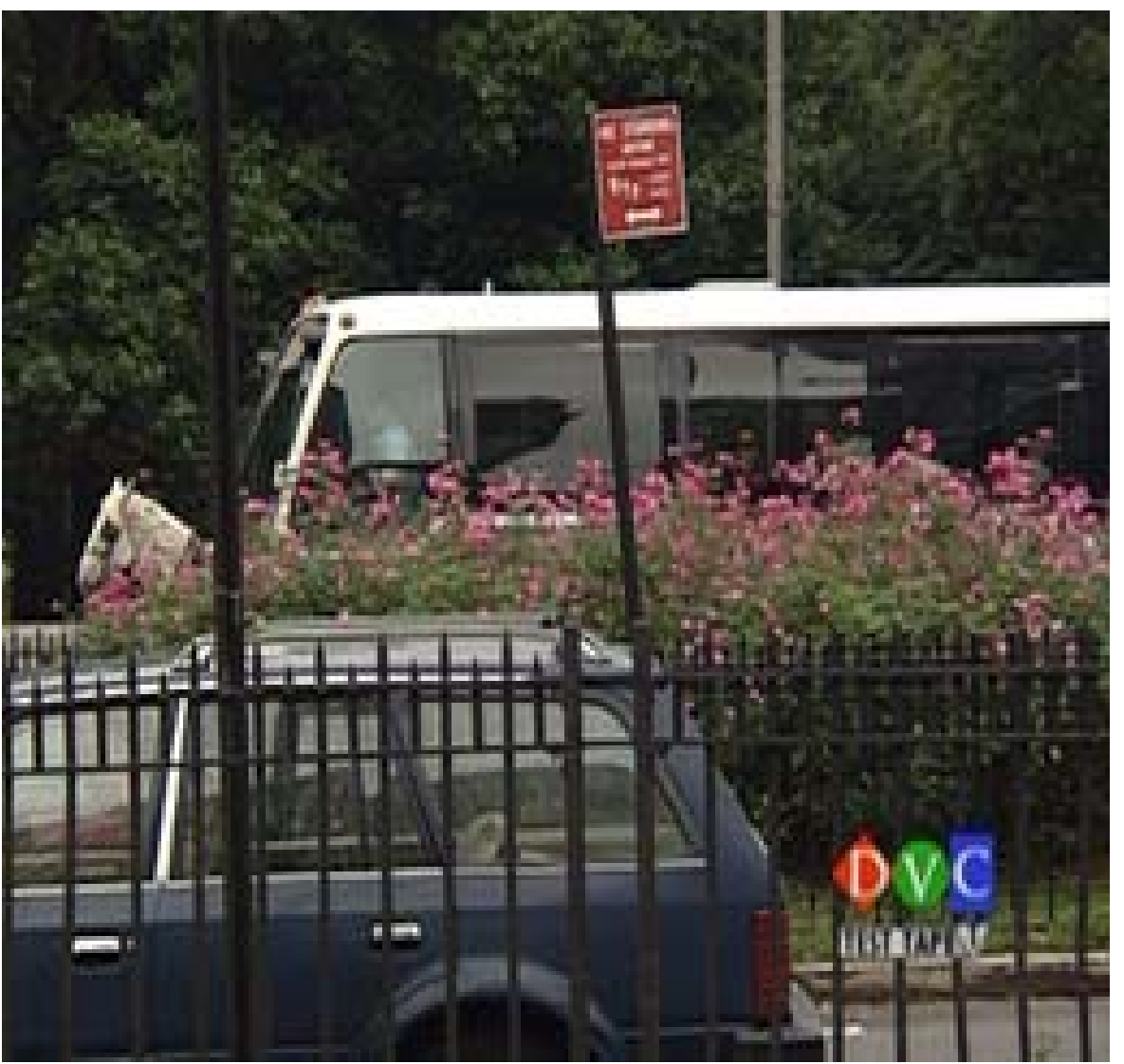}\\
\includegraphics[width=0.14\textwidth]{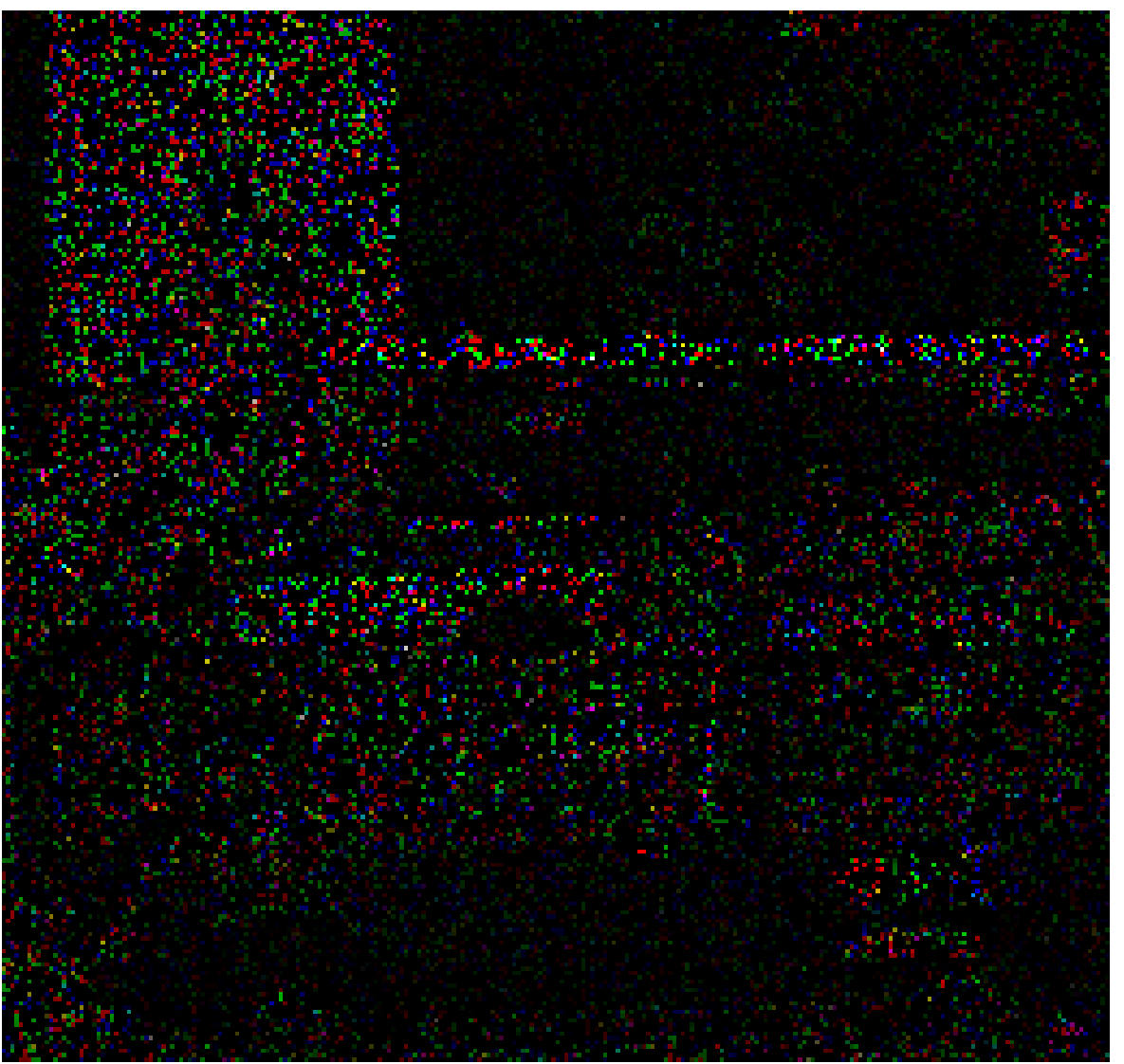}&
\includegraphics[width=0.14\textwidth]{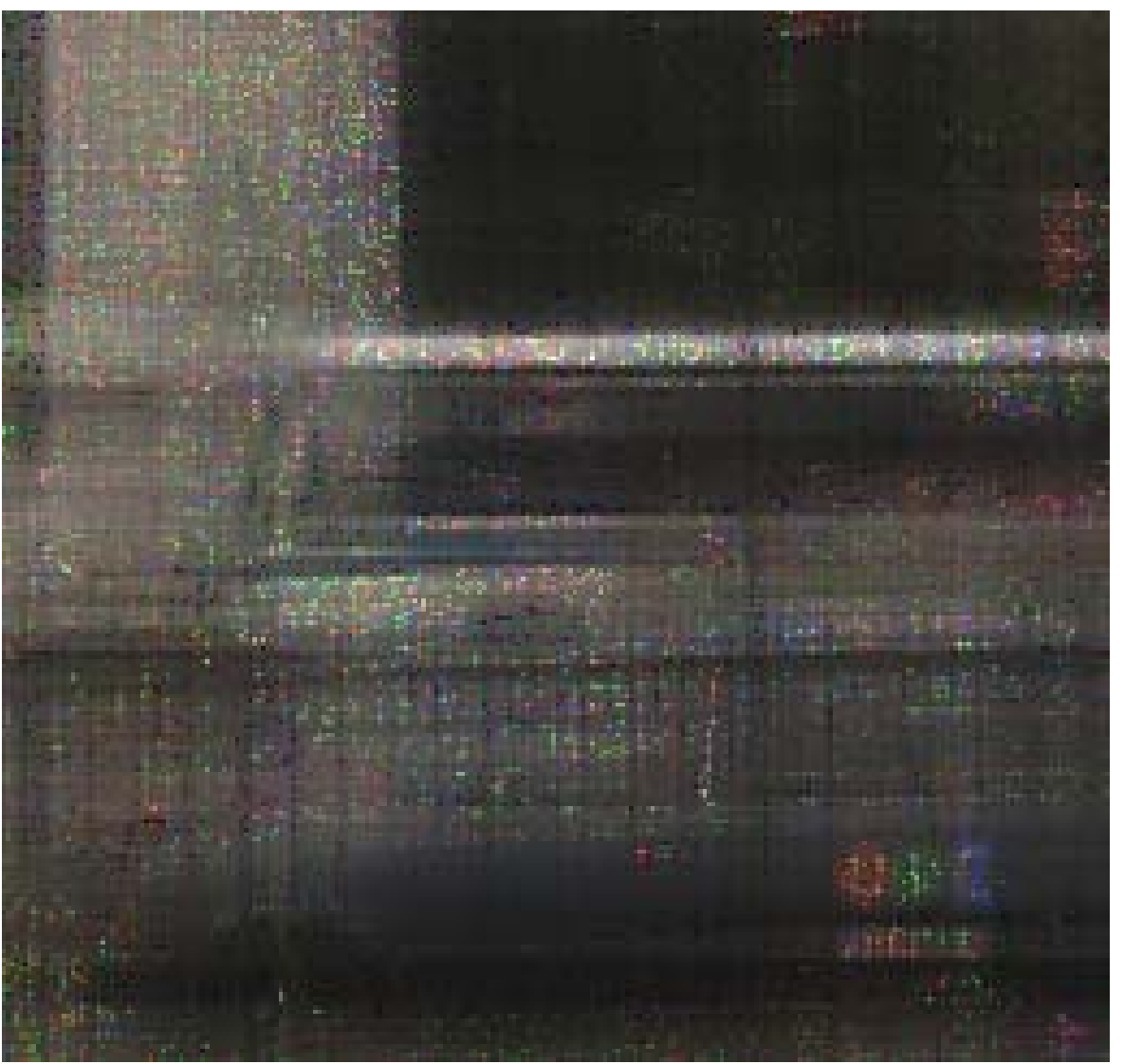}&
\includegraphics[width=0.14\textwidth]{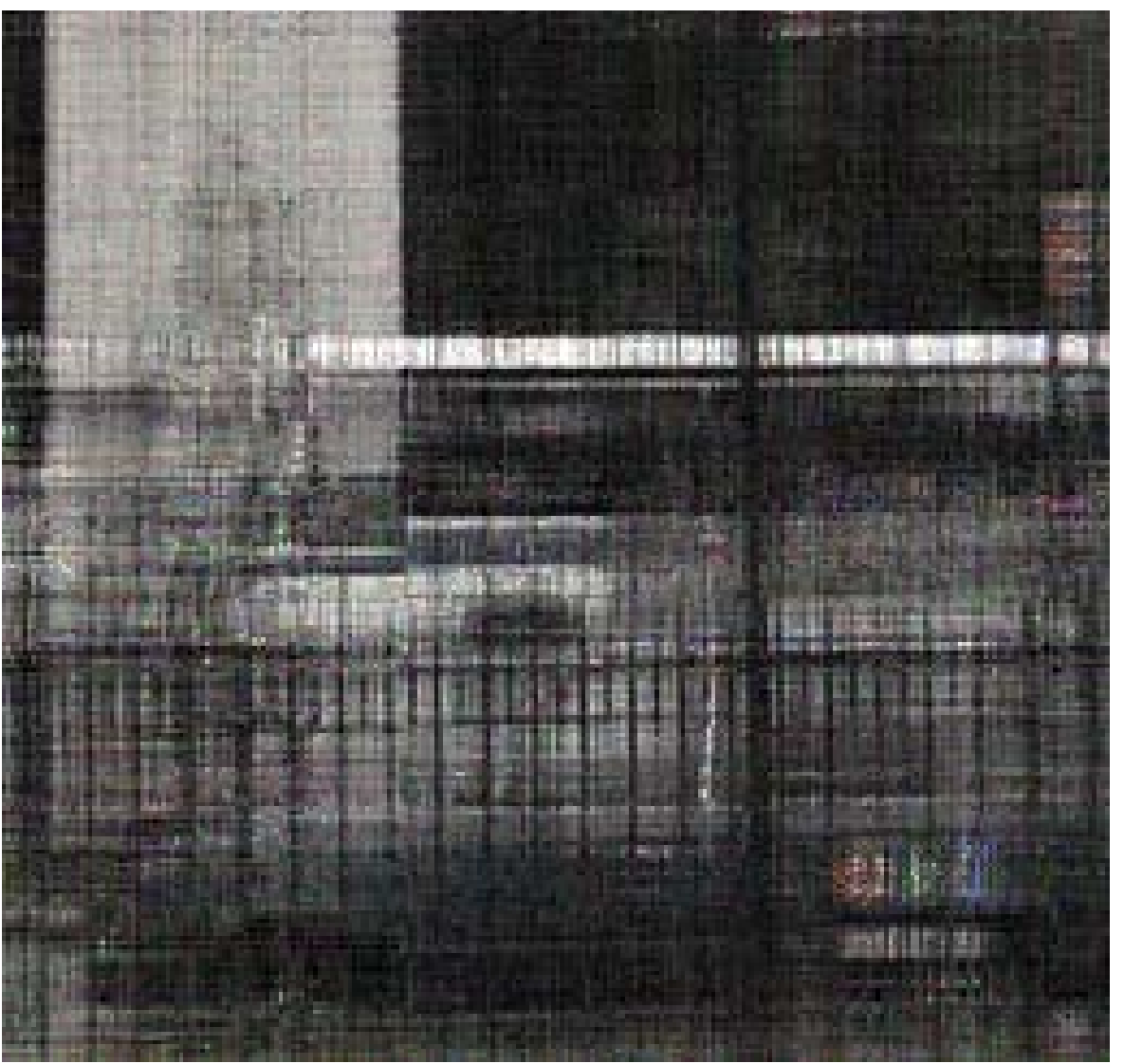}&
\includegraphics[width=0.14\textwidth]{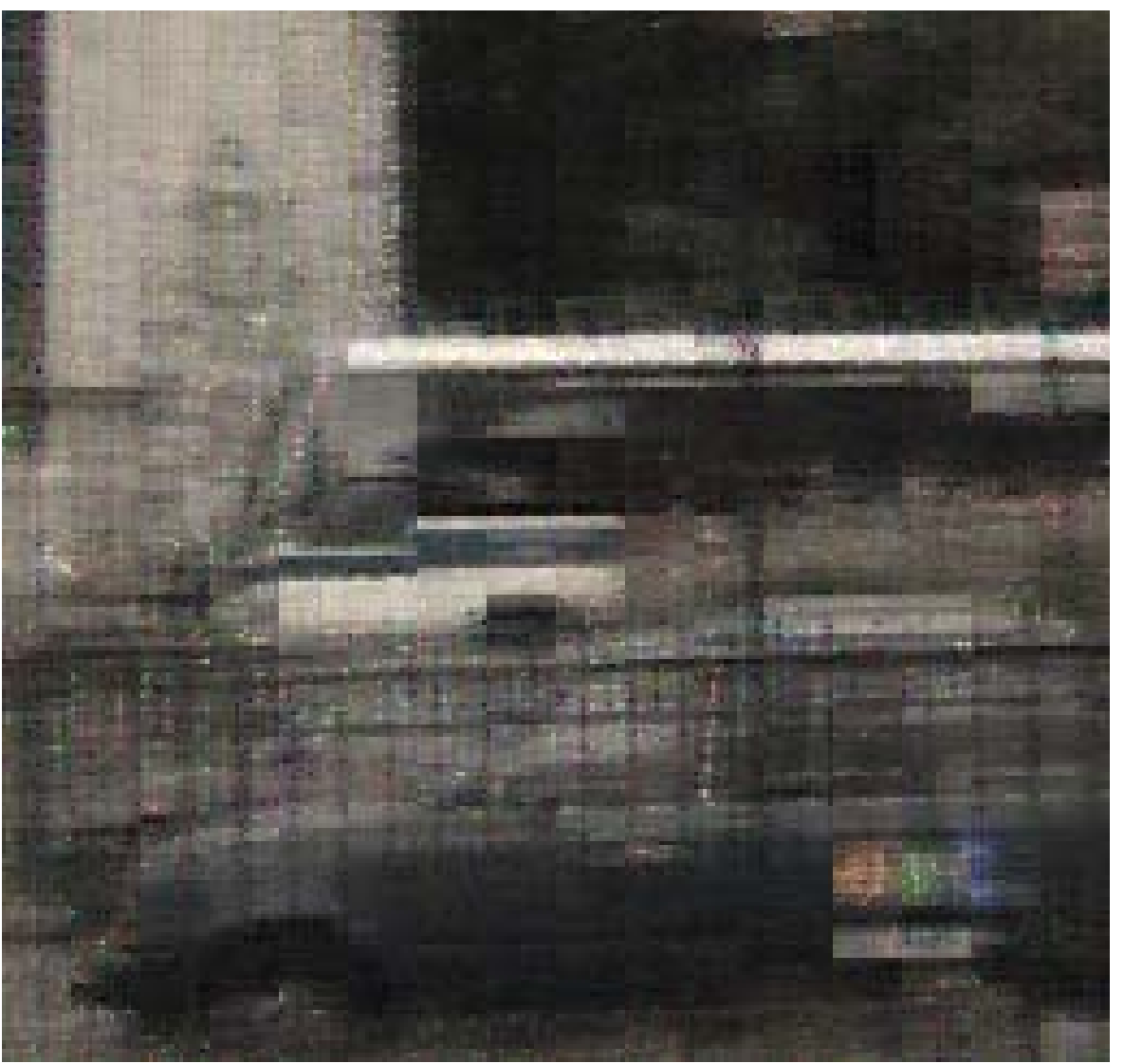}&
\includegraphics[width=0.14\textwidth]{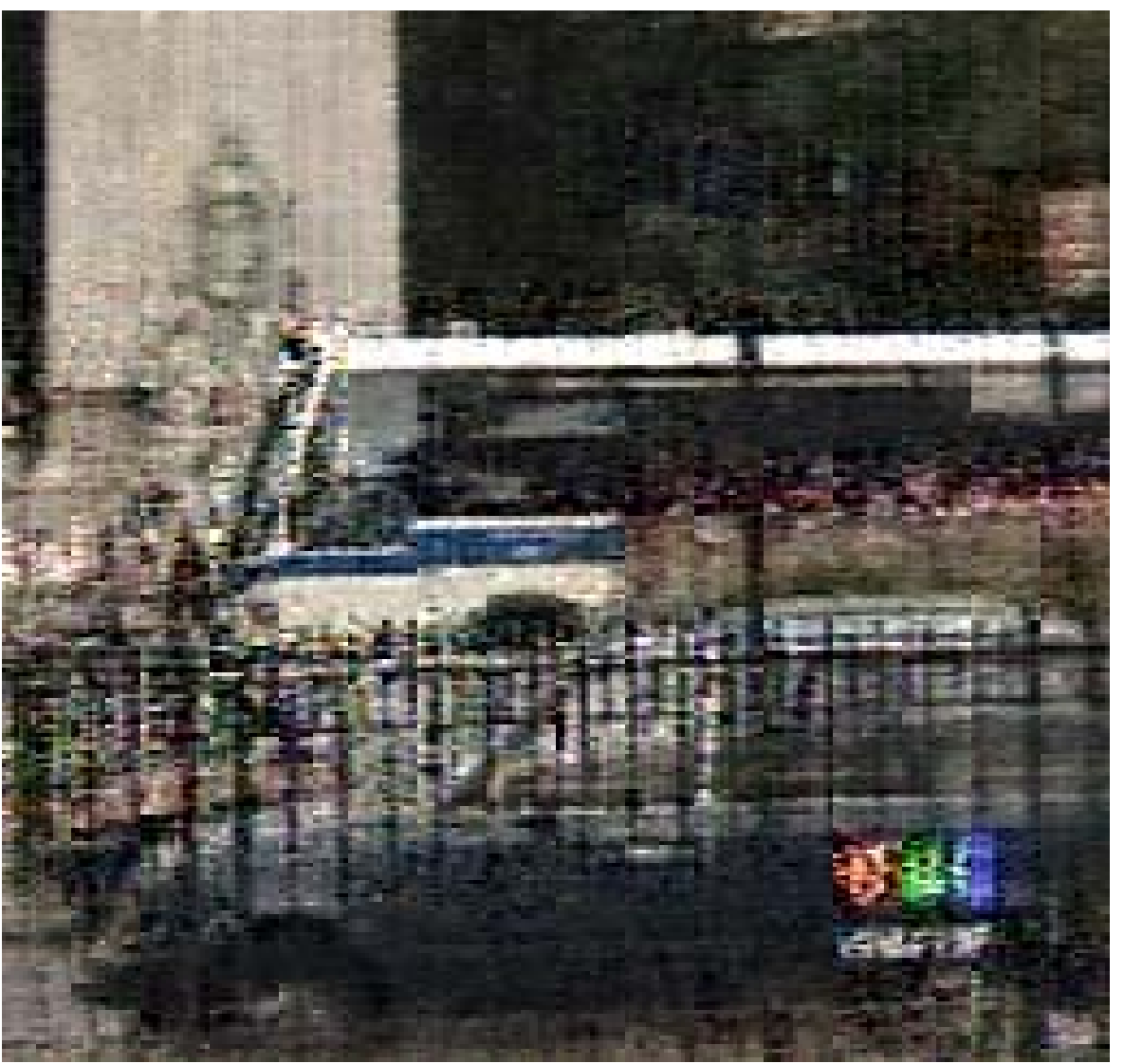}&
\includegraphics[width=0.14\textwidth]{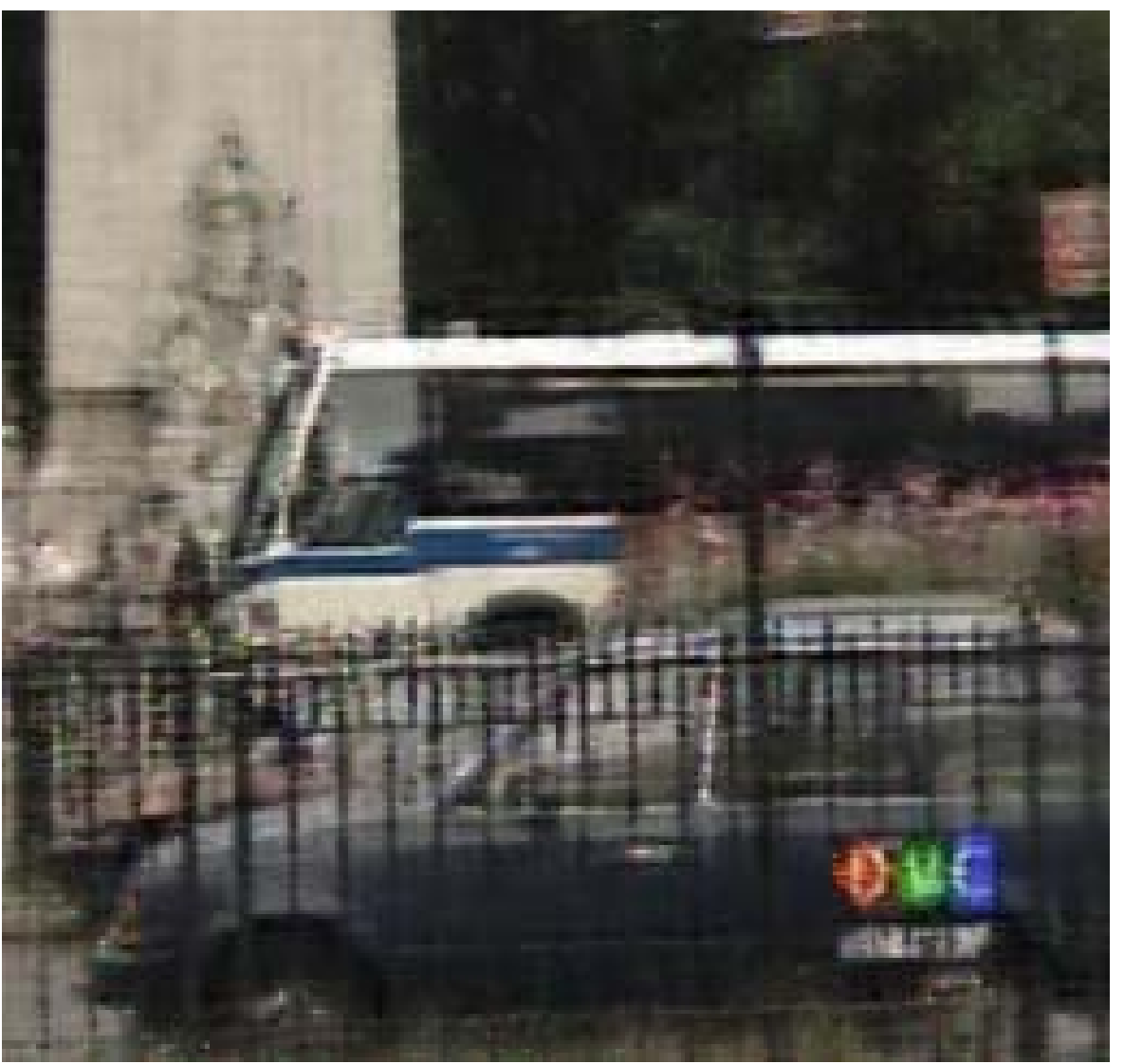}&
\includegraphics[width=0.14\textwidth]{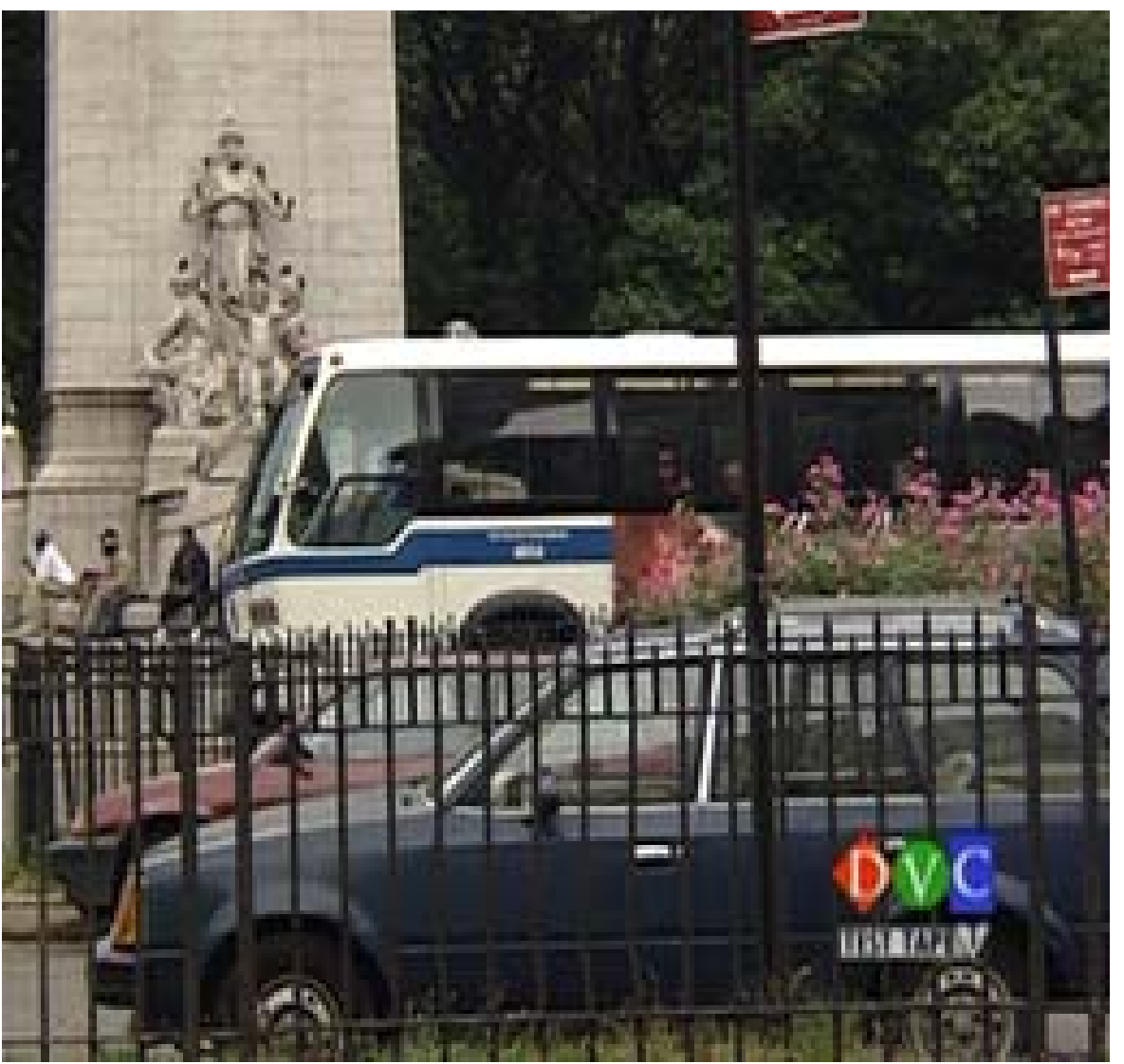}\\
\includegraphics[width=0.14\textwidth]{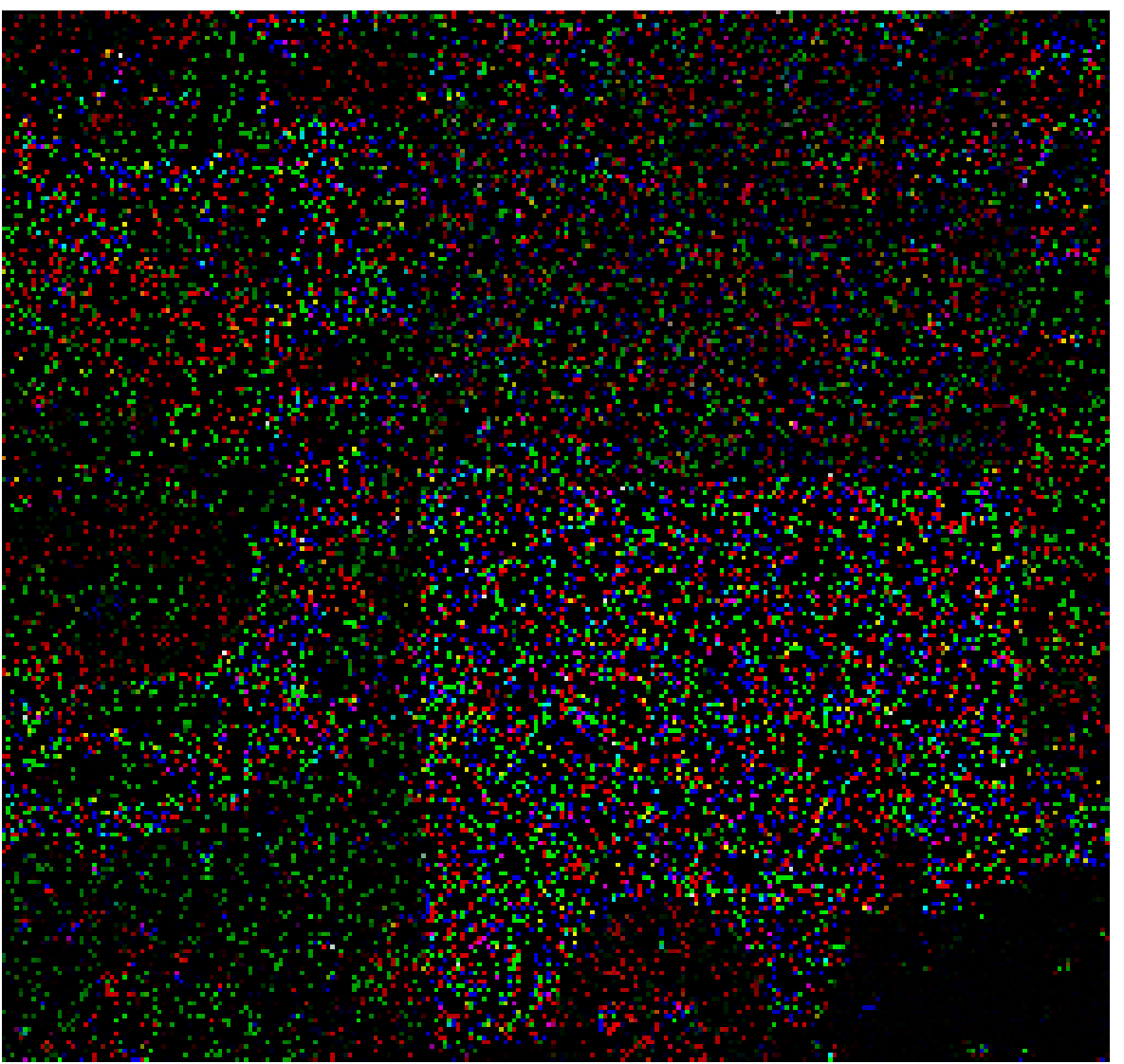}&
\includegraphics[width=0.14\textwidth]{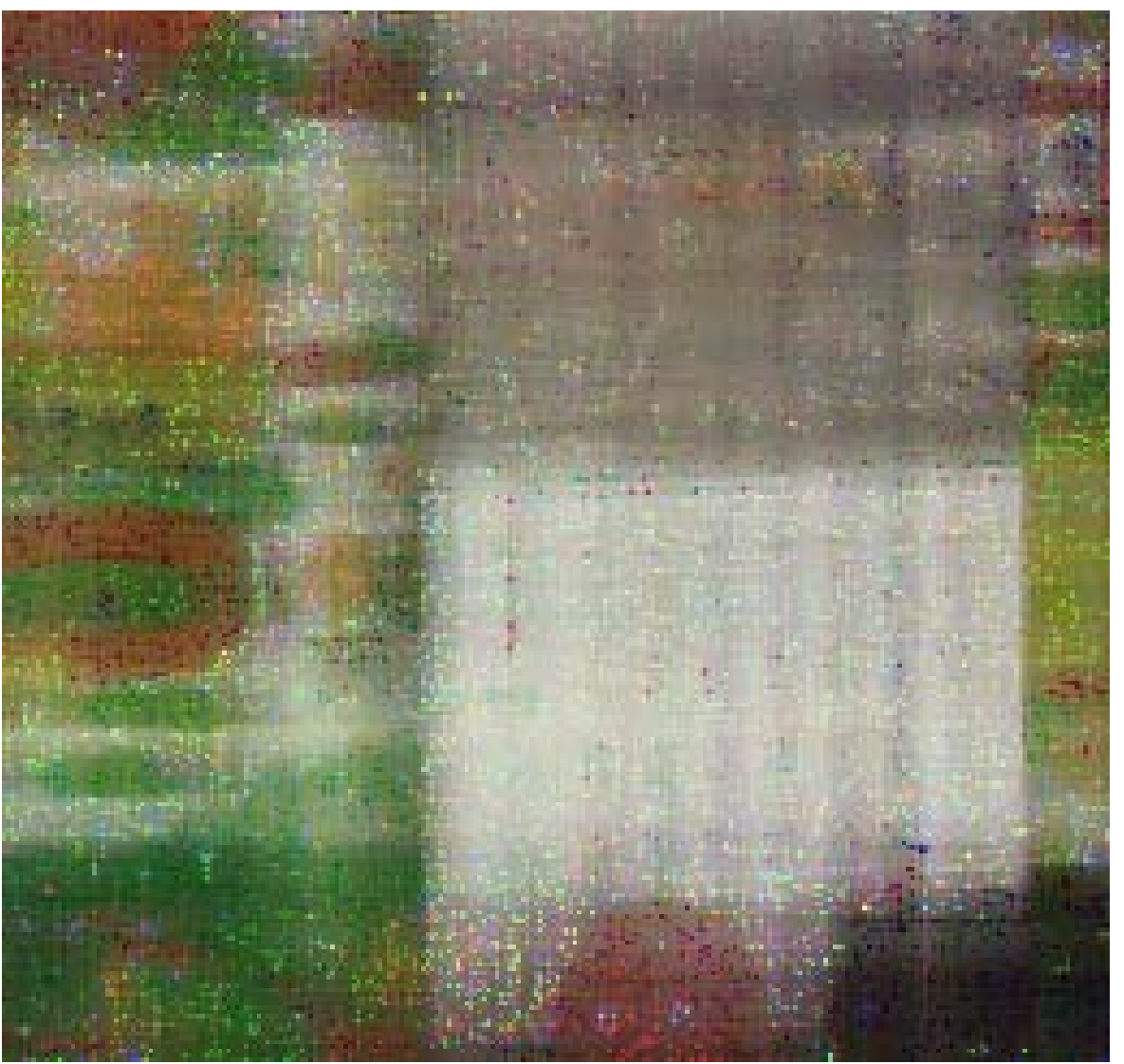}&
\includegraphics[width=0.14\textwidth]{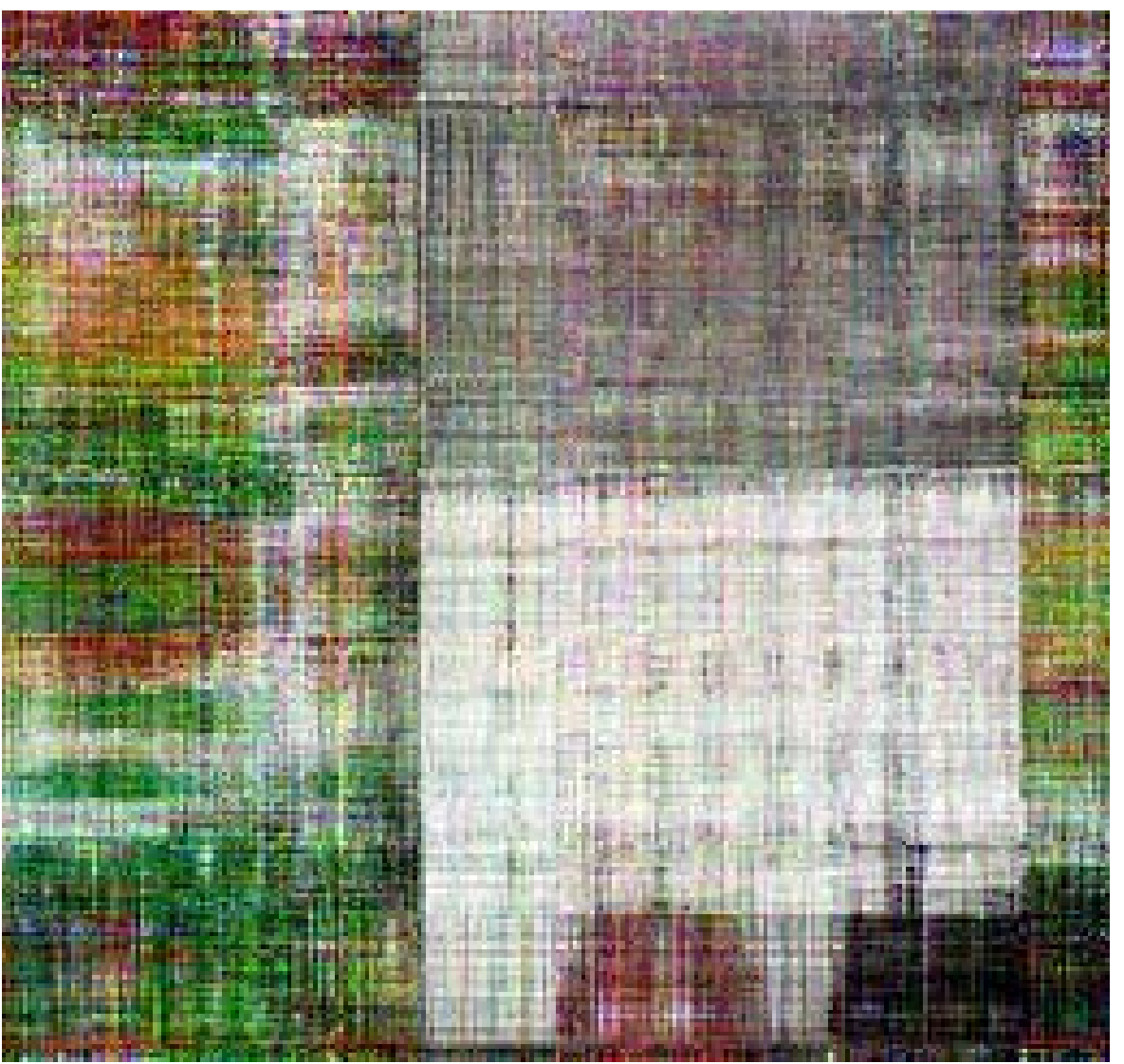}&
\includegraphics[width=0.14\textwidth]{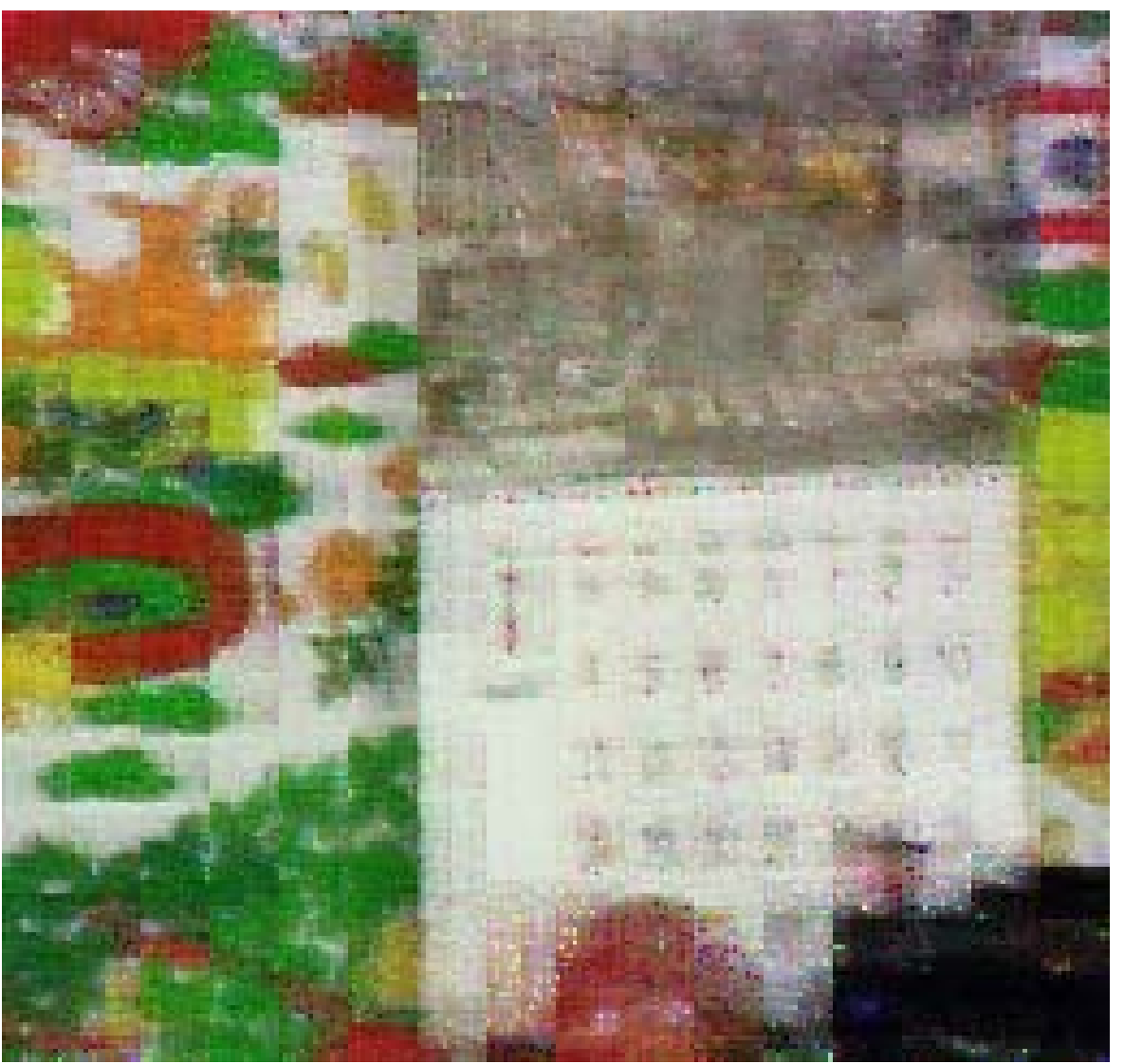}&
\includegraphics[width=0.14\textwidth]{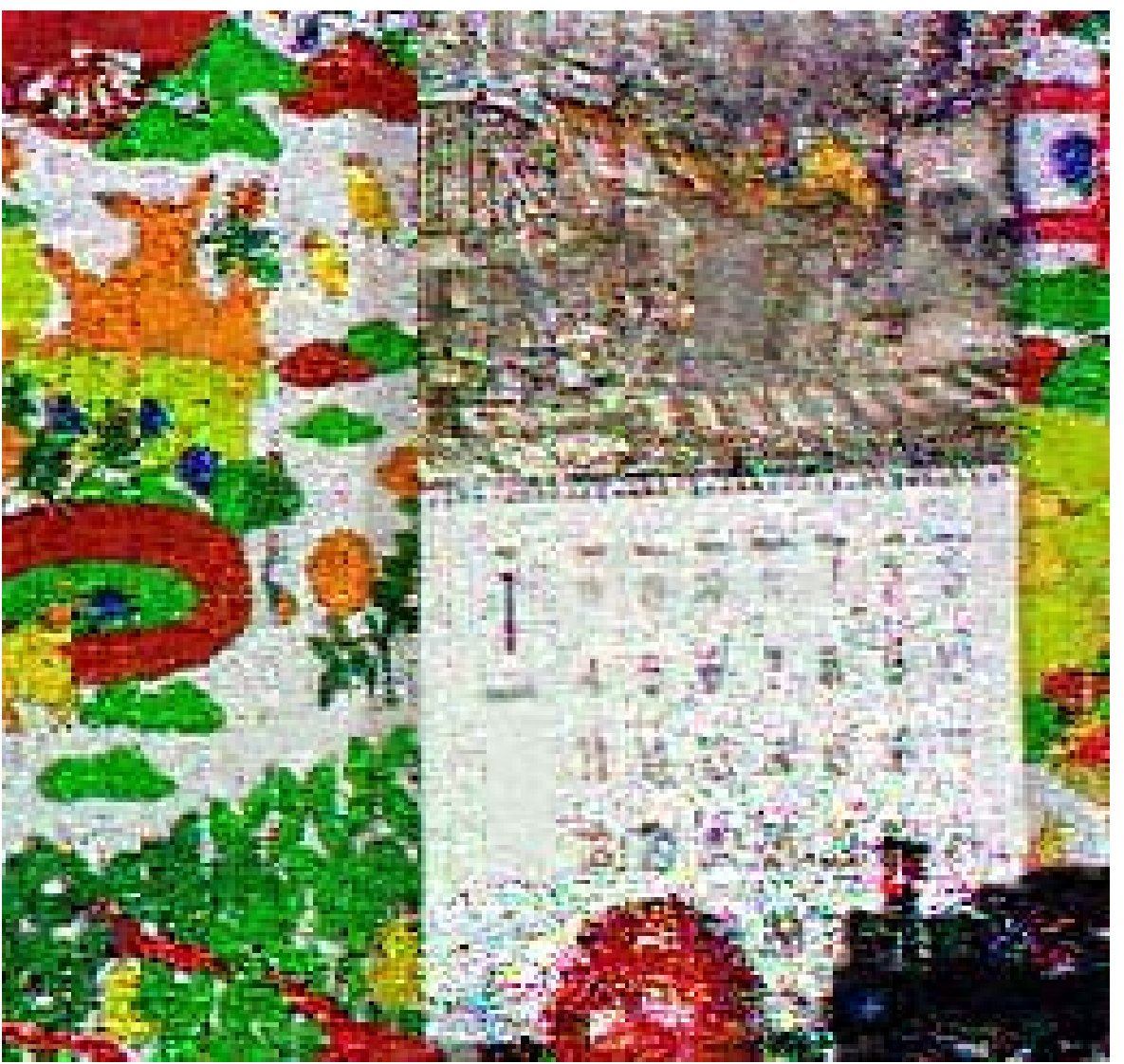}&
\includegraphics[width=0.14\textwidth]{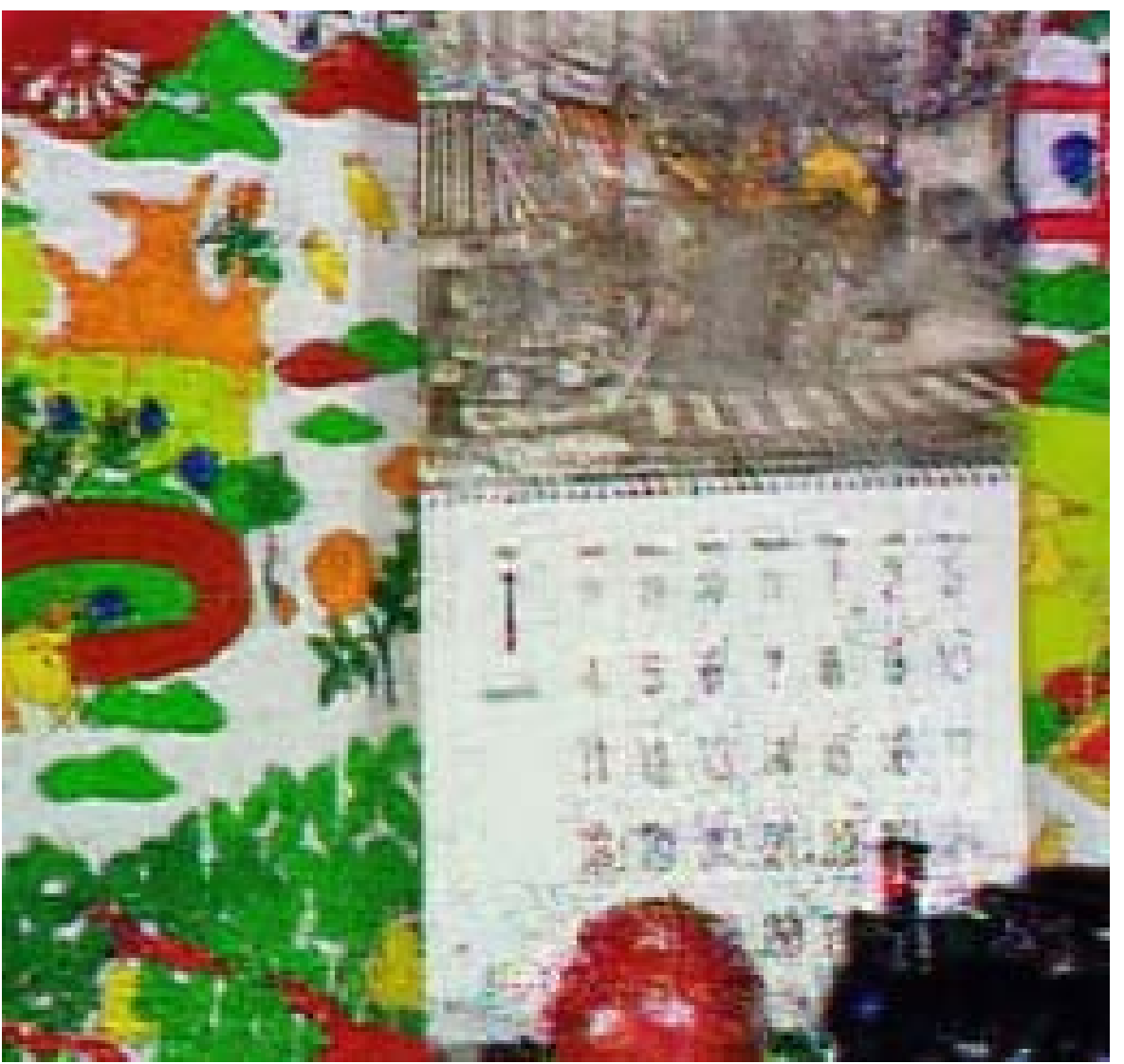}&
\includegraphics[width=0.14\textwidth]{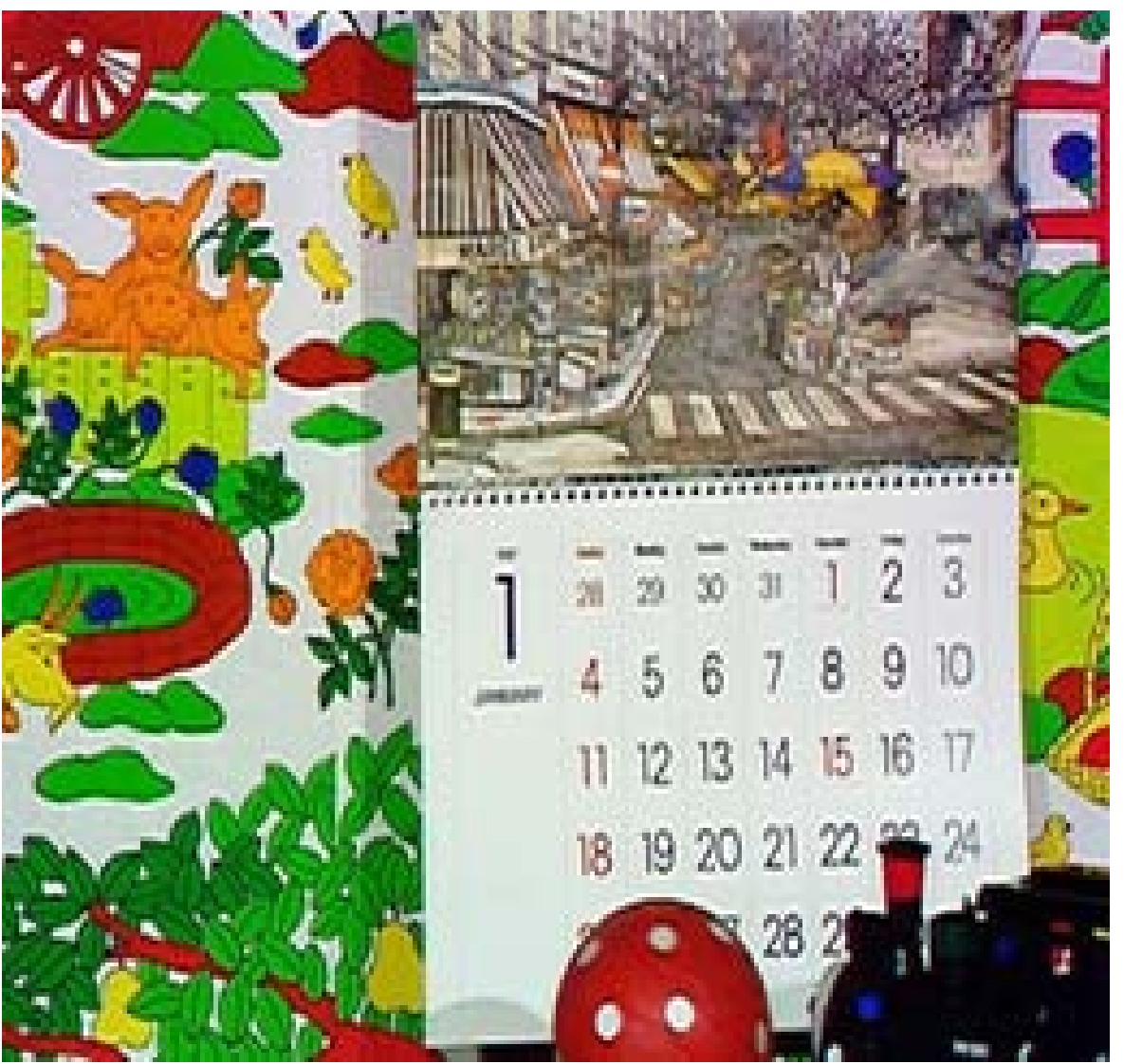}\\
\includegraphics[width=0.14\textwidth]{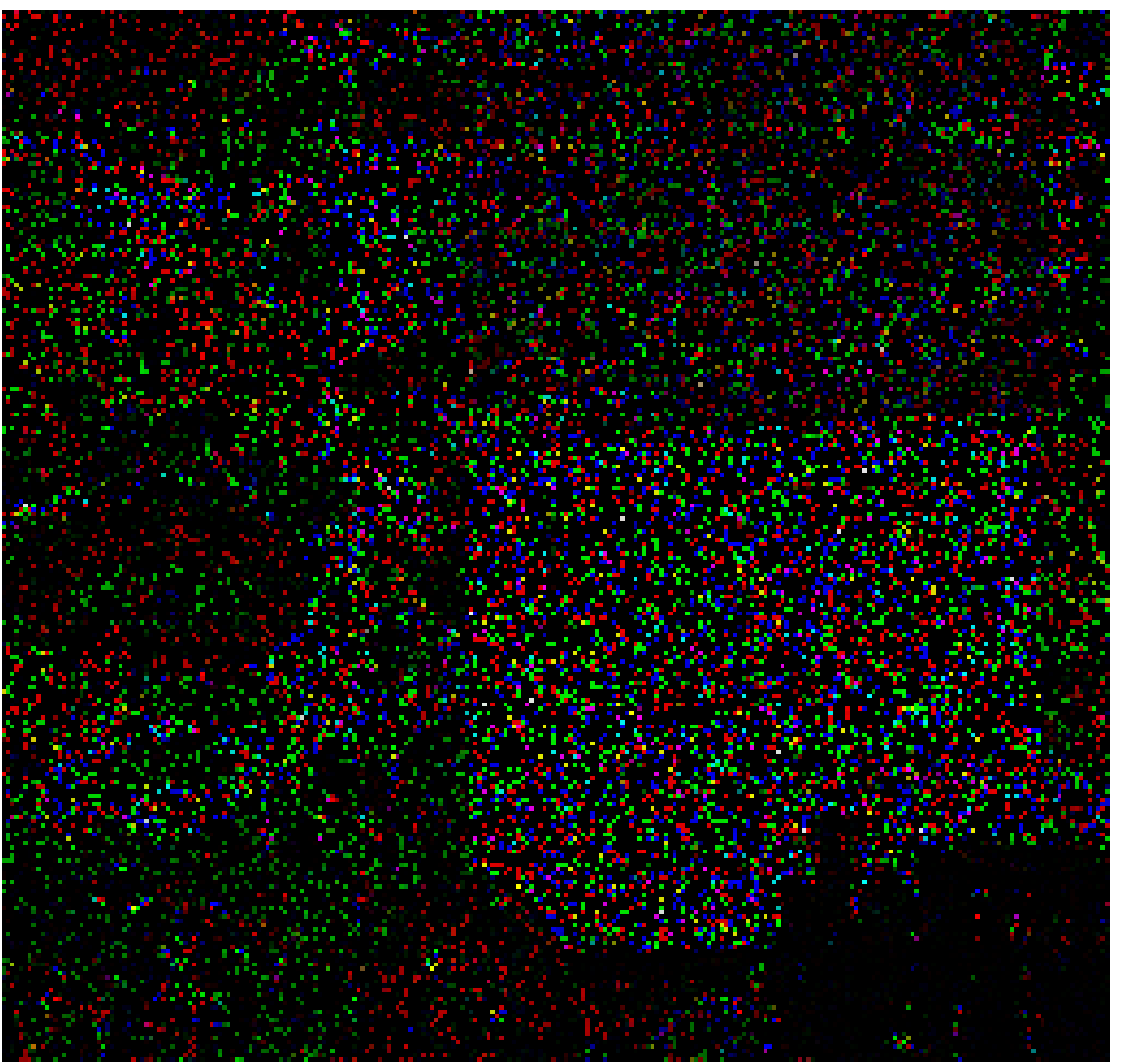}&
\includegraphics[width=0.14\textwidth]{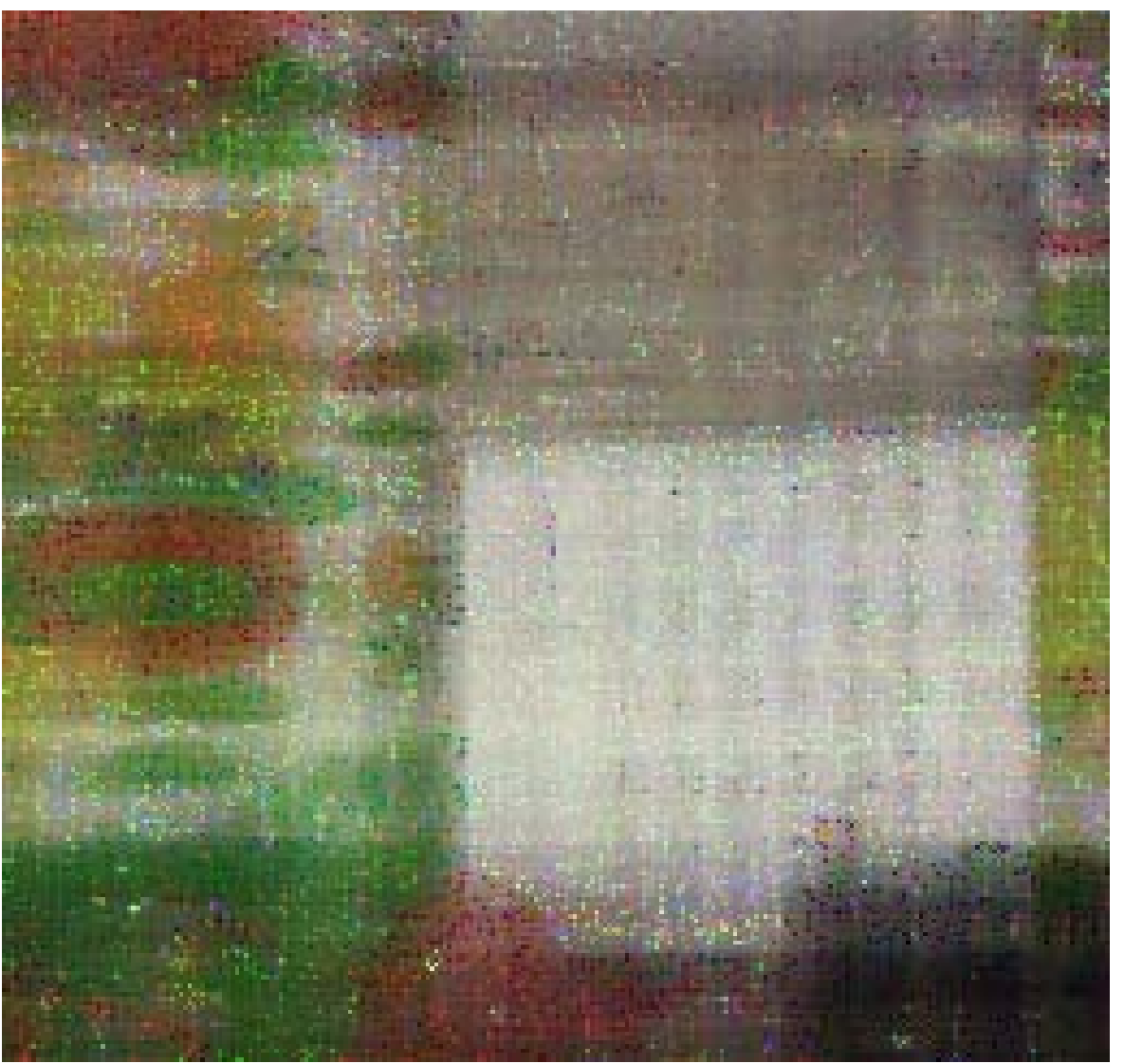}&
\includegraphics[width=0.14\textwidth]{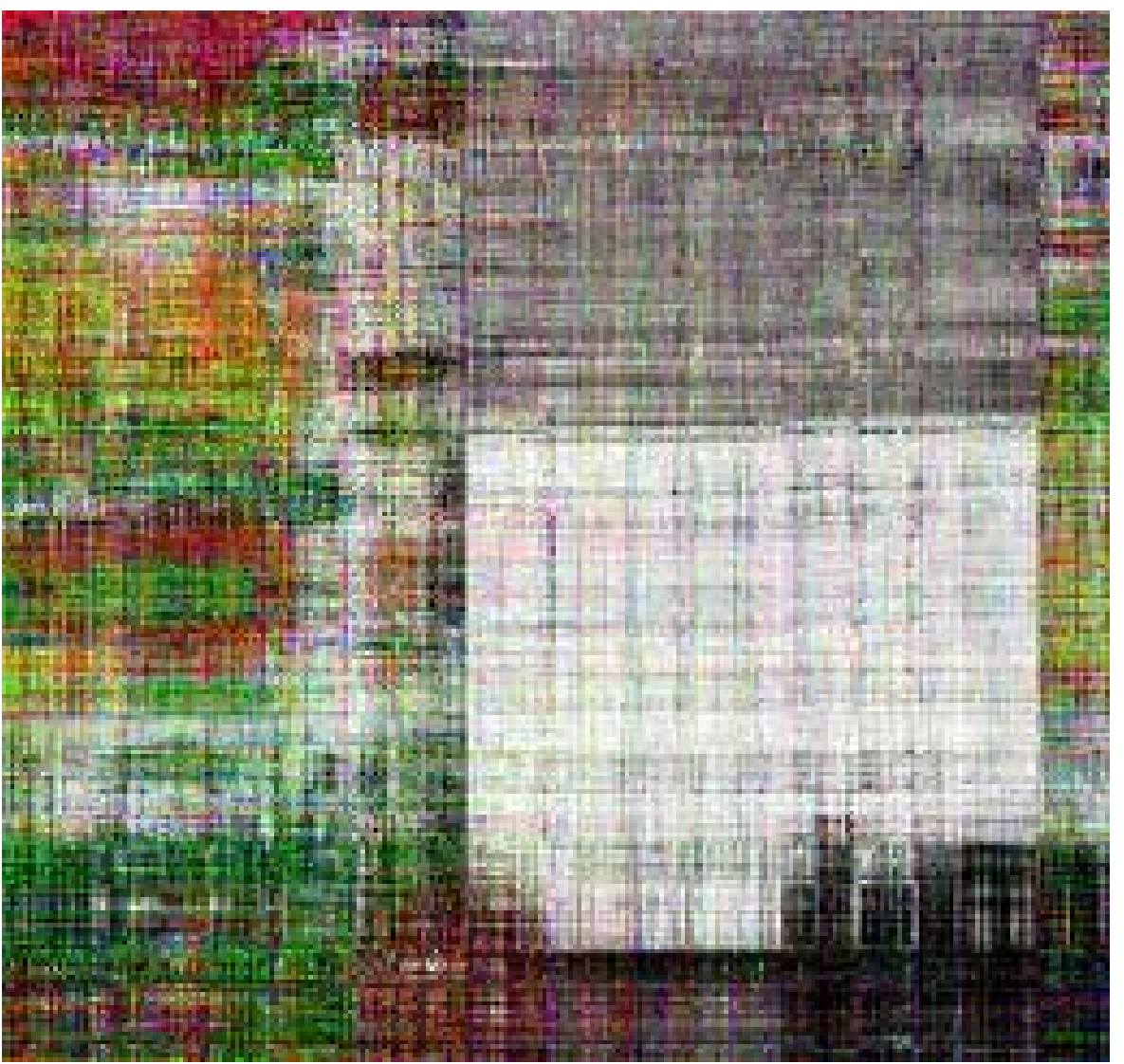}&
\includegraphics[width=0.14\textwidth]{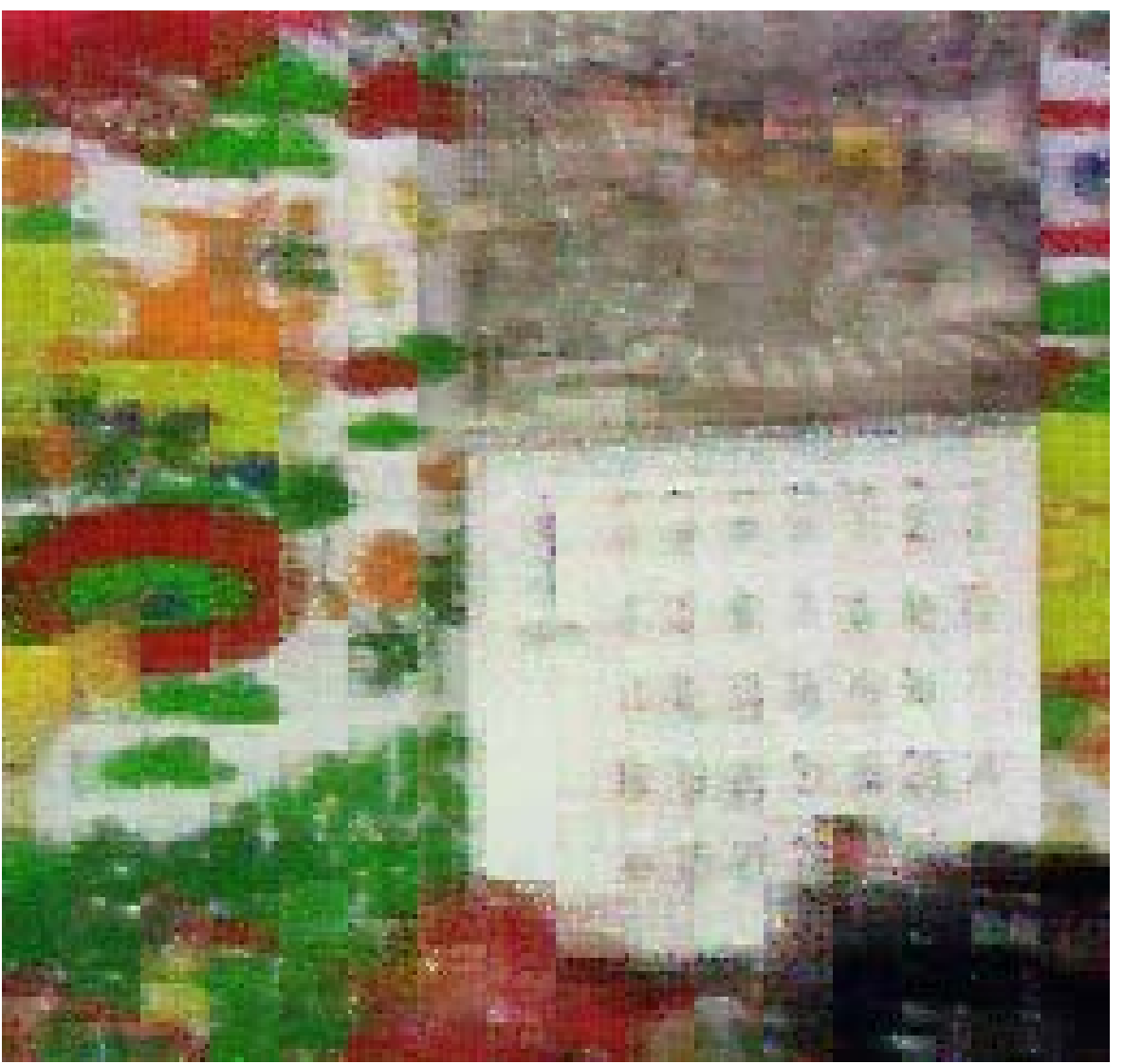}&
\includegraphics[width=0.14\textwidth]{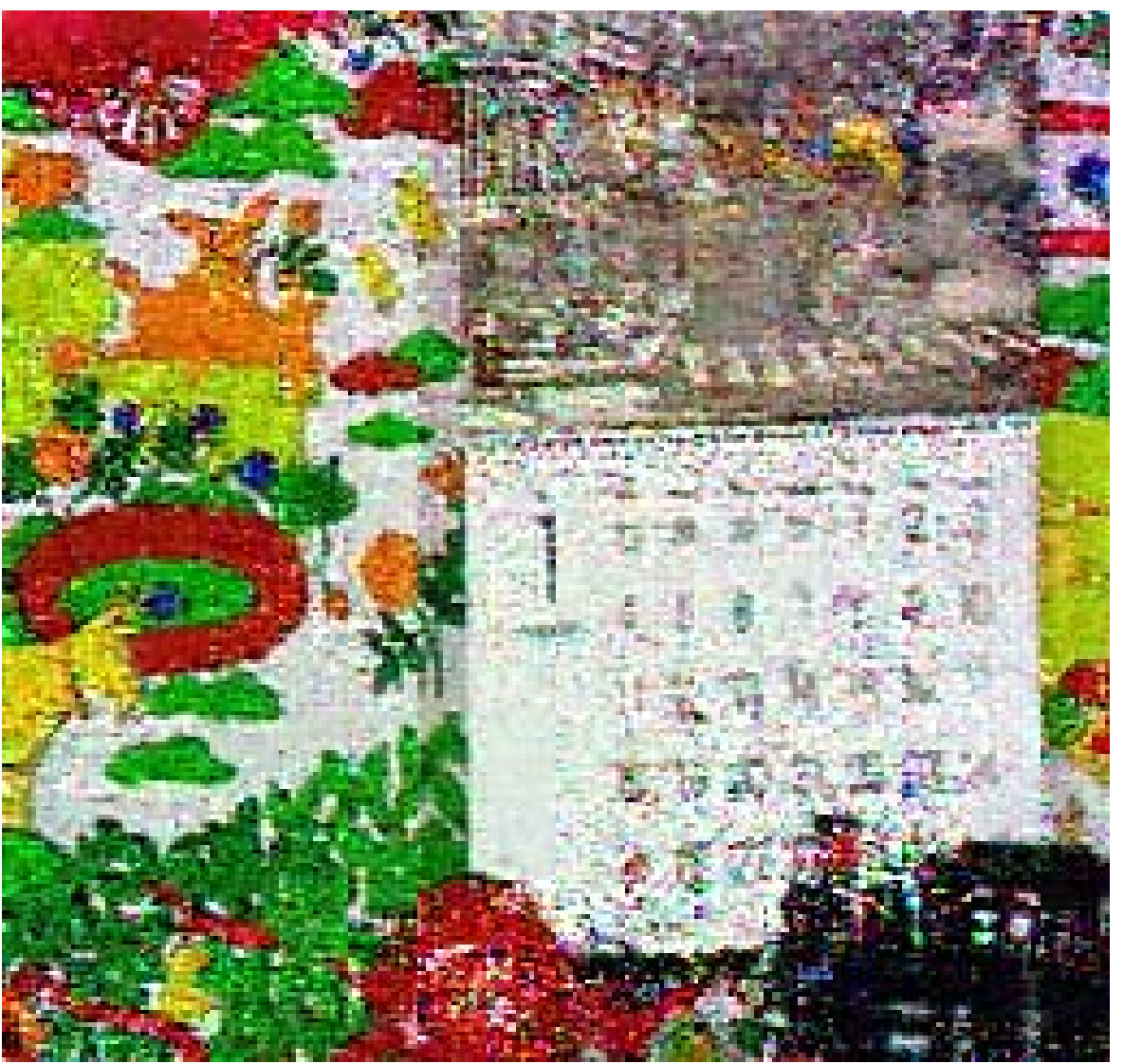}&
\includegraphics[width=0.14\textwidth]{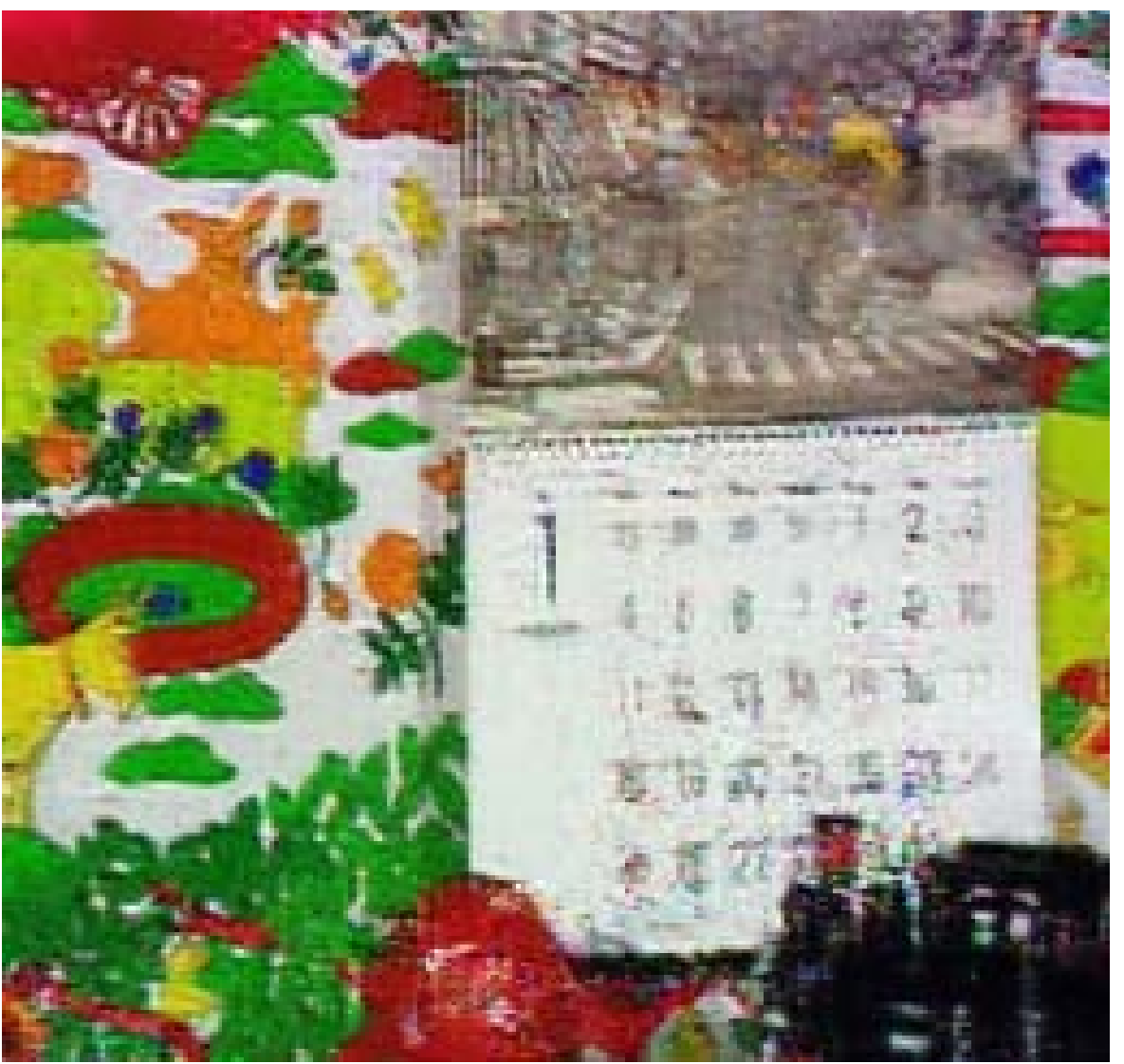}&
\includegraphics[width=0.14\textwidth]{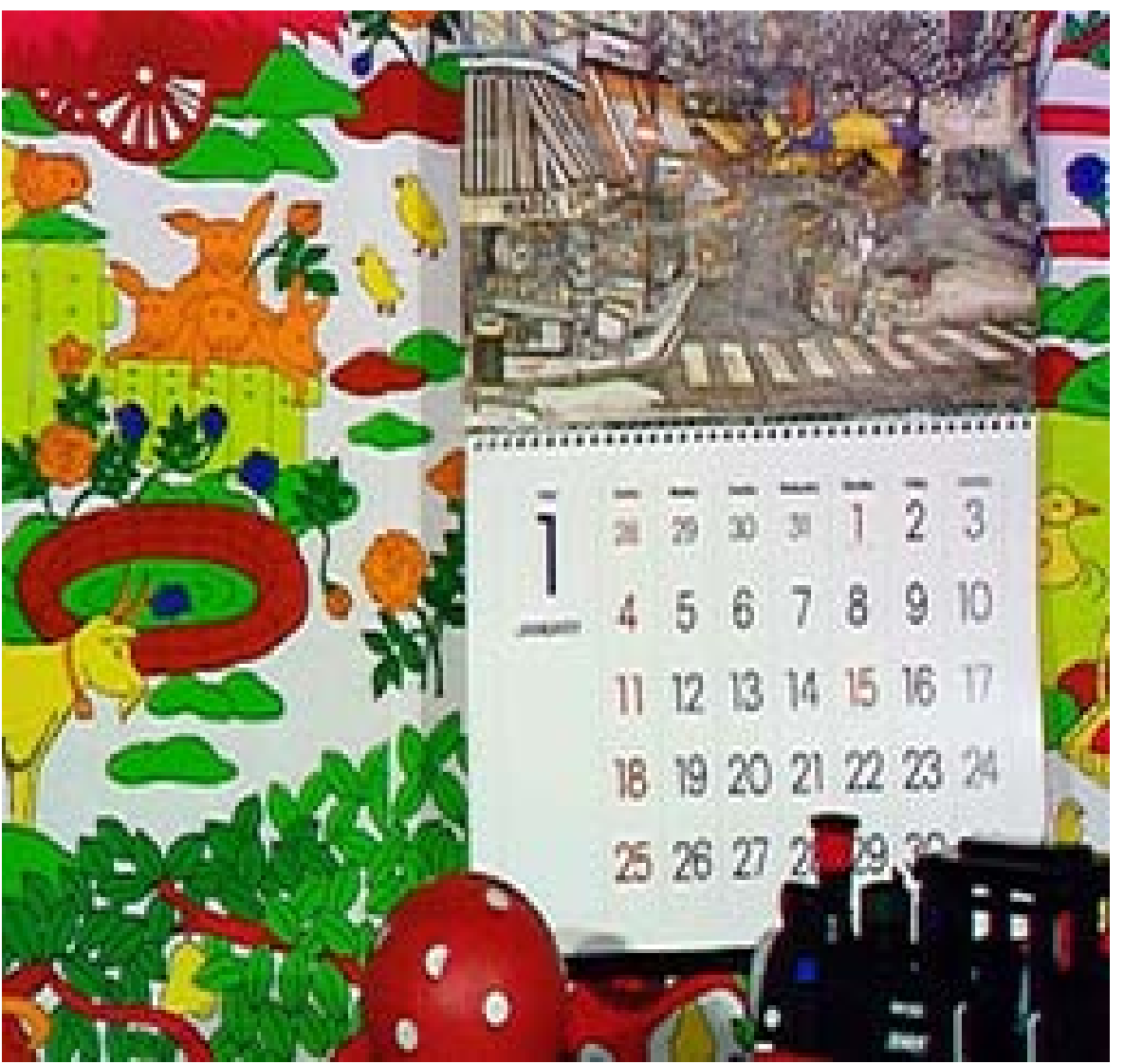}\vspace{0.1cm}\\
 {\footnotesize\textrm{(a) Observed}} & {\footnotesize\textrm{(b) HaLRTC}} & {\footnotesize\textrm{(c) tSVD}} & {\footnotesize\textrm{(d) SiLRTC-TT}} & {\footnotesize\textrm{(e) TMac-TT}} & {\footnotesize\textrm{(f) NL-TT}}& {\footnotesize\textrm{(g) Original}}\\
\end{tabular}
\caption{\small{The results of two frames of testing color videos recovered by different methods. The first (third) and second (fourth) rows: the results of color videos \emph{bus} (\emph{mobile}), respectively. From left to right: (a) the observed image, the results by (b) HaLRTC, (c) tSVD, (d) SiLRTC-TT, (e) TMac-TT, (f) NL-TT, and (g) the original image.}}
  \label{fig:video}
  \end{center}\vspace{-0.3cm}
\end{figure*}

\begin{figure}[!ht]
\scriptsize\setlength{\tabcolsep}{0.1pt}
\begin{center}
\begin{tabular}{cccc}
\includegraphics[width=0.45\textwidth]{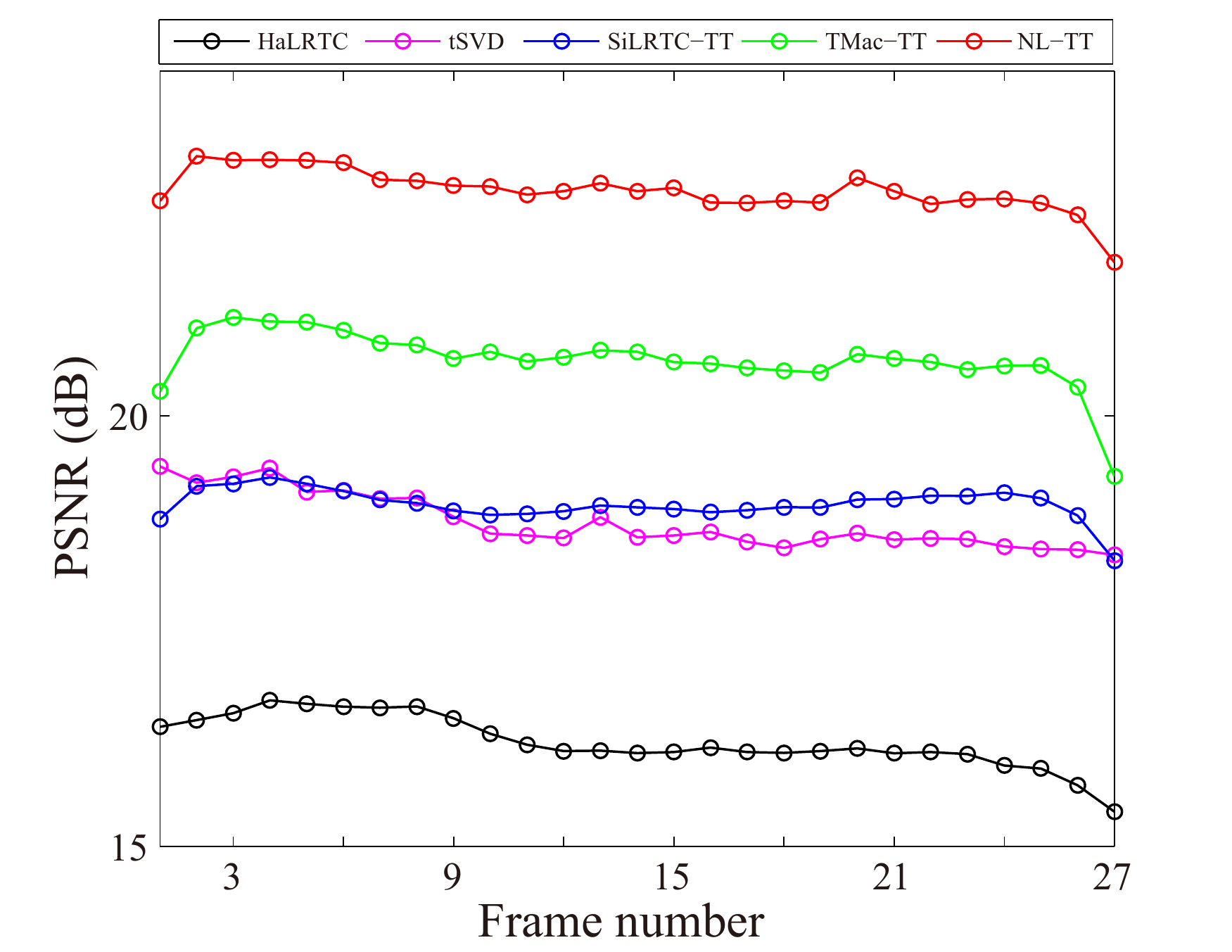}&
\includegraphics[width=0.45\textwidth]{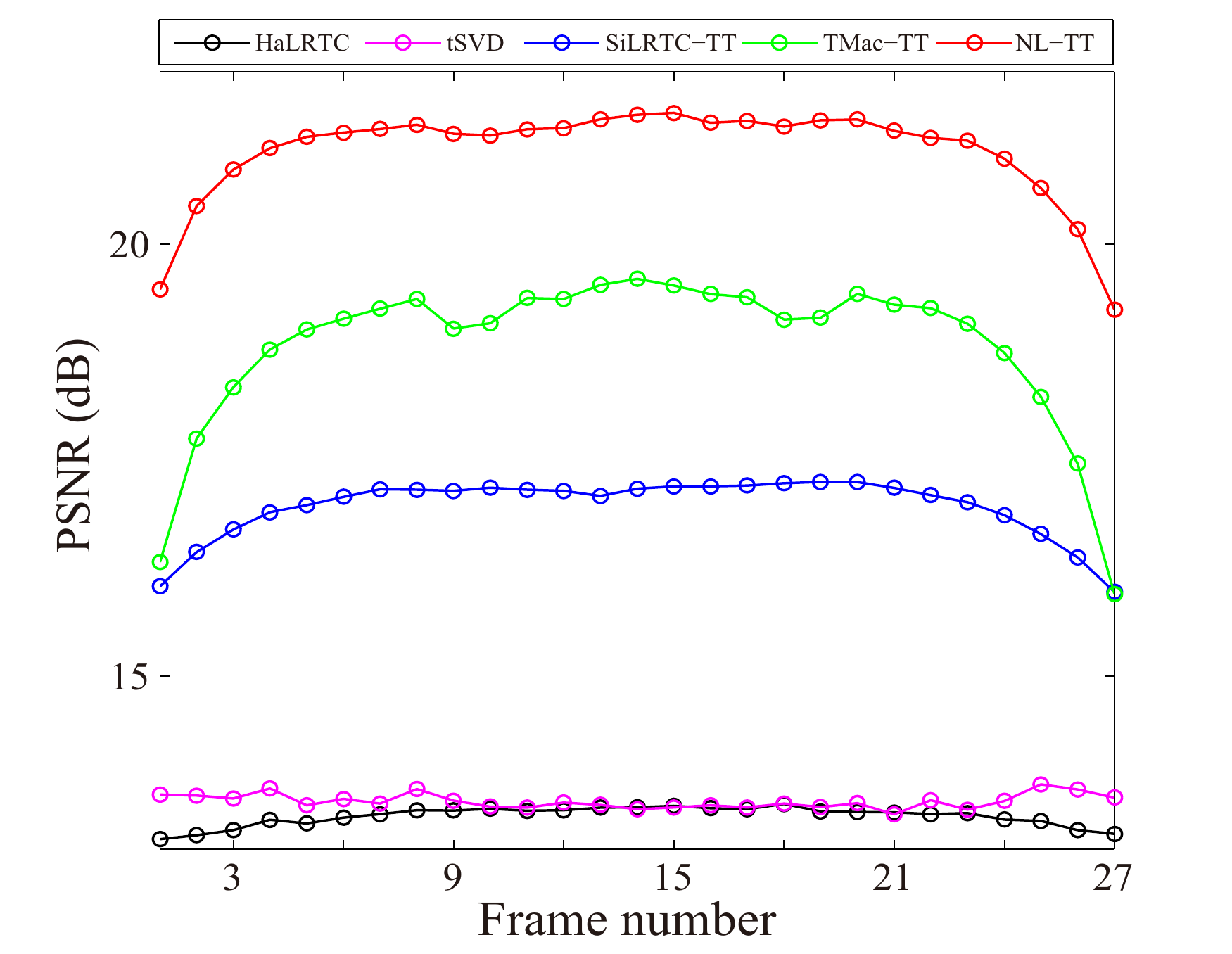}\\
\includegraphics[width=0.45\textwidth]{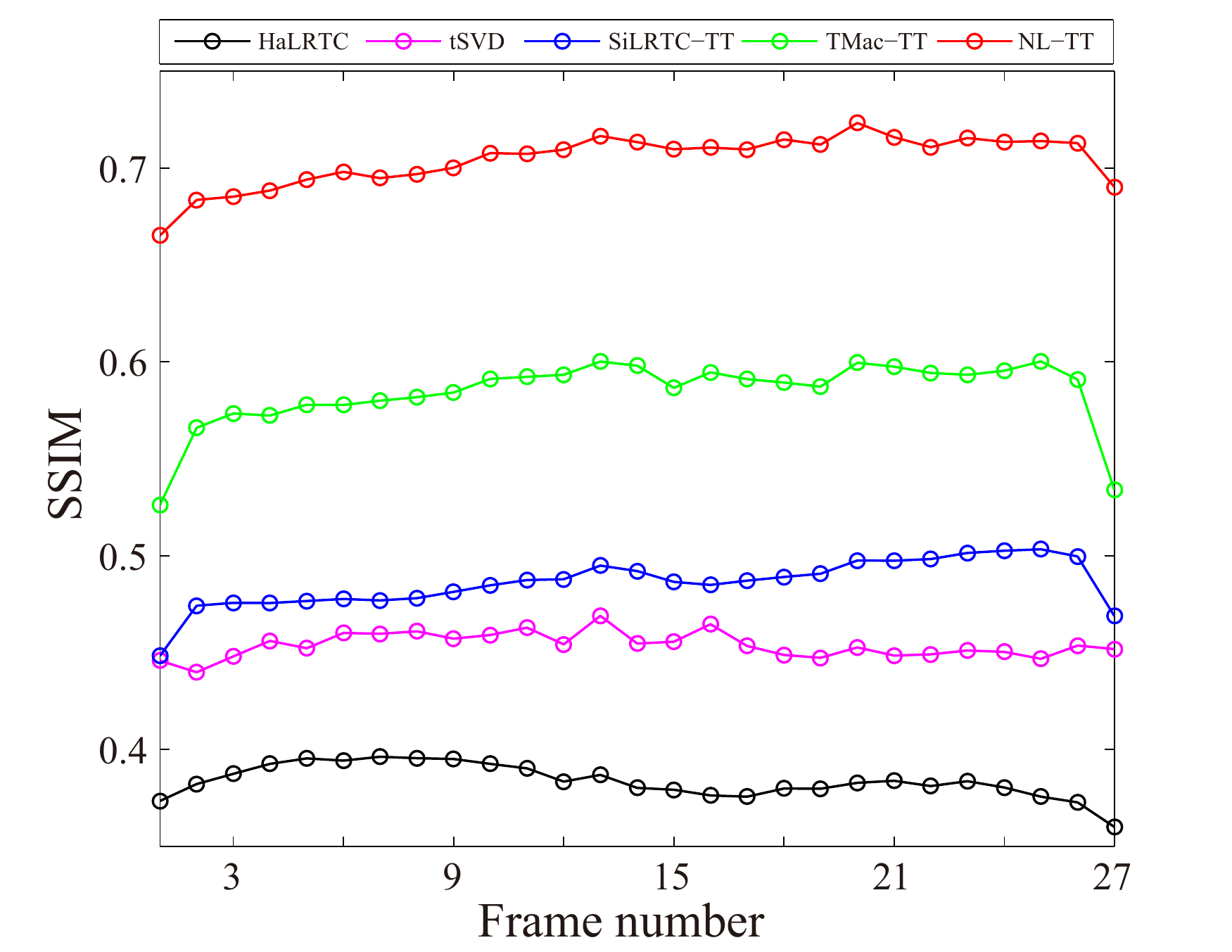}&
\includegraphics[width=0.45\textwidth]{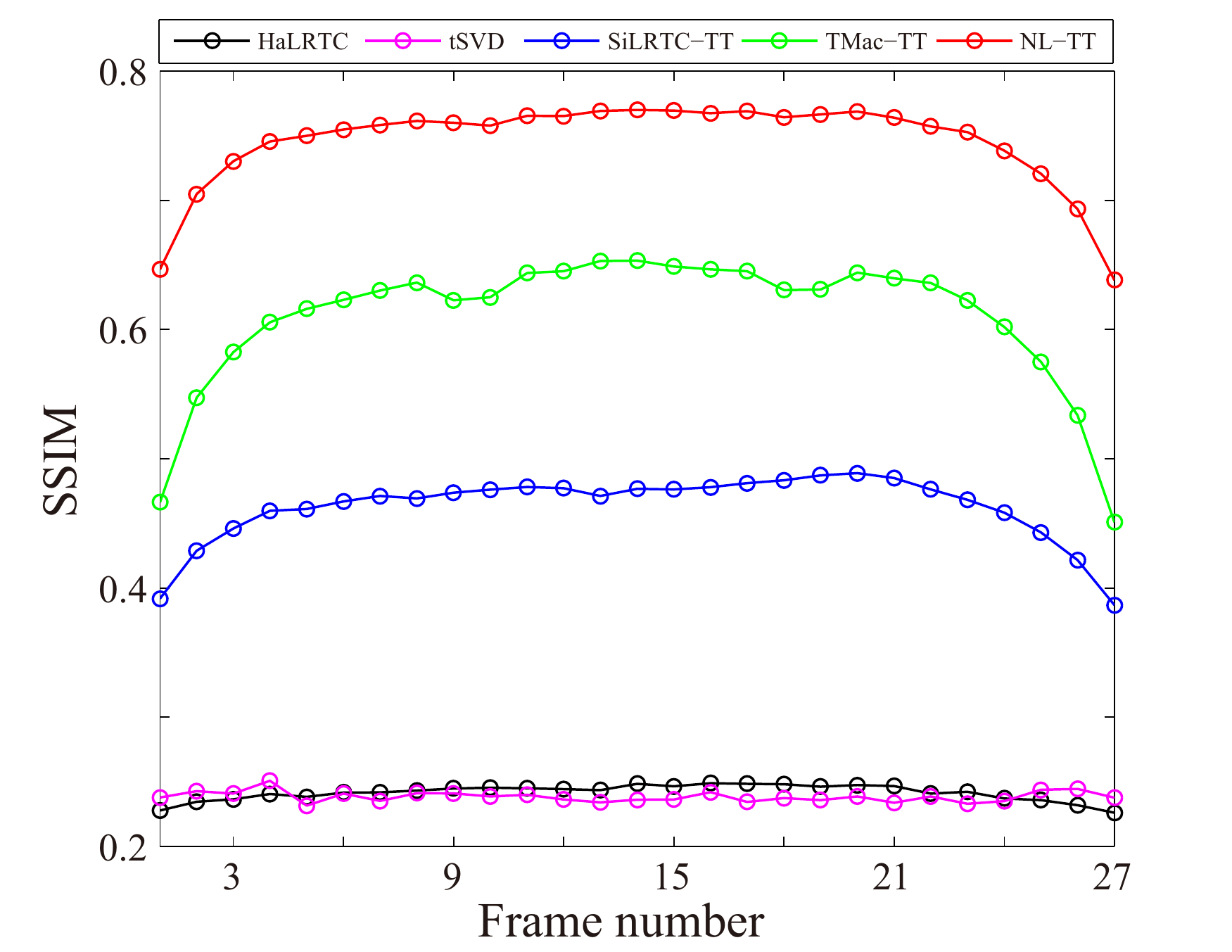}\\
{\footnotesize(a)} {\small\emph{bus}} & {\footnotesize(b)} {\small\emph{mobile}} \\
\end{tabular}
\caption{\small{The PSNR and SSIM values of all frames of color videos recovered by different methods.}}
  \label{fig:video_psnr}
  \end{center}\vspace{-0.3cm}
\end{figure}

\subsection{Color videos}
In this subsection, we test the proposed method on two color videos $bus$ and $mobile$ \footnote{https://media.xiph.org/video/derf/} with random sampling. The size of testing videos is $243\times 256 \times 3 \times 27$. The SR is set as 0.1 in this task.

Fig. \ref{fig:video} shows the visual results by using different methods. Obviously, the results by HaLRTC and SiLRTC-TT appear dark and have color distortion, the results by tSVD have undesirable thorns, and the results by our method visually outperforms TMac-TT in keeping smoothness and details of recovered images. The PSNR and SSIM values of each frame of two reconstructed color videos are plotted in Fig. \ref{fig:video_psnr}. We note that the PSNR and SSIM values of each frame recovered by the proposed method are higher than all compared methods.

\section{Discussions}
\label{section:Discussion}
In this section, we test the effects of parameters of the proposed NL-TT method and show the numerical convergence of the ADMM solver. All tests in this section are based on image \emph{lena} with $SR=0.3$ tube sampling and image \emph{house} with missing curves as examples.
\begin{figure}[!ht]
\scriptsize\setlength{\tabcolsep}{0.9pt}
\begin{center}
\begin{tabular}{ccc}
\includegraphics[width=0.45\textwidth]{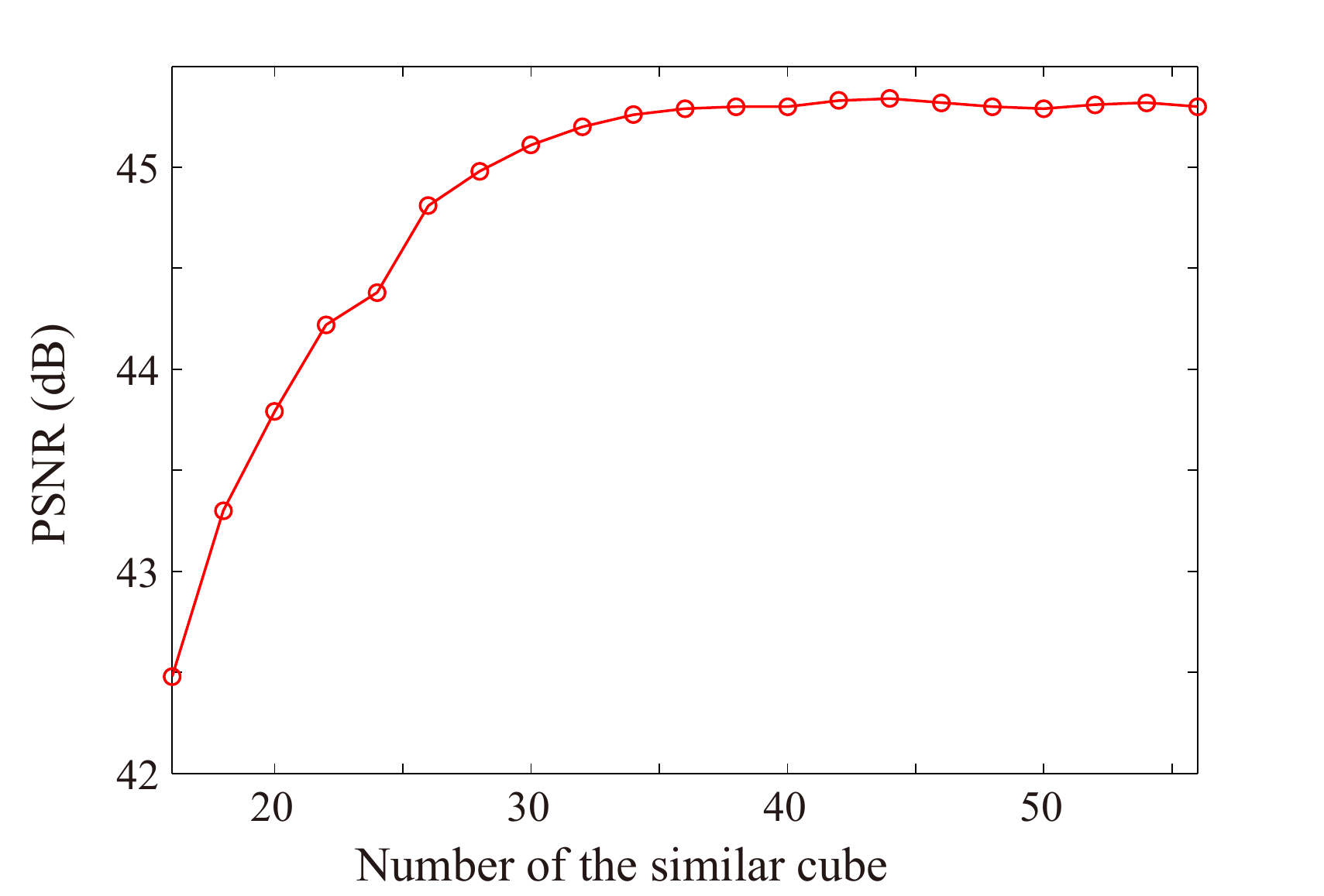}&
\includegraphics[width=0.45\textwidth]{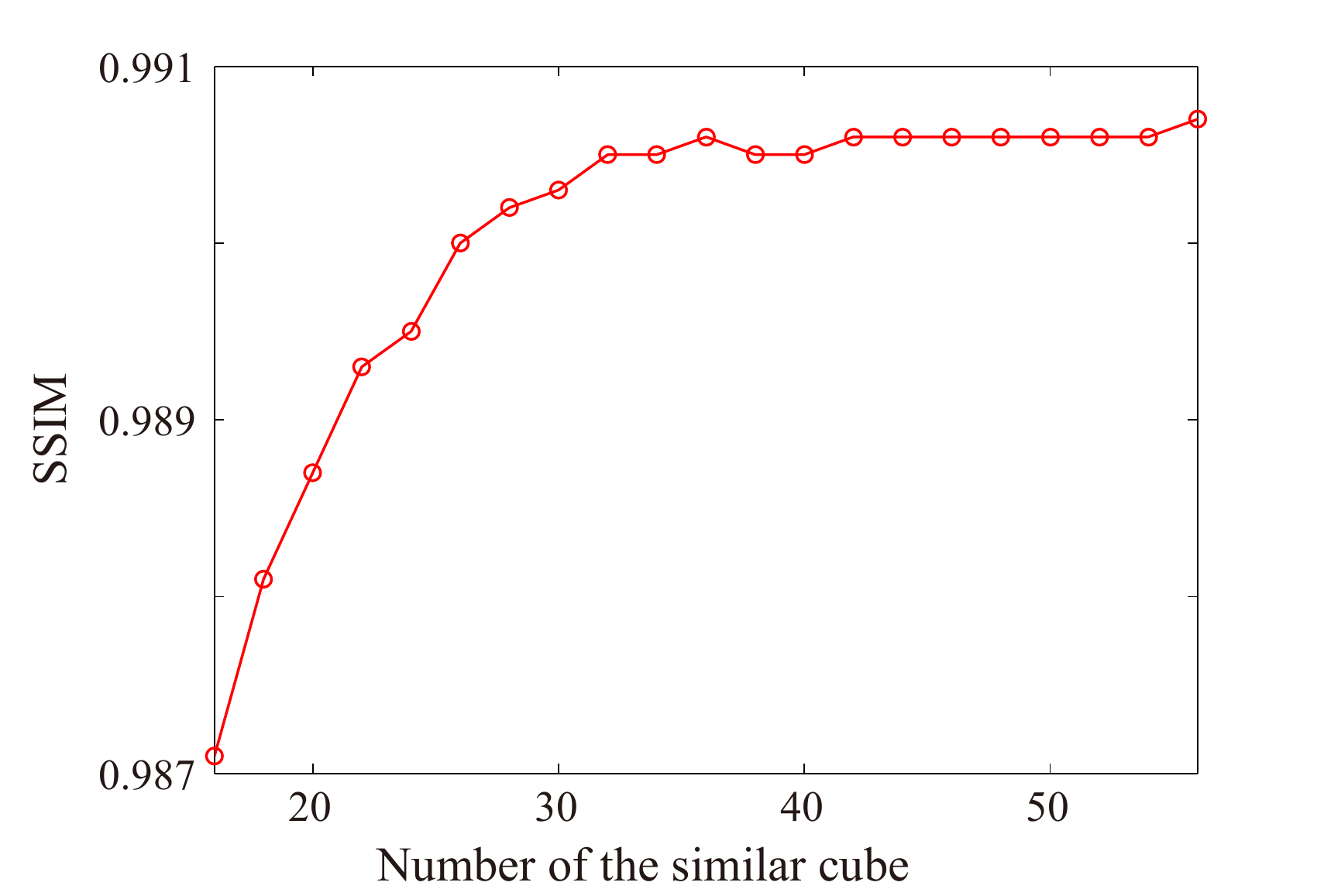} \vspace{0.2cm}\\
{\footnotesize\textrm{(a) PSNR}} & {\footnotesize\textrm{(b) SSIM}}\\
\end{tabular}
\caption{\small{The PSNR and SSIM curves as the function of the number of the similar cube $h$. (a) change in the PSNR value, (b) change in the SSIM value.}}
  \label{fig:patch_number}
  \end{center}\vspace{-0.3cm}
\end{figure}

\begin{figure}[!ht]
\scriptsize\setlength{\tabcolsep}{0.9pt}
\begin{center}
\begin{tabular}{cc}
\includegraphics[width=0.44\textwidth]{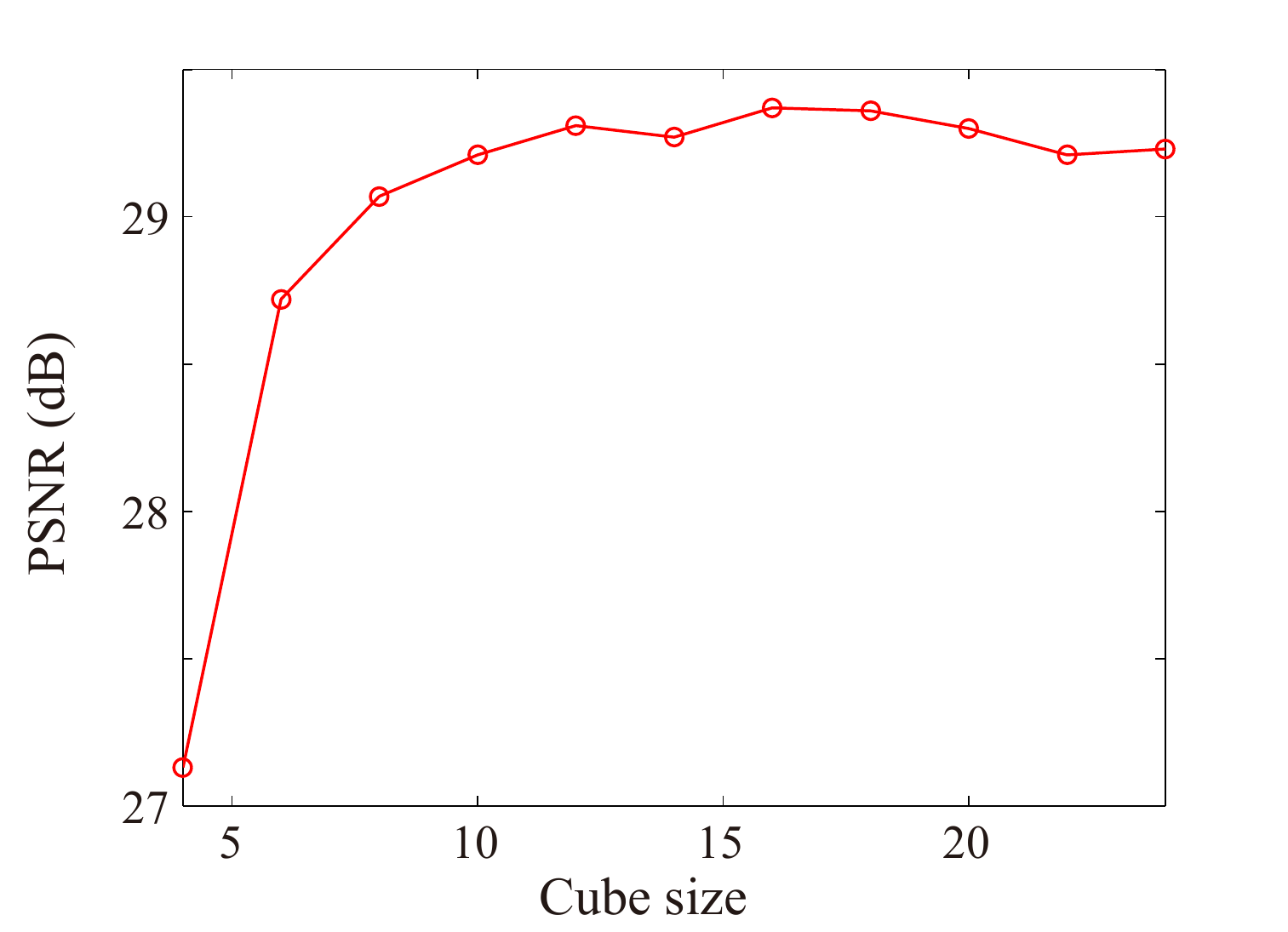}&
\includegraphics[width=0.44\textwidth]{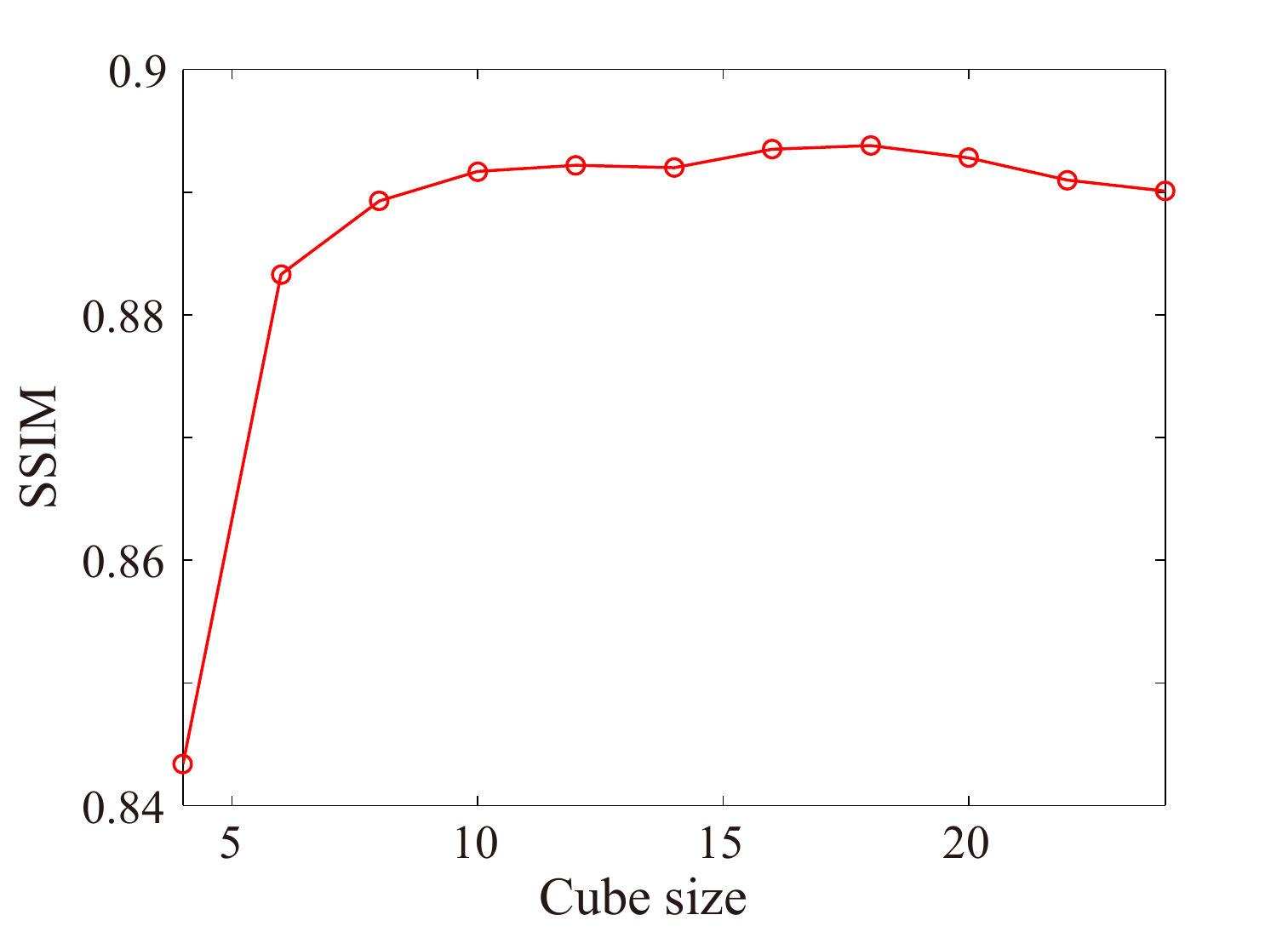}\vspace{0.2cm}\\
{\footnotesize\textrm{(a) PSNR}} & {\footnotesize\textrm{(b) SSIM}} \\
\end{tabular}
\caption{\small{The PSNR and SSIM curves as the function of the number of the cube size $s$. (a) change in the PSNR value, (b) change in the SSIM value.}}
  \label{fig:patch_size}
  \end{center}\vspace{-0.3cm}
\end{figure}

\begin{figure}[!ht]
\scriptsize\setlength{\tabcolsep}{0.9pt}
\begin{center}
\begin{tabular}{cc}
\includegraphics[width=0.9\textwidth]{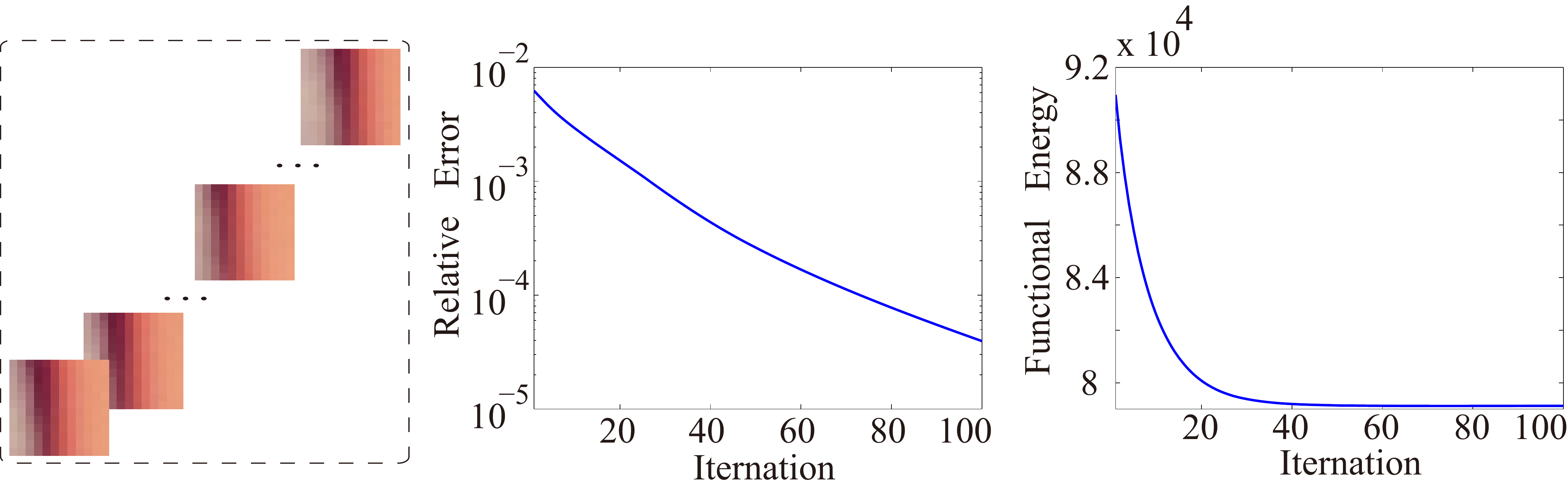}& \vspace{0.2cm} \\
\includegraphics[width=0.9\textwidth]{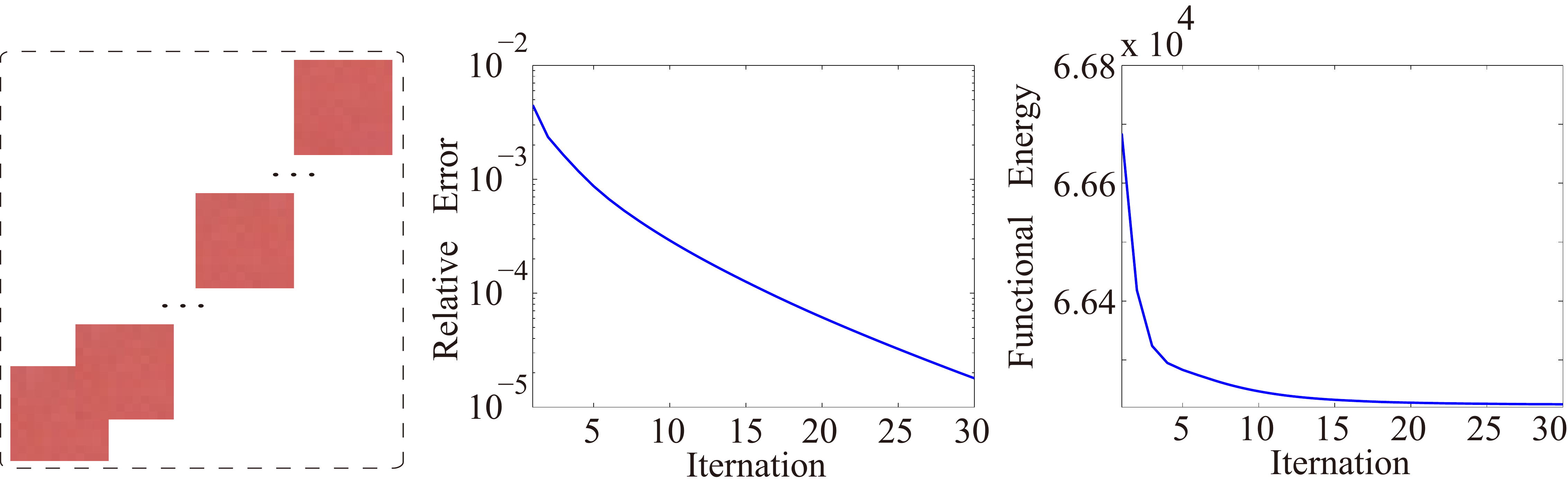}& \vspace{0.2cm} \\
\includegraphics[width=0.9\textwidth]{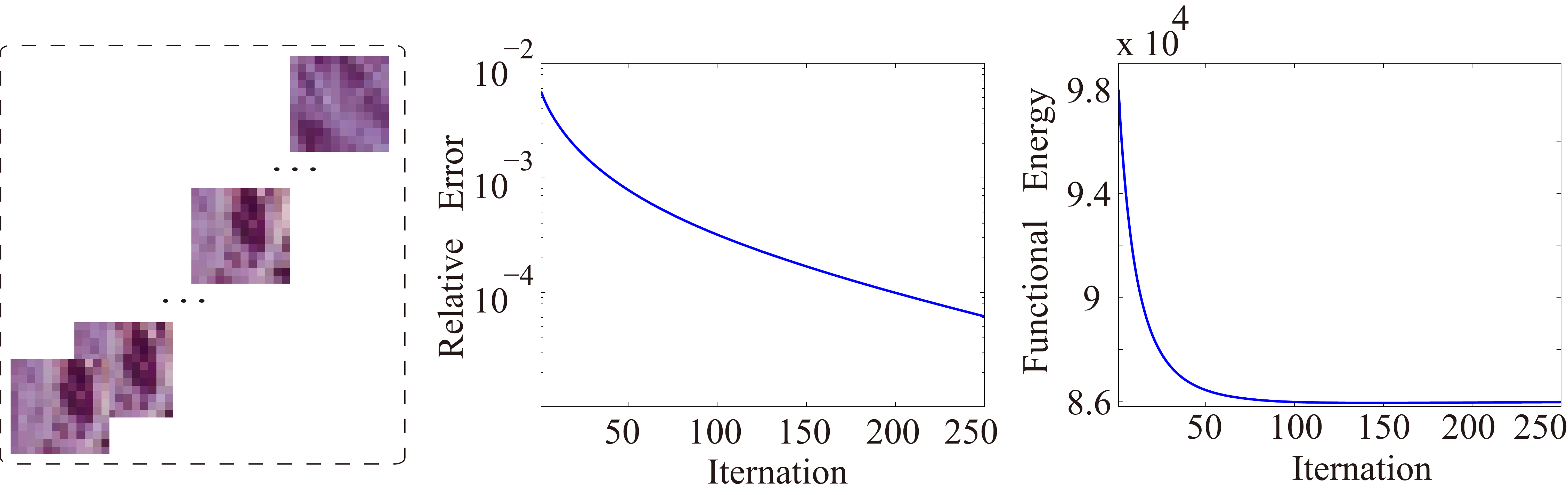}& \vspace{0.2cm} \\
{\footnotesize\textrm{(a) Grouped tensors}}\quad \qquad \qquad \qquad \qquad {\footnotesize\textrm{(b) Relative error}} \qquad \qquad \qquad \qquad \quad {\footnotesize\textrm{(c) Function value}} \\
\end{tabular}
\caption{\small{The relative error and objective function values curve versus the iteration number for grouped tubes. (a) grouped tensors, (b) change in the relative error value, (c) change in the objective function value.}}
  \label{fig:convergence}
  \end{center}\vspace{-0.3cm}
\end{figure}
\textbf{Effect of the number of the similar cube $h$}. In our algorithm, the number of the similar cube $h$ controls the number similar cubes of the grouped tensor. In Fig. \ref{fig:patch_number}, we show the experimental results of PSNR and SSIM values as the function of the number $h$ from 16 to 56 with step 2. From Fig. \ref{fig:patch_number}, we can observe that when the number of similar cubes is small, the PSNR values increase extremely fast. Then the growing rate of the curve becomes relatively slow. The highest PSNR and SSIM values are achieved around $h = 42$. Thus, we empirically choose the number of the similar cube between 30 to 50 with the increment 4.

\textbf{Effect of the cube size $s$.} The cube size $s$ controls the number of patterns in an image cube. Fig. \ref{fig:patch_size} provides the completed results of PSNR and SSIM values as the function of the size $s$ from 4 to 24 with step 2. From Fig. \ref{fig:patch_size}, one can observe that a too small $s$ performs poorly. One possible reason is that small-size cubes do not contain enough well-patterned image texture. The highest PSNR and SSIM values are achieved around $s = 16$. Therefore, we empirically set the parameter $s\in [10, 20]$ with the increment 2.

\textbf{Numerical convergence.} We empirically demonstrate the convergence of the proposed algorithm. Fig. \ref{fig:convergence} displays the relative error value ($\|\mathcal{X}^{l+1}_{p}-\mathcal{X}^{l}_{p}\|_{F}/\|\mathcal{X}^{l}_{p}\|_{F}$) and the objective function value verse the iteration number for several restored groups. We observe that as the iteration number increases, the relative errors converge to zero and the objective function values tend to flat, which empirically indicates the convergence of the proposed algorithm.
\section{Conclusion}
\label{section:Conclusion}
In this paper, we propose a new nonlocal TT rank-based tensor completion method by exploring the NSS prior of tensor data. After block-matching, each group of selected $j$-th order cubes are stacked together into a $(j+1)$-th order data. We apply the low-TT-rank constraint on the grouped tensor, which can simultaneously learn the correlation along the spatial, nonlocal, and temporal/spectral modes. Moreover, we establish a perturbation analysis for the TT low-rankness of groups consisting of similar cubes. An efficient ADMM-based algorithm is developed to solve the proposed model. Experiments on color images, MSIs data, and color videos demonstrate the effectiveness of the proposed method. In the future work, we will try to adaptively choose the penalty parameter of each group to enhance the performance and extend the proposed method to other image processing tasks, such as remotely sensed images recovery \cite{Chen2018Tensor-Decomposition,Fu2016Robust,He2016Total}, rain streaks removal \cite{Jiang2018FastDeRain,Wang2019rain}.

\appendices
\section*{Appendices}
\section{Proof of Proposition \ref{Proposition}}
\label{Appendix A}
\newtheorem{Proof}{\textbf{Proof}}
\begin{proof}
We denote the objective function of \eqref{proposed model} by $E(\mathcal{X})$. It is clear that $E(\mathcal{X})$ is convex, proper, and continuous. According to the Weierstrass' theorem \cite{Bertsekas2003convex}, it remains only to show the coercivity of $E(\mathcal{X})$, i.e., for every sequence $\{\mathcal{X}^{b}\}$ such that $\|\mathcal{X}^{b}\|_{F} \rightarrow \infty$, we have $\lim_{b\rightarrow\infty}E(\mathcal{X}^{b})=\infty$. We prove it by contradiction. Assume that there exists a subsequence of $\{\mathcal{X}^{b}\}$ (also denoted as $\{\mathcal{X}^{b}\}$) that $\{E(\mathcal{X}^{b})\}$ is bounded, we have that $\sum_{k=1}^{j-1}\alpha_{k}\|\textbf{X}_{[k]}\|_{\ast}$ is bounded. Due to the equivalence of norms, we get that $\{\|\mathcal{X}^{b}\|_{F}\}$ is bounded. Then $\{\mathcal{X}^{b}\}$ is a bounded sequence, which is contradictory with the assumption. So the model \eqref{proposed model} has at least one minimizer.
\end{proof}

\section{Proof of Theorems \ref{theorem1}, \ref{theorem2}, \ref{theorem3}, and \ref{theorem4}}
\label{Appendix B}
\emph{Proof of Theorem \ref{theorem1}}

\begin{proof}
Note that
\[
\begin{split}
\|\mathcal{E}\|_{F}^{2}=\|\mathcal{X}-\mathcal{Y}\|_{F}^{2}&=\sum_{i_{2}=1}^{s}\sum_{i_{3}=1}^{n_{3}}\sum_{i_{4}=1}^{h}\|\mathcal{X}(:,i_{2},i_{3},i_{4})-\textbf{x}\|_{2}^{2}\\
&\leq sn_{3}h\varepsilon^{2}.
\end{split}
\]
Therefore, $\|\mathcal{E}\|_{F}\leq\sqrt{sn_{3}h}\varepsilon$. This completes the proof.
\end{proof}

\vspace{0.5cm}
\emph{Proof of Theorem \ref{theorem2}}

\begin{proof}
Note that
\[
\|\hat{\mathcal{E}}\|_{F}^{2}=\|\mathcal{X}-\hat{\mathcal{Y}}\|_{F}^{2}=\sum_{i_{4}=1}^{h}\|\mathcal{X}(:,:,:,i_{4})-\mathcal{\hat{X}}\|_{F}^{2}\leq h\hat{\varepsilon}^{2}.
\]
Therefore, $\|\hat{\mathcal{E}}\|_{F}\leq\sqrt{h}\hat{\varepsilon}$.
\end{proof}

\vspace{0.5cm}
\emph{Proof of Theorem \ref{theorem3}}

\begin{proof}
From the definition of $\mathcal{\tilde{E}}$, we get that
\[
\begin{split}
\|\mathcal{\tilde{E}}\|_{F}^{2}&=\|\mathcal{X}-\mathcal{\tilde{Y}}\|_{F}^{2}\\
&=\sum_{i_{4}=r+1}^{h}\|\mathcal{X}(:,:,:,i_{4})-\mathcal{X}(:,:,:,1)\|_{F}^{2}\\
&\leq (h-r)\tilde{\varepsilon}^{2}.
\end{split}
\]
Hence $\|\mathcal{\tilde{E}}\|_{F}\leq\sqrt{h-r}\tilde{\varepsilon}$.
\end{proof}

\vspace{0.5cm}
\emph{Proof of Theorem \ref{theorem4}}

\begin{proof}
From the definition of $s(\mathcal{X})$, we get that
\[
\begin{split}
|s(\mathcal{X})-s(\mathcal{Y})|&=\bigg|\sum_{k=1}^{3}\alpha_{k}(\|\textbf{X}_{[k]}\|_{\ast}-\|\textbf{Y}_{[k]}\|_{\ast})\bigg|\\
&=\bigg|\sum_{k=1}^{3}\alpha_{k}\Big(\sum_{j_{k}}\big(\sigma_{j_{k}}((\textbf{Y+E})_{[k]})-\sigma_{j_{k}}(\textbf{Y}_{[k]})\big)\Big)\bigg|\\
&\leq\sum_{k=1}^{3}\alpha_{k}\Big(\sum_{j_{k}}\big|\sigma_{j_{k}}((\textbf{Y+E})_{[k]})-\sigma_{j_{k}}(\textbf{Y}_{[k]})\big|\Big)\\
&\leq\sum_{k=1}^{3}\alpha_{k}\Big(\sum_{j_{k}}\big(\|\textbf{E}_{[k]}\|_{F}\big)\Big)\\
&=c\|\mathcal{E}\|_{F}.\\
\end{split}
\]

Therefore, $|s(\mathcal{X})-s(\mathcal{Y})|\leq c\|\mathcal{E}\|_{F}$.
\end{proof}

\section*{Acknowledgments}
The authors would like to thank the authors \cite{Bengua2017Efficient,Liu2013tensor,Zhang2017tSVD} for providing the free download of the source code. This research is supported by the National Science Foundation of China (61772003, 61876203, 11901450), the HKRGC GRF
(12306616, 12200317, 12300519, 12300218), HKU Grant (104005583), the National Postdoctoral Program for Innovative Talents (BX20180252), the Project funded by China Postdoctoral Science Foundation (2018M643611), and Science Strength Promotion Programme of UESTC.

\bibliographystyle{plain}
\bibliography{NLR_TT}
\end{document}